\documentclass{article}

\usepackage[main, final]{neurips_2026}

\usepackage[utf8]{inputenc} 
\usepackage[T1]{fontenc}    
\usepackage{hyperref}       
\usepackage{caption}
\usepackage{url}            
\usepackage{booktabs}       
\usepackage{amsfonts}       
\usepackage{nicefrac}       
\usepackage{microtype}      
\usepackage{xcolor}         
\usepackage{graphicx}       
\usepackage{multirow}       
\usepackage{amsmath}        
\usepackage{xspace}         
\usepackage{tabularx}
\usepackage{xcolor}
\usepackage{wrapfig}
\usepackage{cleveref}
\usepackage{fontawesome5}
\usepackage{roboto-mono}

\newcommand{\model}{\textsc{RefDecoder}\xspace}

\newcommand{\eg}{\textit{e.g.}\xspace}
\newcommand{\ie}{\textit{i.e.}\xspace}

\title{\vspace{-0.45em}\raisebox{-0.4em}{\includegraphics[height=1.6em]{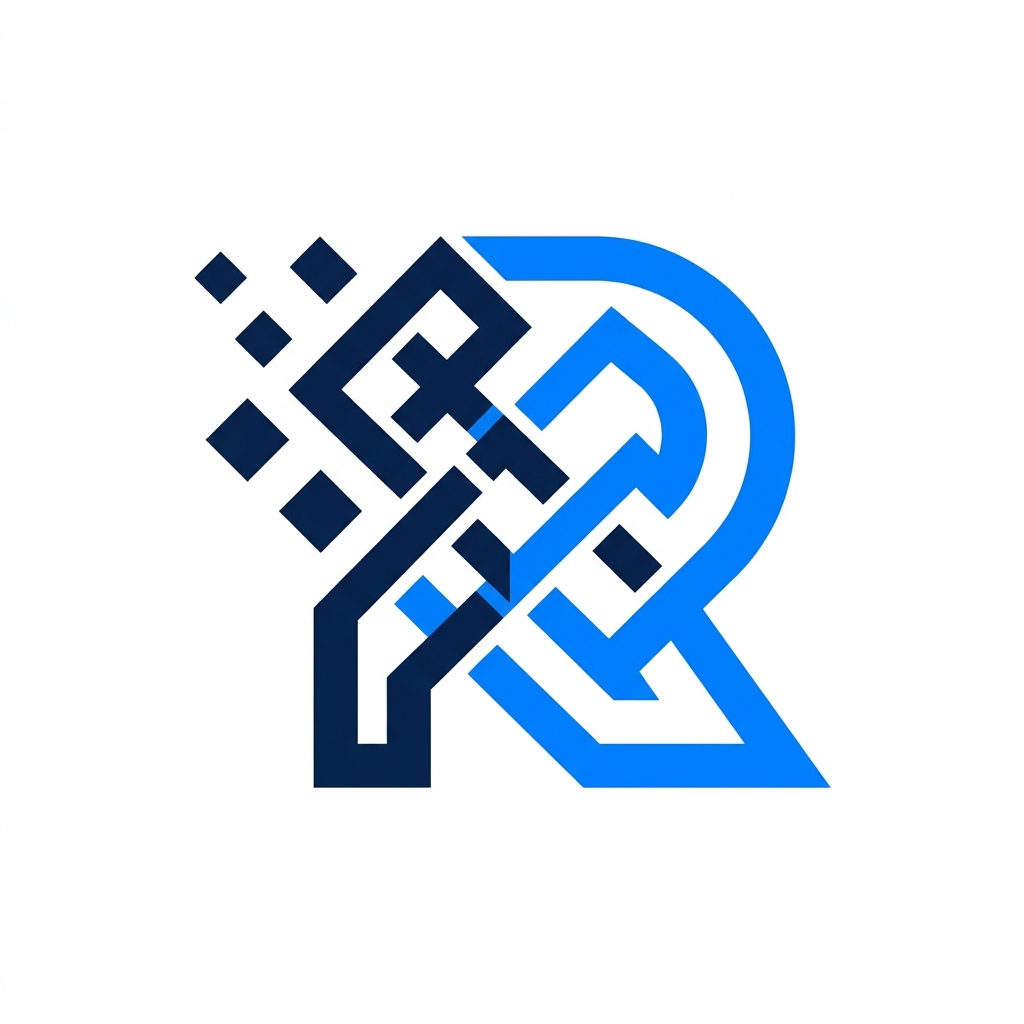}}\,\model: Enhancing Visual Generation with Conditional Video Decoding}

\author{%
  Xiang Fan$^{1*}$\quad Yuheng Wang$^{1*}$\quad Bohan Fang$^{1*}$\quad Zhongzheng Ren$^{1,2}$\quad Ranjay Krishna$^1$ \\
  $^1$University of Washington\quad $^2$University of North Carolina at Chapel Hill \\
  $^*$Equal contribution \\
  {\color[HTML]{E8A317}\faEnvelope}\ \texttt{xiangfan@cs.washington.edu} \\
  {\color[HTML]{367DD4}\faGlobe}\ \url{https://refdecoder.github.io/}
}

\begin{document}

\maketitle
\begin{center}
  \vspace{-2em}
  \includegraphics[width=\linewidth]{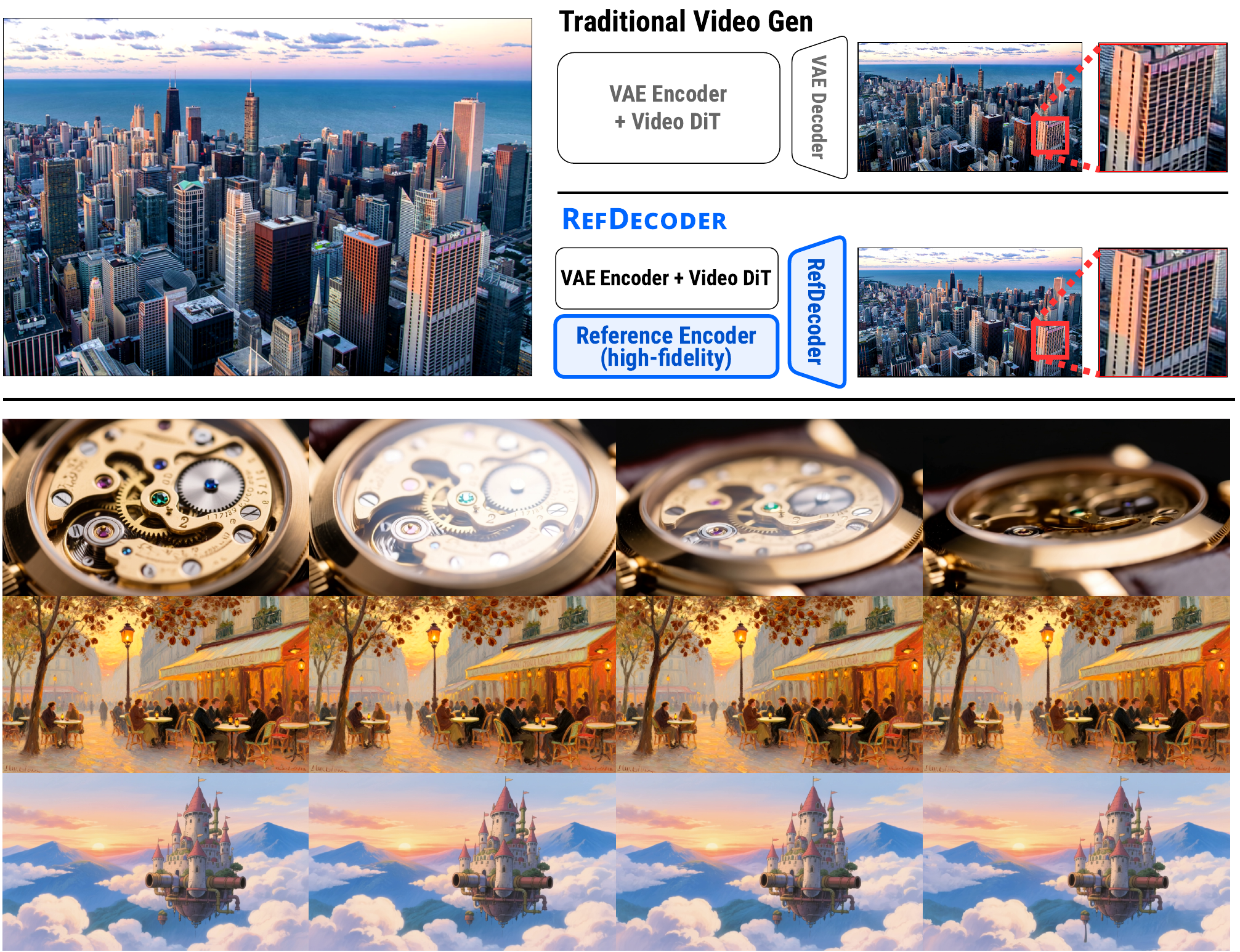}
  \captionof{figure}{(a) \textbf{\model} improves video VAE decoders by conditioning on a high-fidelity reference signal that bypasses the lossy VAE latent round trip. Given the encoded latents, the decoder injects fine-grained details that are not preserved in the VAE latent space, improving reconstruction fidelity and downstream generation quality. (b) \textbf{\model} integrates with existing video generation pipelines (Wan 2.1 shown) without requiring costly re-training of the diffusion model.}
  \label{fig:teaser}
  \vspace{1em}
\end{center}

\begin{abstract}

Video generation powers a vast array of downstream applications.
However, while the \textit{de facto} standard, \ie,  latent diffusion models, typically employ heavily conditioned denoising networks, their decoders often remain unconditional.
We observe that this \textbf{architectural asymmetry} leads to significant loss of detail and inconsistency relative to the input image.
To address this, we argue that the decoder requires equal conditioning to preserve structural integrity.
We introduce \model, a reference-conditioned video VAE decoder by injecting high-fidelity reference image signal directly into the decoding process via reference attention.
Specifically, a lightweight image encoder maps the reference frame into the detail-rich high-dimensional tokens, which are co-processed with the denoised video latent tokens at each decoder up-sampling stage. 
We demonstrate consistent improvements across several distinct decoder backbones (\eg, Wan~2.1 and VideoVAE\textbf{+}), achieving up to $+2.1$\,dB PSNR over the unconditional baselines on the Inter4K, WebVid, and Large Motion reconstruction benchmarks. Notably, \model can be directly swapped into existing video generation systems without additional fine-tuning, and we report across-the-board improvements in subject consistency, background consistency, and overall quality scores on the VBench I2V benchmark. Beyond I2V, \model generalizes well to a wide range of visual generation tasks such as style transfer and video editing refinement.

\end{abstract}

\section{Introduction}
\label{sec:intro}

In the dominant paradigm of video generation, a diffusion backbone denoises latent representations guided by an input conditioning, while a VAE decoder subsequently maps these latents back to pixel space~\cite{blattmann2023stablevideodiffusionscaling,wan2025,yang2024cogvideox}.
While substantial effort has been devoted to improving the conditioning and architecture of the diffusion backbone~\cite{rombach2022highresolutionimagesynthesislatent, ho2020denoising, song2019generative}, \emph{the VAE decoder has received comparatively little attention and remain unconditional in mainstream video diffusion models}. It is typically treated as a standalone reconstruction module that operates without access to the primary conditioning signal at inference time~\cite{blattmann2023stablevideodiffusionscaling,wan2025,yang2024cogvideox,kong2025hunyuanvideosystematicframeworklarge}.

We observe that this \textbf{architectural asymmetry} creates a systematic bottleneck: even when the diffusion model faithfully preserves the reference information within the latent, the decoder must reconstruct fine-grained spatial details from a heavily compressed representation without a detailed anchor.
This leads to two characteristic failure modes:
(1) \emph{Progressive degradation of spatial details}, where textures, edges, and high-frequency content deteriorate in frames that deviate from the conditioning image (e.g.\ \Cref{fig:qual_recon} baseline details); and
(2) \emph{Temporal inconsistency}, where appearance drifts across the sequence (e.g.\ \Cref{fig:qual_gen}(a) human face).
These artifacts emerge during latent-to-pixel decoding rather than the diffusion process itself, exposing the decoder as an under-exploited component that fails to leverage the visual cues already provided by the reference.

However, injecting the reference image into the decoder is non-trivial for several reasons.
(1) Because the decoder operates through hierarchical up-sampling stages, it remains an open question at which specific stage the reference signal should be introduced. 
(2) The conditioning mechanism must maintain compatibility with pretrained decoder weights, ensuring the system remains functional in its original setting, \ie, no reference image. 
(3) The approach should be sufficiently generic to generalize across diverse visual generation tasks and various decoder backbones.

To address aforementioned problems, our key insight is that the decoder's hidden space is far richer than the VAE latent space; encoding the reference image into this high-fidelity space and injecting it as additional self-attention tokens lets the decoder exploit details unavailable from the latent code alone.
Building on this, we propose \textbf{\model}, a reference-conditioned video VAE decoder with two components: 
a \textbf{reference image encoder}, a lightweight convolutional network mapping the reference frame into the decoder's hidden space as spatially aligned tokens (Sec.~\ref{sec:ref_enc}(a));
and a \textbf{conditional token decoder} that, at each up-sampling stage, concatenates reference and video latent tokens along the temporal axis and jointly processes them via a shared transformer block with rotary positional embeddings (RoPE)~\cite{su2021roformer}; the tokens are then separated and up-sampled independently, preserving compatibility with the pretrained decoder (Sec.~\ref{sec:cond_dec}(b)).
To prevent the decoder from ignoring the reference, we further introduce a \textbf{latent token dropout} strategy that randomly zeroes video latent tokens at a variable rate, forcing the model to recover spatial details from reference tokens via attention (Sec.~\ref{sec:dropout}(c)).

Our newly introduced modules are architecture-agnostic and generalize to existing video VAE decoders, with the transformer block shared across decoder stages. Because only the decoder is modified and the encoder remains frozen, \model is a \textbf{drop-in replacement} for any existing video VAE decoder and can immediately benefit downstream diffusion models without retraining.

We test \model on different video VAE backbones including Wan~2.1~\cite{wan2025} and VideoVAE\textbf{+}~\cite{xing2025videovae+}, demonstrating consistent improvements across all of them.
On the Inter4K reconstruction benchmark, \model achieves over 1\,dB PSNR improvement compared to the unconditional baseline.
In generation tasks, on the VBench I2V benchmark~\cite{huang2023vbench,huang2025vbench++}, our method improves subject consistency, background consistency, motion smoothness, aesthetic quality, and overall scores.
We show that the same reference injection mechanism naturally extends to \emph{decode-time style transfer}. By supplying a style image as the reference, \model produces stylized video without any task-specific architecture modification, validating the generality of conditional token decoding. Furthermore, in video editing tasks, \model enhances the fidelity of edited videos to the input reference while preserving the intended edits, demonstrating that reference conditioning as a powerful tool to mitigate the trade-off between editability and fidelity~\cite{fan2024videoshoplocalizedsemanticvideo}.
These results suggest a broader message: in image conditioned generation pipelines, the decoder is not merely a passive reconstruction module but an active participant that can---and should---leverage conditioning signals. Don't forget the decoder.

\section{Method}
\label{sec:method}

\begin{figure}[t]
    \vspace{-3em}
  \centering
  \includegraphics[width=\textwidth]{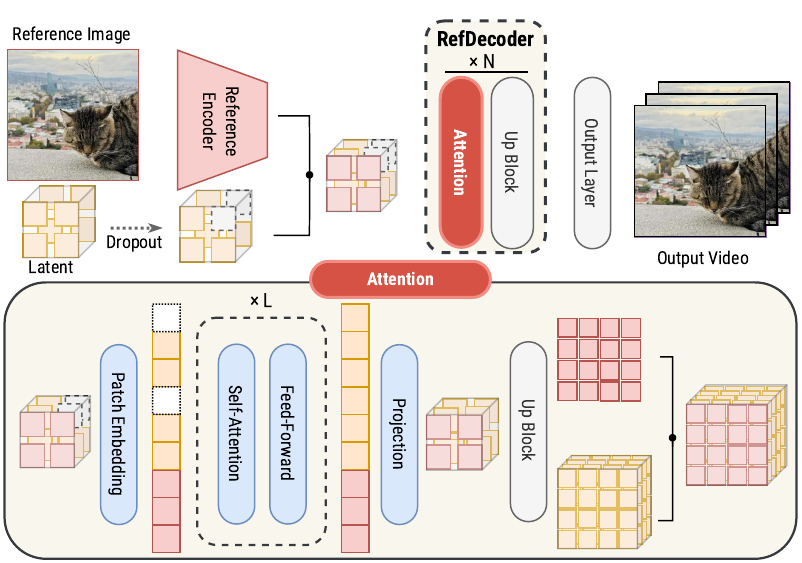}
  \caption{\textbf{Architecture overview of the \model.} The reference image is encoded into tokens and injected into the decoder through shared attention blocks, enabling interactions between reference and latent tokens during decoding.}
  \label{fig:arch}
  \vspace{-1.5em}
\end{figure}

We propose \textbf{\model}, a reference-conditioned video VAE decoder that injects reference image tokens directly into the decoding process via joint attention over concatenated reference and video tokens, enhancing spatial fidelity and temporal coherence. The overall architecture is illustrated in Fig.~\ref{fig:arch}. We build upon a standard video VAE backbone and augment its decoder with conditional token interactions, which we detail below.

\noindent\textbf{Reference image encoding.}
\label{sec:ref_enc}
We start with a reference image $I_{\mathrm{ref}} \in \mathbb{R}^{3 \times H \times W}$. Typical methods encode the reference into a low-dimensional feature space comparable to the VAE latent bottleneck (\eg 16 channels), and obtain hierarchical features derived therefrom. To preserve high-fidelity information, we instead project the original image patches directly into a high-dimensional feature space (\eg 512 dimensions), which is fed to the decoder at the very first stage, upsampled alongside the video features throughout decoding stages, and interacted with through injected attention layers at every stage (\Cref{sec:cond_dec}(b)). This setup allows multiresolution high-dimensional feature extraction within the decoding process.

We keep the reference encoder minimal---applying a single convolution followed by normalization---to avoid any issue with a deeper encoder smoothing out high-frequency details.

After encoding, we obtain reference tokens $\mathbf{z}_{\text{ref}} \in \mathbb{R}^{C  \times \frac{H}{p} \times \frac{W}{p}}$, where $C$ is the first decoder stage's channel dimension and $p$ is the VAE's spatial compression ratio. This aligns reference tokens with the decoder's initial feature map, enabling direct position-aware information transfer through attention.

\noindent\textbf{Conditional token decoding.}
\label{sec:cond_dec}
Standard video VAE decoders rely on causal 3D convolutional upsampling layers with limited local receptive fields, which cannot recover fine-grained details lost to the narrow latent bottleneck. To let the decoder selectively retrieve details from the reference, we insert Transformer blocks into the upsampling stages of the pretrained decoder, enabling each video token to query reference tokens and extract relevant information.

Our conditional token decoding involves three design goals: (1) combining reference and video tokens for joint processing; (2) preserving compatibility with the existing VAE upsampling path; and (3) minimizing parameter overhead.

\noindent\textit{Token concatenation.}
At decoder stage $s$, we concatenate video tokens $\mathbf{z}^{(s)} \in \mathbb{R}^{C_s \times T_s \times H_s \times W_s}$ and reference tokens $\mathbf{z}_{\mathrm{ref}}^{(s)} \in \mathbb{R}^{C_s \times 1 \times H_s \times W_s}$ along the temporal dimension, yielding a tensor of shape $\mathbb{R}^{C_s \times (1 + T_s) \times H_s \times W_s}$. This is then processed through self attention in a Transformer block.

\noindent\textit{Separation and upsampling.}
After attention, the output is split back into reference tokens 
$\hat{\mathbf{z}}_{\mathrm{ref}}^{(s)}$ and video tokens 
$\hat{\mathbf{z}}^{(s)}$. 
Each branch is then passed through the pre-trained upsampling modules independently. 
The reference tokens are upsampled only spatially, 
while the video tokens undergo both spatial and temporal upsampling.

\noindent\textit{Weight sharing across stages.}
To reduce the parameter overhead of Transformer blocks, Transformer weights are shared across all stages. 
Stage-specific patch embedding layers project varying channel dimensions 
$C_s$ to a unified transformer hidden dimension, enabling weight sharing without architectural constraints. We ablate the effect of the number of weight-shared blocks in \Cref{sec:ablation}.

\noindent\textbf{Latent token dropout.}
\label{sec:dropout}
To encourage the decoder to actively rely on the reference rather than solely on the video latents, we apply token dropout to the latents before decoding. At each training step, every spatiotemporal position in $\mathbf{z}$ is independently zeroed with probability $r$, where $r$ is sampled uniformly from $[0, r_{\mathrm{max}})$, forcing the model to retrieve missing details from the reference tokens via attention. We use $r_{\mathrm{max}} = 0.7$ by default.

\noindent\textbf{Training.}
We initialize the VAE from a pretrained video VAE and freeze the encoder entirely. The decoder is fine-tuned at a reduced learning rate, while the newly added modules (reference image encoder and Transformer blocks) are trained from scratch. The training objective is:
\begin{equation*}
    \mathcal{L} = \|\mathbf{x} - \hat{\mathbf{x}}\|_1 +\mathcal{L}_{\text{LPIPS}}(\mathbf{x}, \hat{\mathbf{x}})
\end{equation*}
where $\mathbf{x}$ and $\hat{\mathbf{x}}$ denote the ground-truth and reconstructed video frames respectively, and $\mathcal{L}_{\text{LPIPS}}$ is the LPIPS perceptual loss~\cite{zhang2018perceptual}.

\noindent\textit{Random frame reference selection.} While the reference image typically serves as the first frame at inference, the model should ideally be robust to a wider range of reference content. We therefore randomly select a frame from the input video as the reference during training, preventing overfitting to a fixed temporal relationship.

\noindent\textit{Two-stage curriculum training.} We first train on short clips of the minimum supported number of frames, then extend to the maximum frame count to generalize to long temporal contexts.

\noindent\textbf{Causal variant for long videos.}
\label{sec:causal}
We also fine-tune a causal variant (4K steps, same data) that decodes the latents chunk by chunk, like the Wan~2.1 VAE decoder, rather than all at once.

\vspace{-.5em}
\section{Experiments}
\vspace{-.5em}
\label{sec:experiments}

We evaluate \model on \textbf{video reconstruction} and \textbf{image-to-video generation}. Throughout, \model is used as a plug-and-play \emph{drop-in replacement} for the baseline VAE decoder: the encoder, the diffusion backbone, and the inference pipeline remain unchanged. We also ablate the number of Transformer blocks, latent token dropout, and the training curriculum.

\subsection{Experimental Setup}
\noindent\textbf{Training data.}
Our training set contains approximately 100K videos including videos from MiraData9K~\cite{li2025realcam, zheng2024cami2v}, DL3DV~\cite{li2025realcam, zheng2024cami2v}, and a subset of OpenVidHD-0.4M~\cite{nan2024openvid}.

\noindent\textbf{Benchmarks.}
For reconstruction, we use the three test sets provided by VideoVAE\textbf{+}~\cite{xing2025videovae+}---an Inter4K~\cite{stergiou2022adapoolexponentialadaptivepooling} test split of 500 high-quality videos, a subset of WebVid~\cite{Bain21}, and a Large Motion subset of 100 videos (80 from WebVid, 20 from Inter4K) curated for complex camera and object motion---using identical splits to ensure direct comparability with~\cite{xing2025videovae+} (see \Cref{sec:supp_recon_benchmarks}). For long videos, we reconstruct 81 frames on 500 videos randomly sampled from Inter4K (Inter4K-81; drawn independently of the Inter4K test list, with which it may overlap) and run VBench at 81 frames. We report PSNR, SSIM, and LPIPS for reconstruction, and the standard 12 VBench dimensions~\cite{huang2023vbench, huang2025vbench++, zheng2025vbench2} for generation.

\noindent\textbf{Implementation details.}
We apply \model to two architecturally distinct backbones, Wan~2.1~\cite{wan2025} and VideoVAE\textbf{+}~\cite{xing2025videovae+}, which differ substantially in their channel widths and decoder depth. We freeze the encoder and train only the decoder, and drop up to 70\% latent video tokens during training. We adopt a two-stage curriculum: for Wan~2.1, 5-frame clips at $480\times832$ followed by 17-frame clips; for VideoVAE\textbf{+}, 4-frame clips at $216\times216$ (as does \cite{xing2025videovae+}) followed by 16-frame clips at the same resolution. All 81-frame results use the causal variant \model{}$^\dagger$ (\Cref{sec:causal}), fine-tuned from the default checkpoint for 4K steps (about 12 hours) on the same data with the same reference sampling, and compared against the original Wan~2.1 decoder.
\label{sec:supp_long_video}

\noindent\textbf{Baselines.} We compare our method with the vanilla VAE autoencoders from frontier open source video models, including Wan~2.1~\cite{wan2025} and HunyuanVAE~\cite{kong2025hunyuanvideosystematicframeworklarge}, as well as state-of-the-art academic VAE models including VideoVAE\textbf{+}~\cite{xing2025videovae+} and Reducio~\cite{tian2025reducio}.

\begin{figure}[t]
\centering
    \centering
    \setlength{\tabcolsep}{1pt}
    \renewcommand{\arraystretch}{0.5}
    \begin{tabular}{ccccc}
        \rotatebox{90}{\small\hspace{10pt}Wan 2.1} &
        \includegraphics[width=0.235\textwidth]{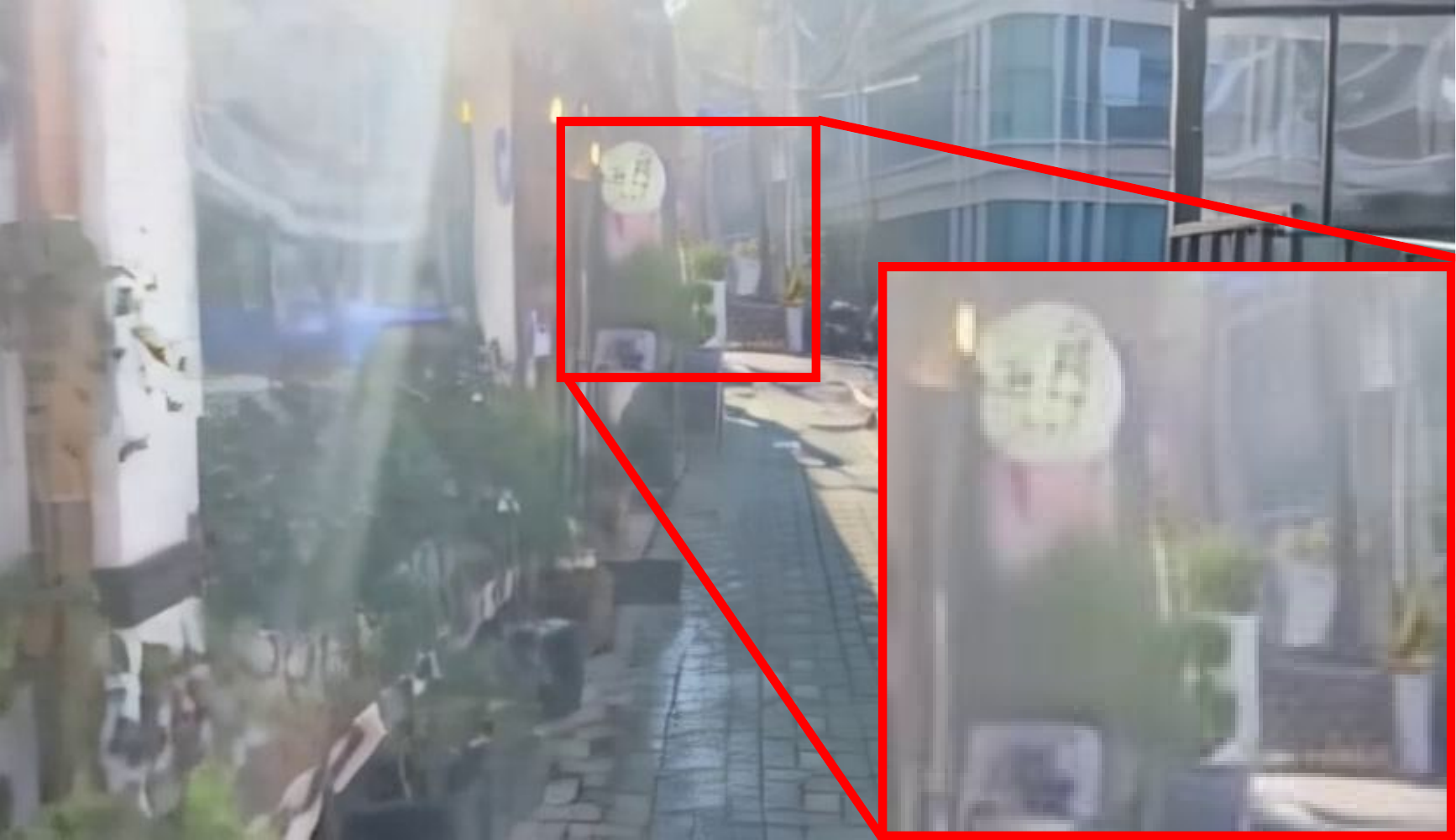} &
        \includegraphics[width=0.235\textwidth]{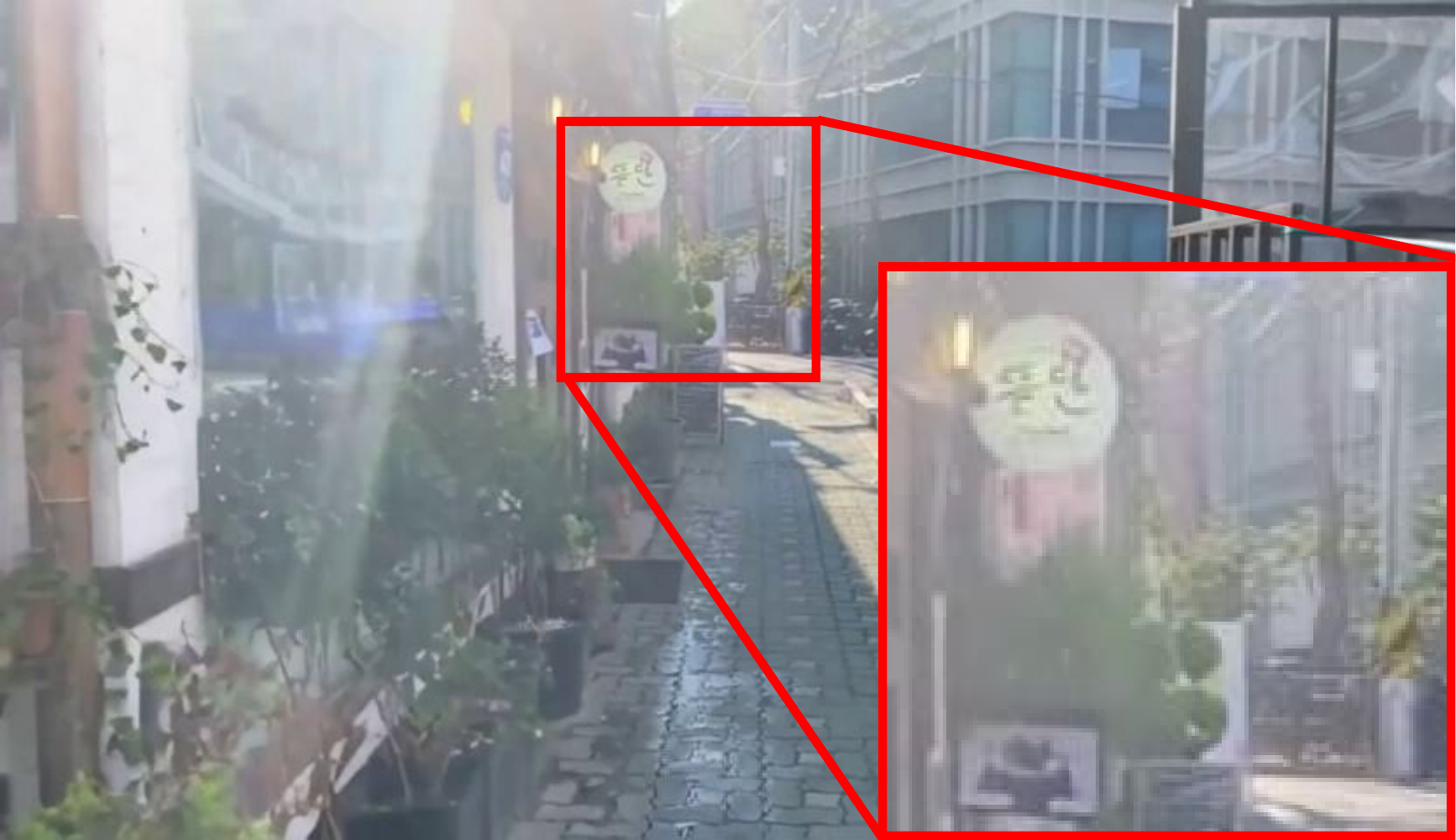} &
        \includegraphics[width=0.235\textwidth]{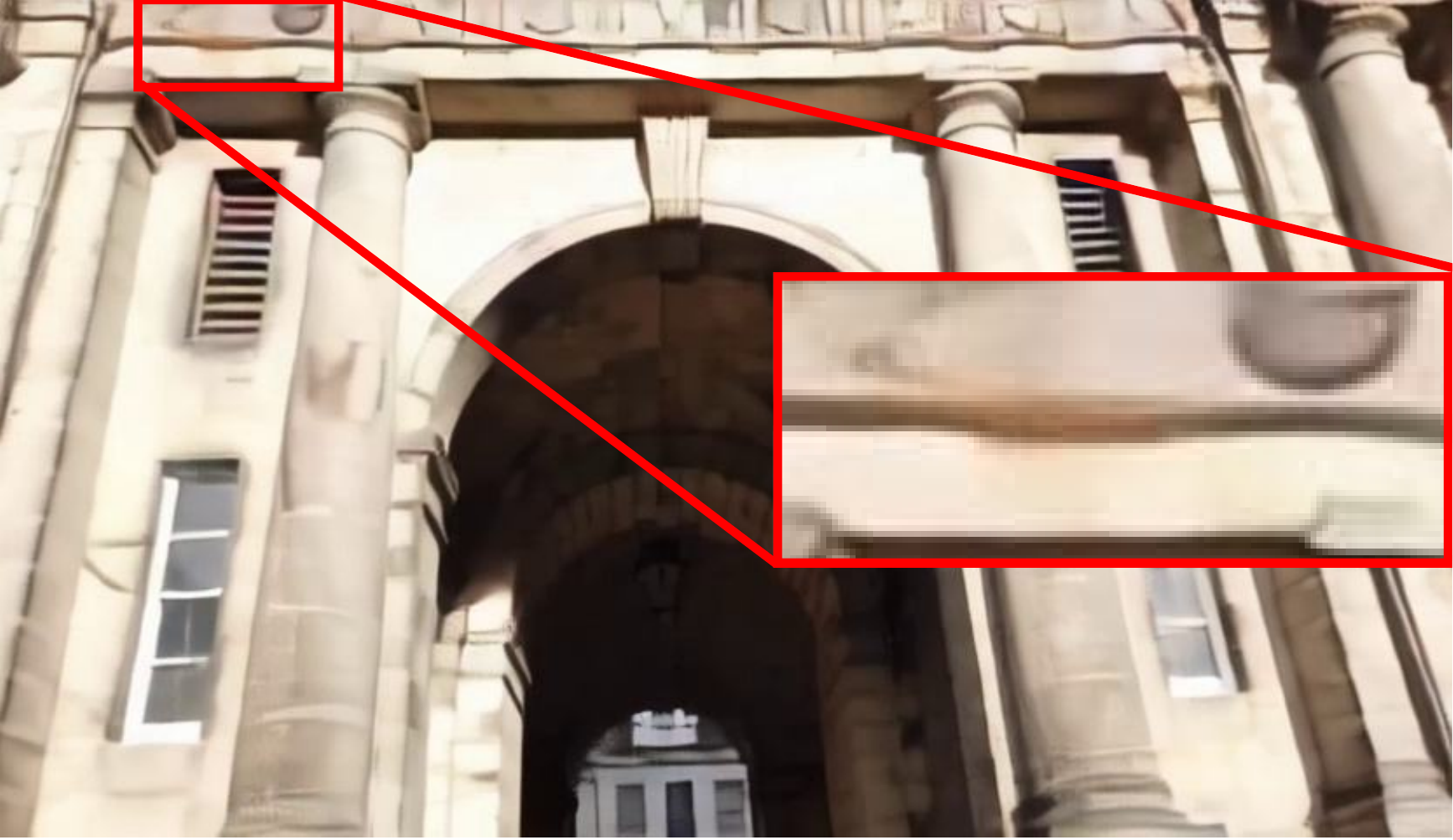} &
        \includegraphics[width=0.235\textwidth]{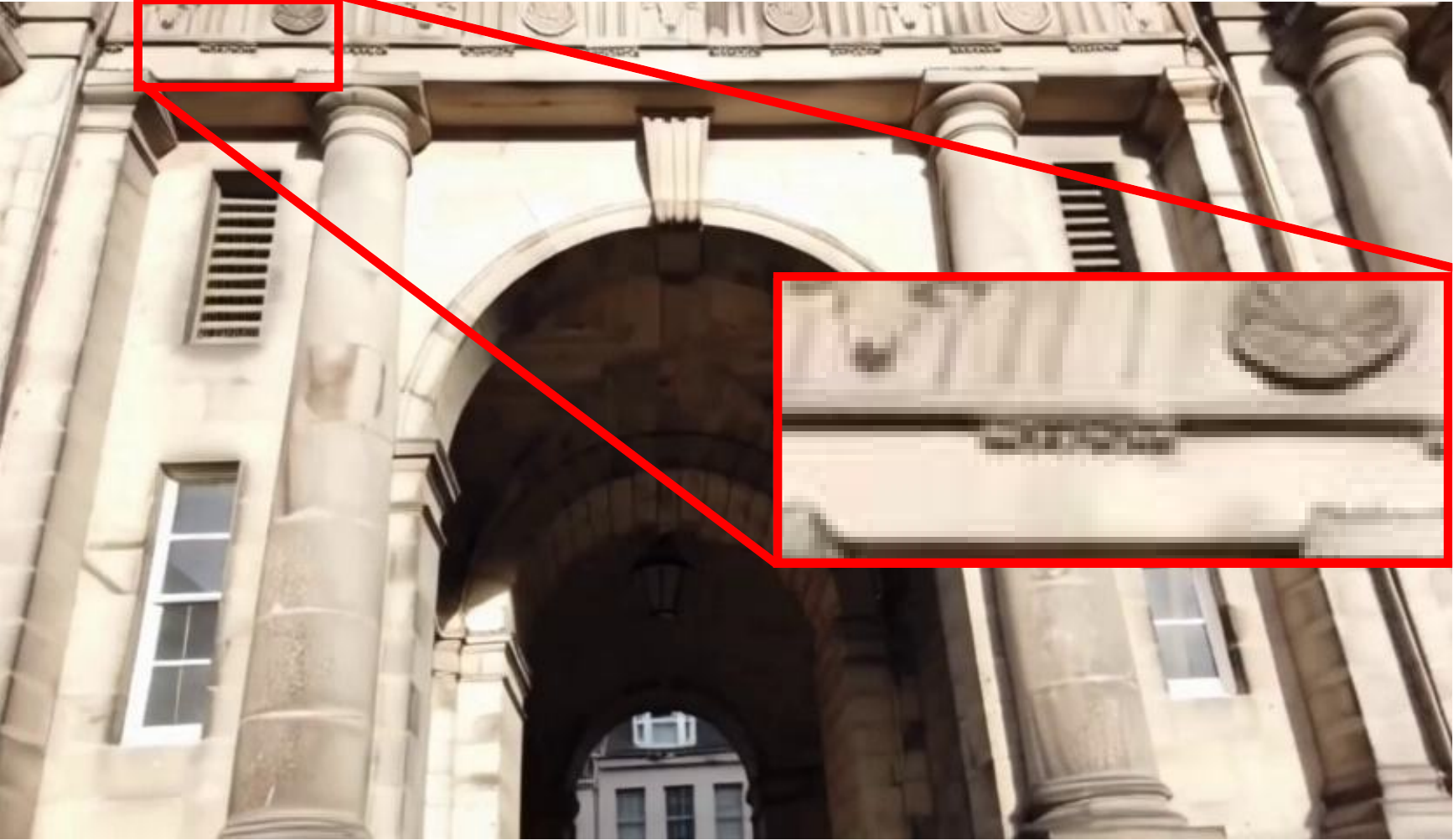} \\
        \rotatebox{90}{\small\hspace{10pt}Wan 2.1} &
        \includegraphics[width=0.235\textwidth]{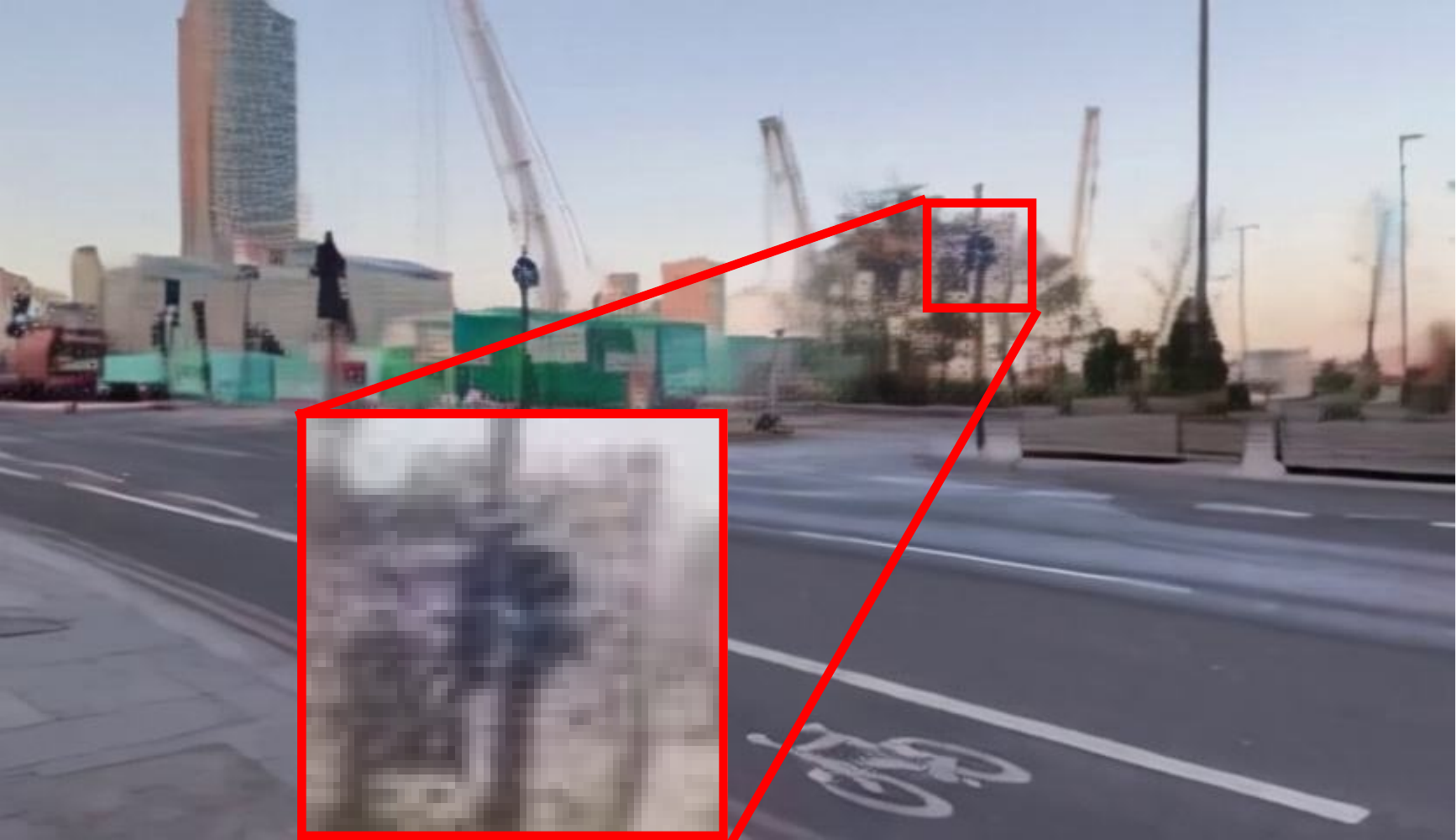} &
        \includegraphics[width=0.235\textwidth]{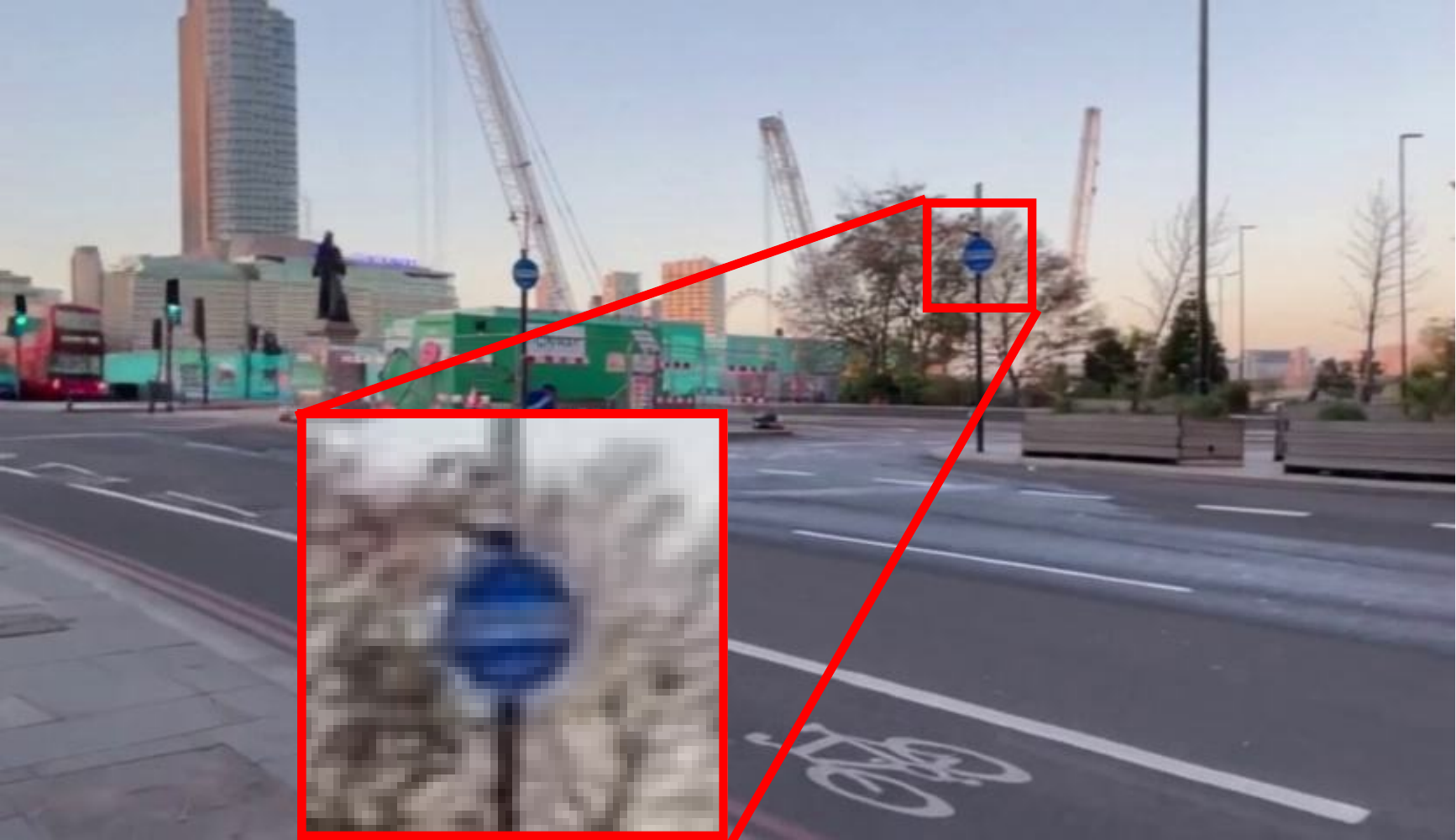} &
        \includegraphics[width=0.235\textwidth]{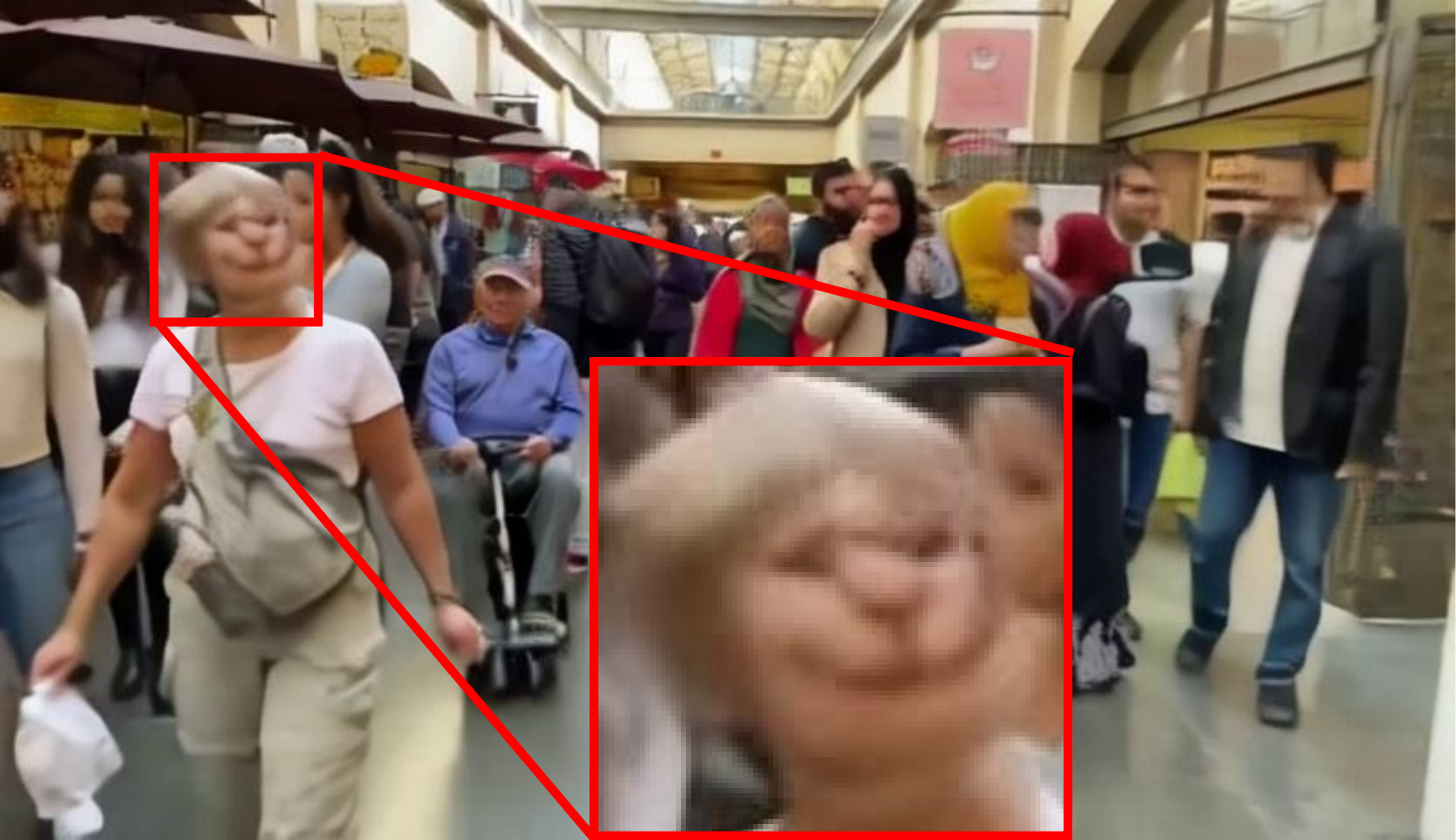} &
        \includegraphics[width=0.235\textwidth]{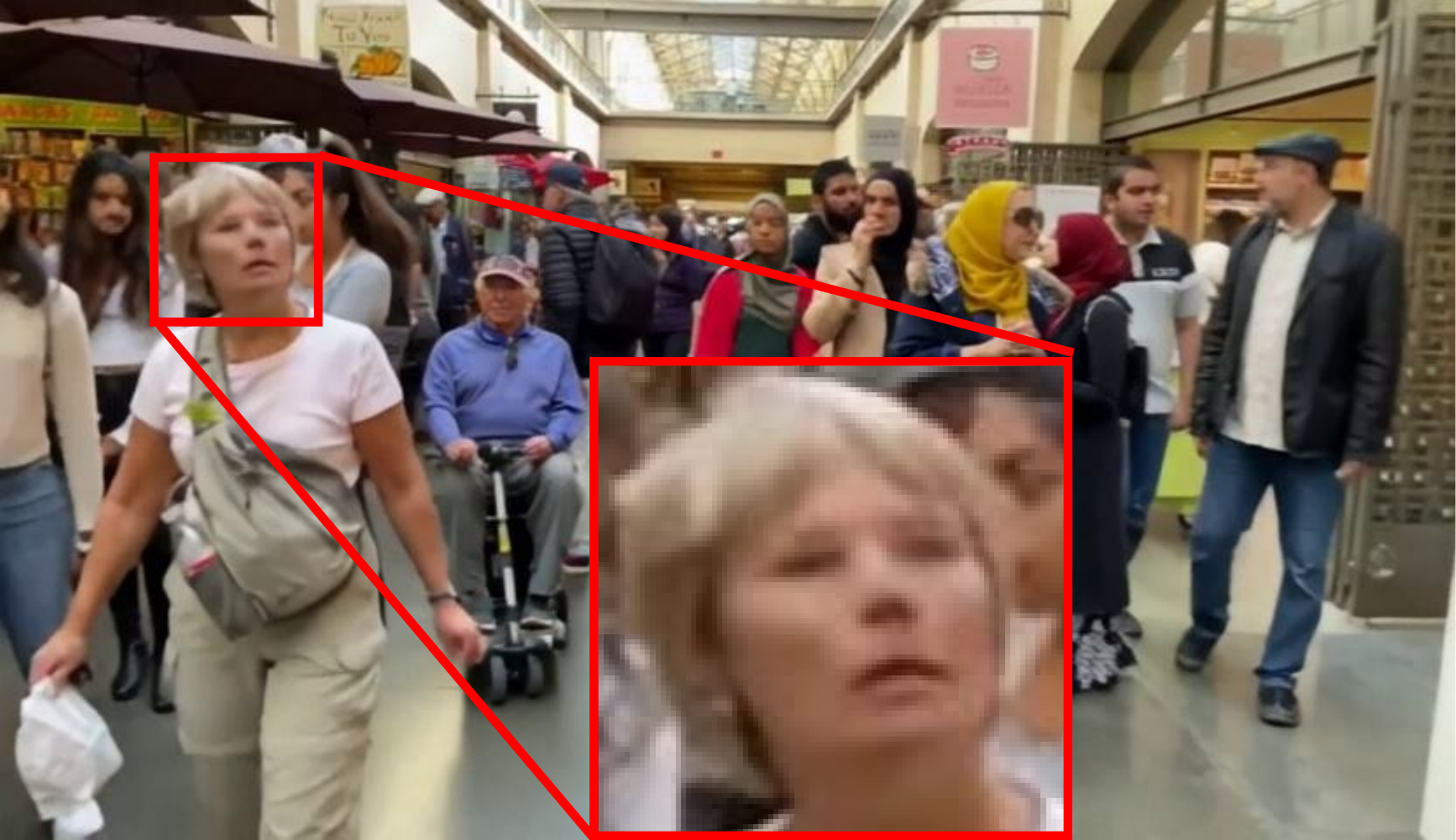} \\
        \rotatebox{90}{\small\hspace{10pt}Wan 2.1} &
        \includegraphics[width=0.235\textwidth]{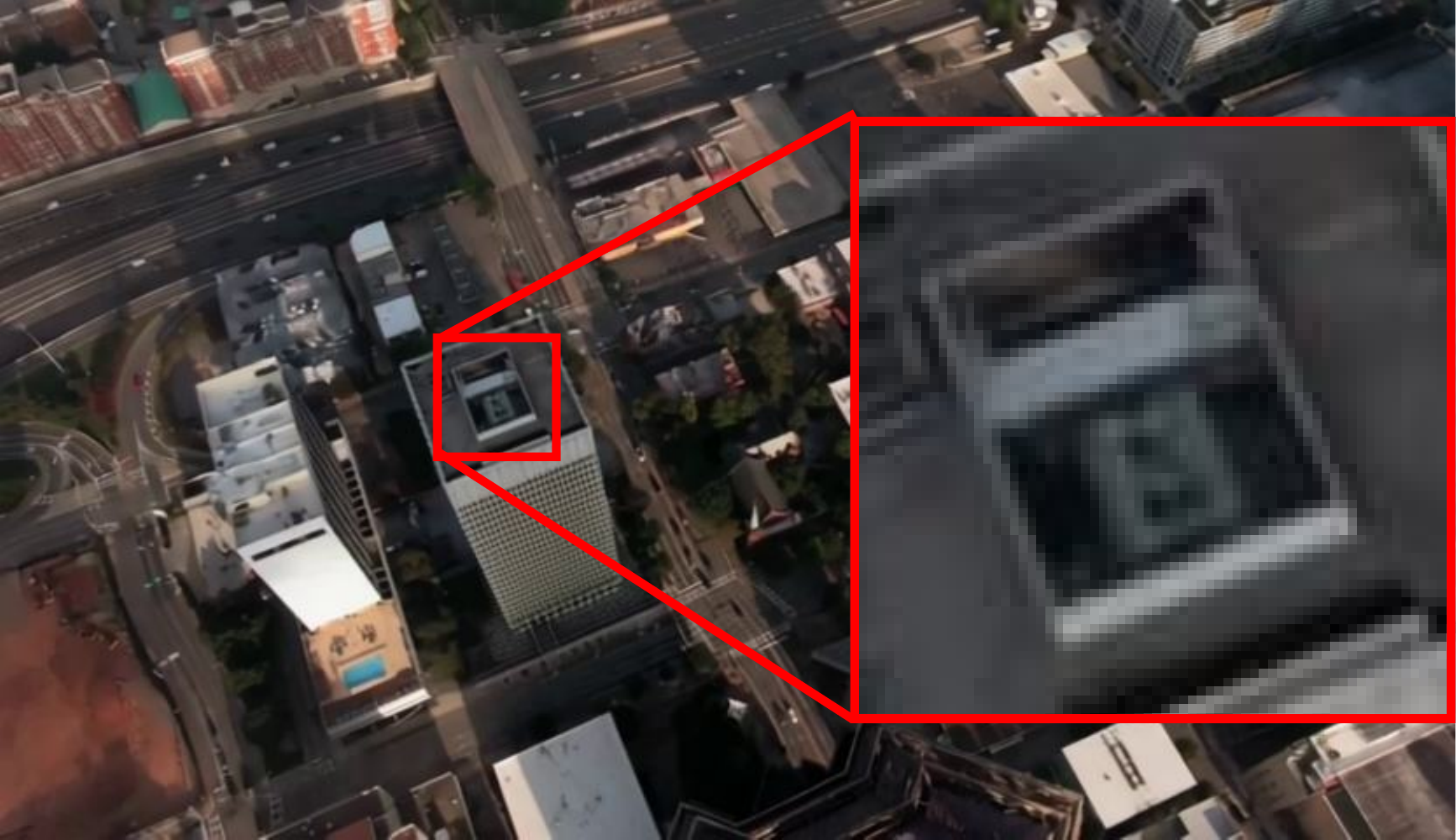} &
        \includegraphics[width=0.235\textwidth]{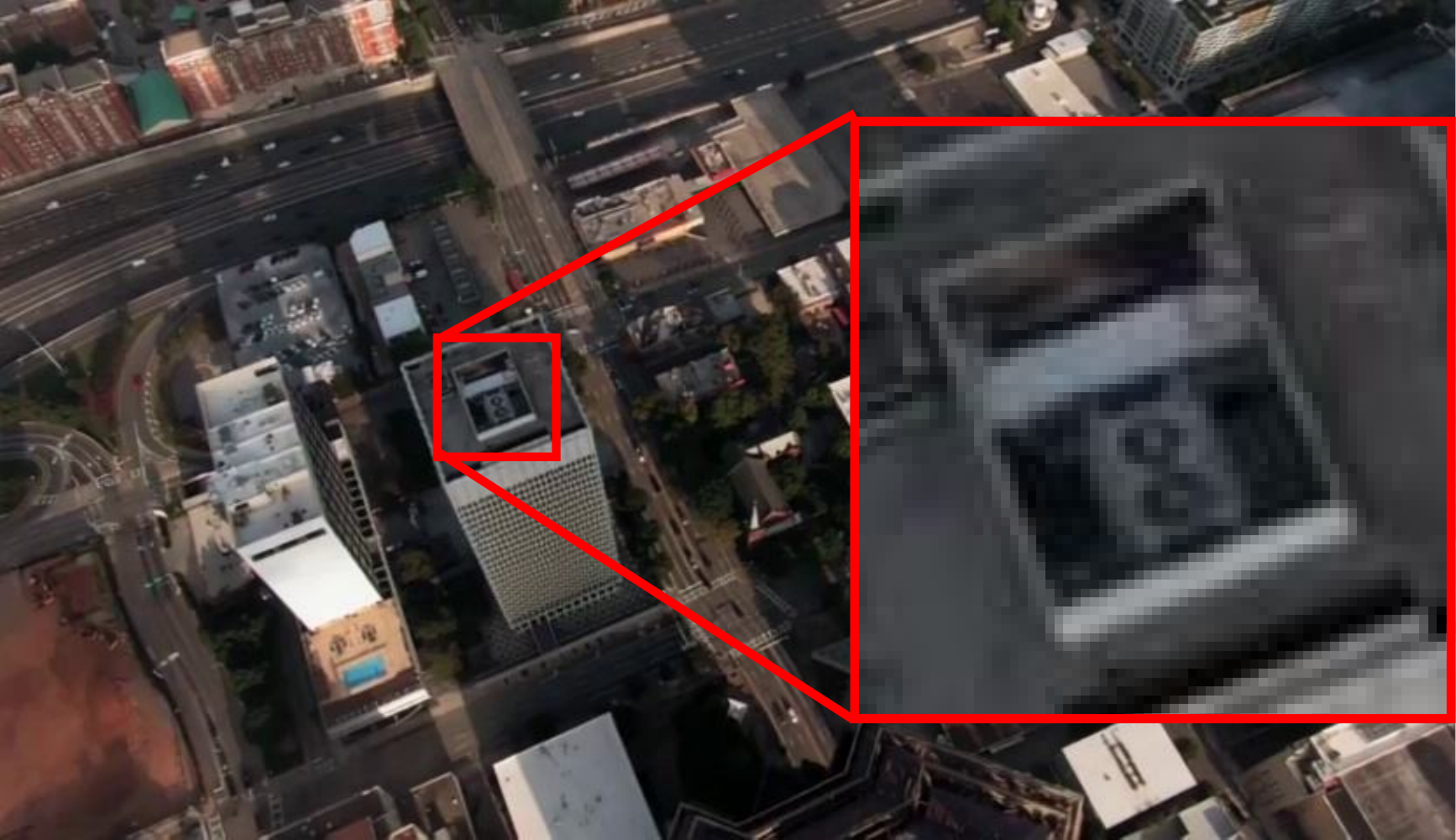} &
        \includegraphics[width=0.235\textwidth]{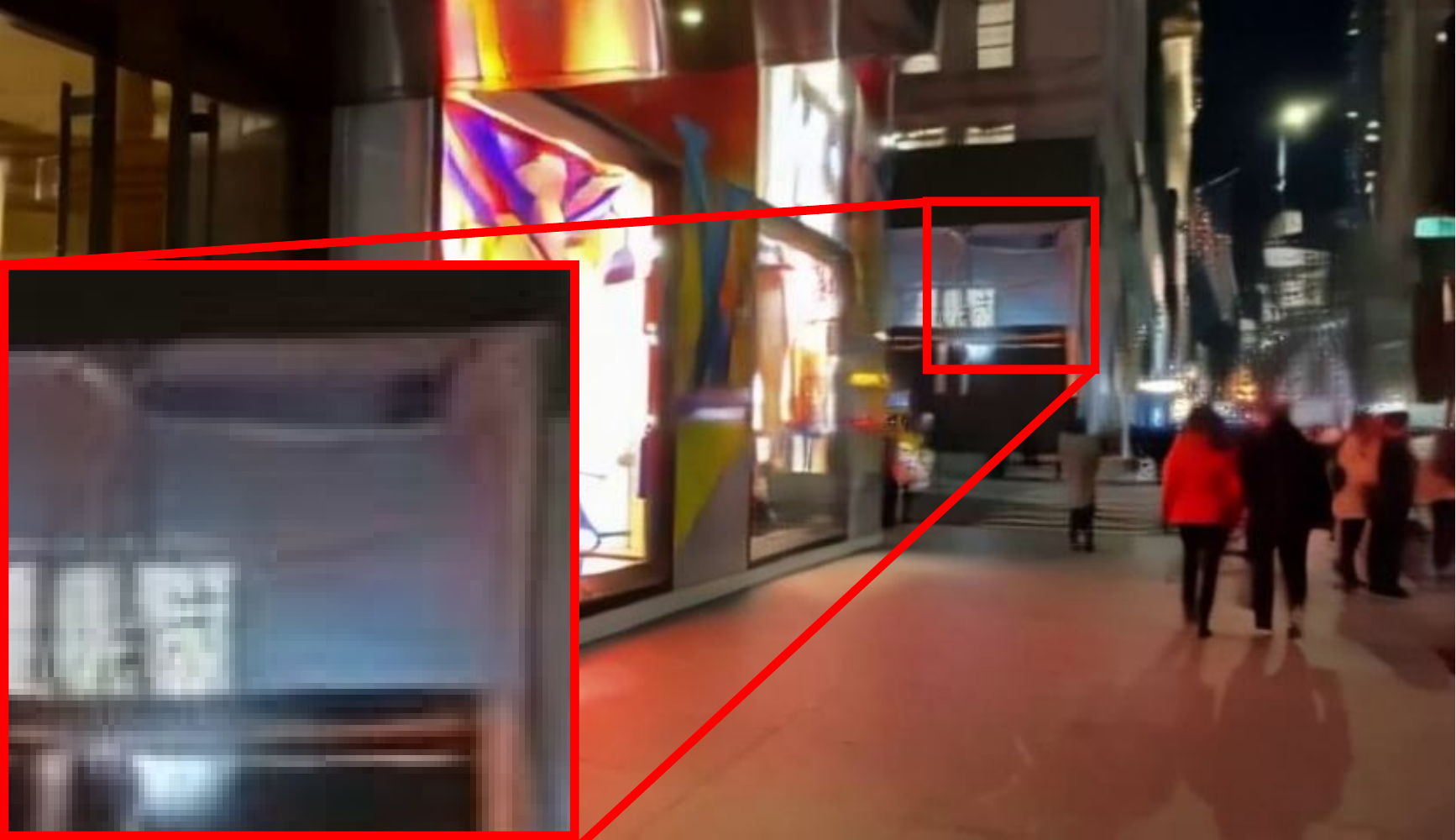} &
        \includegraphics[width=0.235\textwidth]{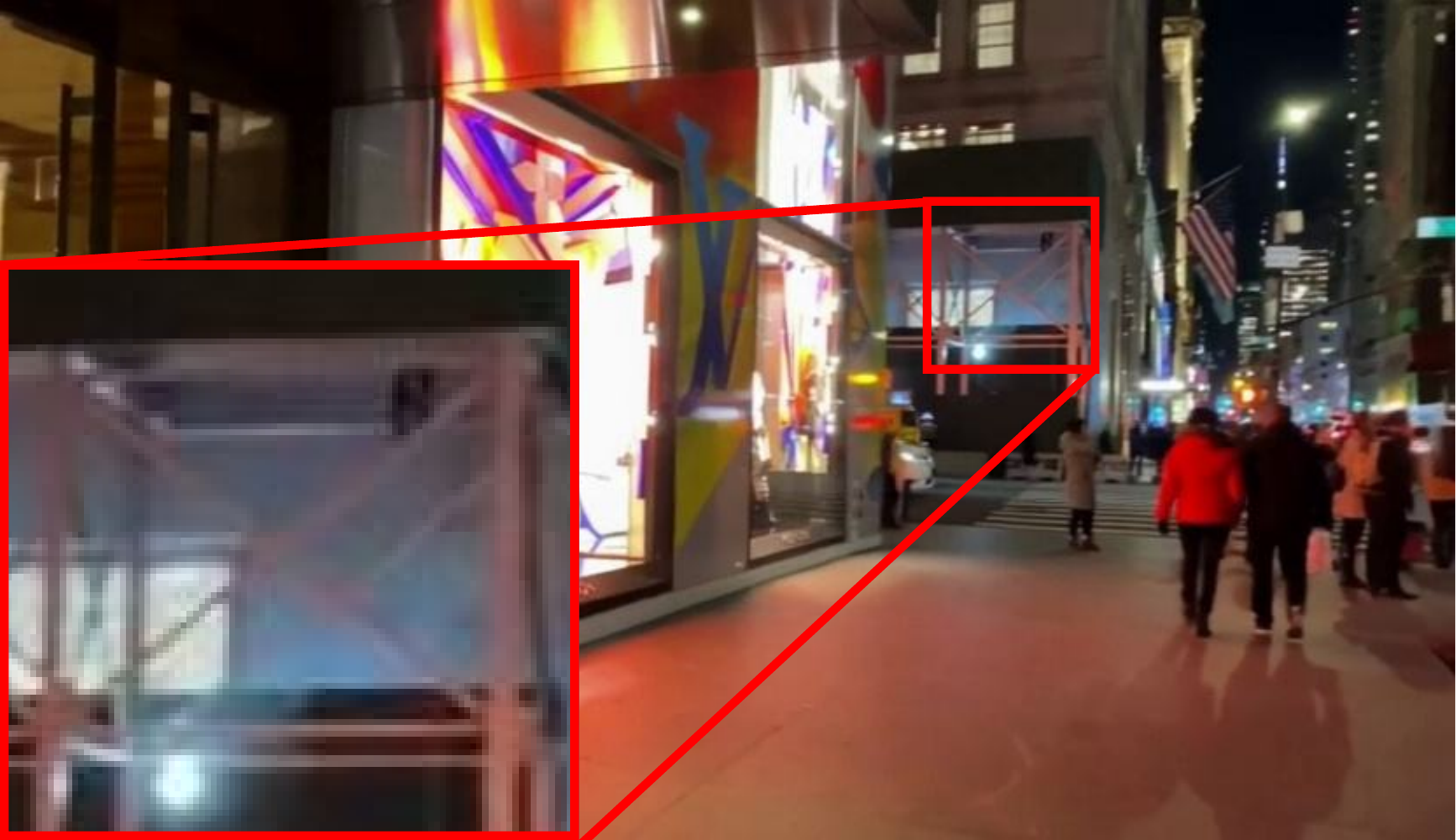} \\

        \rotatebox{90}{\small\hspace{4pt}VideoVAE+} &
        \includegraphics[width=0.235\textwidth]{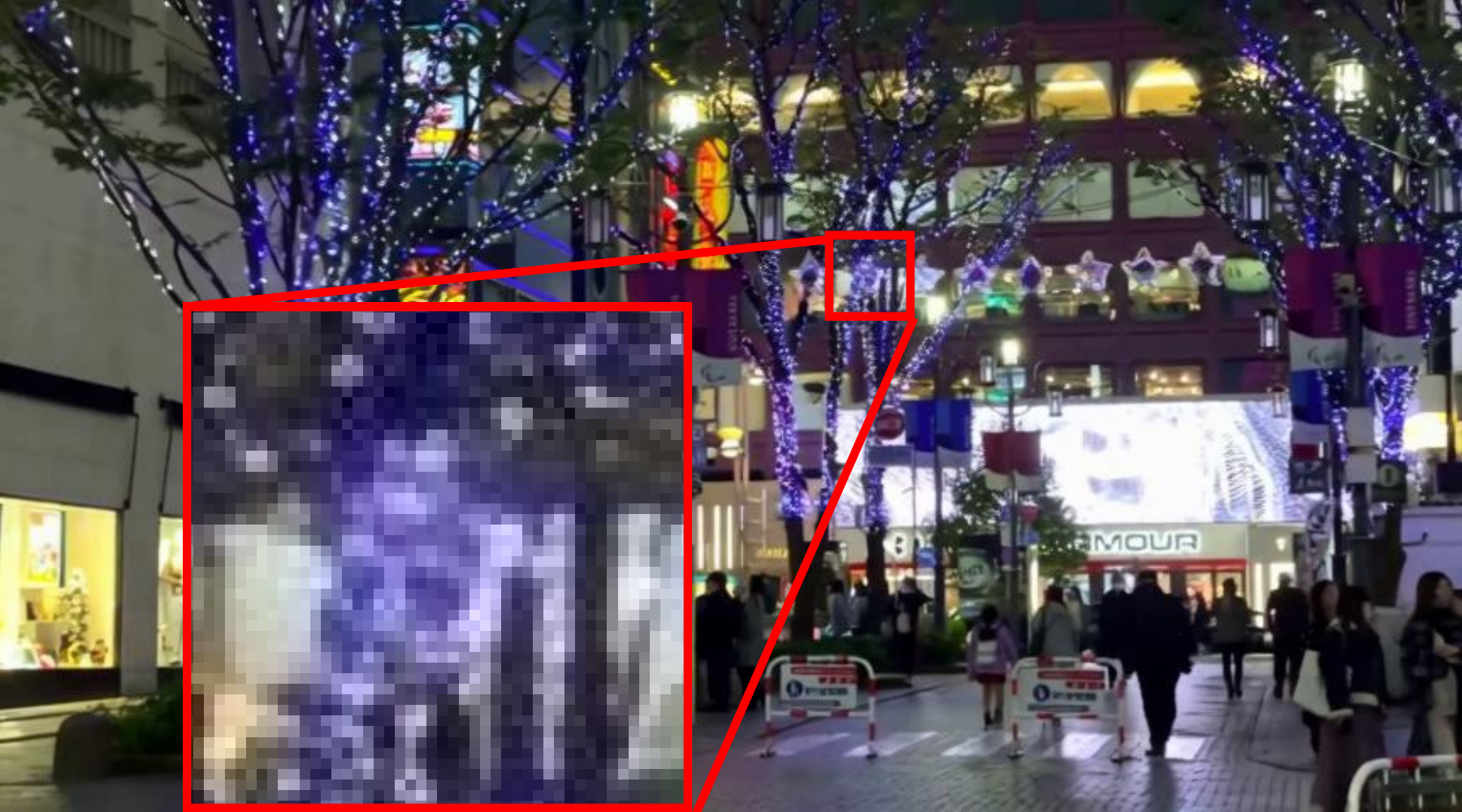} &
        \includegraphics[width=0.235\textwidth]{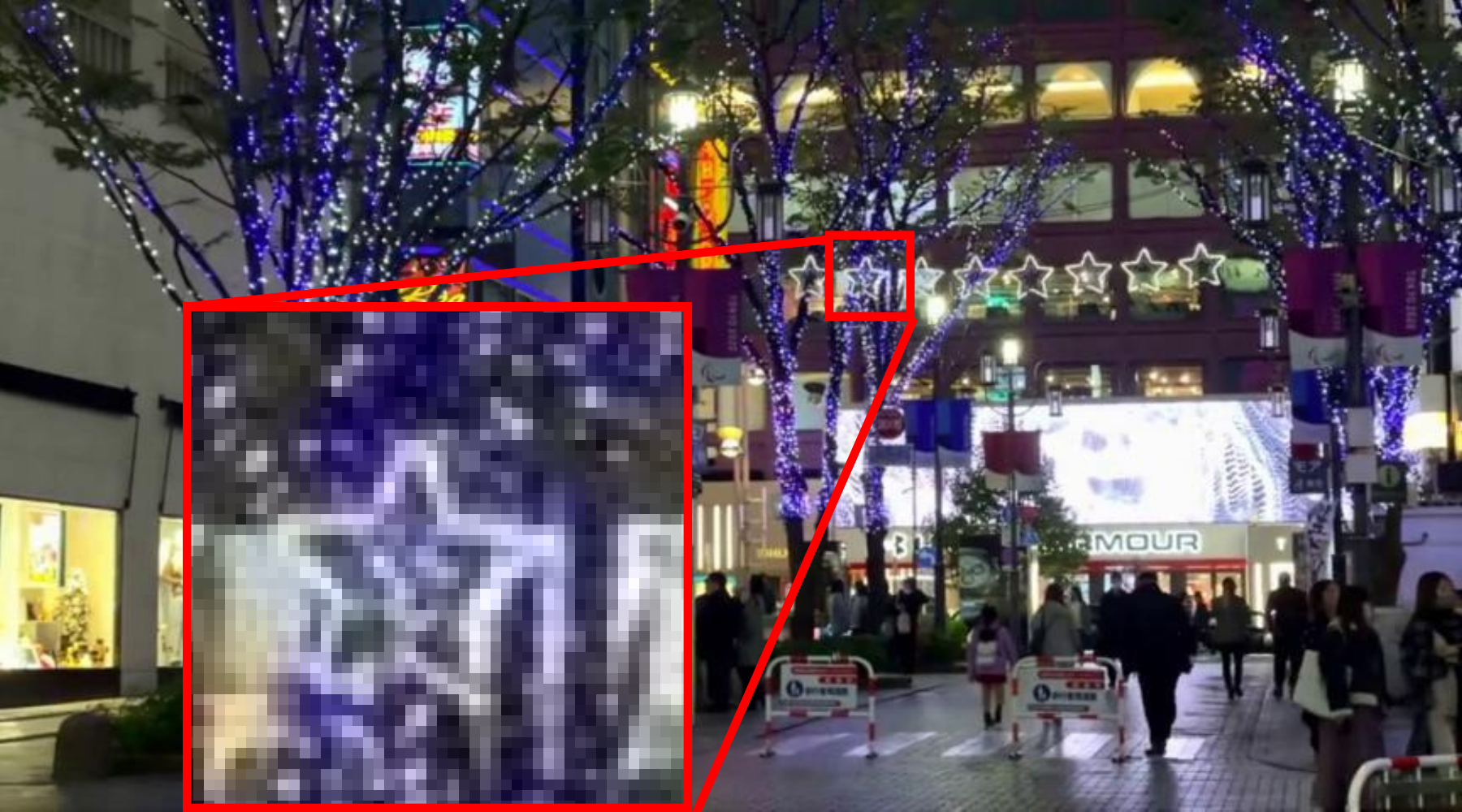} &
        \includegraphics[width=0.235\textwidth]{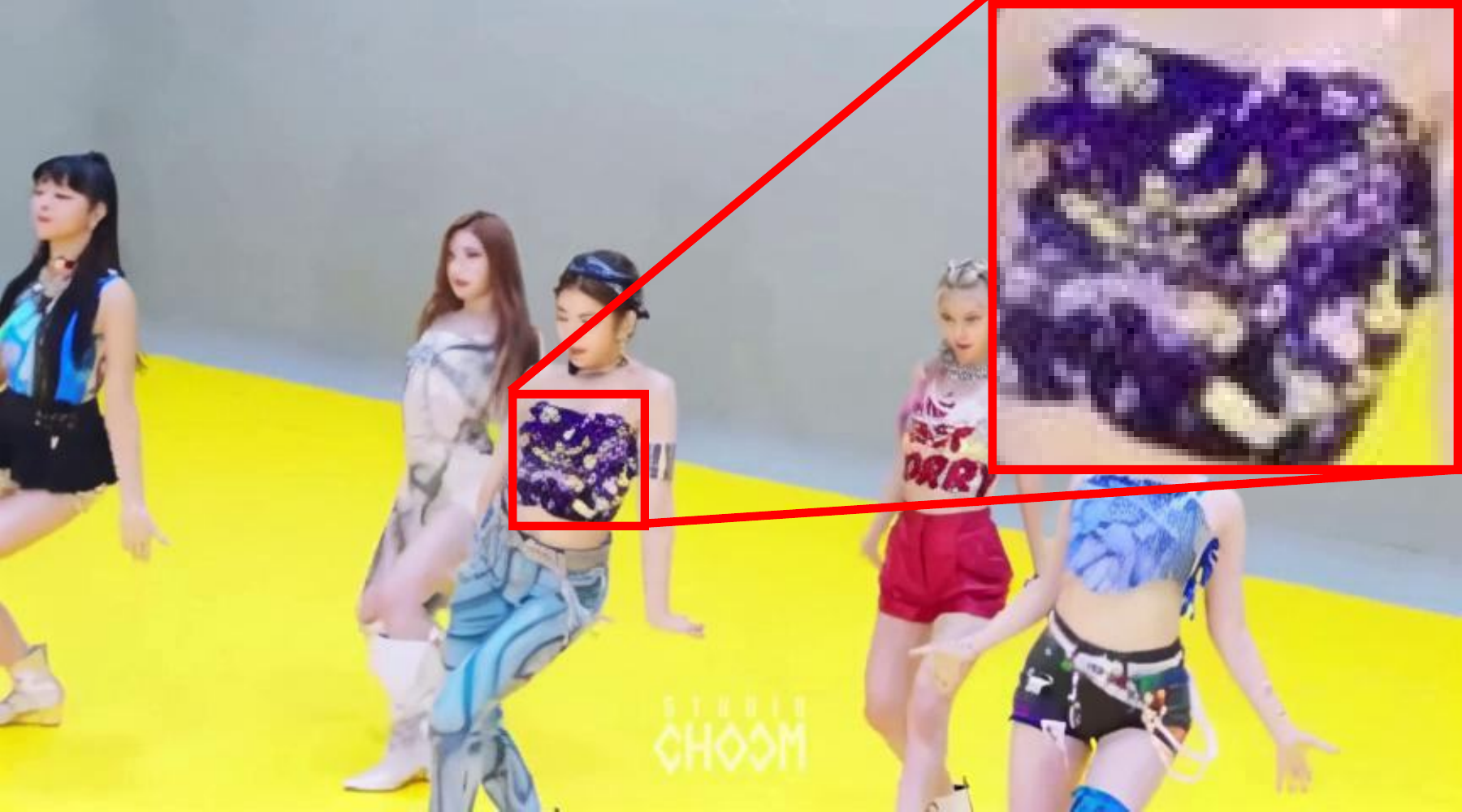} &
        \includegraphics[width=0.235\textwidth]{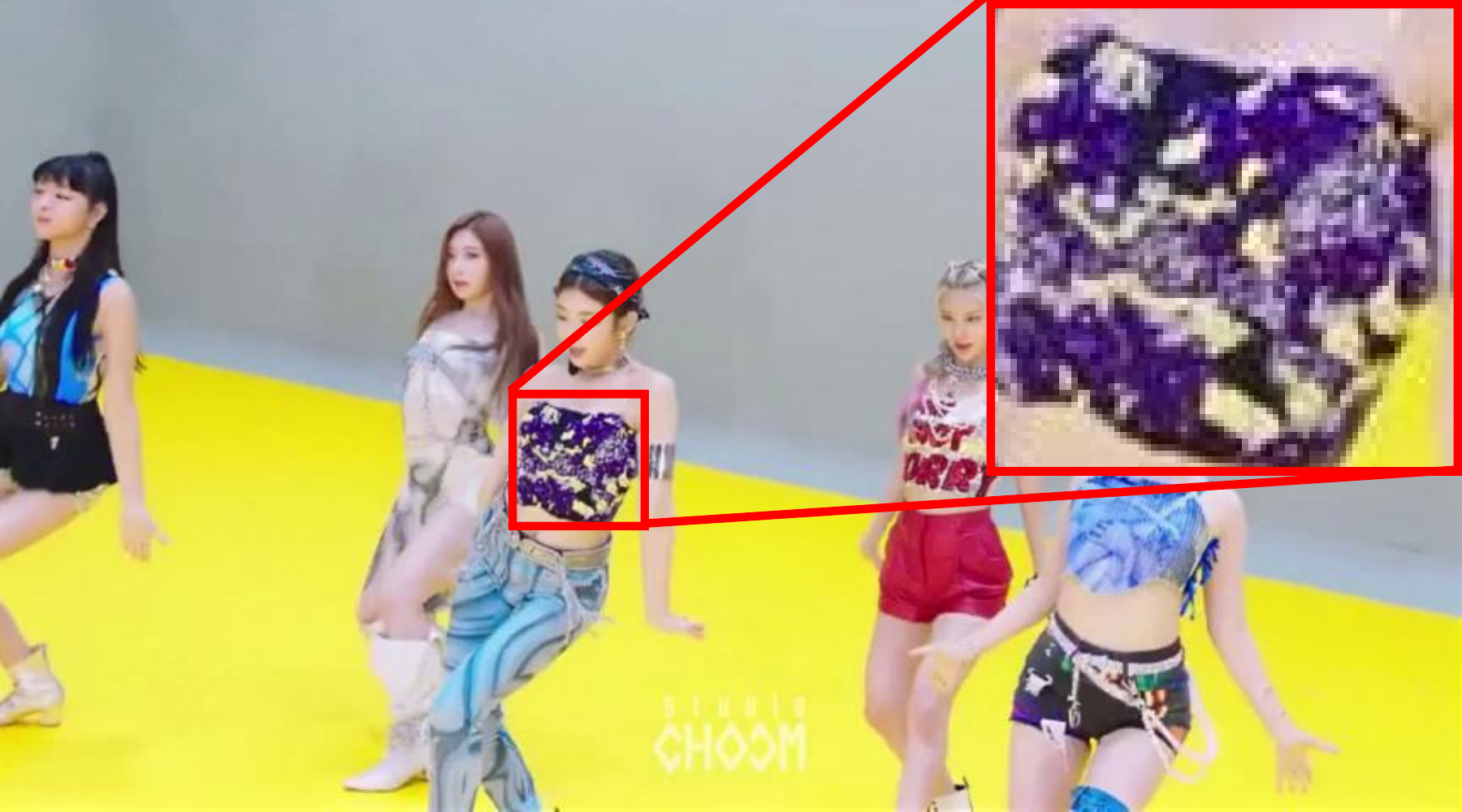} \\
        \rotatebox{90}{\small\hspace{4pt}VideoVAE+} &
        \includegraphics[width=0.235\textwidth]{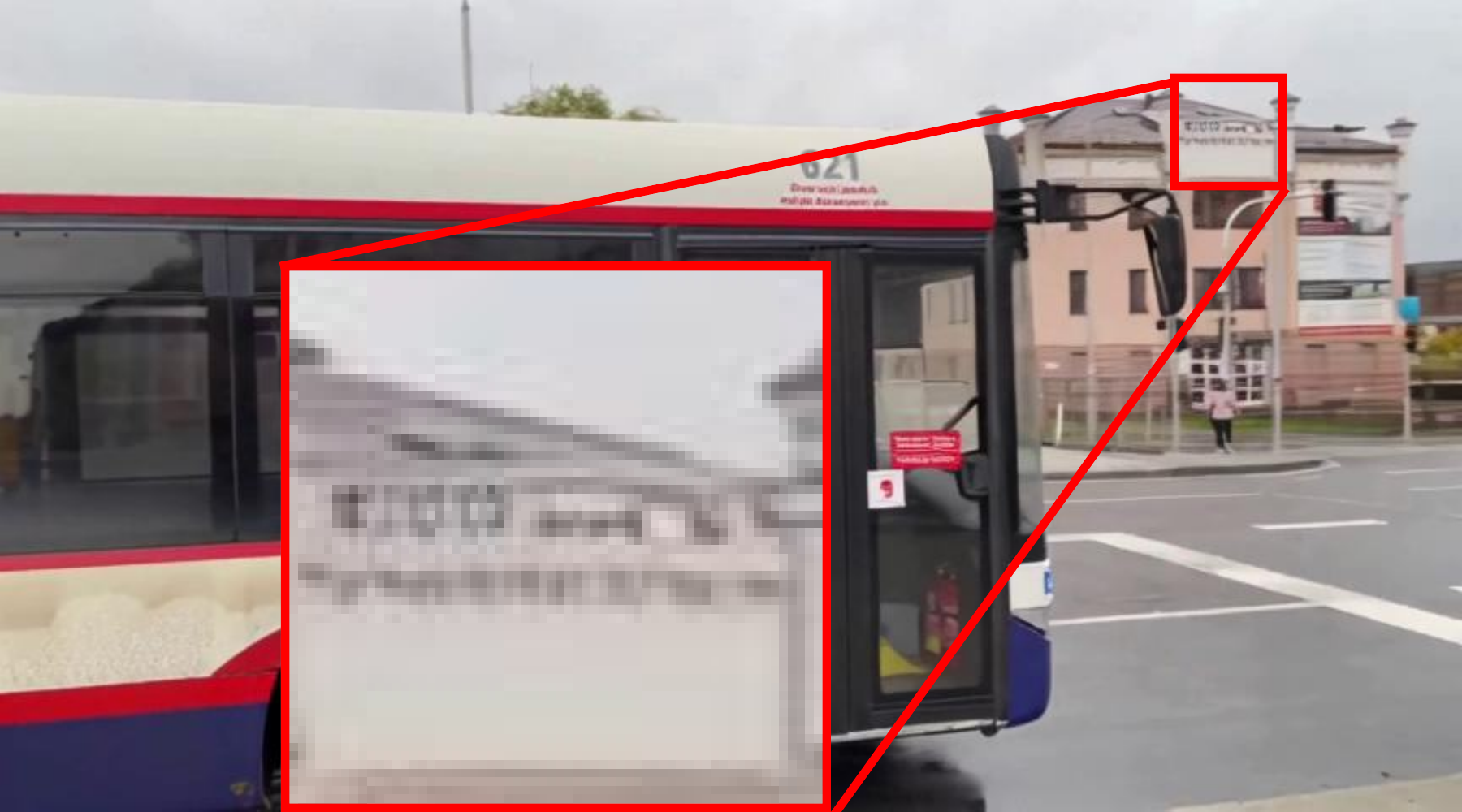} &
        \includegraphics[width=0.235\textwidth]{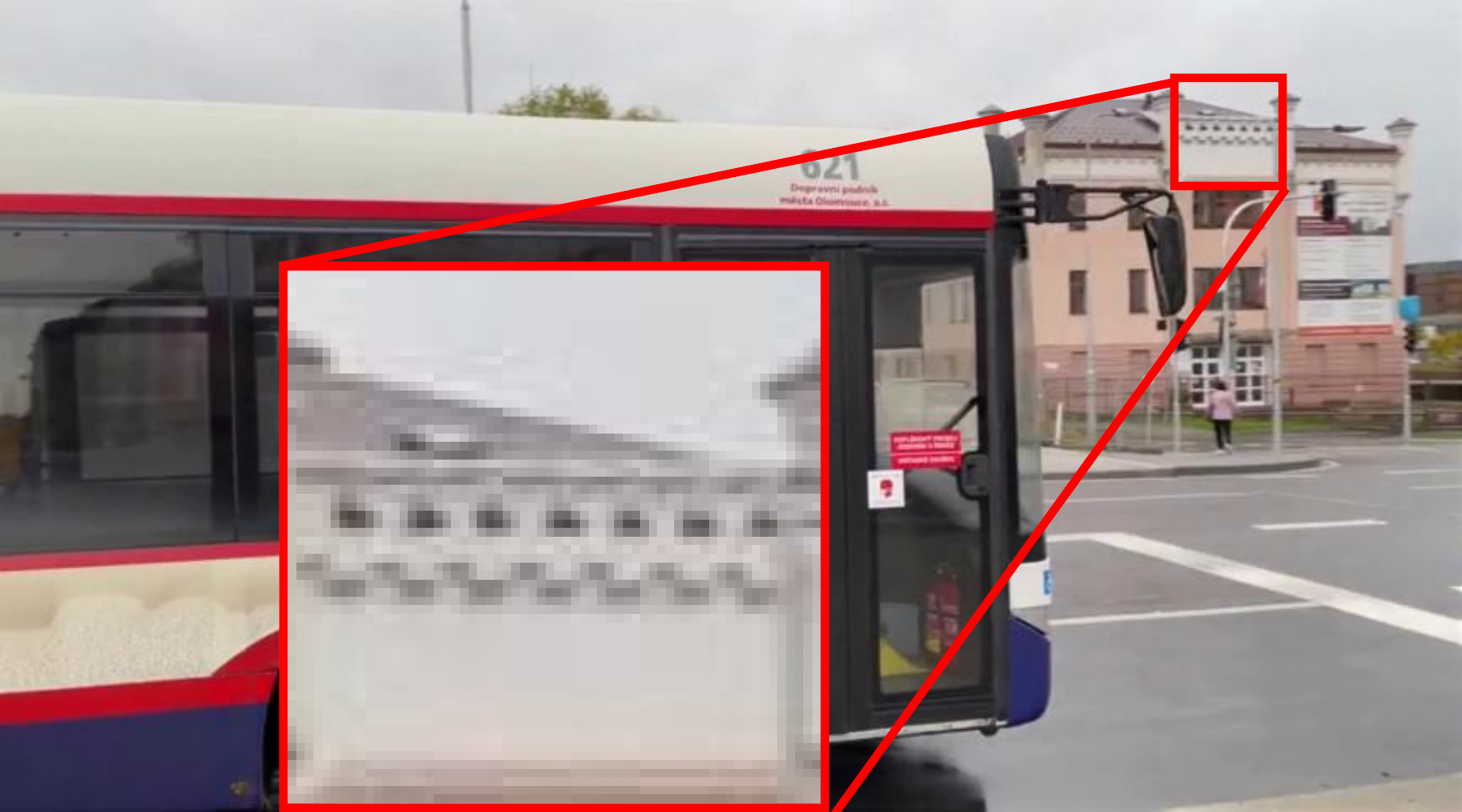} &
        \includegraphics[width=0.235\textwidth]{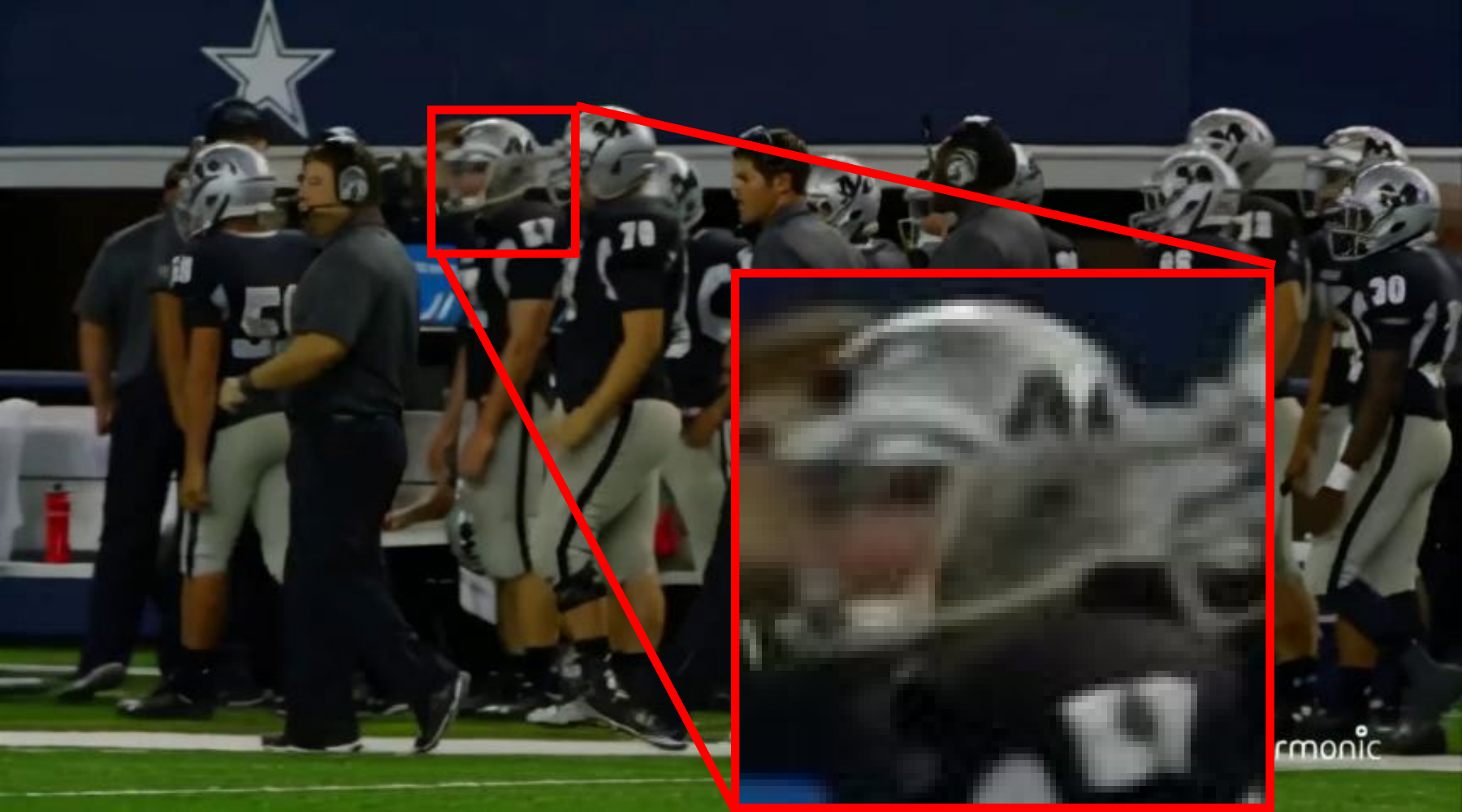} &
        \includegraphics[width=0.235\textwidth]{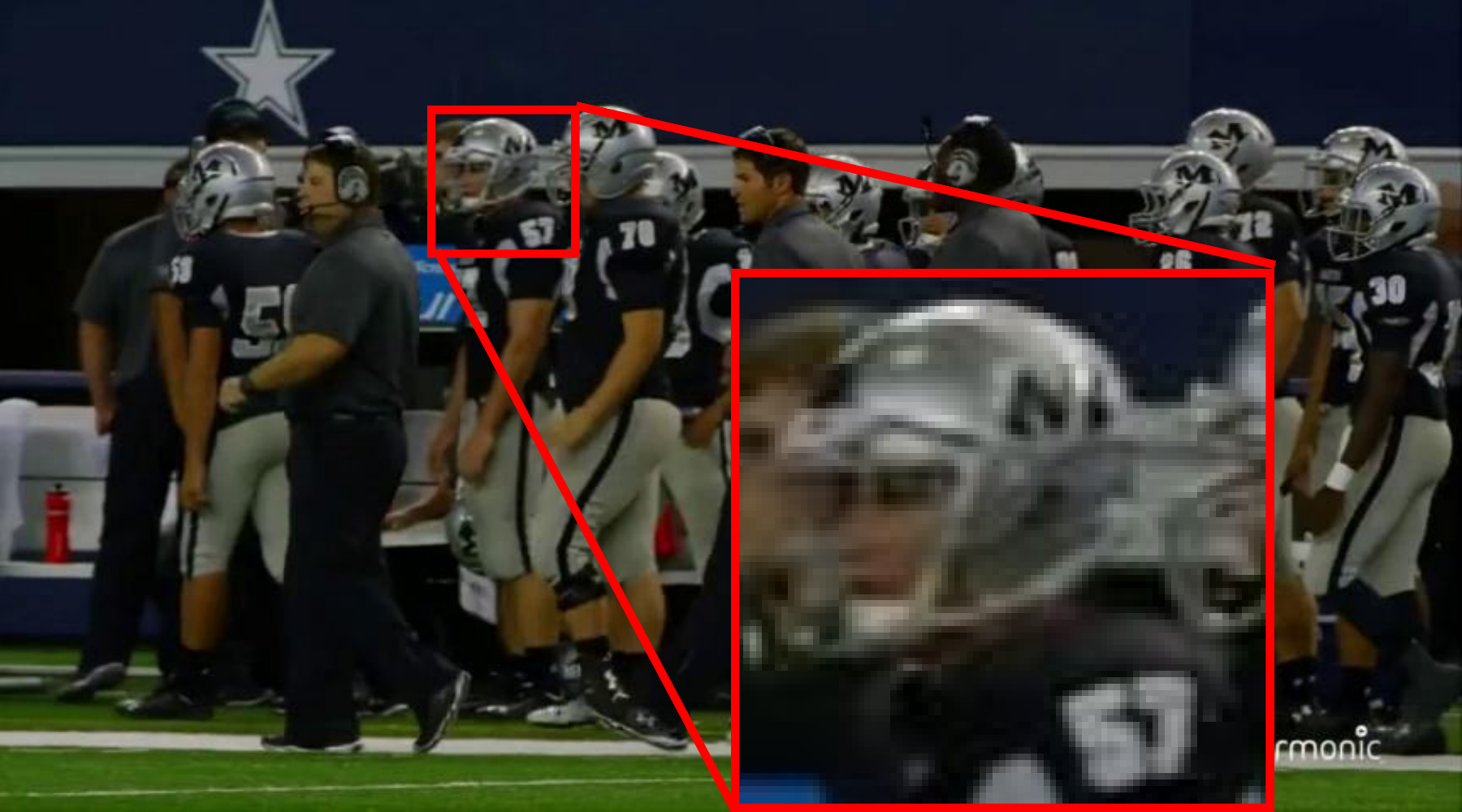} \\
        \rotatebox{90}{\small\hspace{4pt}VideoVAE+} &
        \includegraphics[width=0.235\textwidth]{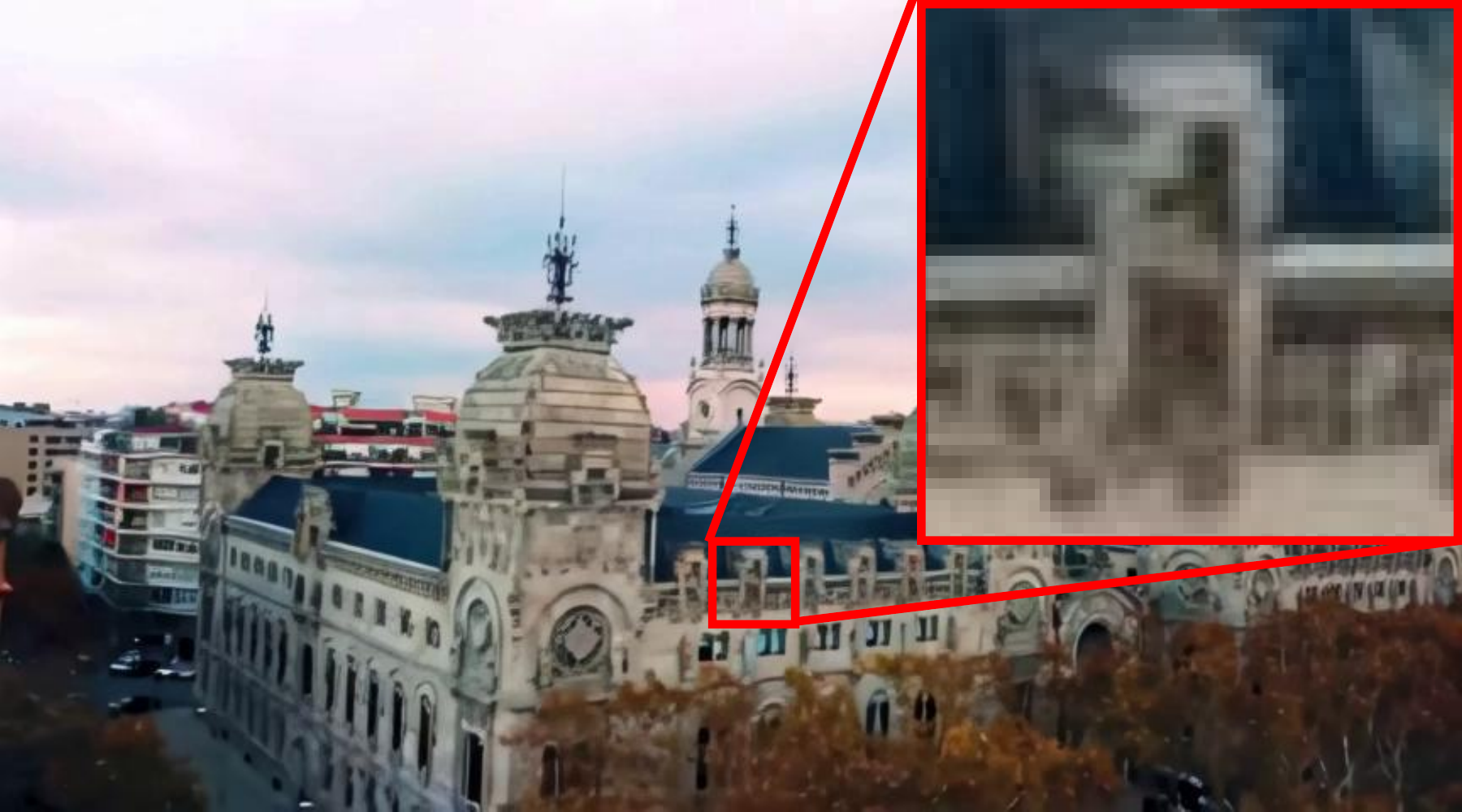} &
        \includegraphics[width=0.235\textwidth]{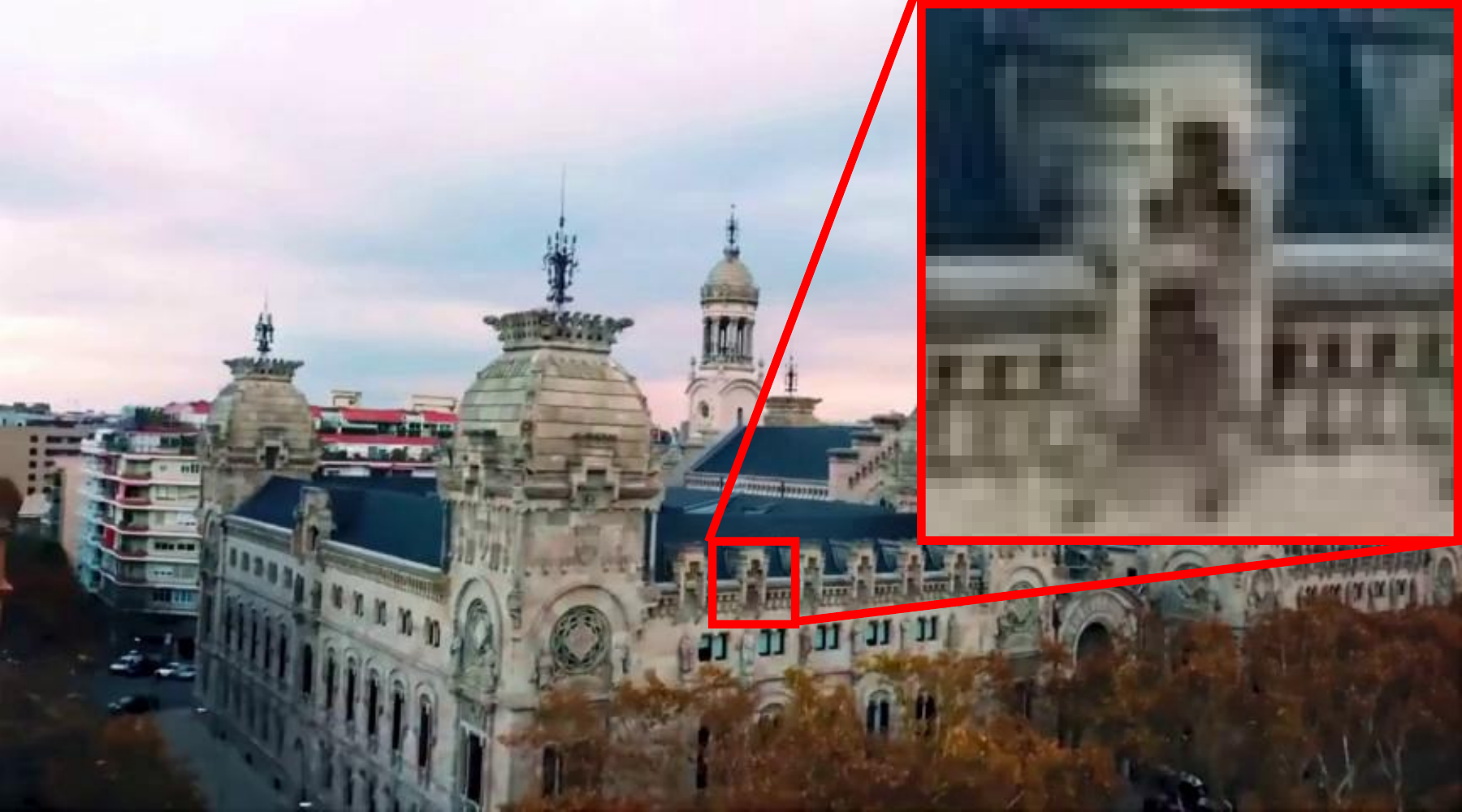} &
        \includegraphics[width=0.235\textwidth]{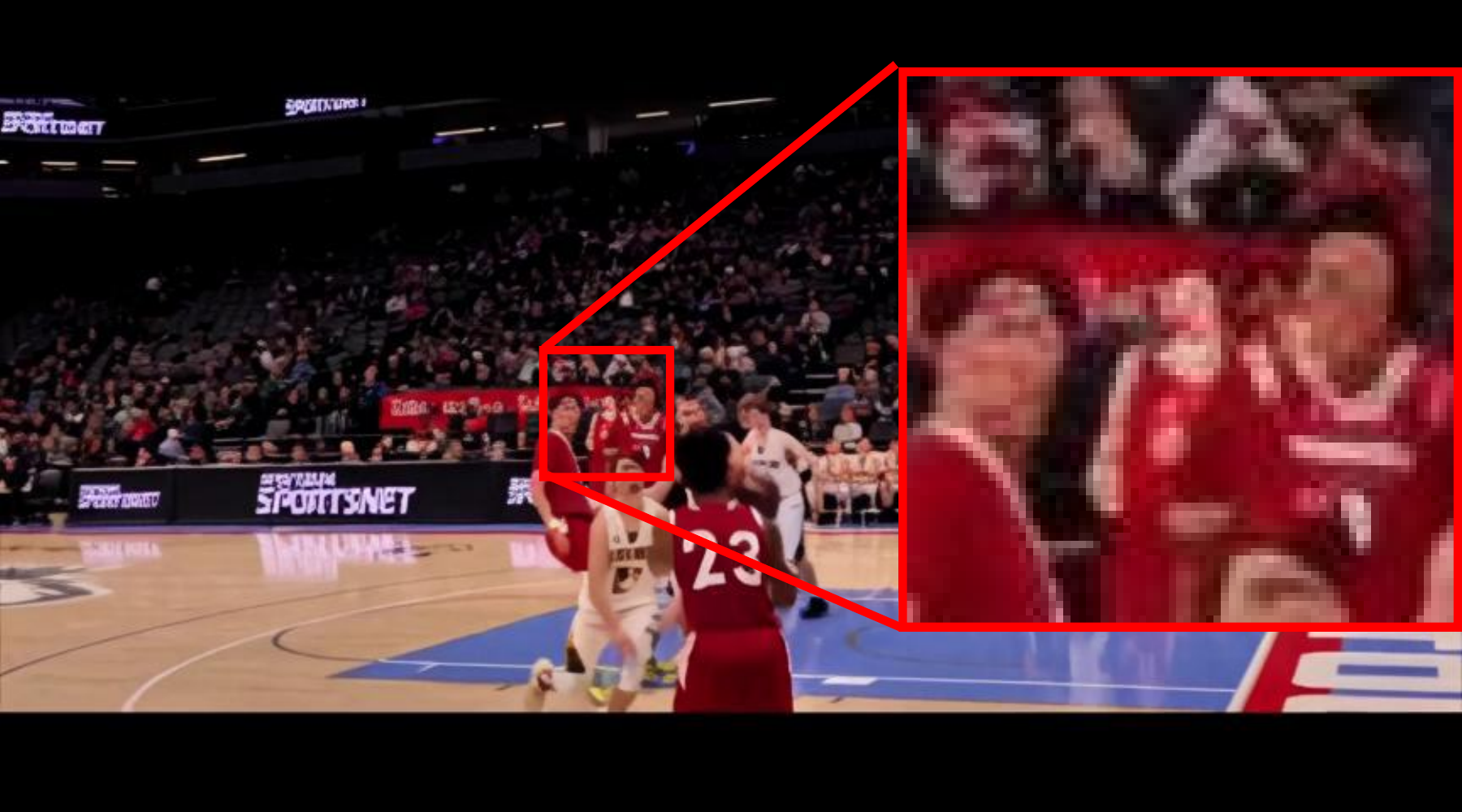} &
        \includegraphics[width=0.235\textwidth]{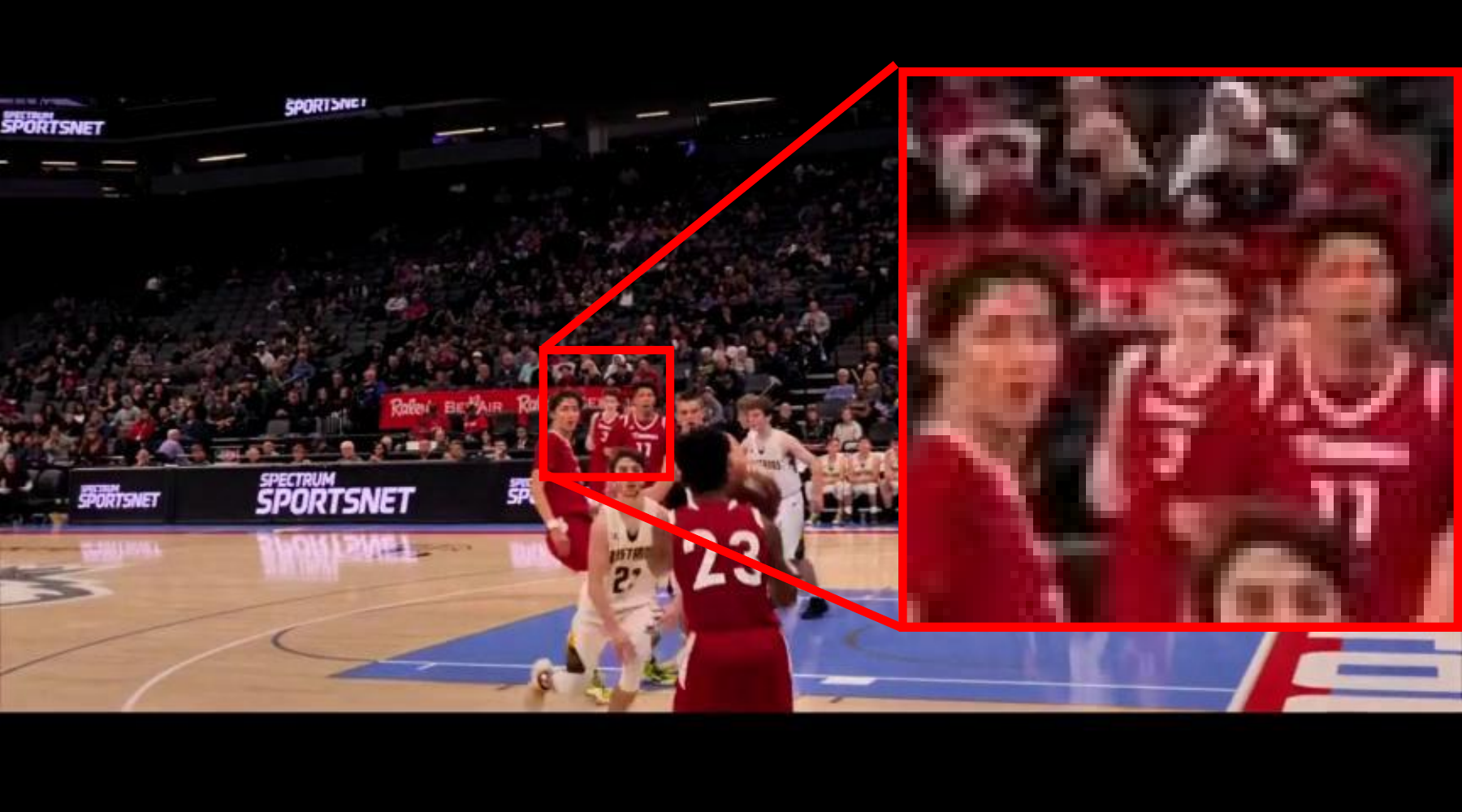} \\[2pt]
        & \small{Baseline} & \small{Ours} & \small{Baseline} & \small{Ours} \\
    \end{tabular}
    \caption{\textbf{Reconstruction comparisons.} Cropped regions from two backbones (Wan~2.1 and VideoVAE\textbf{+}) are shown. \model produces sharper reconstructions and better preserves fine details than the baseline, particularly for high-frequency content like text, human faces, and structural patterns.}
    \label{fig:qual_recon}
    \vspace{-1em}
\end{figure}

\subsection{Reconstruction Results}
\label{sec:recon_results}
Table~\ref{tab:recon_metrics} reports reconstruction metrics on Inter4K, WebVid, and the Large Motion subset from \cite{xing2025videovae+}. \model consistently outperforms existing baselines, with gains of +1.2\,dB PSNR on Inter4K (Wan~2.1) and +2.1\,dB PSNR on WebVid (VideoVAE\textbf{+}), and reduced LPIPS across the majority of the settings, demonstrating fine detail recovery and architectural generalization. A per-category breakdown shows that gains are positive across every category and are largest for content-rich scenes (Appendix~\ref{sec:supp_per_category}), a setting that many video generation pipelines struggle with~\citep{wan2025,kong2025hunyuanvideosystematicframeworklarge,yang2024cogvideox}.

We additionally compare against three prior reference-conditioned VAEs: H3AE~\cite{wu2025h3ae}, Reducio-VAE~\cite{tian2025reducio}, and RefTok~\cite{fan2025reftok}. Quantitative numbers for Reducio-VAE on our benchmarks are included in Table~\ref{tab:recon_metrics}. Since H3AE and RefTok have not released code or weights, we follow each method's reported evaluation protocol and run \model (VideoVAE\textbf{+} backbone) on the same dataset---DAVIS~\cite{Perazzi2016, Pont-Tuset_arXiv_2017} for H3AE and BAIR~\cite{ebert2017selfsupervisedvisualplanningtemporal} for RefTok. As shown in Table~\ref{tab:concurrent_comparison}, \model outperforms both under their own evaluation settings: it matches H3AE in PSNR while improving SSIM by $1.5$ on DAVIS, and surpasses RefTok by $7.4$\,dB PSNR and $4.1$ SSIM on BAIR, with LPIPS reduced from $0.013$ to $0.005$.

\noindent\textbf{Long and high-motion videos.}
On 81-frame reconstruction (Inter4K-81), \model{}$^\dagger$ improves every metric over the Wan~2.1 decoder ($+0.4$\,dB PSNR, $+1.3$ SSIM, $-0.012$ LPIPS; Table~\ref{tab:long_motion}). To test whether the gains hold under strong motion, we rank the 500 videos by a per-video motion score computed with RAFT~\cite{teed2020raftrecurrentallpairsfield} optical flow and keep the top quartile, yielding 125 videos with a mean (median) motion of 16.28 (14.08)\,px/frame. On this subset the gains grow to $+0.8$\,dB PSNR, $+2.0$ SSIM, and $-0.023$ LPIPS. From the full set to the high-motion subset, Wan~2.1 drops by $0.8$\,dB PSNR and $1.9$ SSIM and its LPIPS rises by $0.017$, whereas \model{}$^\dagger$ drops by only $0.4$\,dB PSNR and $1.2$ SSIM, with LPIPS rising by $0.006$. \model is therefore less sensitive to strong motion and is not limited to static or low-motion videos.

We further examine robustness to temporal distance and reference
mismatch. At a distance of $10$ frames from the reference,
\model retains a PSNR gain of approximately $0.4$\,dB.
On a scene-boundary stress test, it improves PSNR and SSIM,
although LPIPS worsens, revealing a trade-off under reference
mismatch. Full results are provided in
Appendix~\ref{sec:supp_reference_mismatch}.

\begin{table}[t]
    \centering
    \caption{\textbf{Video reconstruction results on WebVid, Inter4K (test), and Large Motion datasets.} Best results are shown in bold. Our method consistently outperforms the Wan~2.1 baseline and VideoVAE\textbf{+} across most metrics.}
    \label{tab:recon_metrics}
    \resizebox{\textwidth}{!}{%
        \setlength\tabcolsep{4pt}
        \renewcommand{\arraystretch}{1.2}
        \begin{tabular}{lccccccccc}
            \toprule
            & \multicolumn{3}{c}{\textbf{WebVid}} & \multicolumn{3}{c}{\textbf{Inter4K (test)}} & \multicolumn{3}{c}{\textbf{Large Motion}} \\
            \cmidrule(lr){2-4} \cmidrule(lr){5-7} \cmidrule(lr){8-10}
            \textbf{Model} (Backbone) & \textbf{PSNR$\uparrow$} & \textbf{SSIM$\uparrow$} & \textbf{LPIPS$\downarrow$} & \textbf{PSNR$\uparrow$} & \textbf{SSIM$\uparrow$} & \textbf{LPIPS$\downarrow$} & \textbf{PSNR$\uparrow$} & \textbf{SSIM$\uparrow$} & \textbf{LPIPS$\downarrow$} \\
            \midrule
            Cosmos-Tokenizer~\cite{cosmos_token}           & 31.2 & 89.6 & 0.107 & 31.3 & 88.6 & 0.103 & 30.2 & 86.8 & 0.119 \\
            CogVideoX-VAE~\cite{yang2024cogvideox}              & 32.5 & 92.3 & 0.053 & 32.9 & 92.1 & 0.050 & 31.1 & 89.8 & 0.069 \\
            EasyAnimate-VAE~\cite{xu2024easyanimatehighperformancelongvideo}            & 31.5 & 90.5 & 0.057 & 32.1 & 90.9 & 0.041 & 30.5 & 88.5 & 0.060 \\
            CV-VAE~\cite{zhao2024cv}                     & 31.6 & 90.6 & 0.064 & 32.3 & 90.8 & 0.055 & 30.7 & 88.7 & 0.073 \\
            Open-Sora-Plan (WF-VAE-L)~\cite{li2025wfvaeenhancingvideovae}  & 32.5 & 91.5 & 0.051 & 33.1 & 92.1 & 0.040 & 31.2 & 89.6 & 0.060 \\
            IV-VAE~\cite{wu2024improvedvideovaelatent}                     & 33.2 & 92.7 & 0.041 & 33.2 & 92.3 & 0.038 & 31.9 & 90.9 & 0.049 \\
            VidTok~\cite{tang2024vidtokversatileopensourcevideo}                     & 33.4 & 93.1 & 0.043 & 34.4 & 93.6 & 0.034 & 31.9 & 90.8 & 0.053 \\
            HunyuanVAE~\cite{kong2025hunyuanvideosystematicframeworklarge}              & 33.2 & 92.7 & 0.037 & 33.4 & 92.7 & 0.032 & 32.1 & 91.2 & 0.045 \\
            Reducio-VAE~\cite{tian2025reducio} (T/4)                                   & 32.4 & 88.8 & 0.057 & 29.7 & 86.3 & 0.096 & 26.8 & 78.6 & 0.126 \\
            \midrule
            Wan~2.1 VAE~\cite{wan2025}                  & 32.8 & 91.4 & 0.032 & 33.7 & 93.0 & 0.034 & 30.8 & 89.1 & 0.048 \\
            \textbf{\model} (Wan~2.1)            & 33.5 & 92.6 & 0.037 & \textbf{34.9} & \textbf{94.9} & \textbf{0.031} & 31.4 & 90.4 & 0.053 \\[-2pt]
            & {\scriptsize\color{teal}(+0.7)} & {\scriptsize\color{teal}(+1.2)} & {\scriptsize\color{lightgray}(+.005)} & {\scriptsize\color{teal}(+1.2)} & {\scriptsize\color{teal}(+1.9)} & {\scriptsize\color{teal}(-.003)} & {\scriptsize\color{teal}(+0.6)} & {\scriptsize\color{teal}(+1.3)} & {\scriptsize\color{lightgray}(+.005)} \\
            \midrule
            VideoVAE\textbf{+}~\cite{xing2025videovae+}         & 32.9 & 93.0 & 0.041 & 33.9 & 93.3 & 0.034 & 31.9 & 90.7 & 0.050 \\
            \textbf{\model} (VideoVAE\textbf{+}) & \textbf{35.1} & \textbf{93.5} & \textbf{0.031} & 34.7 & 94.7 & 0.033 & \textbf{32.9} & \textbf{92.3} & \textbf{0.043} \\[-2pt]
            & {\scriptsize\color{teal}(+2.1)} & {\scriptsize\color{teal}(+0.5)} & {\scriptsize\color{teal}(-.010)} & {\scriptsize\color{teal}(+0.8)} & {\scriptsize\color{teal}(+1.5)} & {\scriptsize\color{teal}(-.001)} & {\scriptsize\color{teal}(+1.1)} & {\scriptsize\color{teal}(+1.6)} & {\scriptsize\color{teal}(-.007)} \\
            \bottomrule
        \end{tabular}
    }
    \vspace{-1em}
\end{table}

\begin{table}[t]
    \centering
    \caption{\textbf{Comparison with prior reference-conditioned VAEs} on each method's own benchmark: H3AE on DAVIS~\cite{Perazzi2016, Pont-Tuset_arXiv_2017} and RefTok on BAIR~\cite{ebert2017selfsupervisedvisualplanningtemporal}. \model uses the VideoVAE\textbf{+} backbone.}
    \label{tab:concurrent_comparison}
    \small
    \setlength\tabcolsep{5pt}
    \renewcommand{\arraystretch}{1.15}
    \begin{tabular}{lcc|lccc}
        \toprule
        \multicolumn{3}{c|}{\textbf{DAVIS}} & \multicolumn{4}{c}{\textbf{BAIR}} \\
        \cmidrule(lr){1-3} \cmidrule(lr){4-7}
        \textbf{Model} & PSNR$\uparrow$ & SSIM$\uparrow$ & \textbf{Model} & PSNR$\uparrow$ & SSIM$\uparrow$ & LPIPS$\downarrow$ \\
        \midrule
        H3AE~\cite{wu2025h3ae} & 33.0 & 89.9 & RefTok~\cite{fan2025reftok} & 28.8 & 95.0 & 0.013 \\
        \textbf{\model} & \textbf{33.1} & \textbf{91.4} & \textbf{\model} & \textbf{36.2} & \textbf{99.1} & \textbf{0.005} \\
        \bottomrule
    \end{tabular}
    \vspace{-1em}
\end{table}

\begin{figure}[t]
    \centering
    \setlength{\tabcolsep}{1pt}
    \renewcommand{\arraystretch}{0.5}
    \begin{tabular}{cc@{\hspace{10pt}}cccc}
        \multirow{2}{*}[-2pt]{\rotatebox{90}{\normalsize\textbf{GT}}} &
        \multirow{2}{*}[15pt]{\includegraphics[width=0.23\textwidth]{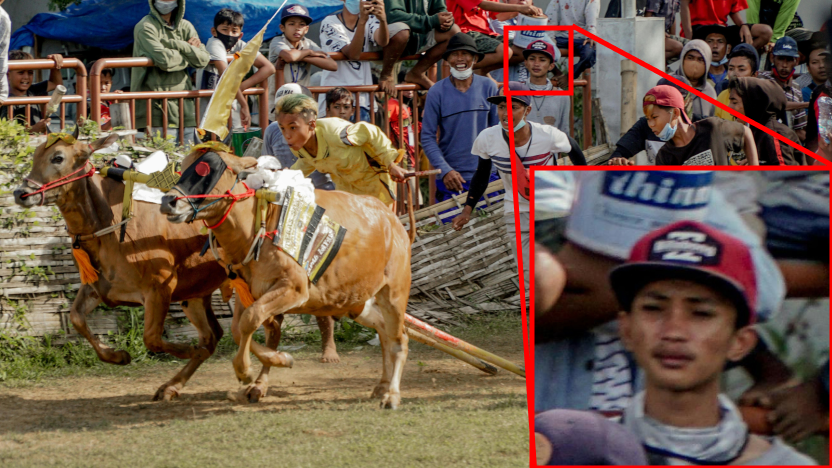}} &
        \rotatebox{90}{\small\hspace{15pt}Wan} &
        \includegraphics[width=0.23\textwidth]{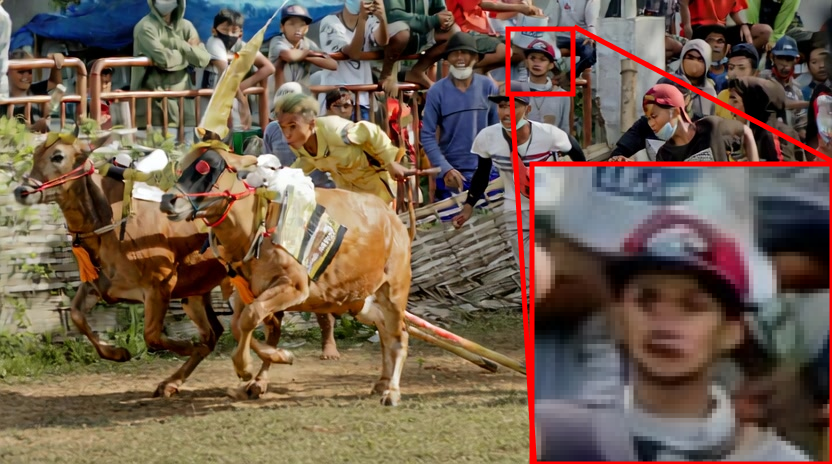} &
        \includegraphics[width=0.23\textwidth]{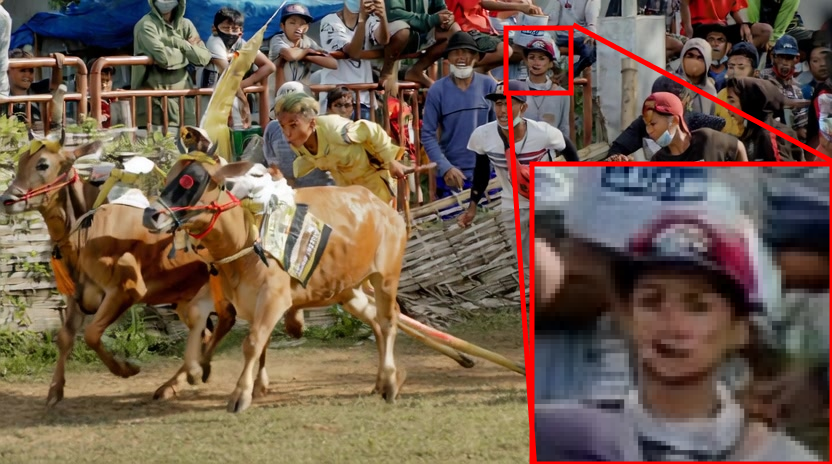} &
        \includegraphics[width=0.23\textwidth]{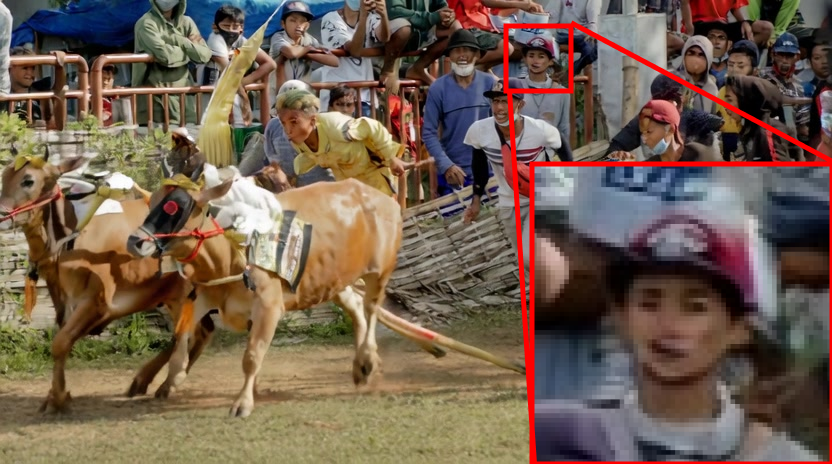} \\

        &
        &
        \rotatebox{90}{\small\hspace{8pt}Wan Ours} &
        \includegraphics[width=0.23\textwidth]{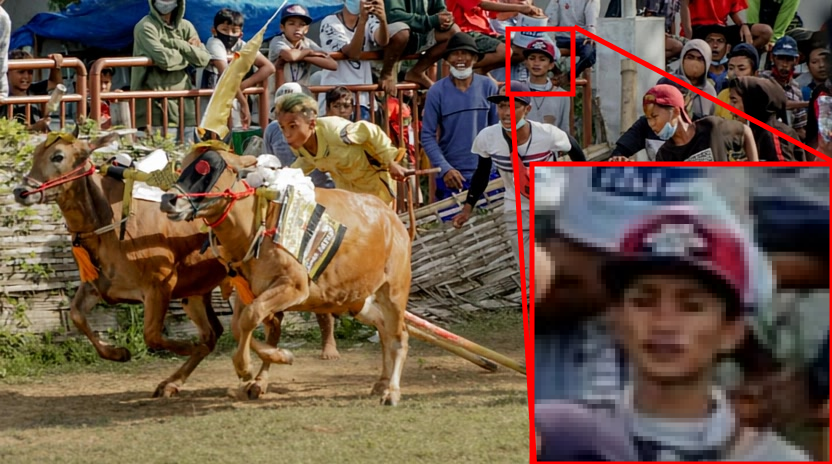} &
        \includegraphics[width=0.23\textwidth]{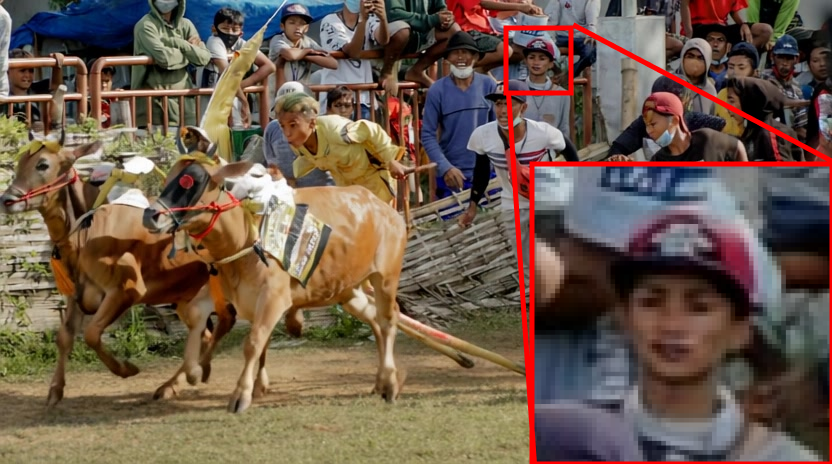} &
        \includegraphics[width=0.23\textwidth]{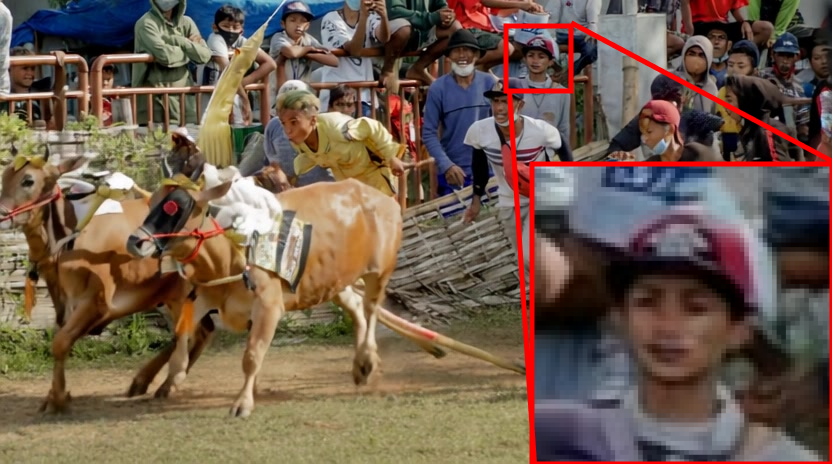} \\

        \multirow{2}{*}[-2pt]{\rotatebox{90}{\normalsize\textbf{GT}}} &
        \multirow{2}{*}[15pt]{\includegraphics[width=0.23\textwidth]{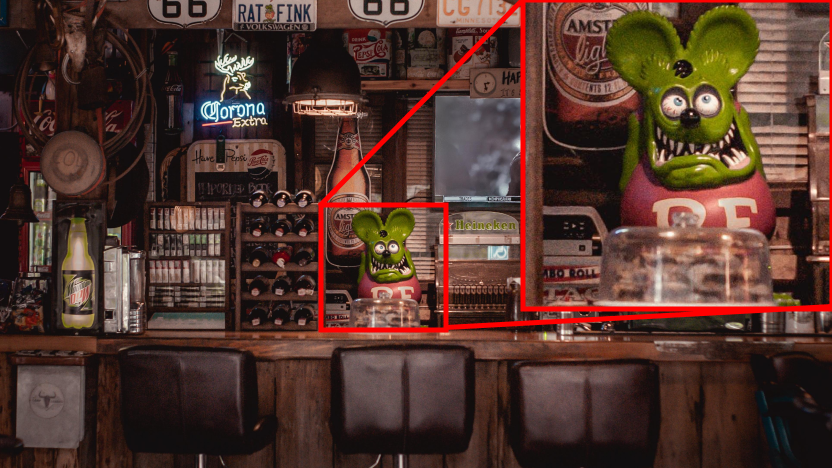}} &
        \rotatebox{90}{\small\hspace{15pt}Wan} &
        \includegraphics[width=0.23\textwidth]{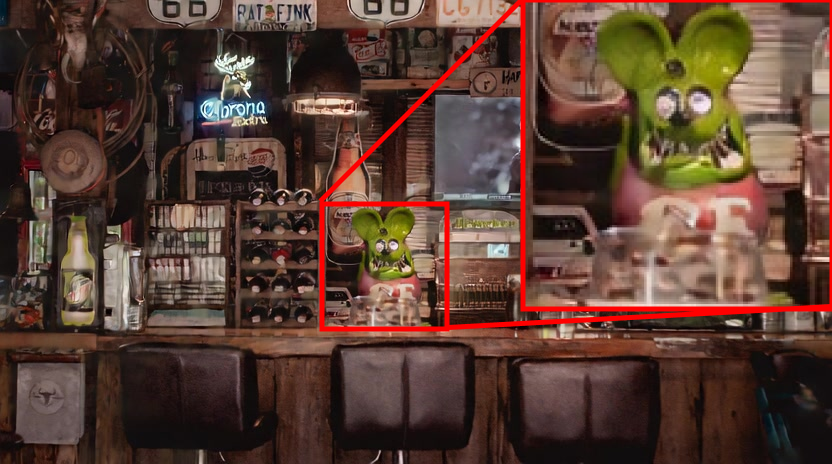} &
        \includegraphics[width=0.23\textwidth]{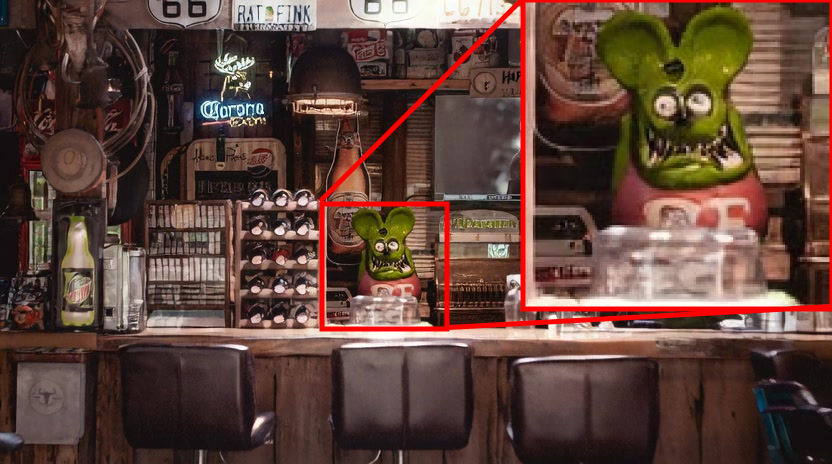} &
        \includegraphics[width=0.23\textwidth]{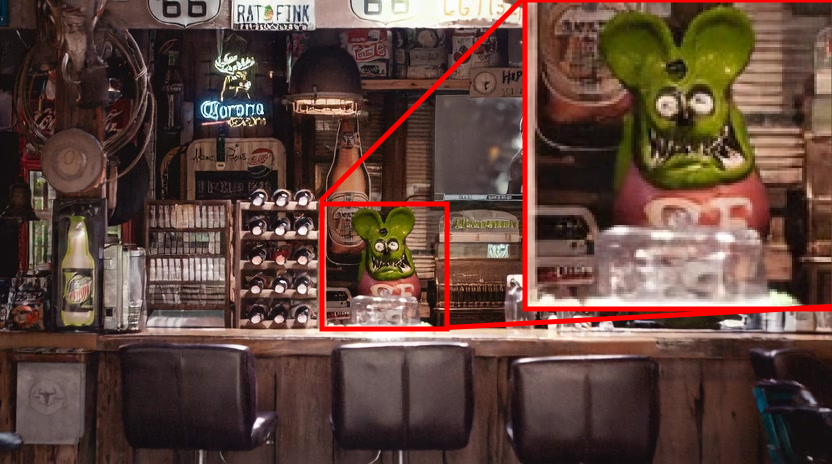} \\

        &
        &
        \rotatebox{90}{\small\hspace{8pt}Ours} &
        \includegraphics[width=0.23\textwidth]{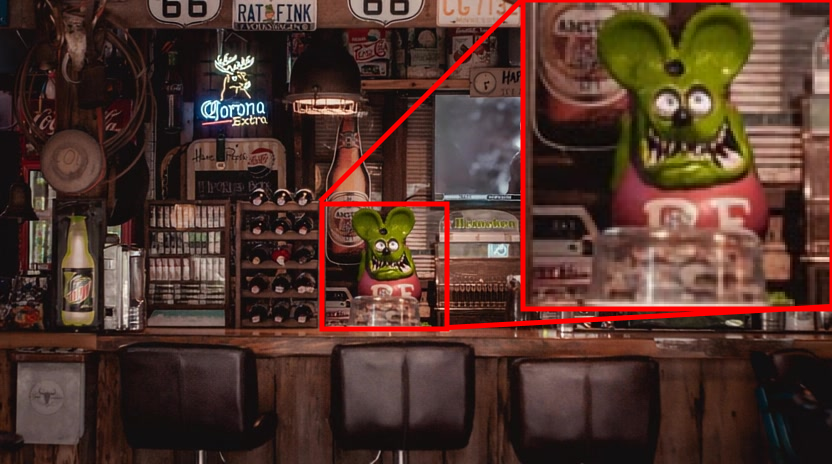} &
        \includegraphics[width=0.23\textwidth]{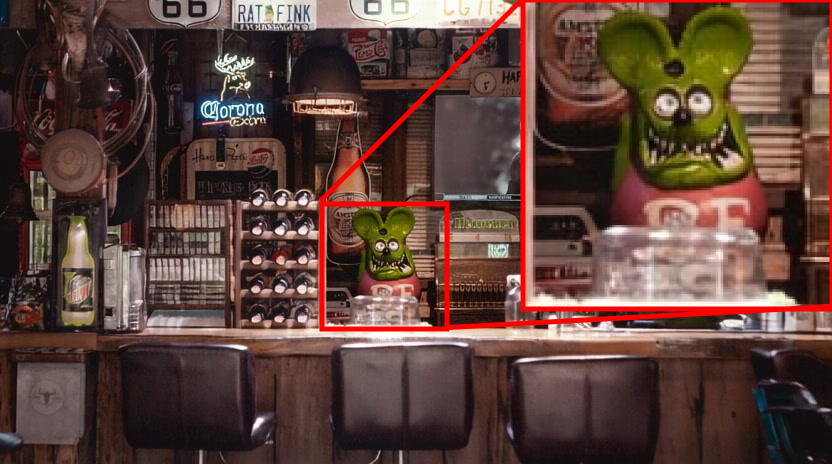} &
        \includegraphics[width=0.23\textwidth]{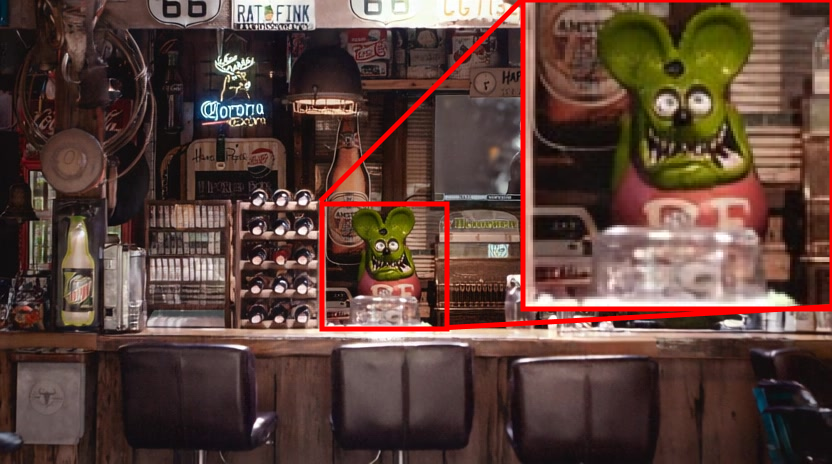} \\
    \end{tabular}
    \caption{\textbf{Video generation comparisons.} For the same input reference, three frames generated at different timesteps are shown. The baseline distorts the underlying scene structure, whereas \model\ preserves the structure while producing sharper and more consistent details across frames.}
    \label{fig:qual_gen}
    \vspace{-1.5em}
\end{figure}

\subsection{Generation results}
We plug \model into the unmodified Wan~2.1 pipeline as a drop-in decoder replacement, without any retraining of the diffusion model. To isolate the decoder's contribution, we adopt a \emph{fixed-seed} protocol: for each prompt we draw $5$ random seeds once, generate diffusion latents with the unmodified Wan~2.1 pipeline, and decode the same latents with both Wan~2.1 baseline decoder and \model, so that diffusion sampling variance does not affect the pairwise comparison and the improvement is attributable solely to the decoder. Evaluation details are in Appendix~\ref{sec:supp_vbench_protocol}.

As shown in Table~\ref{tab:video_metrics} (17 frames), \model outperforms the baseline on 10 out of 12 VBench dimensions and is on par in imaging quality, with the largest gains in subject and background consistency, and raises the total VBench score from 87.9 to 88.2 ($+0.3$).

\noindent\textbf{Long and high-motion videos.}
Repeating the fixed-seed protocol with 81 frames instead of 17, \model{}$^\dagger$ raises the total score from $87.9$ to $88.1$ and improves all three aggregate scores and 6 of the 9 dimensions, with the largest gains in imaging quality and subject consistency (Table~\ref{tab:video_metrics}, 81 frames). We further evaluate the highest-motion quartile of the original, unmodified VBench set, selected using a per-video mean RAFT optical-flow threshold of $14.883$. As shown in Table~\ref{tab:high_motion_vbench}, \model{}$^\dagger$ improves Quality, I2V, and Total by $0.17$, $0.35$, and $0.26$ percentage points, compared with $0.13$, $0.25$, and $0.19$ on the full set, so the generation improvements persist under substantial motion.

\noindent\textbf{Temporal stability.}
We also assess temporal stability under the same fixed-seed protocol with flow warping error ($E_\text{warp}$)~\cite{lai2018learning}, temporal LPIPS ($E_\text{tLPIPS}$)~\cite{zhang2018perceptual}, flicker ($E_\text{flicker}$)~\cite{bonneel2015blind}, and CLIP consistency (CLIP$_\text{cons}$)~\cite{esser2023structurecontentguidedvideosynthesis}. \model reduces all three degradation metrics by $8$--$15\%$ and slightly improves CLIP consistency, producing less flicker and more temporally coherent videos (Appendix~\ref{sec:supp_temporal_stability}).

\begin{table}[t]
    \centering
    \caption{\textbf{VBench generation eval results (12 dimensions) on the Wan~2.1 backbone at 17 and 81 frames.} Subj.~/~BG denote subject/background consistency; I2V~Subj./BG are their I2V-specific variants. \model{}$^\dagger$: causal variant (\Cref{sec:causal}). Best results per frame count in bold.}
    \label{tab:video_metrics}
    \small
    \setlength\tabcolsep{3pt}
    \renewcommand{\arraystretch}{1.15}
    \resizebox{\textwidth}{!}{%
    \begin{tabular}{lccccccccccccc}
        \toprule
        & & \multicolumn{6}{c}{\textbf{Quality Dimensions}} & \multicolumn{3}{c}{\textbf{I2V Dimensions}} & \multicolumn{3}{c}{\textbf{Aggregate Scores}} \\
        \cmidrule(lr){3-8} \cmidrule(lr){9-11} \cmidrule(lr){12-14}
        \textbf{Model} & \textbf{Frames}
        & Subj. & BG & Motion & Dyna. & Aesth. & Imag.
        & I2V~Subj. & I2V~BG & Camera
        & Quality & I2V & Total \\
        \midrule
        Wan~2.1~\cite{wan2025} & \multirow{2}{*}{17} & 96.6 & 97.9 & 98.1 & 47.5 & 65.6 & \textbf{70.1} & 98.0 & 99.1 & \textbf{19.3} & 81.3 & 94.5 & 87.9 \\
        \model                 &                     & \textbf{96.8} & \textbf{98.2} & \textbf{98.3} & \textbf{47.6} & \textbf{66.0} & \textbf{70.1} & \textbf{98.3} & \textbf{99.3} & 19.1 & \textbf{81.6} & \textbf{94.8} & \textbf{88.2} \\
        \midrule
        Wan~2.1~\cite{wan2025} & \multirow{2}{*}{81} & 94.9 & \textbf{97.7} & 98.2 & \textbf{46.6} & \textbf{65.1} & 70.2 & 98.0 & 99.3 & 28.9 & 80.8 & 95.0 & 87.9 \\
        \model{}$^\dagger$     &                     & \textbf{95.3} & 97.5 & \textbf{98.5} & 46.3 & 64.4 & \textbf{70.9} & \textbf{98.2} & \textbf{99.4} & \textbf{29.1} & \textbf{80.9} & \textbf{95.3} & \textbf{88.1} \\
        \bottomrule
    \end{tabular}}
    \vspace{0.6em}
    \caption{\textbf{81-frame generation on high-motion VBench} (highest-motion quartile of the original VBench set by RAFT optical flow; Wan~2.1 backbone). \model{}$^\dagger$: causal variant (\Cref{sec:causal}).}
    \label{tab:high_motion_vbench}
    \setlength\tabcolsep{8pt}
    \begin{tabular}{lccc}
        \toprule
        \textbf{Model} & Quality$\uparrow$ & I2V$\uparrow$ & Total$\uparrow$ \\
        \midrule
        Wan~2.1~\cite{wan2025} & 81.93 & 93.22 & 87.57 \\
        \model{}$^\dagger$ (Wan~2.1) & \textbf{82.10} & \textbf{93.56} & \textbf{87.83} \\[-2pt]
        & {\scriptsize\color{teal}($+0.17$)} & {\scriptsize\color{teal}($+0.35$)} & {\scriptsize\color{teal}($+0.26$)} \\
        \bottomrule
    \end{tabular}
    \vspace{0.6em}
    \caption{\textbf{81-frame reconstruction on Inter4K-81 and its high-motion subset} (top quartile by RAFT motion score; Wan~2.1 backbone). \model{}$^\dagger$: causal variant (\Cref{sec:causal}).}
    \label{tab:long_motion}
    \setlength\tabcolsep{5pt}
    \begin{tabular}{lcccccc}
        \toprule
        & \multicolumn{3}{c}{\textbf{All (500 videos)}} & \multicolumn{3}{c}{\textbf{High motion (125 videos)}} \\
        \cmidrule(lr){2-4} \cmidrule(lr){5-7}
        \textbf{Model} & PSNR$\uparrow$ & SSIM$\uparrow$ & LPIPS$\downarrow$ & PSNR$\uparrow$ & SSIM$\uparrow$ & LPIPS$\downarrow$ \\
        \midrule
        Wan~2.1~\cite{wan2025} & 31.0 & 90.0 & 0.042 & 30.2 & 88.1 & 0.059 \\
        \model{}$^\dagger$ (Wan~2.1) & \textbf{31.4} & \textbf{91.3} & \textbf{0.030} & \textbf{31.0} & \textbf{90.1} & \textbf{0.036} \\[-2pt]
        & {\scriptsize\color{teal}($+0.4$)} & {\scriptsize\color{teal}($+1.3$)} & {\scriptsize\color{teal}($-0.012$)} & {\scriptsize\color{teal}($+0.8$)} & {\scriptsize\color{teal}($+2.0$)} & {\scriptsize\color{teal}($-0.023$)} \\
        \bottomrule
    \end{tabular}
    \vspace{-1.2em}
\end{table}

\subsection{Human evaluation}
We conduct a blind preference study comparing \model with four VAE baselines: Wan~2.1, HunyuanVAE, VideoVAE\textbf{+}, and Reducio-VAE. Participants use a slider-overlay interface in which the two videos are shown on randomized sides and model identities are hidden. Overall, \model is preferred over every baseline. In reconstruction tasks, \model is preferred over Wan~2.1, HunyuanVAE, VideoVAE\textbf{+}, and Reducio-VAE, with preference rates of $92.3\%$, $69.2\%$, $63.4\%$, and $89.6\%$, respectively. In generation tasks, \model is preferred over Wan~2.1 with a preference rate of $82.4\%$. Appendix~\ref{sec:supp_human_eval} provides further detail.

\subsection{Qualitative results}
Figure~\ref{fig:qual_recon} shows reconstruction results. Our decoder recovers sharper edges and finer textures than the baseline, particularly in high-frequency regions such as text, human faces, and fine structural details. These improvements are consistent across backbones. Figure~\ref{fig:qual_gen} compares generation results produced using the baseline decoder vs.\ \model. As we can see, \model better preserves the structure and produces sharper, more consistent details across frames.

\subsection{Efficiency}
\model adds overhead only to the lightweight VAE decoding stage, whereas the DiT denoising stage dominates the diffusion pipeline. When run on Wan~2.1, the total pipeline latency is nearly equal with minimal overhead ($72{,}037 \pm 16$ ms vs.\ $74{,}584 \pm 40$ ms).

\subsection{Ablation study}
\label{sec:ablation}

\noindent \textbf{Effect of the number of blocks.}
We ablate the number of Transformer blocks on Wan~2.1 with 3, 5, 7, and 10. Reconstruction quality improves with depth: the 10-blocks model achieves $34.9$\,dB PSNR on Inter4K, up from $34.3$ for the 3-blocks variant, and is best across all benchmarks (Appendix~\ref{sec:supp_layer_ablation}).

\noindent \textbf{Latent token dropout and two-stage curriculum.}
We further ablate two training-time choices: latent token dropout and the two-stage curriculum. We find that higher latent-token dropout consistently improves reconstruction. Gains appear on both the reference and non-reference frames, indicating that stronger dropout encourages better use of reference information. The two-stage curriculum substantially improves over one-stage training (e.g., overall PSNR $30.5 \rightarrow 34.3$\,dB), helping the model adapt from short to longer-video decoding. Full training details and results are in Appendix~\ref{sec:supp_dropout_curriculum}.

\noindent \textbf{Alternative reference injection strategy.}
We also compare against a ControlNet-style~\cite{zhang2023adding} alternative that adds residual features from a parallel reference encoder to the decoder at each stage. This approach lacks temporal reasoning and converges to lower quality than our attention-based design. Details and qualitative results are in Appendix~\ref{sec:supp_controlnet}.

\noindent\textbf{Reference injection stages.}
We evaluate stage-specific reference injection by retaining conditioning at only one decoder stage and fine-tuning each variant under the same settings. Injecting reference information at all three stages achieves the best reconstruction quality among the tested configurations, indicating that conditioning across multiple decoding scales provides complementary benefits. The stage-specific variants offer lower per-chunk decoding latency. Full results are provided in Appendix~\ref{sec:supp_stage_injection}.
\section{Applications}
\label{sec:applications}

\subsection{Style transfer}
\label{sec:style_transfer}
\begin{figure}[t]
    \centering
    \setlength{\tabcolsep}{2pt}
    \renewcommand{\arraystretch}{0.8}
    \begin{tabular}{cccc}
        \includegraphics[width=0.24\textwidth]{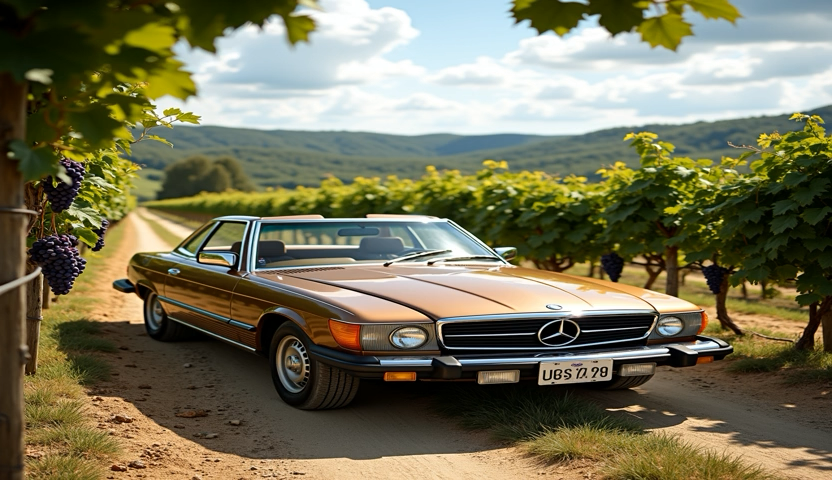} &
        \includegraphics[width=0.24\textwidth]{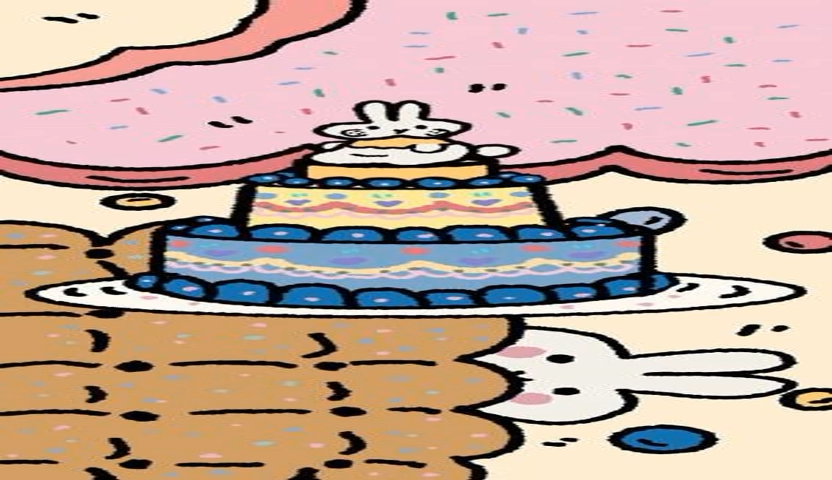} &
        \includegraphics[width=0.24\textwidth]{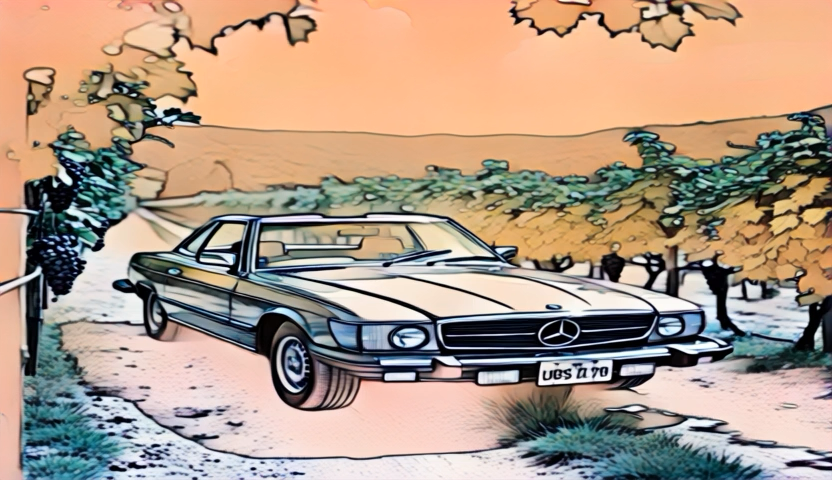} &
        \includegraphics[width=0.24\textwidth]{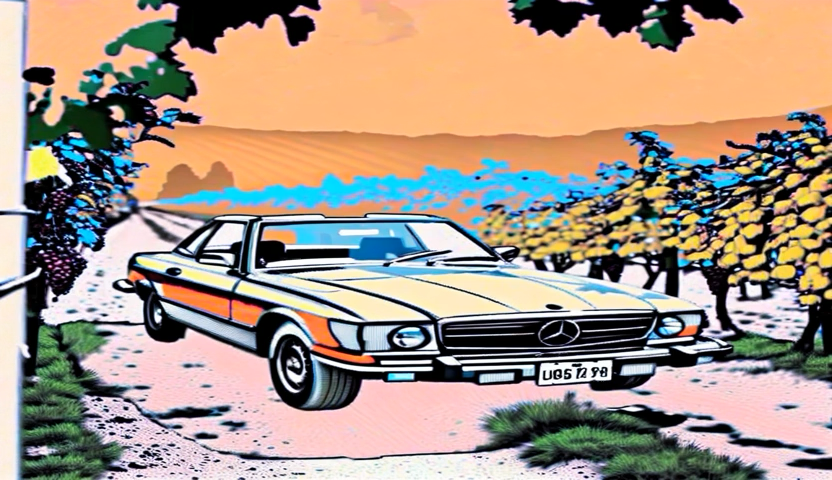} \\
        \includegraphics[width=0.24\textwidth]{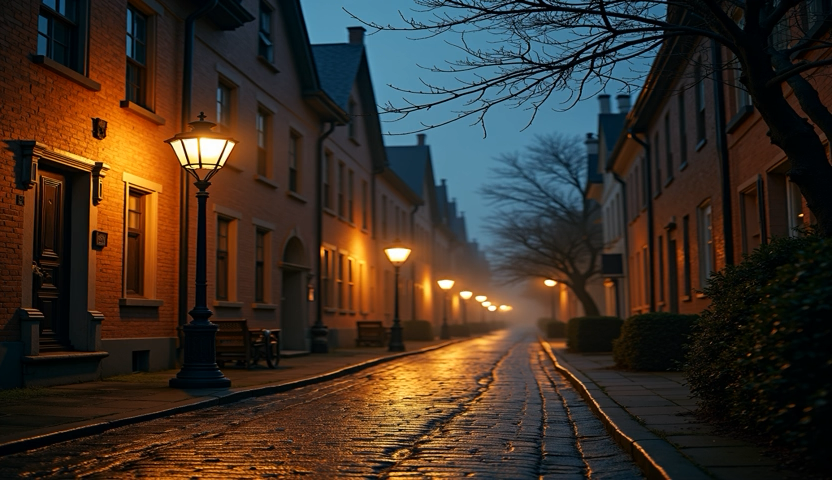} &
        \includegraphics[width=0.24\textwidth]{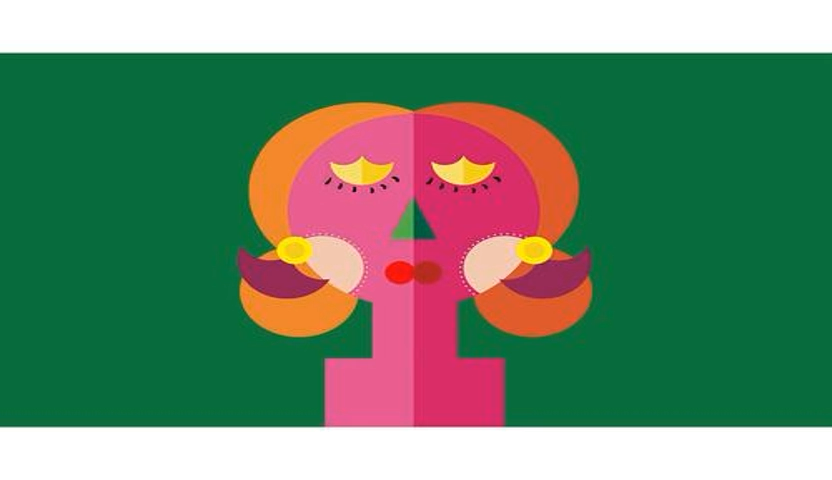} &
        \includegraphics[width=0.24\textwidth]{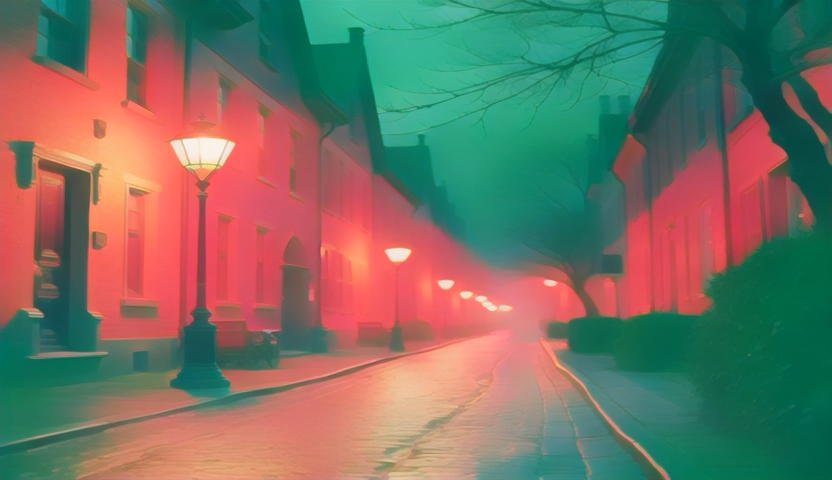} &
        \includegraphics[width=0.24\textwidth]{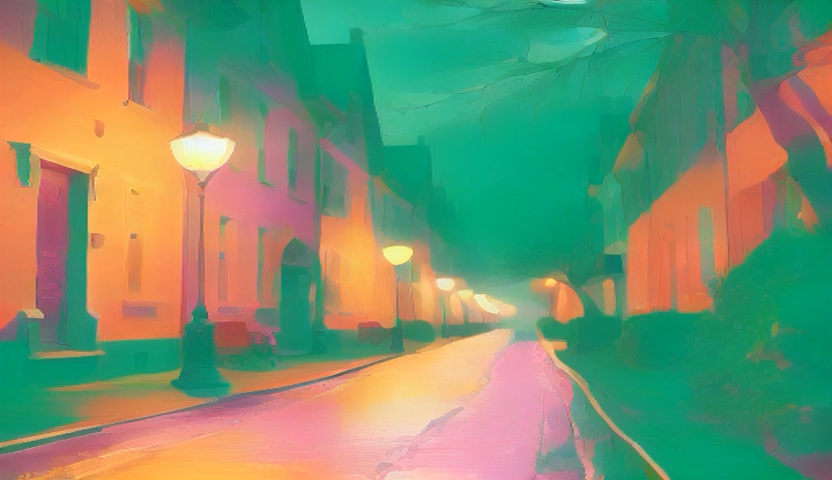} \\
        \includegraphics[width=0.24\textwidth]{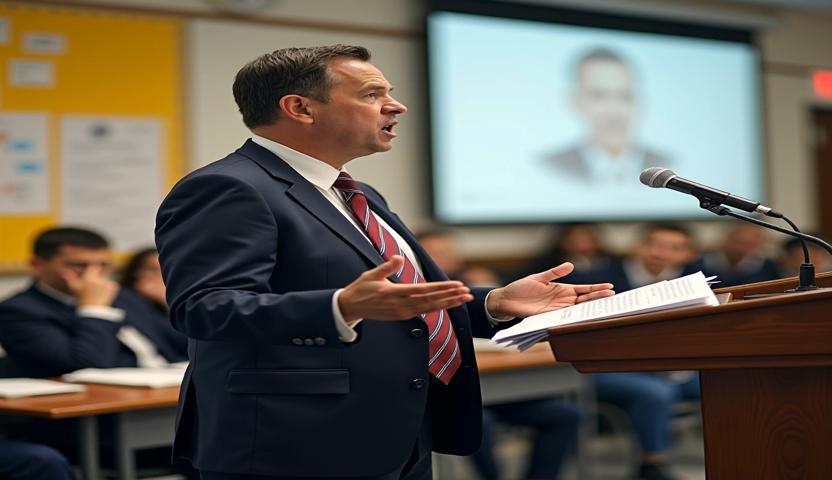} &
        \includegraphics[width=0.24\textwidth]{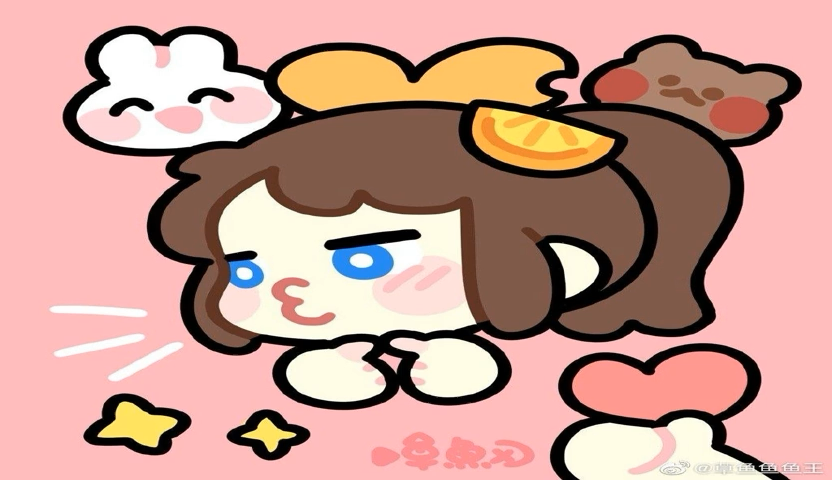} &
        \includegraphics[width=0.24\textwidth]{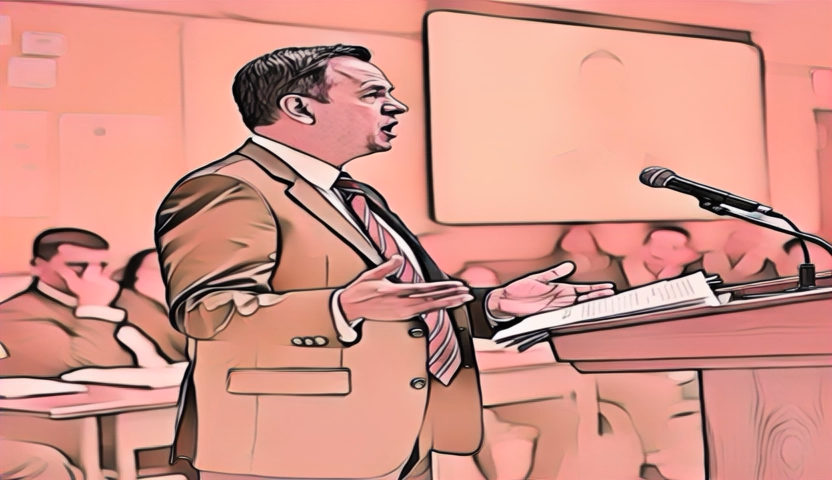} &
        \includegraphics[width=0.24\textwidth]{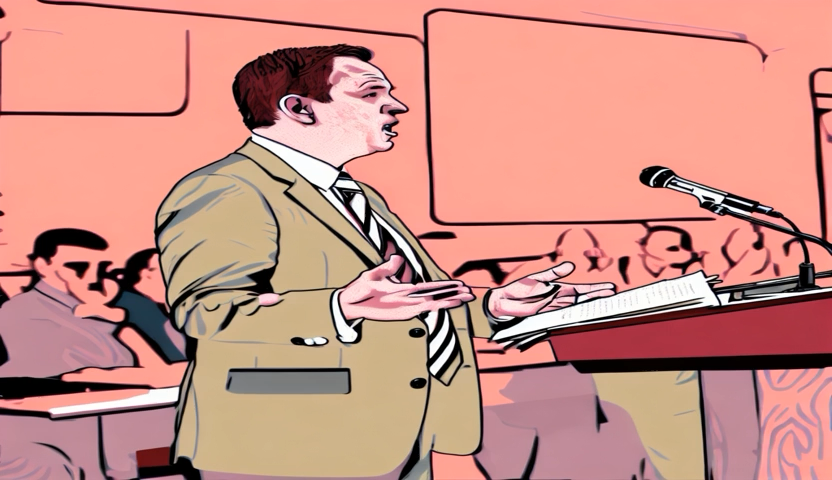} \\
        \small{Input} & \small{Reference} & \small{Ours} & \small{Ground Truth} \\
    \end{tabular}
    \caption{\textbf{Qualitative results on the style transfer task.} The reconstructed images follow the reference style while preserving the structural content of the input images.}
    \label{fig:style_transfer}
\end{figure}

To demonstrate the generality of our reference injection design, we apply \model to decode-time style transfer: by supplying a style image as the reference and training on OmniStyle150K~\cite{wang2025omnistylefilteringhighquality}, \model transfers the reference style to the input video at inference while preserving its structural content. Figure~\ref{fig:style_transfer} shows qualitative results.

\subsection{Video editing}
\label{sec:loraedit}
First-frame-guided video editors take a source video together with an edited version of its first frame, and produce a video in which the user-specified edit is applied consistently across all frames while everything else is expected to stay identical to the source. In practice, however, the editing model often noticeably degrades the \emph{non-edited} regions.

We experiment with video editing on LoRA-Edit~\cite{gao2025lora}, a first-frame-guided editor that fine-tunes the Wan~2.1 model with a mask-aware LoRA. We compare decoding the latents with the original Wan~2.1 decoder vs.\ \model. We follow LoRA-Edit's pipeline to track the edited object across frames, and score masked PSNR, SSIM, and LPIPS over the non-edited region, averaged over all frames. As shown in Table~\ref{tab:loraedit}, \model improves the non-edited region on every metric ($+1.4$\,dB PSNR, $+1.9$ SSIM (\%), $-0.018$ LPIPS). Figure~\ref{fig:loraedit_qual} further shows that \model preserves high-quality details that the baseline decoder blurs out, despite significant movement from the reference frame.

\begin{table}[t]
    \centering
    \caption{\textbf{Evaluation of \model on LoRA-Edit~\cite{gao2025lora}.} We decode the same latents with the original decoder vs.\ \model and score the non-edited region against the source video using a per-frame mask. Numbers in {\color{teal}teal} are gains of \model over the baseline.}
    \label{tab:loraedit}
    \small
    \setlength\tabcolsep{8pt}
    \renewcommand{\arraystretch}{1.15}
    \begin{tabular}{lccc}
        \toprule
        \textbf{Decoder} & \textbf{PSNR$\uparrow$} & \textbf{SSIM$\uparrow$} & \textbf{LPIPS$\downarrow$} \\
        \midrule
        Wan~2.1~\cite{wan2025} & 30.2 & 90.7 & 0.0477 \\
        \textbf{\model} (Wan~2.1) & \textbf{31.6} & \textbf{92.6} & \textbf{0.0297} \\[-2pt]
        & {\scriptsize\color{teal}($+1.4$)} & {\scriptsize\color{teal}($+1.9$)} & {\scriptsize\color{teal}($-0.018$)} \\
        \bottomrule
    \end{tabular}
\end{table}

\begin{figure}[t]
    \centering
    \setlength{\tabcolsep}{2pt}
    \renewcommand{\arraystretch}{0.8}
    \begin{tabular}{cc}
        \includegraphics[width=0.33\textwidth]{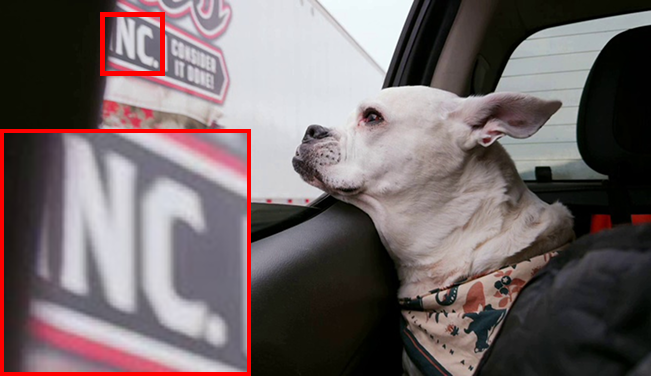} &
        \includegraphics[width=0.33\textwidth]{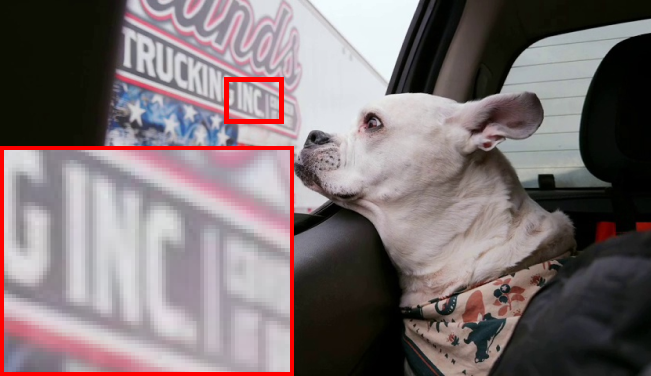} \\
        \small{Reference} & \small{Ground truth} \\
        \includegraphics[width=0.33\textwidth]{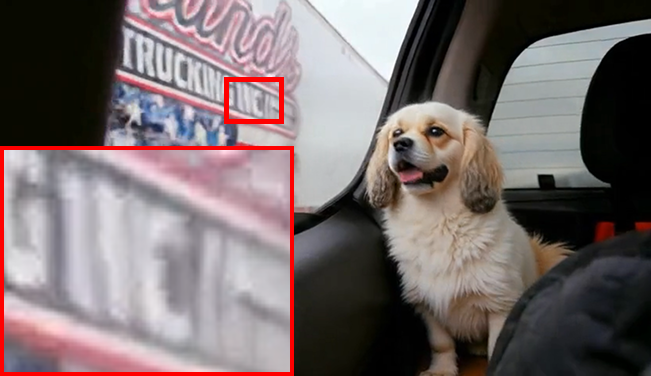} &
        \includegraphics[width=0.33\textwidth]{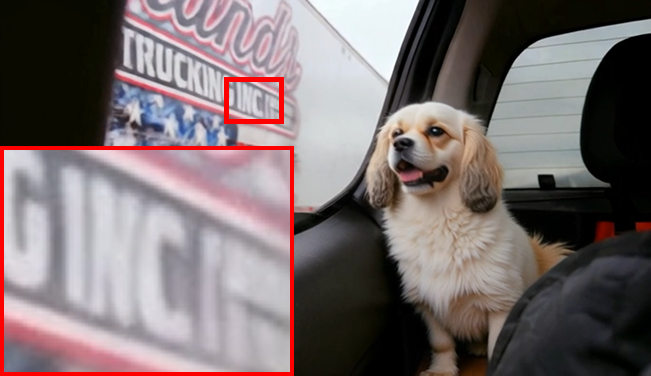} \\
        \small{LoRA-Edit} & \small{LoRA-Edit + \model} \\
    \end{tabular}
    \caption{\textbf{Qualitative comparison on LoRA-Edit~\cite{gao2025lora}.} \emph{Reference}: first frame of the source video. The baseline decoder distorts non-edited regions, while \model regenerates them faithfully from the reference, preserving high-quality details such as the text.}
    \label{fig:loraedit_qual}
    \vspace{-1.5em}
\end{figure}
\section{Discussion}
\label{sec:discussion}

We introduced \model, a reference-conditioned video VAE decoder that addresses a fundamental asymmetry in conditional generation: while the diffusion backbone is richly conditioned, the decoder remains generally unconditional. By injecting reference image tokens via self-attention, \model serves as a lightweight, architecture-agnostic drop-in replacement for existing decoders. Consistent improvements across several backbones, including mainstream open-source foundation video models, on both reconstruction and generation benchmarks, together with its natural extension to style transfer and video editing, demonstrate the generality and task-agnostic nature of conditional decoding. We hope this work encourages the community to rethink the decoder's role in conditional generation.

Several limitations suggest future directions: extending from a single reference to multiple references (\eg, for interpolation or multi-view tasks), exploring richer reference encoders beyond a single convolution layer, automating hyperparameter selection across architectures, and scaling beyond 81 frames, where the reference becomes more temporally distant. Because \model amplifies the perceptual fidelity of existing video generation models, it also inherits the same dual-use risks as the underlying model (\eg, deepfake misuse).

\begin{ack}
This work was supported in part by the NSF through award IIS-2211133. The authors declare no additional competing interests.
\end{ack}

\clearpage
{\small
\bibliography{main}
\bibliographystyle{plain}
}

\clearpage
\appendix
\vspace{-0.5em}
\vspace{-0.5em}
\section{Related Work}
\label{sec:related}
\vspace{-0.5em}
\vspace{-0.5em}

\noindent\textbf{Visual tokenization for generation.}
Variational autoencoders (VAEs)~\cite{kingma2022autoencodingvariationalbayes} learn latent-space probabilistic models for efficient representation.
VQ-VAE~\cite{van2017neural} introduces a vector-quantized discrete latent space, and VQGAN~\cite{esser2021tamingtransformershighresolutionimage} extends it to high-resolution images with adversarial training.
Latent diffusion models (LDMs)~\cite{rombach2022highresolutionimagesynthesislatent} leverage these encoded latent spaces for high-quality image generation.
For video, MAGVIT~\cite{yu2023magvitmaskedgenerativevideo} employs a 3D tokenizer for efficient video encoding, and MAGVIT v2~\cite{yu2024languagemodelbeatsdiffusion} introduces a causal 3D CNN.
ElasticTok~\cite{yan2024elastictokadaptivetokenizationimage} offers adaptive tokenization for long videos, and OmniTokenizer~\cite{wang2024omnitokenizerjointimagevideotokenizer} proposes joint image-video tokenization.
Recent continuous video autoencoders such as Cosmos Tokenizer~\cite{reda2024cosmostokenizer} and VideoVAE\textbf{+}~\cite{xing2025videovae+} serve as the visual backbone for state-of-the-art video generation models.

\noindent\textbf{Video generation.}
Early video generation relied on GANs~\cite{skorokhodov2022styleganvcontinuousvideogenerator,tulyakov2017mocogandecomposingmotioncontent,clark2019adversarialvideogenerationcomplex,yu2022generatingvideosdynamicsawareimplicit}, but these are limited to short, low-resolution outputs.
The field has since shifted to video diffusion models~\cite{blattmann2023stablevideodiffusionscaling,singer2022makeavideotexttovideogenerationtextvideo,blattmann2023alignlatentshighresolutionvideo,menapace2024snapvideoscaledspatiotemporal,chen2024videocrafter2overcomingdatalimitations,wang2023videocomposercompositionalvideosynthesis,zhang2023show1marryingpixellatent}, which produce high-quality results at the cost of many denoising steps.
Frontier open-source systems such as Wan~2.1~\cite{wan2025}, HunyuanVideo~\cite{kong2025hunyuanvideosystematicframeworklarge}, and CogVideoX~\cite{yang2024cogvideox} adopt the latent diffusion paradigm with powerful VAE backbones, achieving state-of-the-art quality. Image-to-video (I2V) models condition the generation process on a reference frame to produce temporally coherent video.
Stable Video Diffusion~\cite{blattmann2023stablevideodiffusionscaling} conditions a video diffusion model on a single image via concatenation in latent space.
DynamiCrafter~\cite{xing2023dynamicrafteranimatingopendomainimages} animates open-domain images by injecting them as visual context into the diffusion backbone.
IP-Adapter~\cite{ye2023ipadapter} introduces decoupled cross-attention to inject image prompt features into diffusion models, enabling image-guided generation with a lightweight adapter.
A common thread across these approaches is that conditioning is applied exclusively to the \emph{diffusion model}; the VAE decoder remains a purely unconditional reconstruction module and does not have access to the reference image.

\noindent\textbf{Reference-conditioned decoding.}
Recent attempts to introduce reference conditioning into the VAE decoder have struggled to achieve full quality recovery using the reference signal. For example, RefTok~\cite{fan2025reftok} achieves improved reconstruction only at low resolutions using a quantized codebook and struggles to yield a high-quality video generation model; H3AE~\cite{wu2025h3ae} instead focuses on high compression ratios, with imperfect reference information forming blurred reconstructions; Reducio-VAE~\cite{tian2025reducio} employs low-dimensional latent features without capturing high-fidelity reference signal. CMD~\cite{yu2024cmd} learns a video autoencoder that decomposes videos into a content frame and compact motion latents.

Prior methods exploring this space develop their own video autoencoder architectures and are tightly coupled to their respective systems. This further limits their quality and applicability, as their corresponding diffusion models are tied to the specific autoencoder representations. For example, compared to the 1.5 billion training videos of Wan 2.1, H3AE and Reducio are only trained on million-scale data sources.
In contrast, \model is designed as a \emph{plug-and-play} decoder module that can be integrated with existing, pretrained frontier video model VAE decoders
without modifying the encoder or retraining the diffusion backbone, making it immediately deployable in production pipelines.

\section{Implementation Details}

We apply \model to two pretrained video VAE backbones: Wan~2.1~\cite{wan2025} and VideoVAE\textbf{+}~\cite{xing2025videovae+}. Below we describe the architecture and training details using Wan~2.1 and  VideoVAE\textbf{+} as backbones.

\subsection{Model Architecture}
The Wan~2.1 decoder operates in three stages at progressively higher spatial resolutions, with channel dimensions of 384, 192, and 192. 
At each stage, spatio-temporal feature tokens are extracted using 3D patch embeddings with temporal stride~1 and spatial strides of $2{\times}2$, $4{\times}4$, and $8{\times}8$, respectively.

For comparison, VideoVAE\textbf{+} also adopts a three-stage decoder but with channel dimensions of 512, 256, and 128. Spatio-temporal tokens are extracted via 3D patch embeddings with temporal stride~1 and spatial strides of $8{\times}8$, $16{\times}16$, and $16{\times}16$, respectively.

For both backbones, Transformer blocks are shared across all three stages. Each uses $H{=}12$ heads with head dimension $d_h{=}128$, resulting in a total hidden dimension of $d{=}1536$. We adopt Rotary Position Embeddings (RoPE)~\cite{su2021roformer} for positional encoding and GELU activations in the feed-forward layers.

The reference image encoder is implemented as a lightweight tokenizer consisting of a single strided 3D convolution layer (patch size~8, Xavier-uniform initialization) that maps the reference frame into tokens matching the channel dimension of the first decoder stage (384 for Wan~2.1 and 512 for VideoVAE\textbf{+}). This is followed by normalization (RMSNorm for Wan~2.1 and GroupNorm for VideoVAE\textbf{+}). During training, the reference frame is sampled uniformly at random from the input video sequence.

\subsection{Training Setup}
\label{sec:supp_train_setup}
We train on 8 NVIDIA H200 GPUs with a per-GPU batch size of~2 (effective batch size~16) for 5 days. Model parameters are divided into two groups: (1)~the attention block and reference image encoder, optimized at a base learning rate of $2 \times 10^{-4}$ (effective learning rate $1.6 \times 10^{-3}$ after linear scaling); and (2)~the pretrained VAE decoder, fine-tuned at $0.1$ of the effective rate. Both groups use AdamW with $\beta_1 = 0.9$ and $\beta_2 = 0.999$.

The learning rate schedule consists of a linear warmup from $1\%$ to $100\%$ of the base rate over the first 1{,}000 steps, followed by cosine annealing over 100{,}000 total steps. The encoder, quantization convolutions, and post-quantization convolutions remain frozen throughout; only the decoder and attention block are updated.

\subsection{Loss Function}
The training objective combines pixel-wise and perceptual losses:
\begin{equation*}
  \mathcal{L} = \|\mathbf{x} - \hat{\mathbf{x}}\|_1 + \mathcal{L}_{\text{LPIPS}}(\mathbf{x}, \hat{\mathbf{x}})
\end{equation*}
where $\mathbf{x}$ and $\hat{\mathbf{x}}$ denote the ground-truth and reconstructed video frames, and $\mathcal{L}_{\text{LPIPS}}$ is the LPIPS perceptual loss~\cite{zhang2018perceptual}. Since the encoder is frozen, KL divergence regularization is disabled.

\subsection{Data and Evaluation}
For Wan~2.1, training videos are resized to $480{\times}832$ pixels, with 5 frames per sample in the first training stage and 17 frames in the second stage. For VideoVAE\textbf{+}, training videos are resized to $216{\times}216$ pixels, with 4 frames per sample in the first stage and 16 frames in the second stage. We evaluate reconstruction quality using PSNR, SSIM, and LPIPS~\cite{zhang2018perceptual}.

\subsection{Reconstruction Benchmarks}
\label{sec:supp_recon_benchmarks}
We use three reconstruction benchmarks. Across all of them we use the exact video lists provided by VideoVAE\textbf{+}~\cite{xing2025videovae+} so that our numbers are directly comparable with those reported in~\cite{xing2025videovae+}.
\begin{itemize}
    \item \textbf{Inter4K (test).} The test split of Inter4K~\cite{stergiou2022adapoolexponentialadaptivepooling}, $500$ high-quality videos.
    \item \textbf{WebVid.} An in-the-wild subset of WebVid~\cite{Bain21}.
    \item \textbf{Large Motion.} $100$ videos ($80$ from WebVid~\cite{Bain21}, $20$ from Inter4K~\cite{stergiou2022adapoolexponentialadaptivepooling}) manually curated to exhibit complex motion dynamics---significant camera motion, fast-moving subjects, and large inter-frame displacements.
\end{itemize}
For all three benchmarks, videos are decoded at the same resolution and frame count as the corresponding training configuration of each backbone, and PSNR / SSIM / LPIPS are averaged over all frames.

\subsection{Per-Category Reconstruction on Inter4K}
\label{sec:supp_per_category}

Table~\ref{tab:per_category} reports per-category PSNR on Inter4K (Wan~2.1 backbone). Gains are positive across every category. They are largest on \emph{content-rich} scenes with strong high-frequency structure (urban streets, neon night, driving POVs, aerial skylines, indoor venues, sports, underwater wildlife, and nature landscapes), and smallest on \emph{content-sparse} scenes that are already smooth or close-up (animation graphics, performance events, close-up macros, and people portraits), where the latent bottleneck loses comparatively less detail and the baseline already reconstructs faithfully. This pattern supports the intuition that the reference signal contributes most where the latent code is least sufficient.

\noindent\textbf{Categorization protocol.}
To group the Inter4K test clips into 12 semantic categories without manual labeling, we use zero-shot CLIP~\cite{radford2021learningtransferablevisualmodels}.

\noindent\textbf{Visual features.} For each clip we sample three frames at evenly-spaced positions. Each frame embedding is $L_2$-normalized, and we mean-pool the three to obtain a single 768-d clip embedding $v_i$.

\noindent\textbf{Category prompts.} We define 12 categories and write 3--4 short natural-language prompts per category (42 prompts in total), encode each with CLIP's text tower, and $L_2$-normalize. Example prompts include ``a busy city street with cars and buildings'' (urban street), ``neon signs and billboards at night in a city'' (neon night), ``first-person view of driving down a road'' (driving pov), and ``an aerial drone view of a city'' (skyline aerial); the remaining categories are indoor venue, sports action, nature landscape, close up macro, people portrait, underwater wildlife, performance event, and animation graphics. The full prompt list is included in the released code.

\noindent\textbf{Category score and assignment.} For clip $i$ and category $c$ with prompt embeddings $\{t_{c,k}\}$, the score is the mean cosine similarity over prompts in that category, $s_{i,c} = \tfrac{1}{|c|}\sum_{k} v_i^{\top} t_{c,k}$, and each clip is assigned to a single category by $\arg\max_{c}\, s_{i,c}$. Per-category counts are reported in Table~\ref{tab:per_category}.

\begin{table}[h]
    \centering
    \caption{Per-category reconstruction PSNR on Inter4K (Wan~2.1 backbone). $n$ denotes the number of videos per category; $\Delta$ is the PSNR gain of \model over the Wan~2.1 baseline. Categories are grouped into content-rich (top) and content-sparse (bottom) scenes.}
    \label{tab:per_category}
    \small
    \setlength\tabcolsep{6pt}
    \renewcommand{\arraystretch}{1.1}
    \begin{tabular}{lcccc}
        \toprule
        \textbf{Category} & \textbf{$n$} & \textbf{Wan~2.1} & \textbf{\model} & \textbf{$\Delta$ PSNR} \\
        \midrule
        \multicolumn{5}{l}{\emph{Content-rich scenes}} \\
        Urban street            &  43 & 29.0 & 31.4 & \textbf{+2.4} \\
        Neon night              &  34 & 31.6 & 33.6 & \textbf{+2.1} \\
        Driving POV             &  36 & 31.3 & 33.4 & \textbf{+2.1} \\
        Skyline aerial          &  48 & 31.3 & 33.3 & \textbf{+1.9} \\
        Indoor venue (arena/mall) &  6 & 30.8 & 32.6 & +1.8 \\
        Sports action           &  29 & 32.7 & 34.2 & +1.6 \\
        Underwater wildlife     &  37 & 32.8 & 34.3 & +1.4 \\
        Nature landscape        &  14 & 32.3 & 33.7 & +1.4 \\
        \midrule
        \multicolumn{5}{l}{\emph{Content-sparse scenes}} \\
        Animation graphics (CGI / fluid sim) & 164 & 33.8 & 35.0 & +1.2 \\
        Performance event (concert/dance) & 45 & 34.2 & 35.2 & +1.0 \\
        Close-up macro          &  23 & 37.4 & 38.2 & +0.8 \\
        People portrait         &  18 & 38.5 & 39.2 & +0.7 \\
        \bottomrule
    \end{tabular}
\end{table}

\subsection{VBench Evaluation Protocol}
\label{sec:supp_vbench_protocol}
Since \model is a drop-in decoder, our VBench~\cite{huang2023vbench, huang2025vbench++, zheng2025vbench2} evaluation follows a \emph{fixed-seed} protocol that controls for diffusion-side stochasticity. We describe the protocol here so that any per-prompt difference reflects the choice of decoder rather than differences in noise or sampling.

\noindent\textbf{Fixed latent generation.}
For each prompt in the VBench info list we generate $K=5$ samples, matching the default VBench-evaluation. For each (prompt, sample-index) pair we draw a 32-bit seed once via \texttt{random.randint(0, $2^{32}-1$)}, persist it to a per-GPU JSON log, and use it to instantiate a generator. The Wan~2.1 pipeline is then run with that generator at 50 inference steps, classifier-free guidance scale $5.0$, $17$ frames, and the same 16:9 480p reference image used by the official VBench benchmark. The pipeline is invoked with \texttt{output\_type="latent"}, so the noised input, denoising trajectory, and final latents are fully determined by the seed. The latents and the seed used to produce them are saved together to disk.

\noindent\textbf{Decoder swap.}
The same set of saved latents is then decoded twice---once with the original Wan~2.1 VAE decoder and once with our \model\---and both decoded videos are scored with the official VBench-I2V evaluation pipeline. Because both decoders consume exactly the same latents produced from exactly the same seeds, the per-prompt comparison is paired: differences in VBench dimensions can only arise from the decoder, not from a different noise sample, prompt order, or guidance schedule. This is why we report the comparison without an explicit per-seed standard deviation; sample-level variance from the diffusion sampler is shared across the two methods.

\noindent\textbf{Reproducibility.}
The seed log files are kept alongside the saved latents, so any third party with access to the Wan~2.1 checkpoint can reproduce both the latents and the decoded videos bit-for-bit. The decoding scripts and the latent-generation script will be released together with the code.

\section{Temporal Stability}
\label{sec:supp_temporal_stability}

Table~\ref{tab:temporal_stability} reports the temporal-stability metrics summarized in the main paper: flow warping error ($E_\text{warp}$)~\cite{lai2018learning}, temporal LPIPS ($E_\text{tLPIPS}$)~\cite{zhang2018perceptual}, flicker ($E_\text{flicker}$)~\cite{bonneel2015blind}, and CLIP consistency (CLIP$_\text{cons}$)~\cite{esser2023structurecontentguidedvideosynthesis}. The first three measure temporal degradation (lower is better) and the last consistency (higher is better). All metrics are computed on the VBench generations under the fixed-seed protocol of Sec.~\ref{sec:supp_vbench_protocol}, so any per-prompt difference reflects the choice of decoder rather than diffusion-side variance.

\begin{table}[t]
    \centering
    \caption{Temporal-stability metrics on VBench (Wan~2.1 backbone, fixed-seed protocol). $E_\text{warp}$, $E_\text{tLPIPS}$, and $E_\text{flicker}$ measure temporal degradation (lower is better); CLIP$_\text{cons}$ measures temporal consistency (higher is better). \model reduces flickering and warping error across all degradation metrics. $\Delta$ is the relative change vs.~Wan~2.1.}
    \label{tab:temporal_stability}
    \small
    \setlength\tabcolsep{6pt}
    \renewcommand{\arraystretch}{1.15}
    \begin{tabular}{lcccc}
        \toprule
        \textbf{Model} & $E_\text{warp}\downarrow$ & $E_\text{tLPIPS}\downarrow$ & $E_\text{flicker}\downarrow$ & CLIP$_\text{cons}\uparrow$ \\
        \midrule
        Wan~2.1~\cite{wan2025}  & 0.03274 & 0.09136 & 0.02727 & 0.99526 \\
        \model (Wan~2.1)        & \textbf{0.02941} & \textbf{0.08367} & \textbf{0.02303} & \textbf{0.99588} \\[-2pt]
        & {\scriptsize\color{teal}($-10.2\%$)} & {\scriptsize\color{teal}($-8.4\%$)} & {\scriptsize\color{teal}($-15.5\%$)} & {\scriptsize\color{teal}($+0.06\%$)} \\
        \bottomrule
    \end{tabular}
\end{table}

\section{Robustness to Reference Mismatch}
\label{sec:supp_reference_mismatch}

\subsection{Scene-Boundary Stress Test}
\label{sec:supp_scene_boundary}

We construct a stress test using $100$ Inter4K videos containing
scene-boundary cuts. We place the boundary near the middle of
each evaluated clip and randomly select the reference frame,
excluding the boundary frame. The reference and evaluated frames
may therefore come from different shots with substantially
different backgrounds.

\begin{table}[t]
    \centering
    \caption{
        \textbf{Reconstruction under scene-boundary reference mismatch.}
        \model improves PSNR and SSIM, while LPIPS increases.
    }
    \label{tab:scene_boundary}
    \small
    \setlength{\tabcolsep}{8pt}
    \renewcommand{\arraystretch}{1.15}
    \begin{tabular}{lccc}
        \toprule
        Model & PSNR $\uparrow$ & SSIM $\uparrow$ & LPIPS $\downarrow$ \\
        \midrule
        Wan~2.1 & 31.45 & 0.905 & \textbf{0.039} \\
        \model & \textbf{31.82} & \textbf{0.917} & 0.045 \\
        \bottomrule
    \end{tabular}
\end{table}

Table~\ref{tab:scene_boundary} shows that \model improves PSNR
by $0.37$\,dB and SSIM by $0.012$, while LPIPS only increases by
$0.006$. Thus, pixel-level and structural improvements persist
in this reference-mismatch setting.

\subsection{Performance versus Reference Distance}
\label{sec:supp_reference_distance}

We examine reconstruction quality as a function of temporal
distance from the reference on $81$-frame clips from $500$
randomly sampled Inter4K videos. The source videos are recorded
at $60$ fps and sampled with temporal stride~4, yielding an
effective frame rate of $15$ fps. For each clip, we randomly
select one reference frame and measure per-frame PSNR as a
function of the absolute frame-index difference from the reference.

At a distance of $10$ sampled frames from the reference, PSNR
increases from approximately $31.1$ to $31.5$\,dB.
This improvement indicates that the benefit is not confined
to the reference frame or its immediate neighbors.

\section{Comparison with H3AE}
\label{sec:supp_h3ae_reducio}

\begin{figure}[h]
    \centering
    \setlength{\tabcolsep}{1pt}
    \renewcommand{\arraystretch}{0.6}
    \resizebox{\textwidth}{!}{%
    \begin{tabular}{ccccc}
        \rotatebox{90}{\small\hspace{18pt}Example 1} &
        \includegraphics[width=0.25\textwidth]{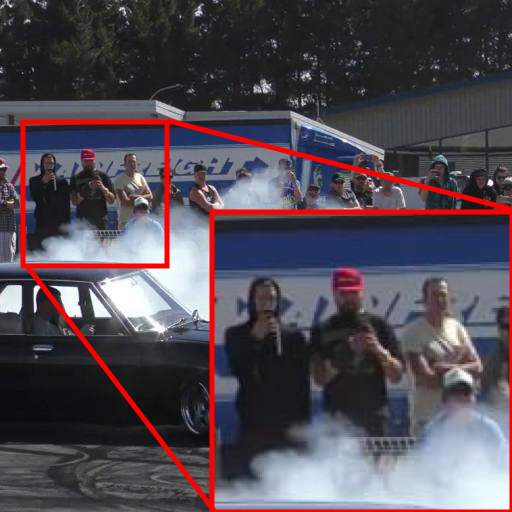} &
        \includegraphics[width=0.25\textwidth]{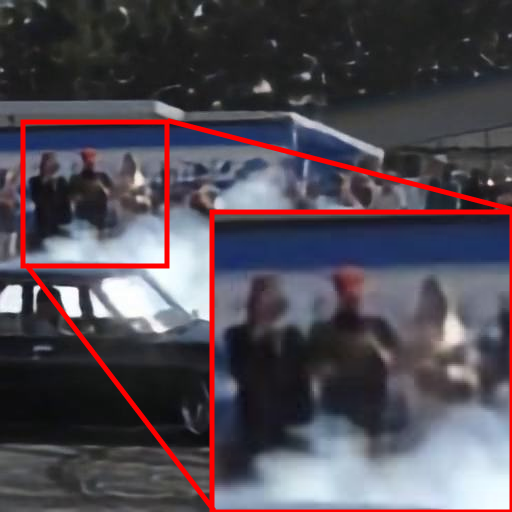} &
        \includegraphics[width=0.25\textwidth]{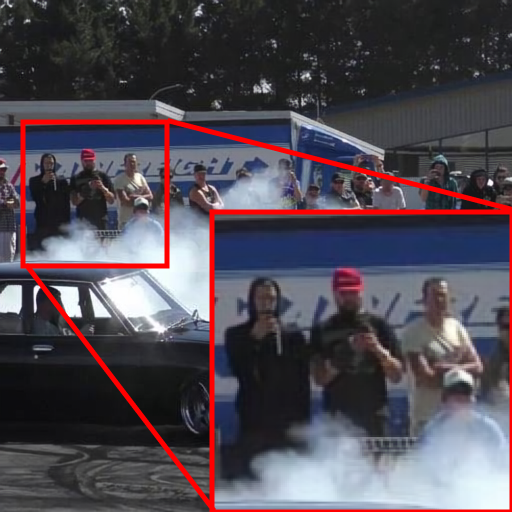} \\
        \rotatebox{90}{\small\hspace{18pt}Example 2} &
        \includegraphics[width=0.25\textwidth]{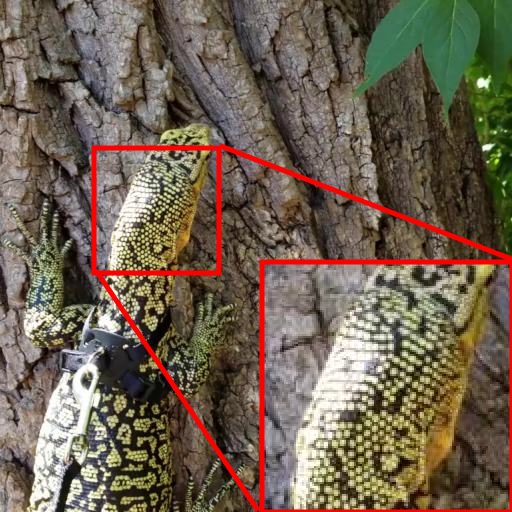} &
        \includegraphics[width=0.25\textwidth]{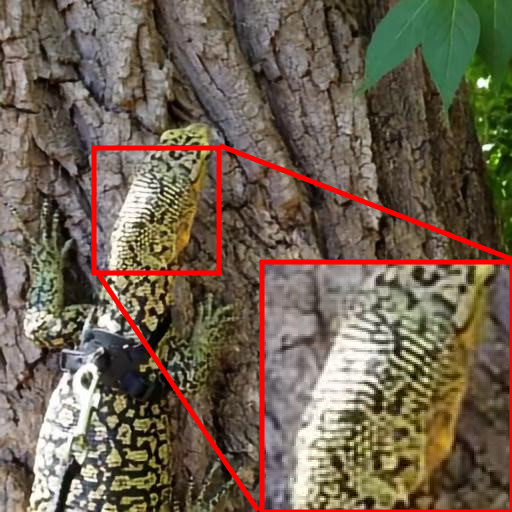} &
        \includegraphics[width=0.25\textwidth]{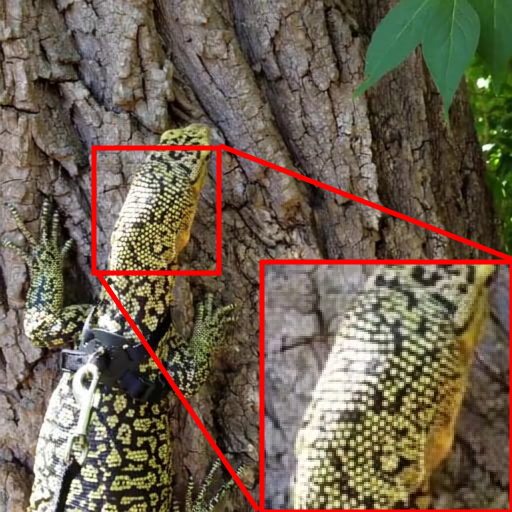} \\[2pt]
        & \small{Ground truth} & \small{H3AE} & \small{Ours} \\
    \end{tabular}}
    \caption{\textbf{Qualitative comparison with H3AE.} The ground-truth and H3AE crops are taken directly from the figures in the H3AE paper, since H3AE's training data, code, and weights are not publicly released. \model reconstructions are produced by us on the \emph{same} input frames so that both methods are evaluated under identical inputs. \model recovers fine textures and structural details (lizard dots, background people) more faithfully than H3AE.}
    \label{fig:h3ae_comparison}
\end{figure}

We provide a qualitative head-to-head comparison with H3AE~\cite{wu2025h3ae}, a prior reference-conditioned VAE. Since H3AE has not released training or evaluation code, training data, or model weights, we cannot run it on our benchmark splits; instead, we take the examples directly from the figures in the H3AE paper and run \model on the same input frames so that both methods are evaluated under identical inputs. As shown in Figure~\ref{fig:h3ae_comparison}, \model preserves fine structures and textures (\eg, the lizard's dots and the spectator faces) noticeably better than H3AE.

\section{Human Evaluation}
\label{sec:supp_human_eval}

\begin{figure}[t]
    \centering
    \includegraphics[width=0.95\textwidth]{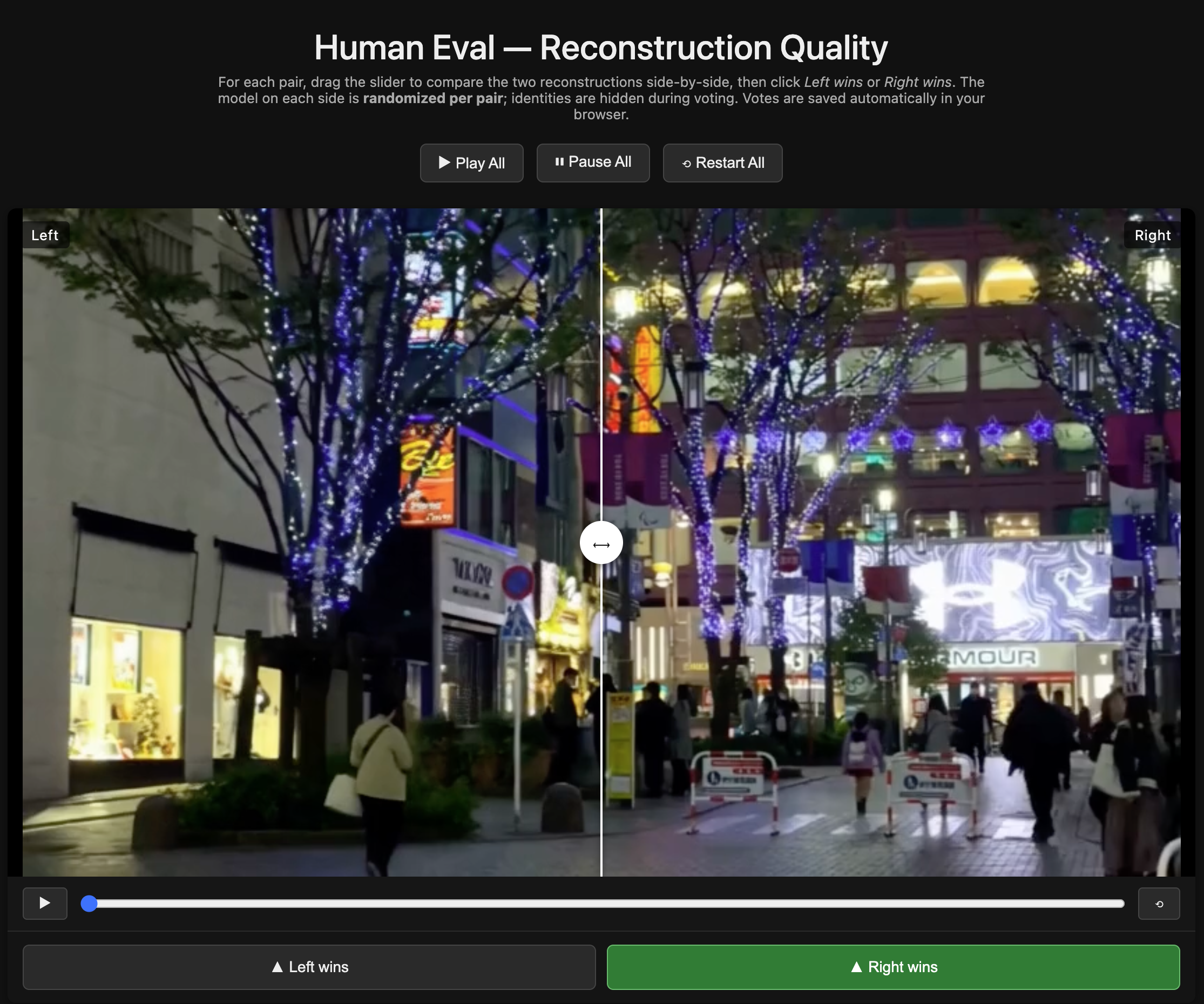}
    \caption{\textbf{Human-evaluation interface.} Each pair of videos is rendered as a slider-overlay: the two videos are stacked and synchronously played, and the evaluator drags a vertical divider to reveal one decoder's output on the left half and the other on the right half. Left/right assignment is randomized per pair and cached across reloads, and model identities are hidden until after voting. The evaluator registers a preference using the ``Left wins'' / ``Right wins'' buttons. Preference rates are normalized over non-tied votes; ties are not shown.}
    \label{fig:human_eval_interface}
\end{figure}

\noindent\textbf{Interface.}
Evaluations are collected through a custom web tool. Each pair is rendered as a single slider-overlay: the two videos are stacked and synchronously played, and the evaluator drags a vertical divider to reveal one model on the left half and the other on the right half. Left/right assignment is randomized per pair (and cached so reloads do not reshuffle), and model identities are hidden until after voting. The evaluator clicks one of two buttons, ``Left wins'' or ``Right wins'', to register a preference. For a tie, the evaluator leaves both options unselected.

\noindent\textbf{Evaluation protocol and votes.}
Evaluators are instructed to view and rate a random subset of the pairs assigned to a given baseline based on overall visual quality. All participants took part voluntarily and received no compensation. We collect $252$ evaluation records in total, including ties: $102$ for Wan~2.1 ($89$ generation, $13$ reconstruction) and $50$ each for VideoVAE\textbf{+}, HunyuanVAE, and Reducio-VAE (all reconstruction). Evaluators leave both preference options unselected when they judge the two videos to be tied. These ties are included in the total count but excluded when computing preference rates.

\noindent\textbf{Results.}
Table~\ref{tab:human_eval} summarizes the outcomes; \model\ is consistently preferred over every baseline: $92.3\%$ vs.~$7.7\%$ on Wan~2.1 reconstruction, $82.4\%$ vs.~$17.6\%$ on Wan~2.1 generation, $89.6\%$ vs.~$10.4\%$ vs.~Reducio-VAE, $69.2\%$ vs.~$30.8\%$ vs.~HunyuanVAE, and $63.4\%$ vs.~$36.6\%$ vs.~VideoVAE\textbf{+}. These human preferences are consistent with the quantitative reconstruction (Table~\ref{tab:recon_metrics}) and VBench (Table~\ref{tab:video_metrics}) results, as well as the temporal-stability results in Table~\ref{tab:temporal_stability}.

\definecolor{VChartColor1}{RGB}{60, 70, 255}
\definecolor{VChartColor2}{RGB}{200, 200, 200}
\definecolor{VChartColor3}{RGB}{237, 237, 237}
\newlength\VChartMax
\setlength\VChartMax{4.5em}
\newcommand*\VChart[4]{~\rlap{\textcolor{VChartColor2}{\rule{1\VChartMax}{1ex}}}\rlap{\textcolor{VChartColor3}{\rule{#3\VChartMax}{1ex}}}\rlap{\textcolor{VChartColor1}{\rule{#2\VChartMax}{1ex}}}\hphantom{\rule{1\VChartMax}{1ex}}~~~#1\% vs.\ #4\%}
\begin{table}[t]
  \caption{Human evaluation. Evaluators compare the visual quality of \model\ against four baselines on video pairs sampled from Inter4K reconstructions and VBench generations, presented via a slider-overlay interface with randomized left/right order and hidden model identities. The \textcolor{VChartColor1}{\textbf{blue}} segment is preference for \model\ and the gray tail is preference for the baseline; the two numbers next to each bar show \model's win percentage versus the baseline.}
  \label{tab:human_eval}
  \centering
  \scriptsize
  \setlength\tabcolsep{6pt}
  \begin{tabular}{l | l}
    \toprule
    \textbf{\model\ (Ours) vs. ...} & \textbf{Preference (Ours \% vs.\ Baseline \%)} \\
    \midrule
    \textbf{Wan~2.1 Generation}~\cite{wan2025}                                        & \VChart{82.4}{.824}{1.0}{17.6} \\
    \textbf{Wan~2.1 Reconstruction}~\cite{wan2025}                                        & \VChart{92.3}{.923}{1.0}{\hphantom{0}7.7} \\
    \textbf{VideoVAE\textbf{+}}~\cite{xing2025videovae+}                   & \VChart{63.4}{.634}{1.0}{36.6} \\
    \textbf{HunyuanVAE}~\cite{kong2025hunyuanvideosystematicframeworklarge} & \VChart{69.2}{.692}{1.0}{30.8} \\
    \textbf{Reducio-VAE}~\cite{tian2025reducio}                            & \VChart{89.6}{.896}{1.0}{10.4} \\
    \bottomrule
  \end{tabular}
  \vspace{-1em}
\end{table}

\section{Latent Training}
\label{sec:latent_train}
To further improve the performance of \model\ on the generation task, we additionally fine-tune the model using latent data. Specifically, we collect 30k images from TIP-I2V~\cite{wang2024tipi2v} and re-caption all images using CogVLM~\cite{wang2023cogvlm}. We then use the Wan~2.1 model~\cite{wan2025} to generate corresponding latent video frames.

During training, we use 30k videos from the original video dataset together with 30k latent videos generated by diffusion models. The latent videos correspond to the output of the diffusion model rather than the output of the VAE encoder. As in the main training stage, the encoder remains frozen and only the decoder is optimized. Training alternates between real videos and latent videos in successive batches.

For real video data, the reconstruction loss is computed between the input video and the reconstructed video. For latent video data, the reconstruction loss is applied only to the first frame by comparing the reconstructed first frame with the input reference frame. Reconstruction loss is not applied to the remaining frames.

In addition, we aim to train the model with latent sequences where the reference frame is not always the first frame, encouraging the model to learn longer temporal dependencies. Empirically, we find that when training only with forward sequences, early frames tend to preserve the reference appearance, while later frames may gradually deviate from it. To alleviate this issue, we construct reversed training sequences so that later frames are encouraged to attend to the reference frame.

Since a single latent video frame encodes multiple video frames, we cannot directly reverse the latent sequence. Instead, we first decode the latent sequence using the original Wan~2.1 decoder to obtain the video frames, reverse their temporal order, and then pass the reversed video through the original Wan~2.1 encoder to obtain the corresponding latent representation used for training.

This training also consists of two stages. In Forward Training, latent sequences are used in the original temporal order. In Bidirectional Training, each batch randomly uses either the forward or reversed sequence with equal probability.

\begin{table}[t]
    \centering
    \caption{VBench evaluation results across 12 evaluation metrics. We compare the Wan~2.1 baseline with \model under different training settings, including latent fine-tuning and bidirectional training. Our chosen strategies improve performance across most dimensions and lead to higher aggregate scores.}
    \label{tab:latent_vbench}
    \resizebox{\textwidth}{!}{%
        \setlength\tabcolsep{4pt}
        \renewcommand{\arraystretch}{1.2}
        \begin{tabular}{lcccccccccccc}
            \toprule
            & \multicolumn{6}{c}{\textbf{Quality Dimensions}} & \multicolumn{3}{c}{\textbf{I2V Dimensions}} & \multicolumn{3}{c}{\textbf{Aggregate Scores}} \\
            \cmidrule(lr){2-7} \cmidrule(lr){8-10} \cmidrule(lr){11-13}
            \textbf{Model}
            & \textbf{\scriptsize Subj.}
            & \textbf{\scriptsize BG}
            & \textbf{\scriptsize Motion}
            & \textbf{\scriptsize Dynamic}
            & \textbf{\scriptsize Aesth.}
            & \textbf{\scriptsize Imaging}
            & \textbf{\scriptsize I2V Subj.}
            & \textbf{\scriptsize I2V BG}
            & \textbf{\scriptsize Camera}
            & \textbf{\scriptsize Quality}
            & \textbf{\scriptsize I2V}
            & \textbf{\scriptsize Total} \\
            \midrule
            Wan~2.1~\cite{wan2025} & 0.9656 & 0.9788 & 0.9814 & 0.4748 & 0.6557 & 0.7011 & 0.9796 & 0.9912 & 0.1934 & 0.8127 & 0.9446 & 0.8786\\
            \model (w/o FT) & 0.9675 & 0.9820 & 0.9828 & \textbf{0.4764} & \textbf{0.6598} & 0.7006 & 0.9830 & 0.9930 & 0.1908 & 0.8156 & 0.9475 & 0.8815\\
            \model (Forward) & 0.9691 & 0.9858 & 0.9829 & 0.4732 & 0.6582 & \textbf{0.7027} & 0.9846 & 0.9949 & 0.1927 &\textbf{ 0.8167} & 0.9497 & 0.8832\\
            \model (Bidirectional) & \textbf{0.9693} & \textbf{0.9859} & \textbf{0.9833} & 0.4724 & 0.6586 & 0.7010 & \textbf{0.9847} & \textbf{0.9950} & \textbf{0.1937} & 0.8167 & \textbf{0.9499} & \textbf{0.8833}\\
            \bottomrule
        \end{tabular}
    }
\end{table}

Table \ref{tab:latent_vbench} presents VBench evaluation results on the Wan~2.1 backbone. Introducing \model already improves most evaluation dimensions compared to the original Wan~2.1 model. Applying latent fine-tuning further improves the performance, indicating that training on diffusion-generated latent videos helps the decoder better adapt to latent-space generation. Finally, the proposed bidirectional training strategy achieves the best overall performance, suggesting that exposing the model to reversed temporal sequences improves reference consistency across frames.

\section{Random Reference Frame}

We further study how the choice of reference frame during training affects model performance.
Specifically, we train two models on Wan~2.1 backbone with different reference-frame strategies: one uses the first frame as the reference during training, while the other randomly samples a reference frame from the video.
Both models use Wan~2.1 as the backbone with 5 Transformer blocks and a dropout rate of 0.7, and are evaluated on the 17 frames $480 \times 832$ Inter4K test set.

During evaluation, we test both models under two settings: using the first frame or a randomly sampled frame as the reference.
This allows us to analyze how each training strategy generalizes to different reference-frame conditions.

Table~\ref{tab:reference_frame_ablation} reports the quantitative results. 
When the model is trained using the first frame as the reference, it achieves the best reconstruction quality when evaluated with the same setting (first-frame reference). 
However, its performance drops significantly when evaluated with a randomly selected reference frame, indicating that the model overfits to the training reference strategy and generalizes poorly to other reference conditions. 
In particular, we observe that the model degenerates when the reference frame is fixed to frame~0, becoming overly dependent on the reference appearance.

In contrast, training with randomly sampled reference frames leads to more robust performance across evaluation settings. The model trained with random references achieves the best overall performance when evaluated with random references and remains competitive when evaluated with the first frame as the reference. Notably, even under the first-frame evaluation setting, random-reference training still outperforms training that always uses the first frame as the reference. These results suggest that random reference training improves the model's ability to adapt to varying reference-frame conditions during inference.

\begin{table}[t]
    \caption{Effect of reference-frame selection during training and evaluation. Models are trained using either the first frame or randomly sampled frames as the reference. Each model is evaluated using both first-frame and random-frame references. Training with random references leads to more robust performance across evaluation settings.
    }
    \label{tab:reference_frame_ablation}
    \resizebox{\textwidth}{!}{%
        \renewcommand{\arraystretch}{1.15}
            \begin{tabular}{c|c|ccc|ccc}
                \toprule
                Training Ref. & Eval Ref. & \multicolumn{3}{c|}{Overall} & \multicolumn{3}{c}{First Frame} \\
                & & PSNR $\uparrow$ & SSIM\,(\%) $\uparrow$ & LPIPS $\downarrow$ & PSNR $\uparrow$ & SSIM\,(\%) $\uparrow$ & LPIPS $\downarrow$ \\
                \midrule
                First frame & First frame & 34.1 & 94.06 & 0.0371 & \textbf{40.8} & \textbf{98.57} & \textbf{0.0046} \\
                First frame & Random frame & 31.9 & 91.06 & 0.0556 & 22.9 & 66.44 & 0.2194 \\
                \midrule
                Random frame & First frame & 34.2 & 94.14 & 0.0371 & 40.0 & 98.24 & 0.0073 \\
                Random frame & Random frame & \textbf{34.6} & \textbf{94.64} & \textbf{0.0337} & 37.2 & 96.82 & 0.0215 \\
                \bottomrule
            \end{tabular}
    }
\end{table}

\section{Effect of the Number of Transformer blocks}
\label{sec:supp_layer_ablation}

Table~\ref{tab:layer_ablation_recon} reports the block-count ablation summarized in the main paper. We train the Wan~2.1 backbone with $\{3, 5, 7, 10\}$ Transformer blocks under the same two-stage curriculum and evaluate on all three reconstruction benchmarks. Reconstruction quality improves consistently with depth across PSNR, SSIM, and LPIPS, and the 10-blocks model attains the best results on every benchmark.

\begin{table}[t]
    \centering
    \caption{Effect of the number of Transformer blocks on reconstruction (Wan~2.1). Increasing number of blocks consistently improves reconstruction quality, with the 10-blocks model achieving the best results.}
    \label{tab:layer_ablation_recon}
    \small
    \resizebox{\textwidth}{!}{%
        \renewcommand{\arraystretch}{1.15}
        \begin{tabular}{cccccccccc}
            \toprule
            & \multicolumn{3}{c}{\textbf{Inter4K (test)}} & \multicolumn{3}{c}{\textbf{WebVid}} & \multicolumn{3}{c}{\textbf{Large Motion}} \\
            \cmidrule(lr){2-4} \cmidrule(lr){5-7} \cmidrule(lr){8-10}
            \textbf{\# Blocks} & \textbf{PSNR$\uparrow$} & \textbf{SSIM\,(\%)$\uparrow$} & \textbf{LPIPS$\downarrow$} & \textbf{PSNR$\uparrow$} & \textbf{SSIM\,(\%)$\uparrow$} & \textbf{LPIPS$\downarrow$} & \textbf{PSNR$\uparrow$} & \textbf{SSIM\,(\%)$\uparrow$} & \textbf{LPIPS$\downarrow$} \\
            \midrule
            3  & 34.3 & 94.4 & 0.035 & 33.2 & 92.3 & 0.038 & 31.1 & 89.9 & 0.055  \\
            5  & 34.6 & 94.6 & 0.034 & 33.4 & 92.6 & 0.038 & 31.3 & 90.2 & 0.055  \\
            7  & 34.5 & 94.7 & 0.032 & 33.3 & 92.3 & 0.037 & 31.2 & 90.1 & 0.054  \\
            10 & \textbf{34.9} & \textbf{94.9} & \textbf{0.031} & \textbf{33.5} & \textbf{92.6} & \textbf{0.037} & \textbf{31.4} & \textbf{90.4} & \textbf{0.053}  \\
            \bottomrule
        \end{tabular}
    }
\end{table}

\section{Latent Token Dropout and Two-Stage Curriculum}
\label{sec:supp_dropout_curriculum}

\noindent\textbf{Setup.}
Table~\ref{tab:ablation_dropout_curriculum} reports the two training-time ablations summarized in the main paper. Both use the Wan~2.1 backbone with 3 Transformer blocks and are evaluated on Inter4K. The top block sweeps the maximum latent-token dropout probability ($0.0$, $0.3$, $0.7$), with each variant trained for $10{,}000$ steps under the two-stage curriculum. The bottom block compares one-stage training (the first curriculum stage only) against the full two-stage curriculum at fixed dropout $0.7$. We additionally report metrics over both the entire reconstructed video (\emph{Overall}) and the reference frame alone (\emph{Reference Frame}) to show whether gains come primarily from regenerating the reference or from propagating reference information to the rest of the video.

\noindent\textbf{Latent token dropout.}
Increasing the maximum dropout from $0.0$ to $0.7$ improves overall PSNR by $+2.71$\,dB and the reference-frame PSNR by $+3.85$\,dB. The reference-frame gain is larger than the overall gain, indicating that with dropout the decoder learns to lean on the reference token when the corresponding latent is missing rather than producing degraded content. Qualitative reconstructions in Figure~\ref{fig:latent_dropout_qualitative} show the same trend: at dropout $0.7$, reconstructions exhibit sharper edges and more faithful textures, while dropout $0.0$ produces blurrier results.

\noindent\textbf{Two-stage curriculum.}
At fixed dropout $0.7$, the two-stage curriculum improves overall PSNR by $+3.80$\,dB and reference-frame PSNR by $+4.64$\,dB compared to one-stage training. The much larger gap on reference-frame PSNR ($+4.64$\,dB vs.\ $+3.80$\,dB overall) suggests that the second stage primarily refines reference-token decoding to longer temporal contexts, supporting the choice to first train on short 5-frame clips and then fine-tune on 17-frame sequences.

\begin{table}[t]
    \centering
    \caption{Ablations on the Wan~2.1 backbone with 3 Transformer blocks, evaluated on Inter4K. \emph{Top}: effect of maximum latent token dropout probability (with two-stage curriculum). \emph{Bottom}: effect of two-stage curriculum training (with dropout 0.7).}
    \label{tab:ablation_dropout_curriculum}
    \small
    \setlength\tabcolsep{4pt}
    \renewcommand{\arraystretch}{1.15}
    \begin{tabular}{ll|ccc|ccc}
        \toprule
        & & \multicolumn{3}{c|}{Overall} & \multicolumn{3}{c}{Reference Frame} \\
        & Setting & PSNR $\uparrow$ & SSIM\,(\%) $\uparrow$ & LPIPS $\downarrow$ & PSNR $\uparrow$ & SSIM\,(\%) $\uparrow$ & LPIPS $\downarrow$ \\
        \midrule
        \multirow{3}{*}{Dropout}
            & 0.0 & 27.4 & 88.3 & 0.083 & 27.3 & 88.6 & 0.082 \\
            & 0.3 & 27.5 & 86.7 & 0.077 & 27.6 & 87.1 & 0.075 \\
            & 0.7 & \textbf{30.1} & \textbf{90.1} & \textbf{0.060} & \textbf{31.1} & \textbf{92.1} & \textbf{0.054} \\
        \midrule
        \multirow{2}{*}{Curriculum}
            & One-stage & 30.5 & 89.7 & 0.058 & 34.9 & 95.7 & 0.018 \\
            & Two-stage & \textbf{34.3} & \textbf{94.4} & \textbf{0.035} & \textbf{39.6} & \textbf{98.2} & \textbf{0.008} \\
        \bottomrule
    \end{tabular}
\end{table}

\begin{figure}[h]
    \centering
    \setlength{\tabcolsep}{1pt}
    \begin{tabular}{cccc}
        \includegraphics[width=0.24\textwidth]{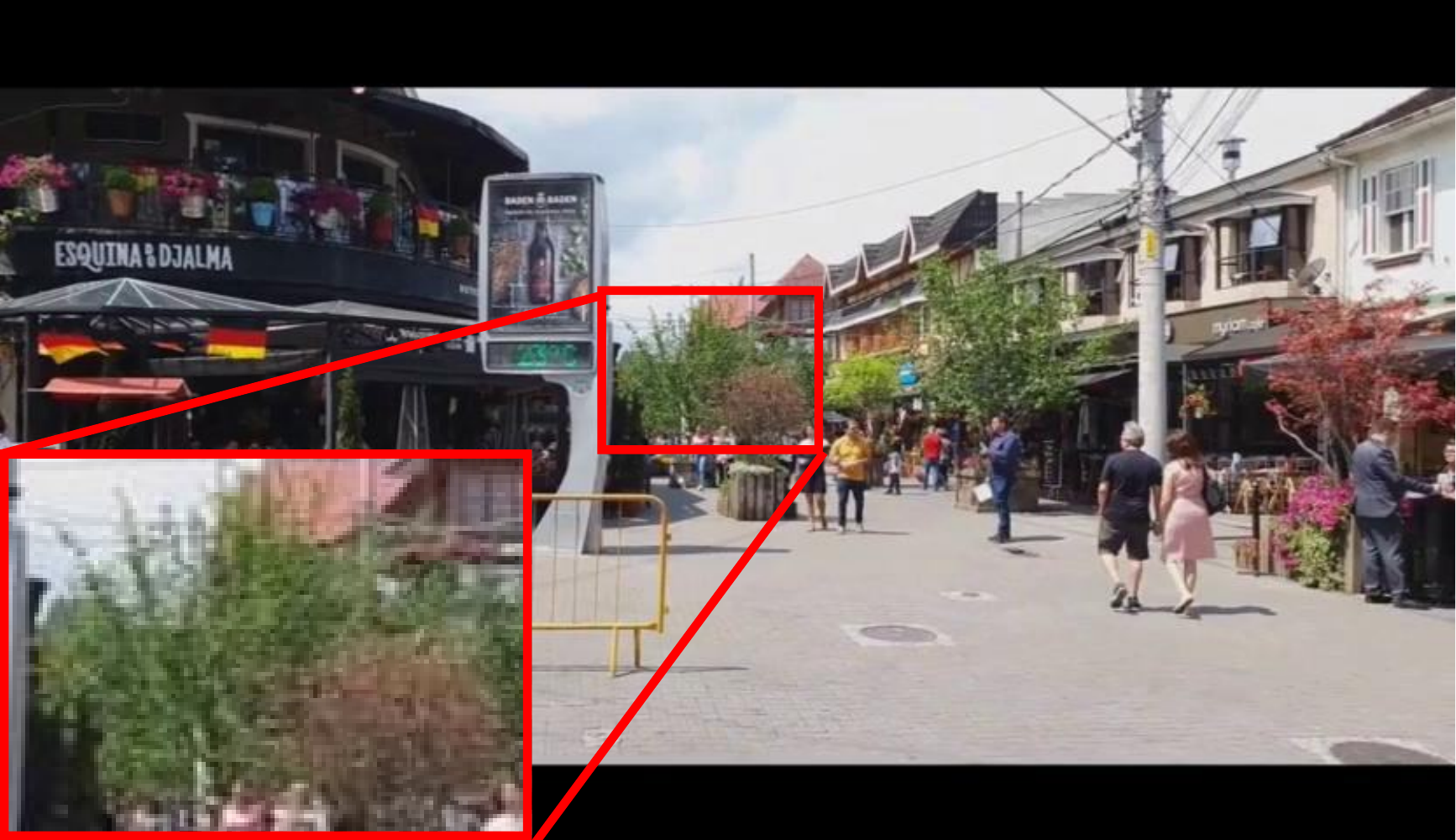} &
        \includegraphics[width=0.24\textwidth]{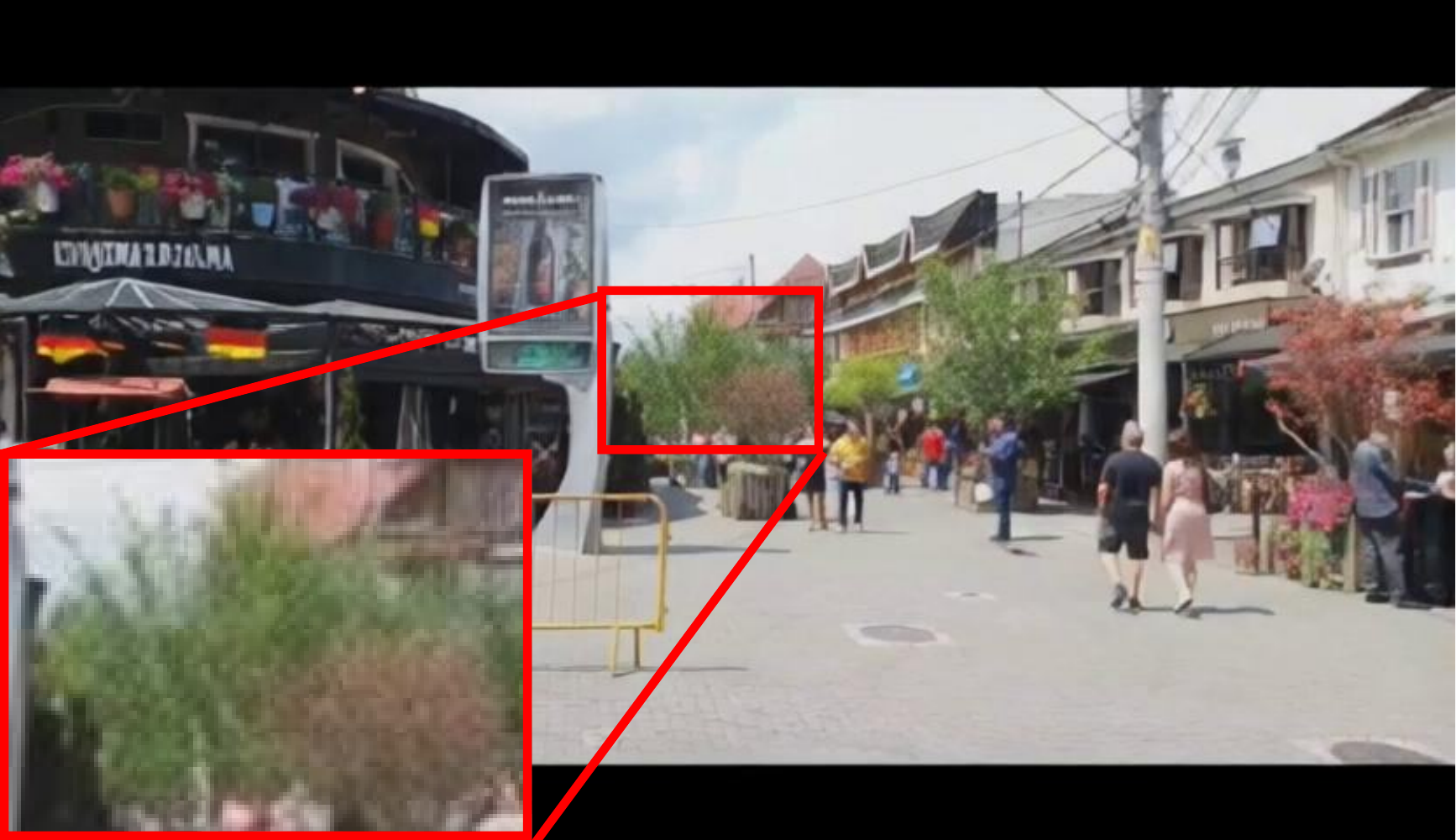} &
        \includegraphics[width=0.24\textwidth]{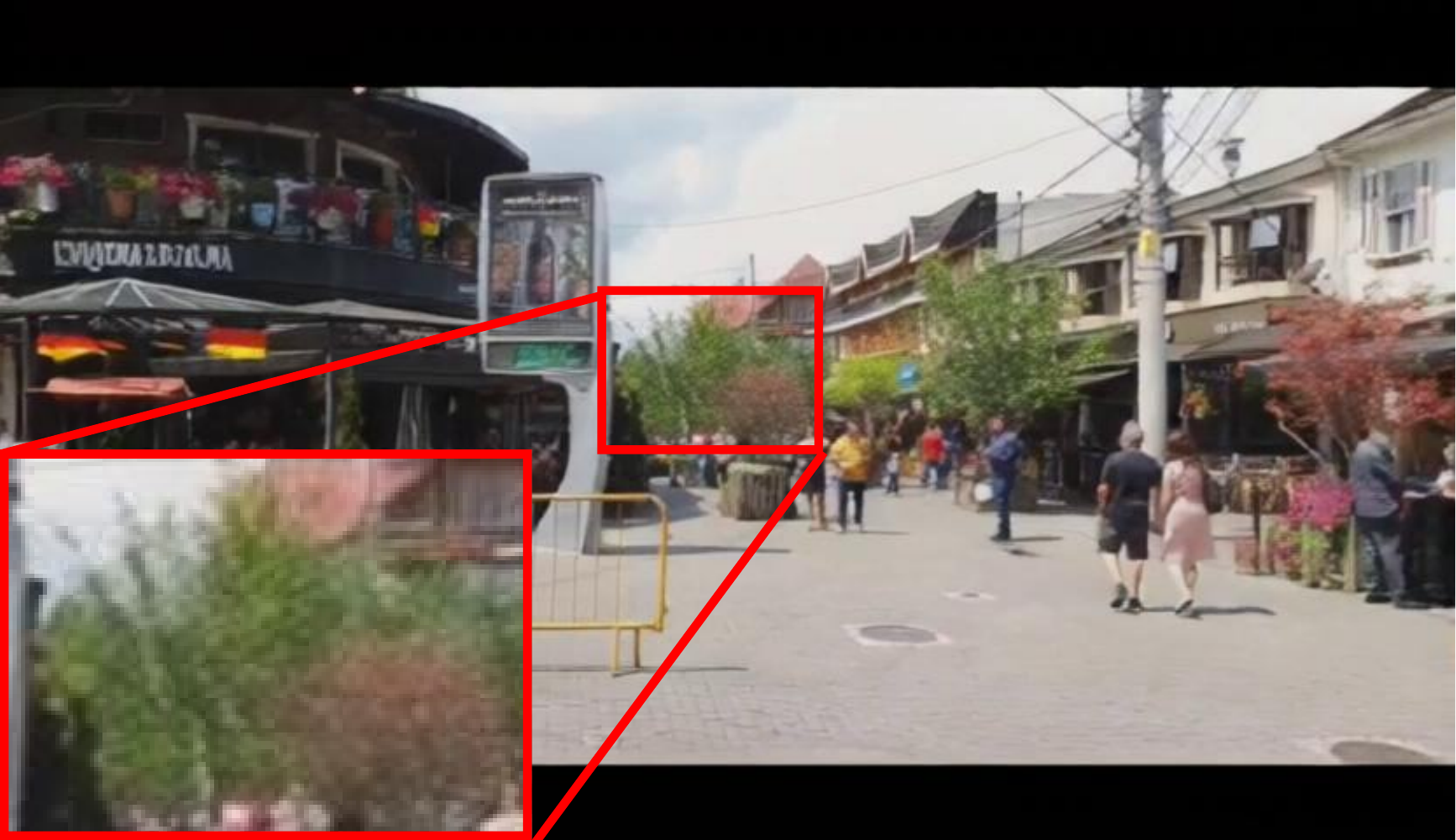} &
        \includegraphics[width=0.24\textwidth]{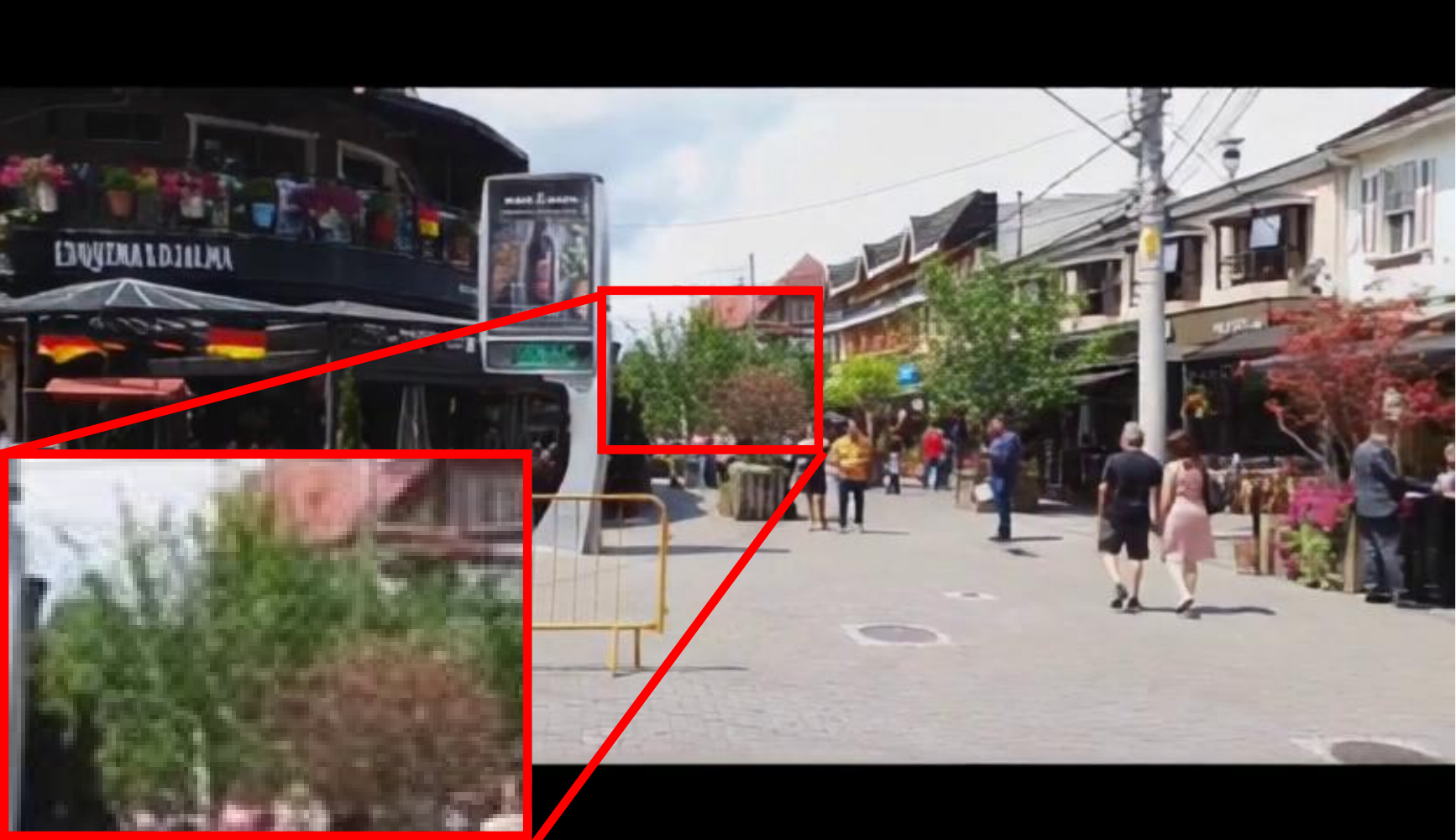} \\[2pt]
        \small{GT} & \small{Dropout 0.0} & \small{Dropout 0.3} & \small{Dropout 0.7} \\
    \end{tabular}
    \caption{\textbf{Qualitative comparison of different dropout rates} on Wan~2.1. Higher dropout encourages the decoder to rely more on reference features, leading to sharper and more detailed reconstructions.}
    \label{fig:latent_dropout_qualitative}
\end{figure}

\section{Alternative Reference Injection Strategy}
\label{sec:supp_controlnet}

We compare our attention-based reference injection with a ControlNet-style~\cite{zhang2023adding} alternative that injects the reference image through a parallel encoder branch, adding residual features to the decoder at each stage. We identify two fundamental limitations of this approach:

\begin{enumerate}
    \item \emph{No temporal reasoning.} ControlNet operates via spatial addition of residual features, applying the reference signal identically to every frame at the same spatial position. Unlike joint attention, it lacks a mechanism to modulate the reference contribution based on temporal context. The decoder cannot distinguish whether a frame is temporally close to or far from the reference, nor can it selectively attend to different reference regions for different frames.
    \item \emph{Suboptimal convergence.} The ControlNet variant converges quickly to a suboptimal reconstruction quality and plateaus. We hypothesize that the spatially rigid injection forces the model into a local minimum where it learns to uniformly blend the reference signal rather than adaptively retrieving fine-grained details.
\end{enumerate}

\begin{figure}[h]
    \centering
    \setlength{\tabcolsep}{1pt}
    \renewcommand{\arraystretch}{0.5}
    \begin{tabular}{cccc}
        \rotatebox{90}{\small\hspace{18pt}Input} &
        \includegraphics[width=0.3\textwidth]{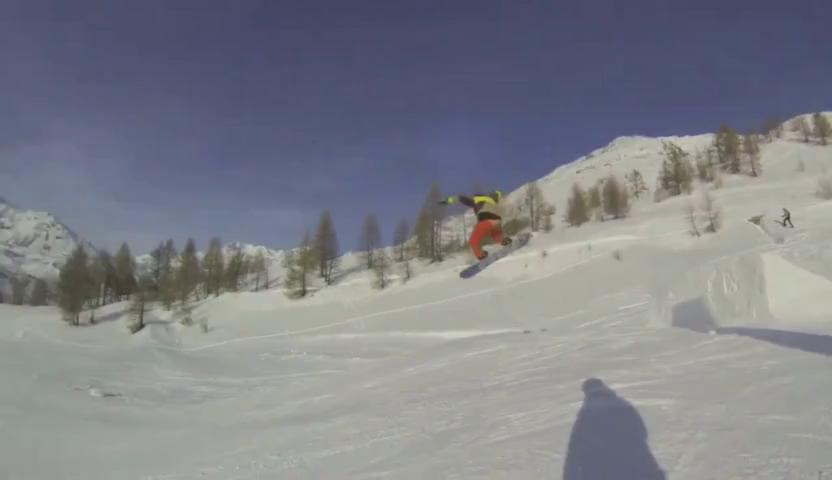} &
        \includegraphics[width=0.3\textwidth]{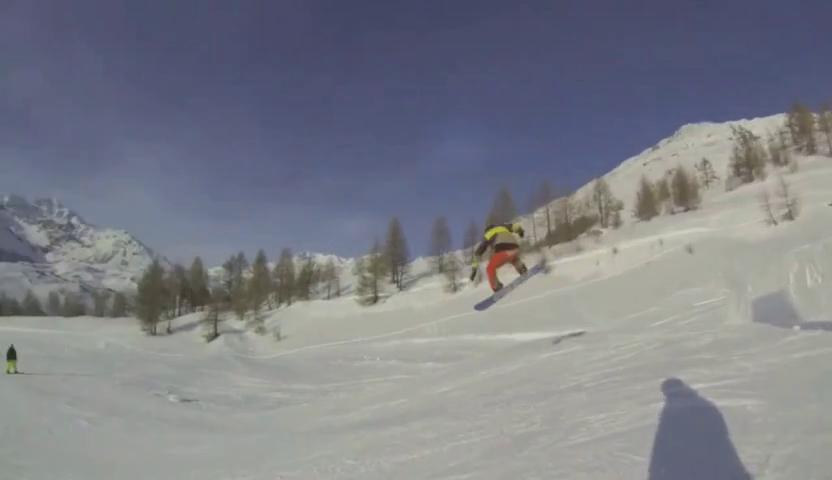} &
        \includegraphics[width=0.3\textwidth]{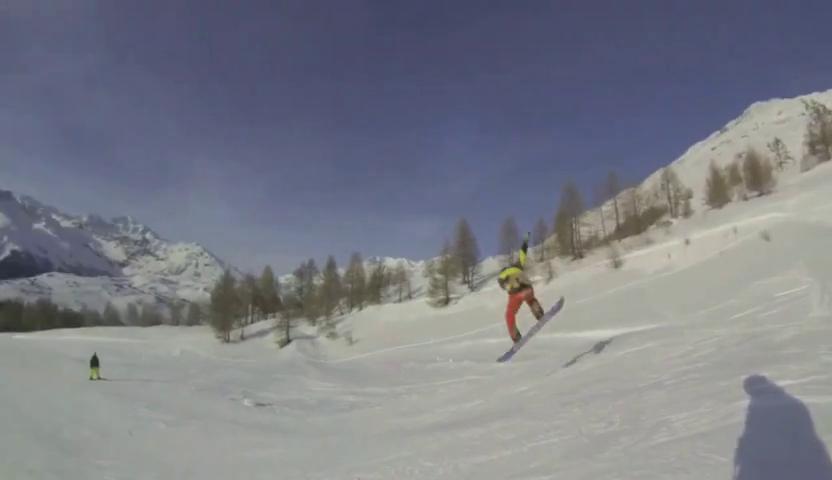} \\
        \rotatebox{90}{\small\hspace{6pt}ControlNet} &
        \includegraphics[width=0.3\textwidth]{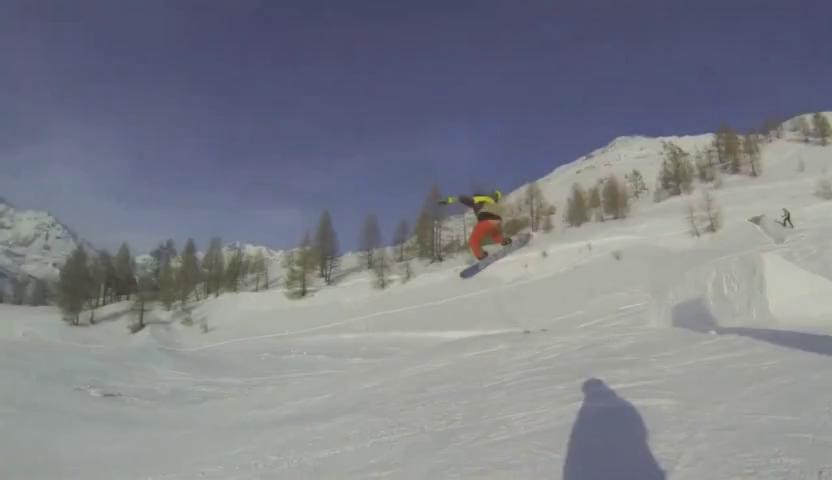} &
        \includegraphics[width=0.3\textwidth]{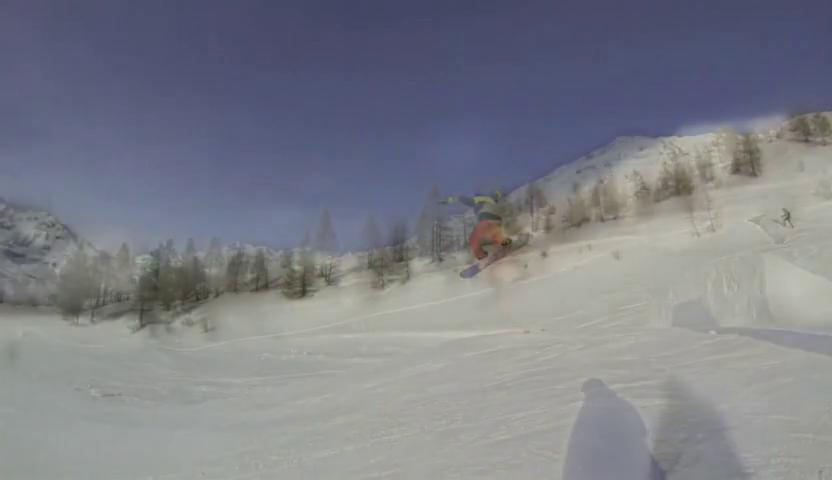} &
        \includegraphics[width=0.3\textwidth]{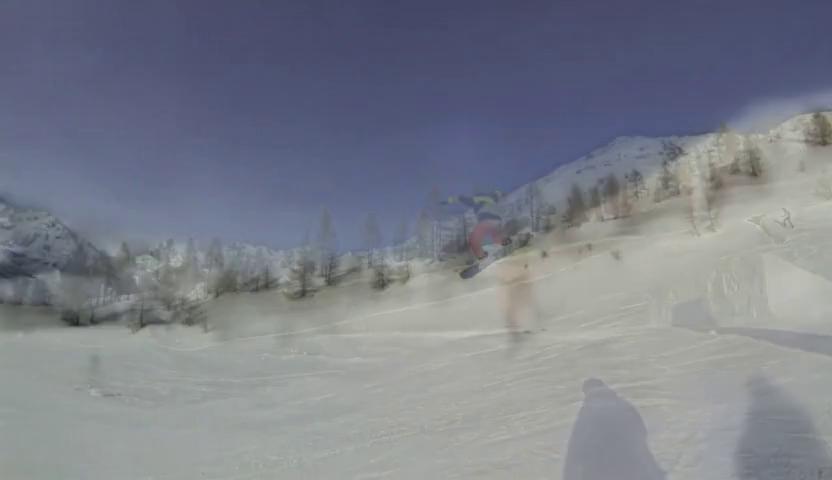} \\
        \rotatebox{90}{\small\hspace{20pt}Ours} &
        \includegraphics[width=0.3\textwidth]{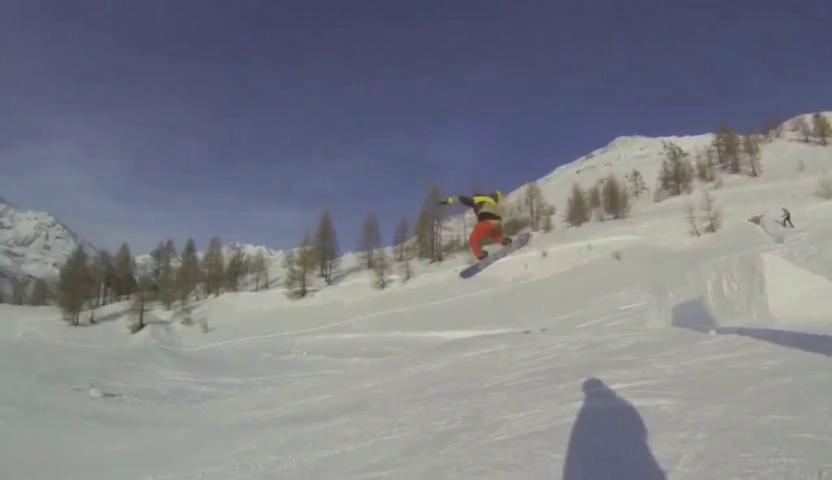} &
        \includegraphics[width=0.3\textwidth]{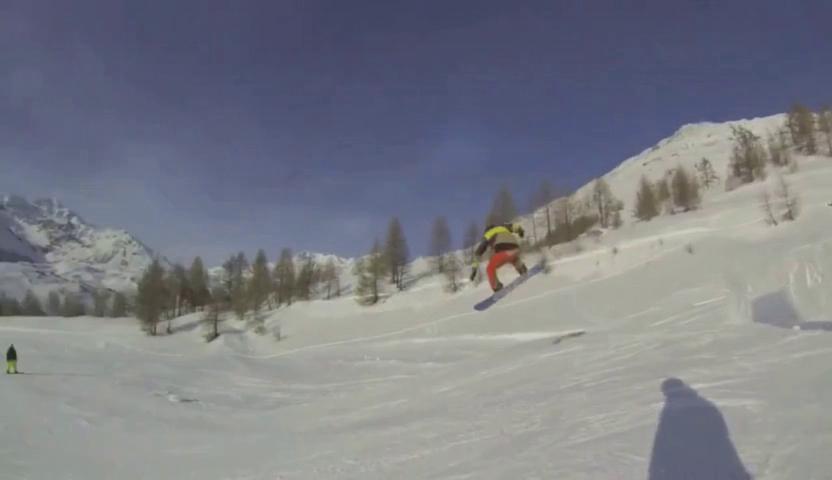} &
        \includegraphics[width=0.3\textwidth]{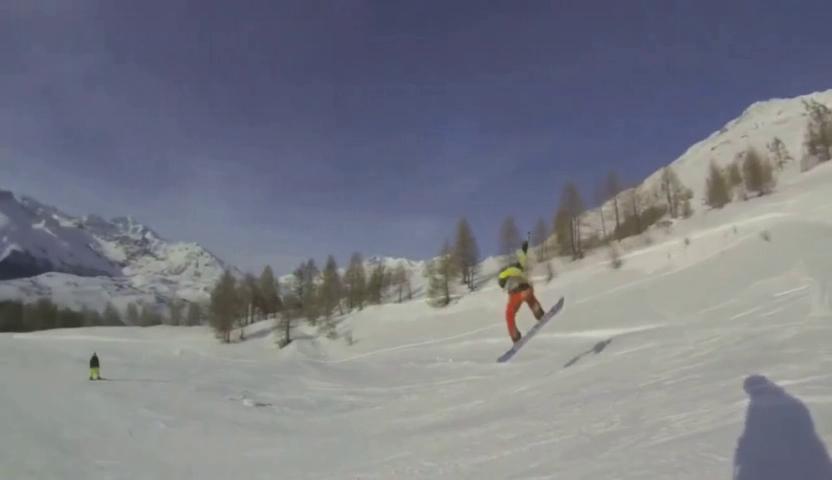} \\[2pt]
        & \small{First frame} & \small{Middle frame} & \small{Last frame} \\
    \end{tabular}
    \caption{\textbf{Qualitative comparison of alternative reference injection methods.} Our attention-based reference injection produces sharper and more temporally consistent reconstructions, while the ControlNet-style approach introduces ghosting artifacts and temporal inconsistency.}
    \label{fig:controlnet_comparison}
\end{figure}

\section{Stage-Specific Reference Injection}
\label{sec:supp_stage_injection}

To evaluate the effect of reference injection at different decoder
stages, we construct stage-specific variants of \model.
Starting from the same pretrained \model\ checkpoint, we keep
reference injection active at only one decoder stage and disable
it at all remaining stages. We then fine-tune each stage-specific
model under the same training settings. This setup allows us to
compare the reconstruction quality and decoding cost of injection
at individual stages with those of the full model, which injects
reference information at all three stages.

\begin{table}[t]
    \centering
    \caption{
        Reconstruction quality and per-chunk decoding latency
        for different reference injection stages.
        ``None'' denotes the original Wan~2.1 decoder.
        Best reconstruction results are shown in bold.
    }
    \label{tab:stage_injection}
    \small
    \setlength{\tabcolsep}{6pt}
    \renewcommand{\arraystretch}{1.15}
    \begin{tabular}{lcccc}
        \toprule
        Injection stages
        & Latency/chunk $\downarrow$
        & PSNR $\uparrow$
        & SSIM $\uparrow$
        & LPIPS $\downarrow$ \\
        \midrule
        None (Wan~2.1)
        & $239.2$\,ms & -- & -- & -- \\
        0
        & $264.1$\,ms & 31.5272 & 0.9093 & 0.0574 \\
        1
        & $279.9$\,ms & 31.6146 & 0.9103 & 0.0582 \\
        2
        & $329.3$\,ms & 32.4959 & 0.9223 & 0.0508 \\
        0, 1, 2 (ours)
        & $393.5$\,ms
        & \textbf{33.2817}
        & \textbf{0.9328}
        & \textbf{0.0414} \\
        \bottomrule
    \end{tabular}
\end{table}

As shown in Table~\ref{tab:stage_injection}, reference injection
at all three stages achieves the best reconstruction performance.
Compared with injection at stage~2 alone, the full model improves
PSNR by $0.7858$\,dB and SSIM by $0.0105$, while reducing LPIPS
by $0.0094$ ($18.5\%$). These results indicate that reference
conditioning across multiple decoding scales provides
complementary benefits.

\section{Code License}
\label{sec:supp_license}

Our work is built upon the HuggingFace Diffusers~\cite{von-platen-etal-2022-diffusers} library, which is licensed under the Apache License 2.0 (\url{https://github.com/huggingface/diffusers/blob/main/LICENSE}). VideoVAE\textbf{+}-related code is licensed under Attribution-NonCommercial-NoDerivatives 4.0 International, per the VideoVAE\textbf{+}~\cite{xing2025videovae+} license (\url{https://github.com/VideoVerses/VideoVAEPlus/blob/main/LICENSE}).

\section{More Qualitative Results}
We present more qualitative comparisons for both video reconstruction and image-to-video (I2V) generation.

\begin{figure*}[t]
\centering
\setlength{\tabcolsep}{1pt}
\renewcommand{\arraystretch}{0.5}
\begin{tabular}{cccc}
    \includegraphics[width=0.235\textwidth]{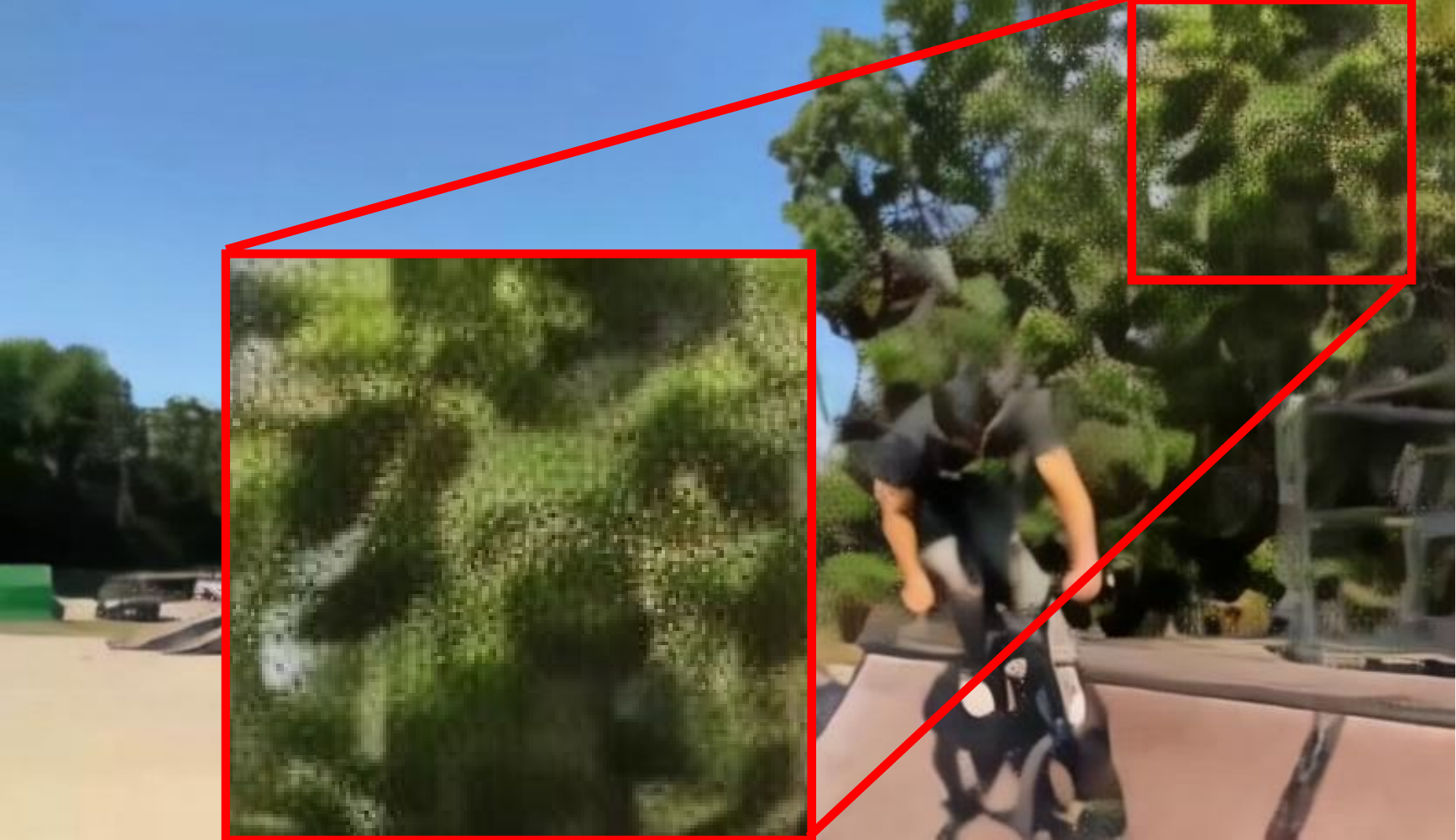} &
    \includegraphics[width=0.235\textwidth]{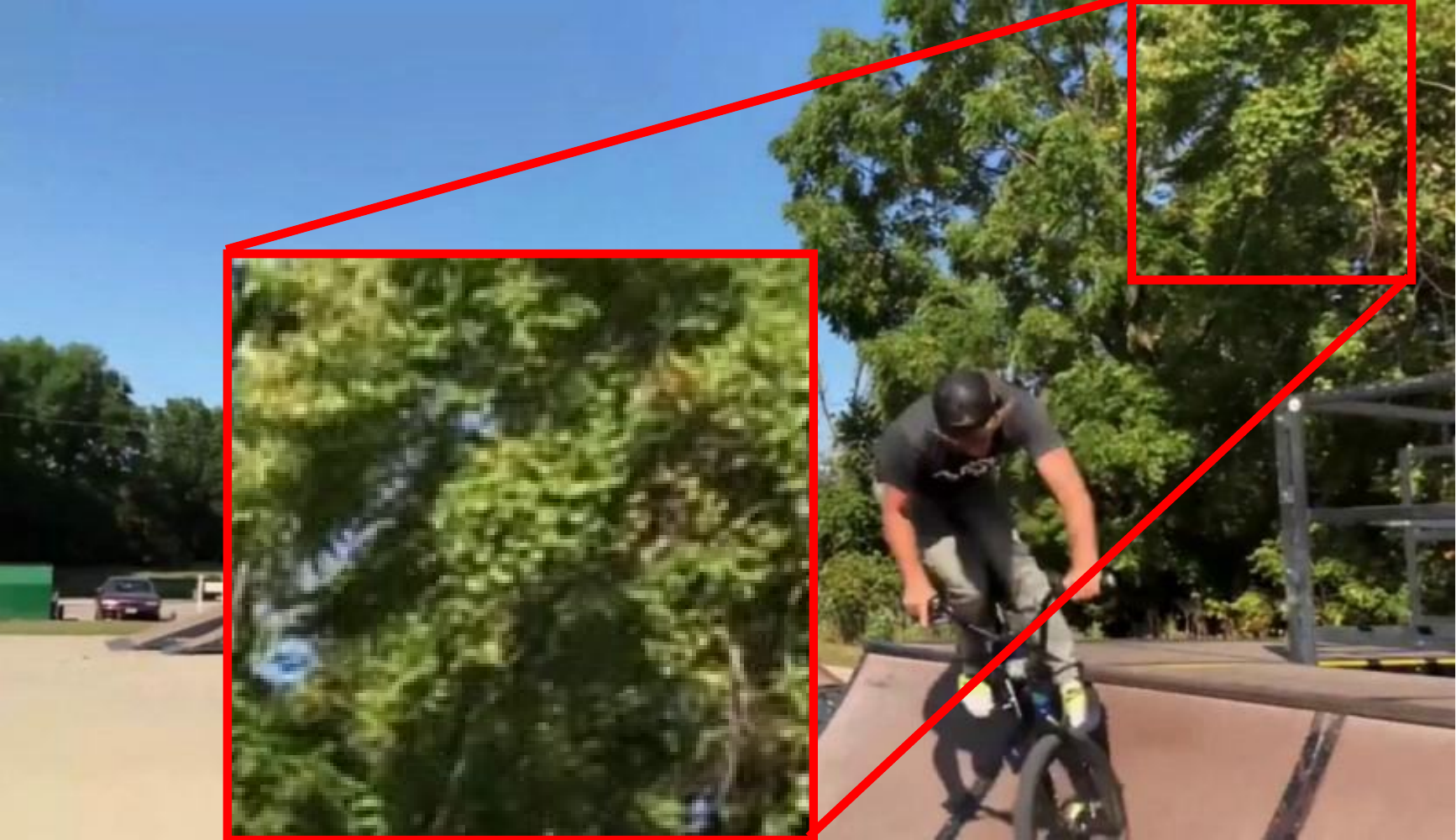} &
    \includegraphics[width=0.235\textwidth]{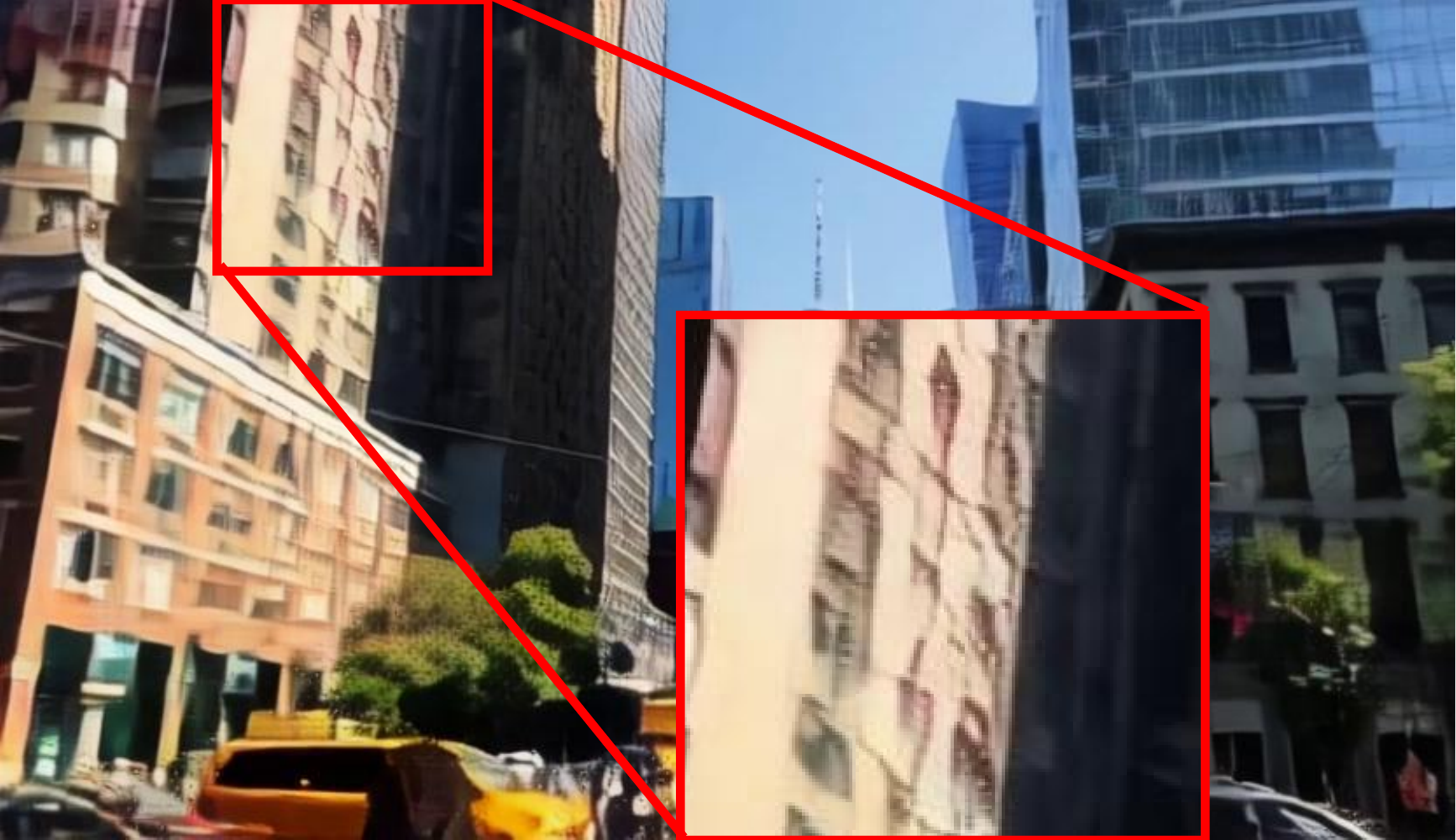} &
    \includegraphics[width=0.235\textwidth]{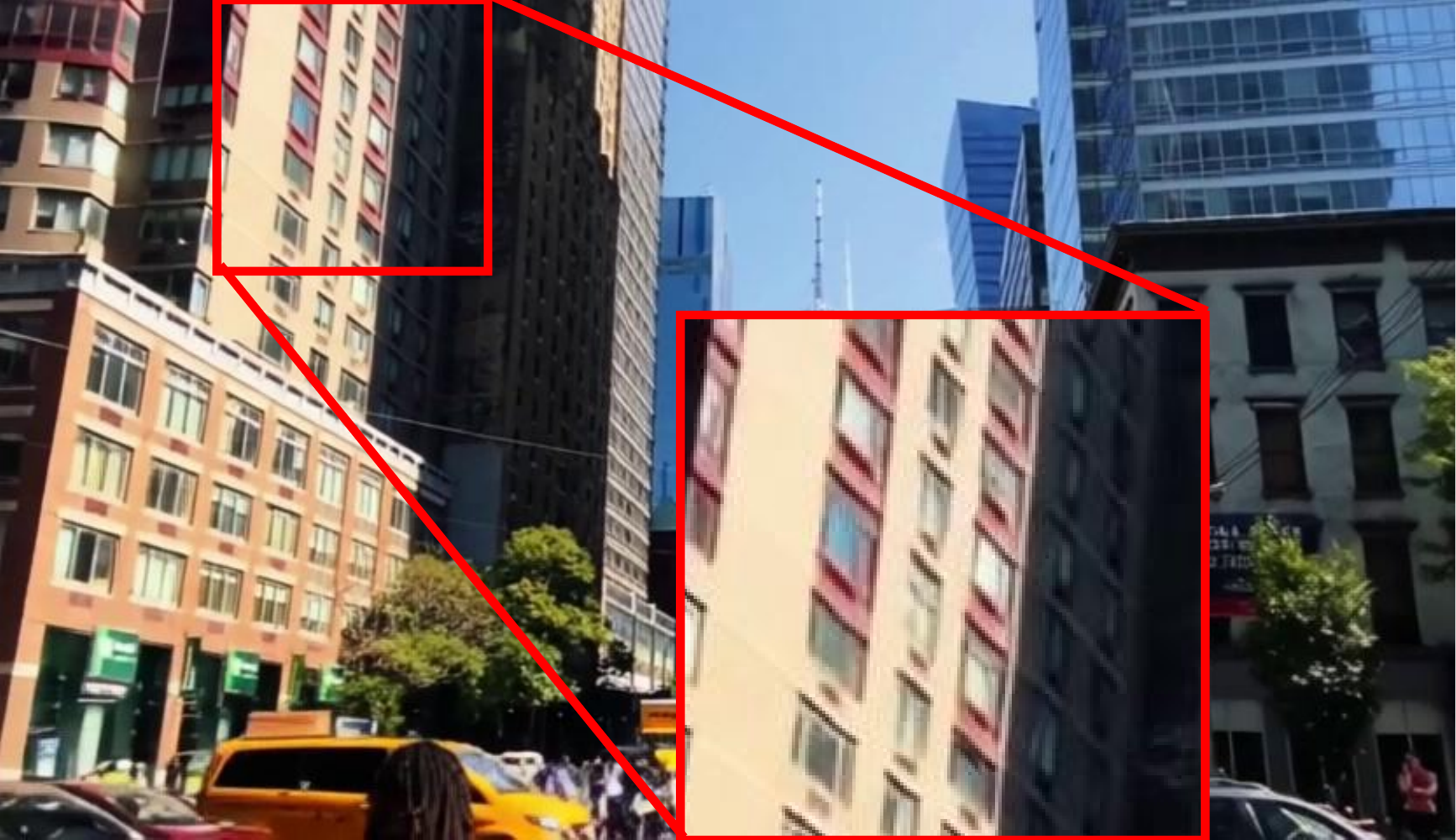} \\
    \includegraphics[width=0.235\textwidth]{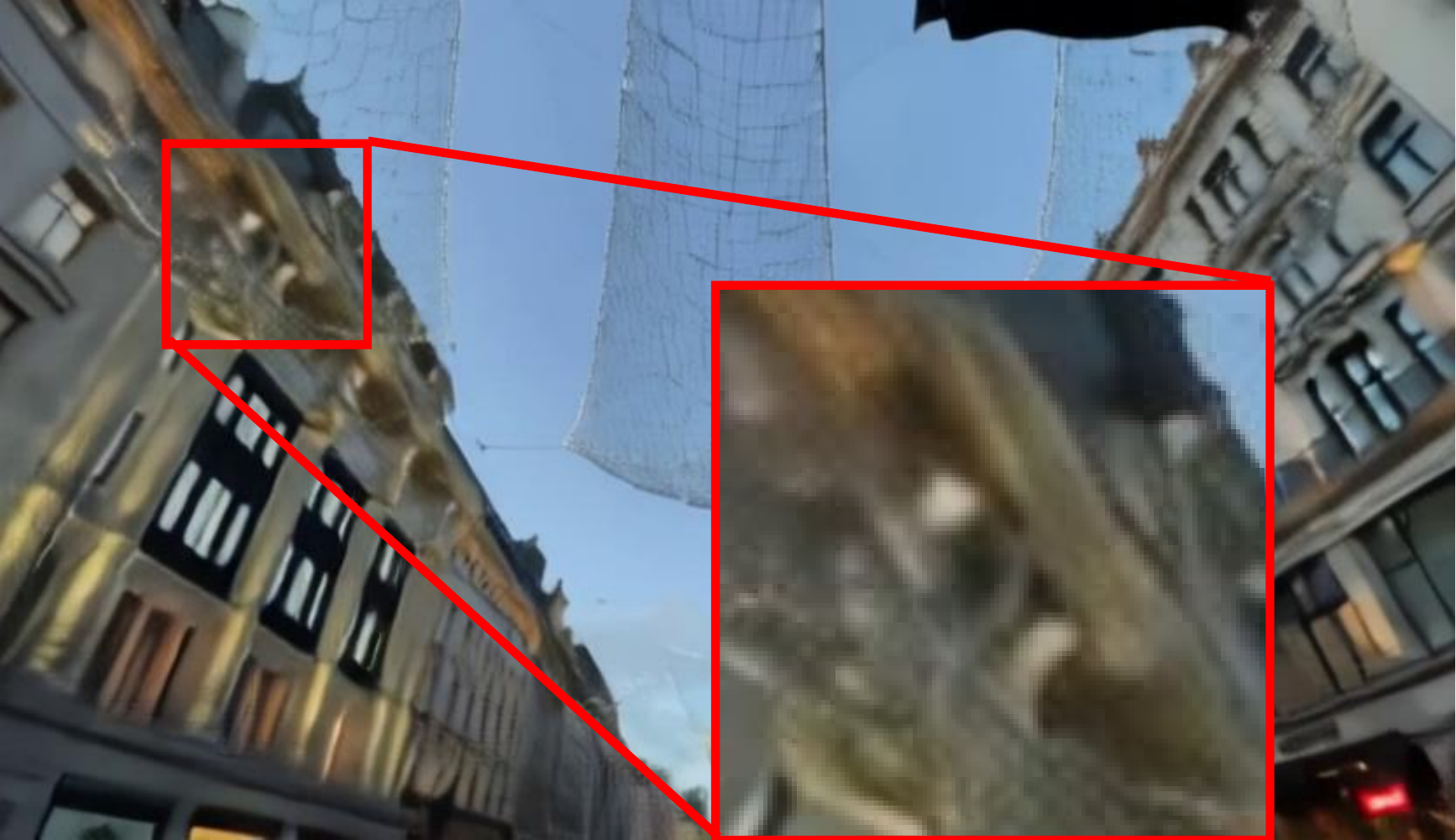} &
    \includegraphics[width=0.235\textwidth]{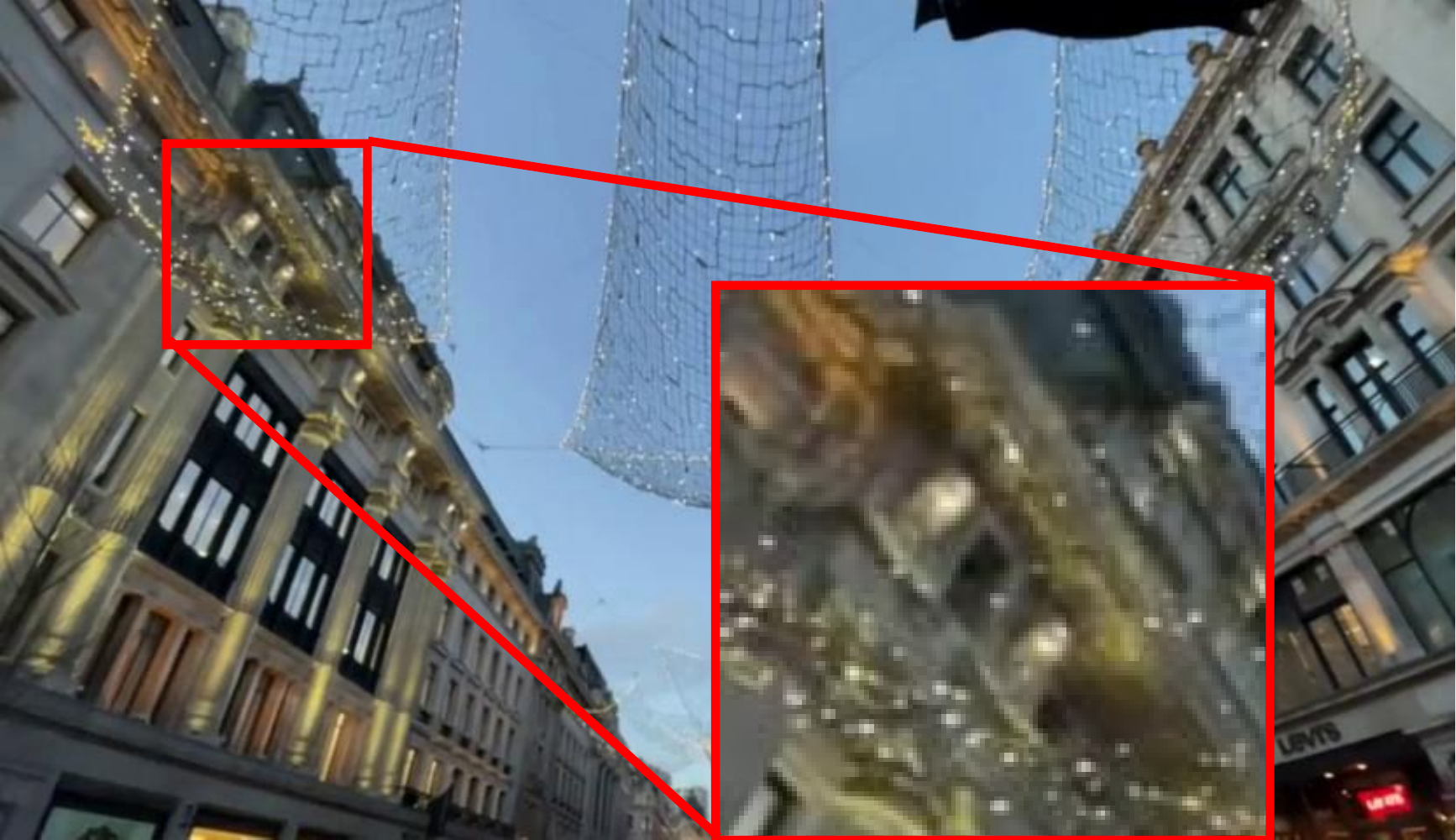} &
    \includegraphics[width=0.235\textwidth]{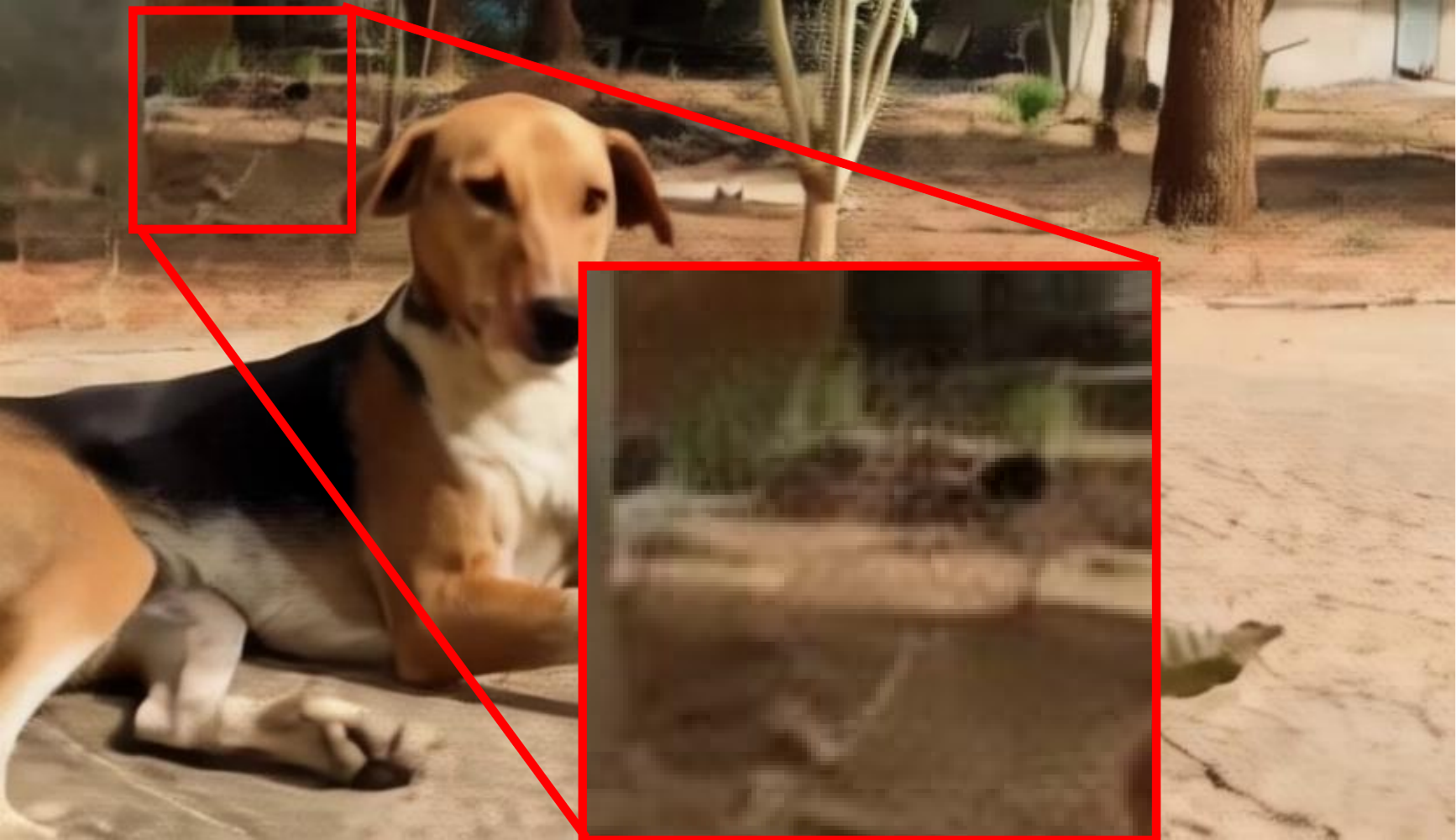} &
    \includegraphics[width=0.235\textwidth]{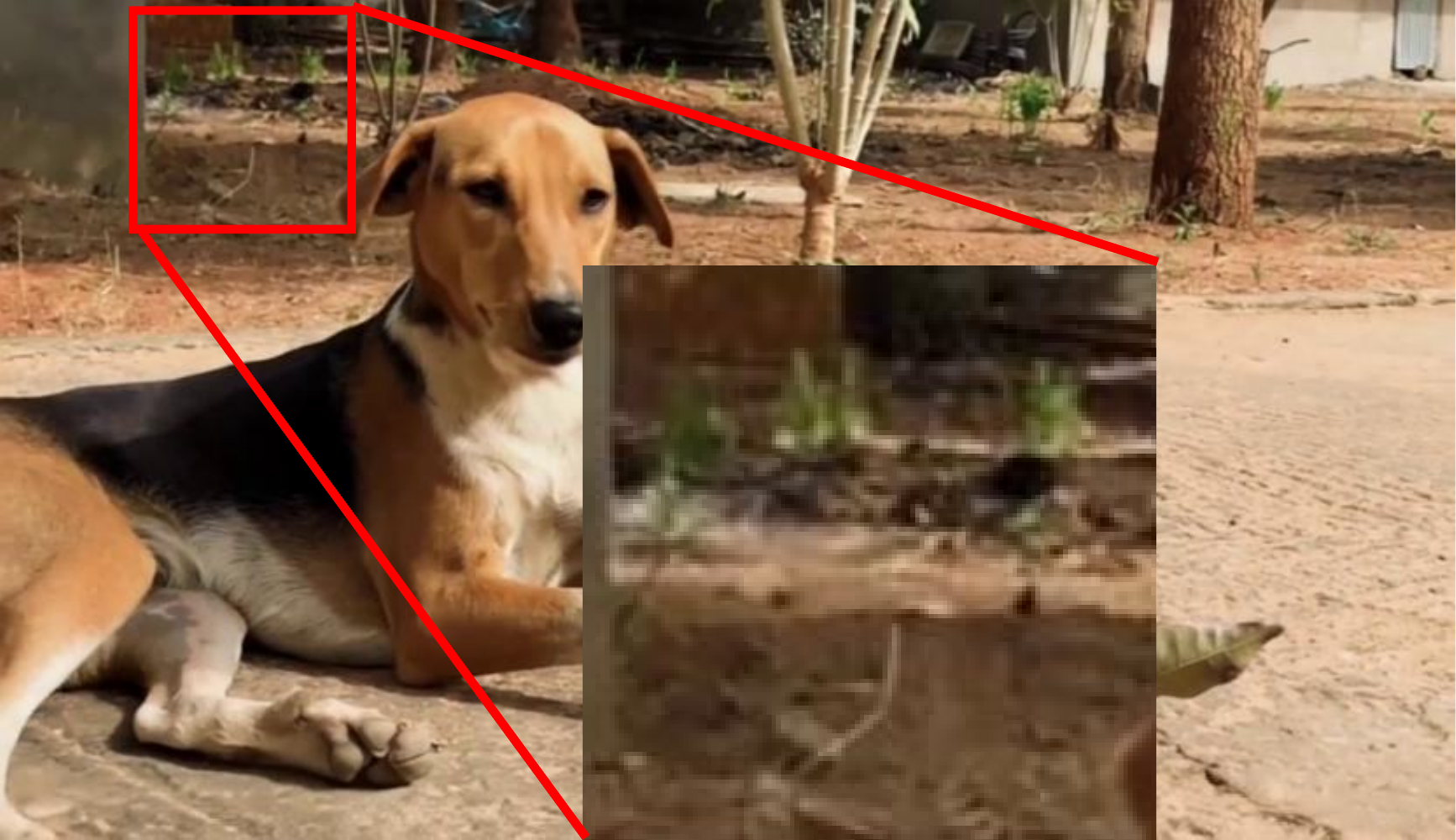} \\
    \includegraphics[width=0.235\textwidth]{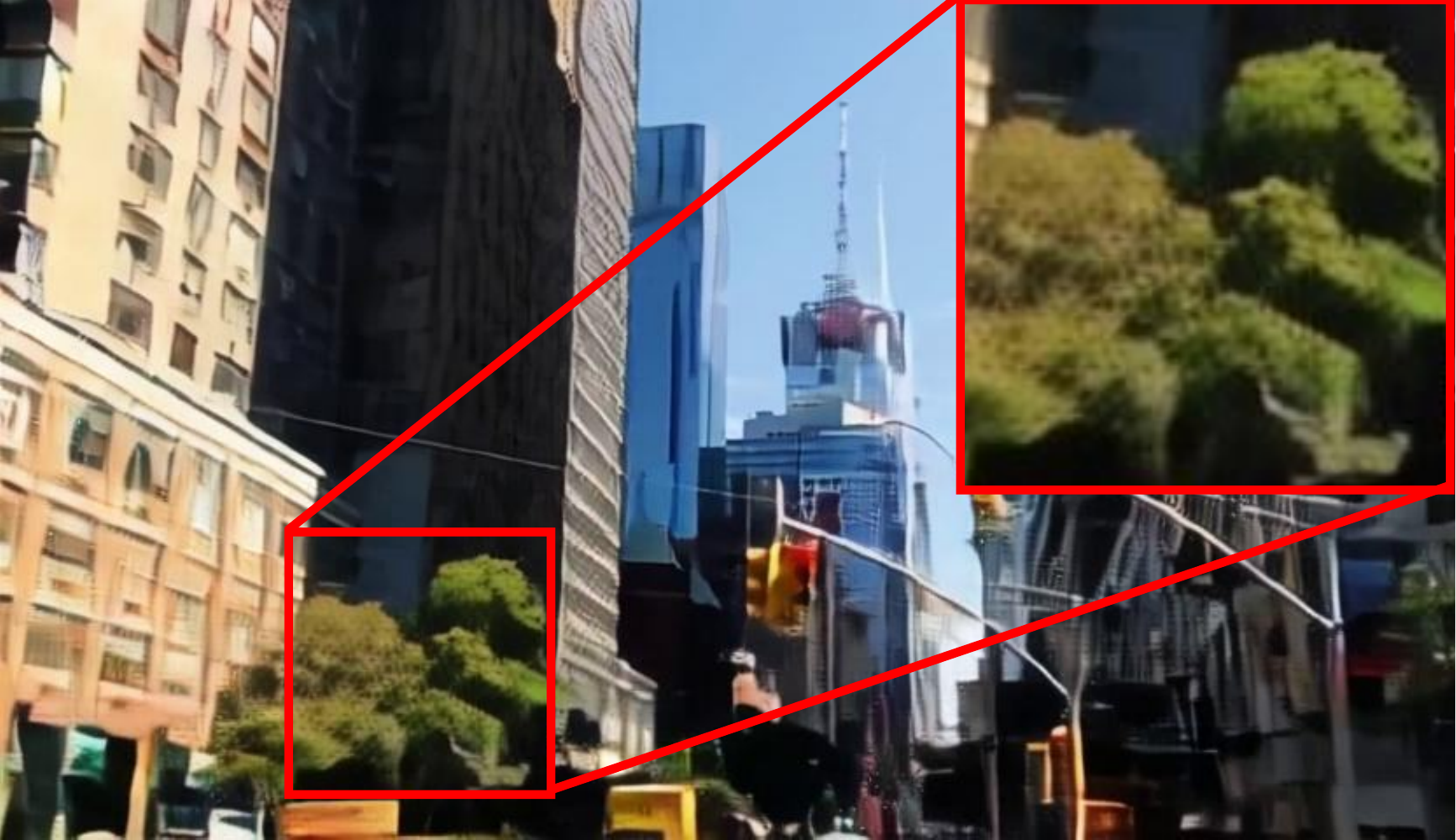} &
    \includegraphics[width=0.235\textwidth]{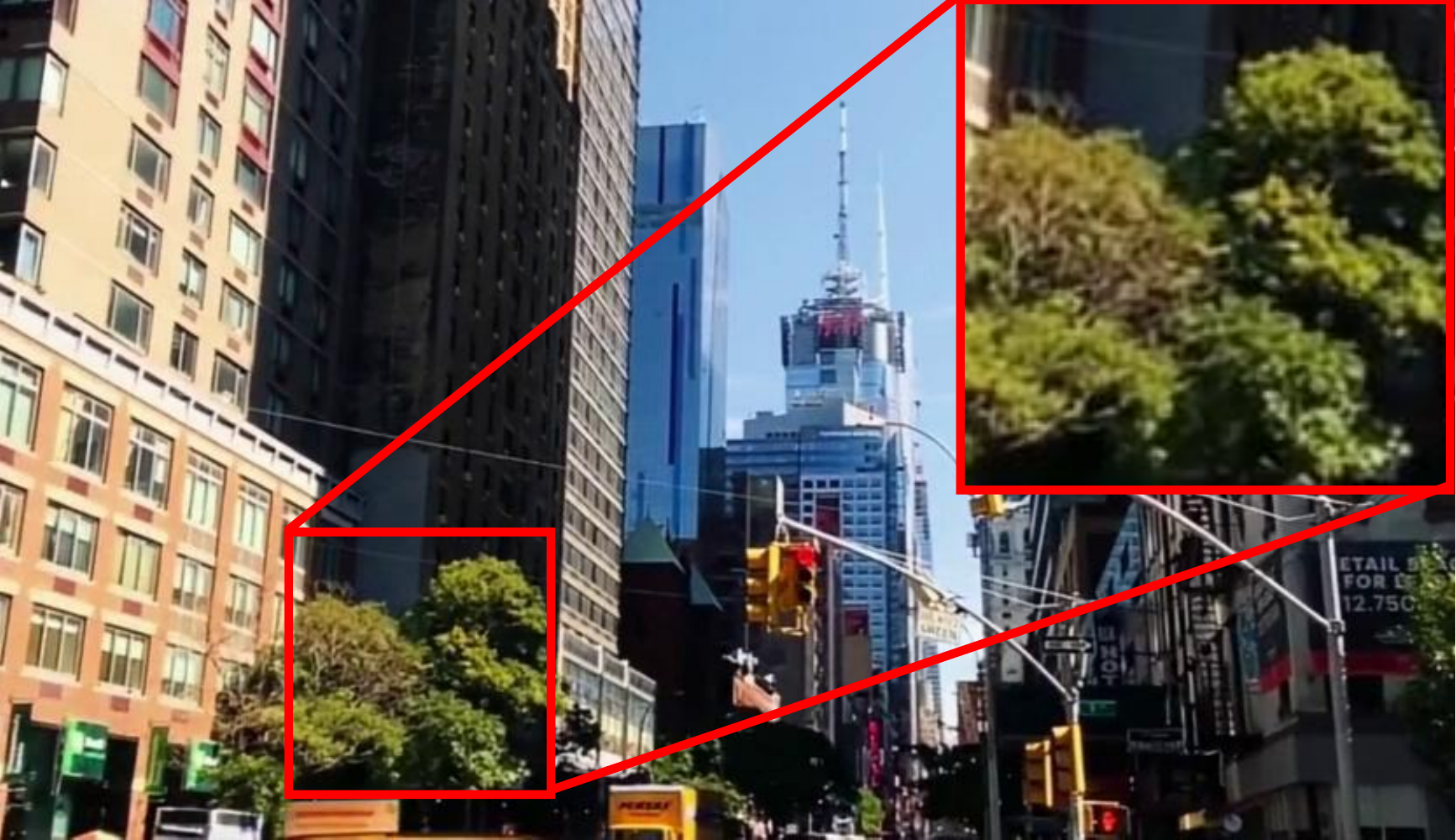} &
    \includegraphics[width=0.235\textwidth]{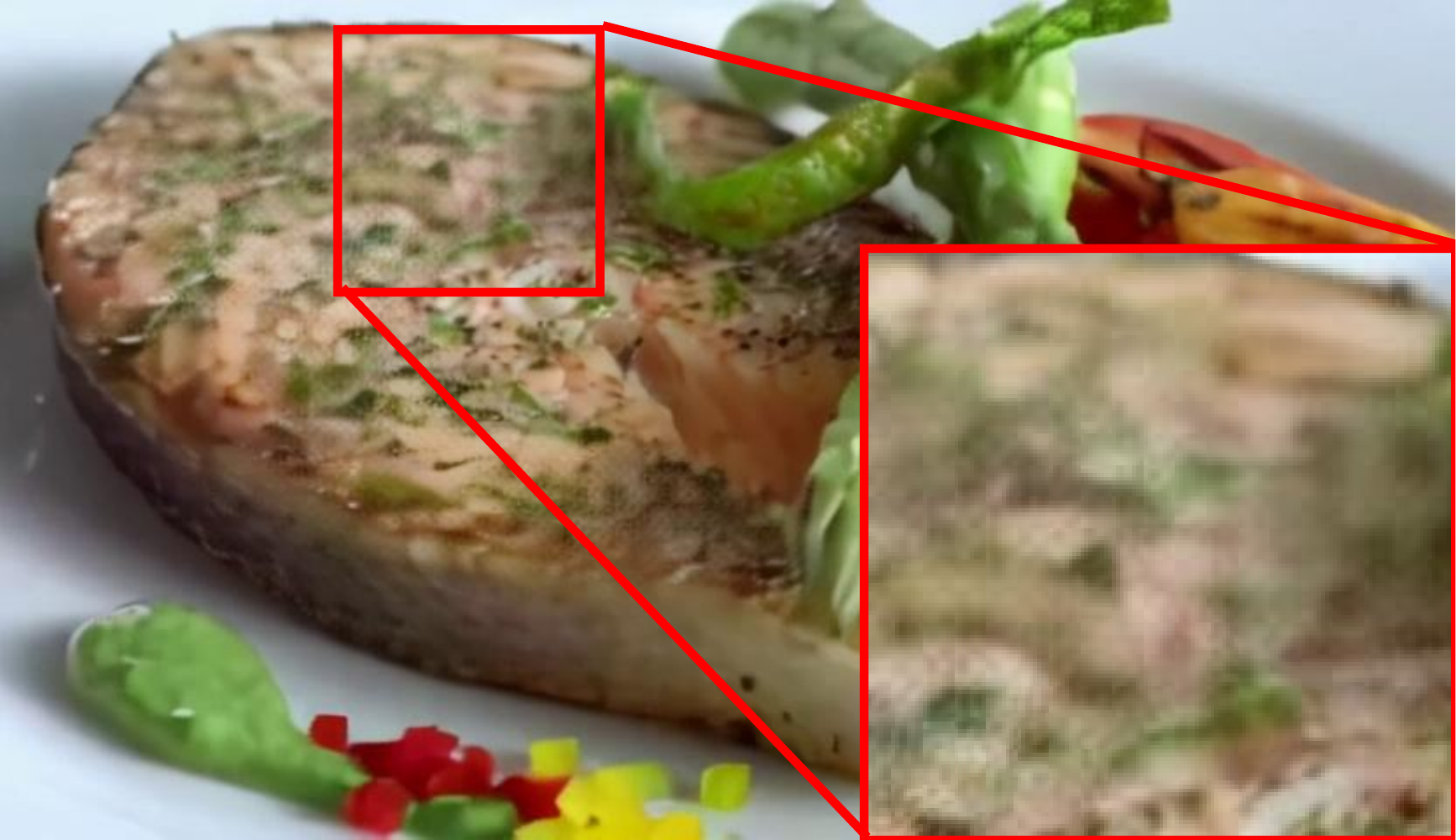} &
    \includegraphics[width=0.235\textwidth]{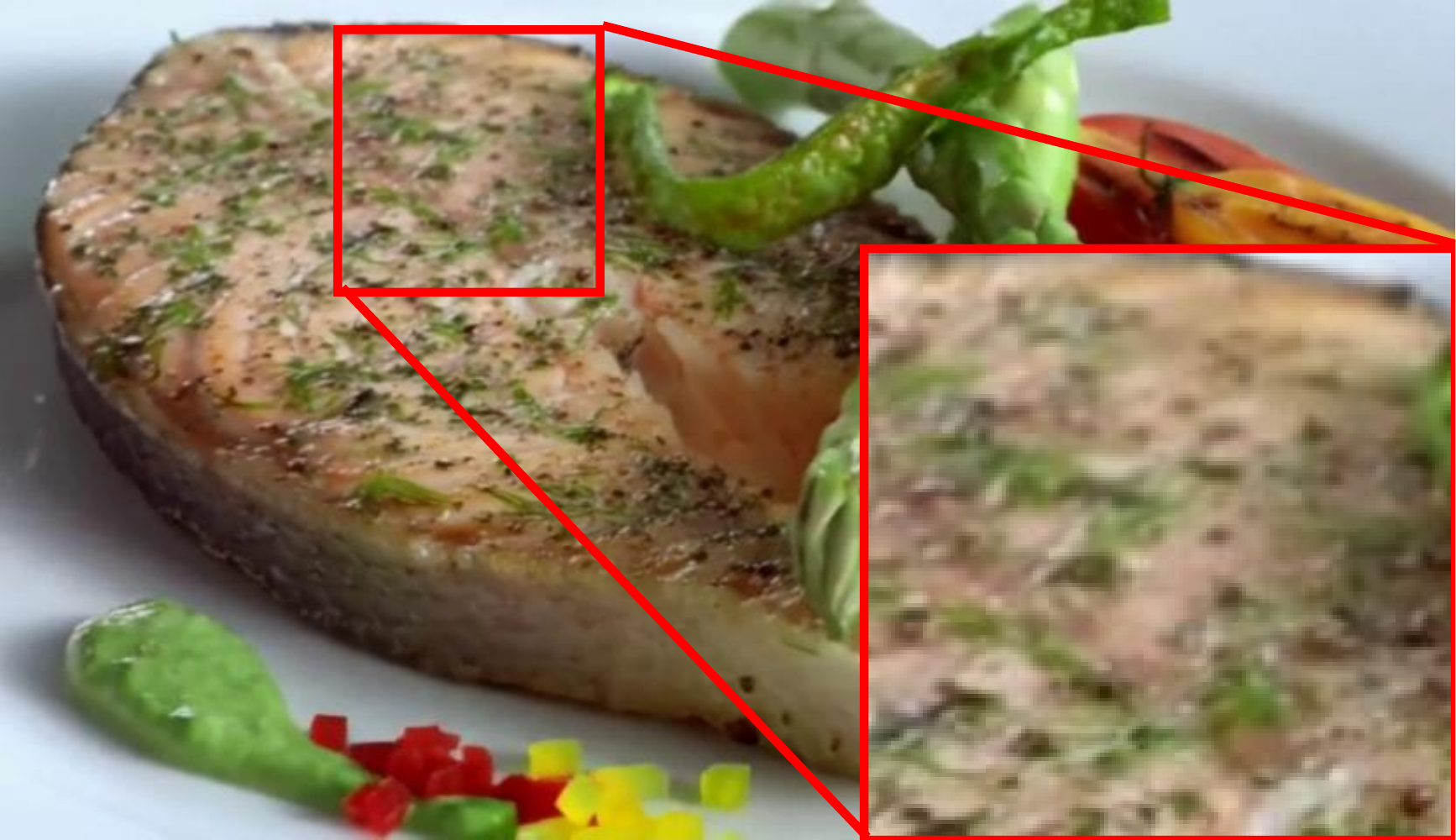} \\
    \includegraphics[width=0.235\textwidth]{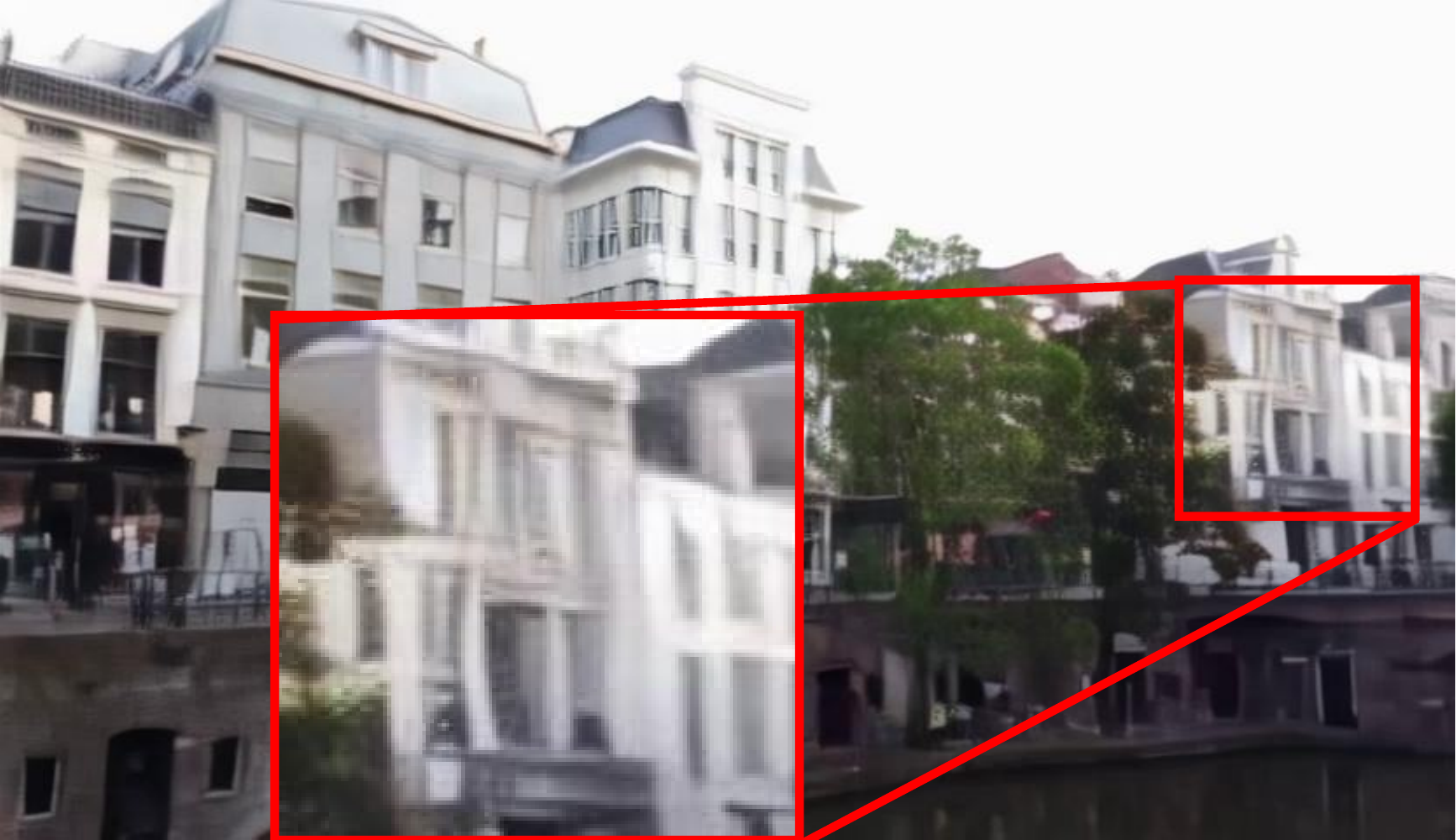} &
    \includegraphics[width=0.235\textwidth]{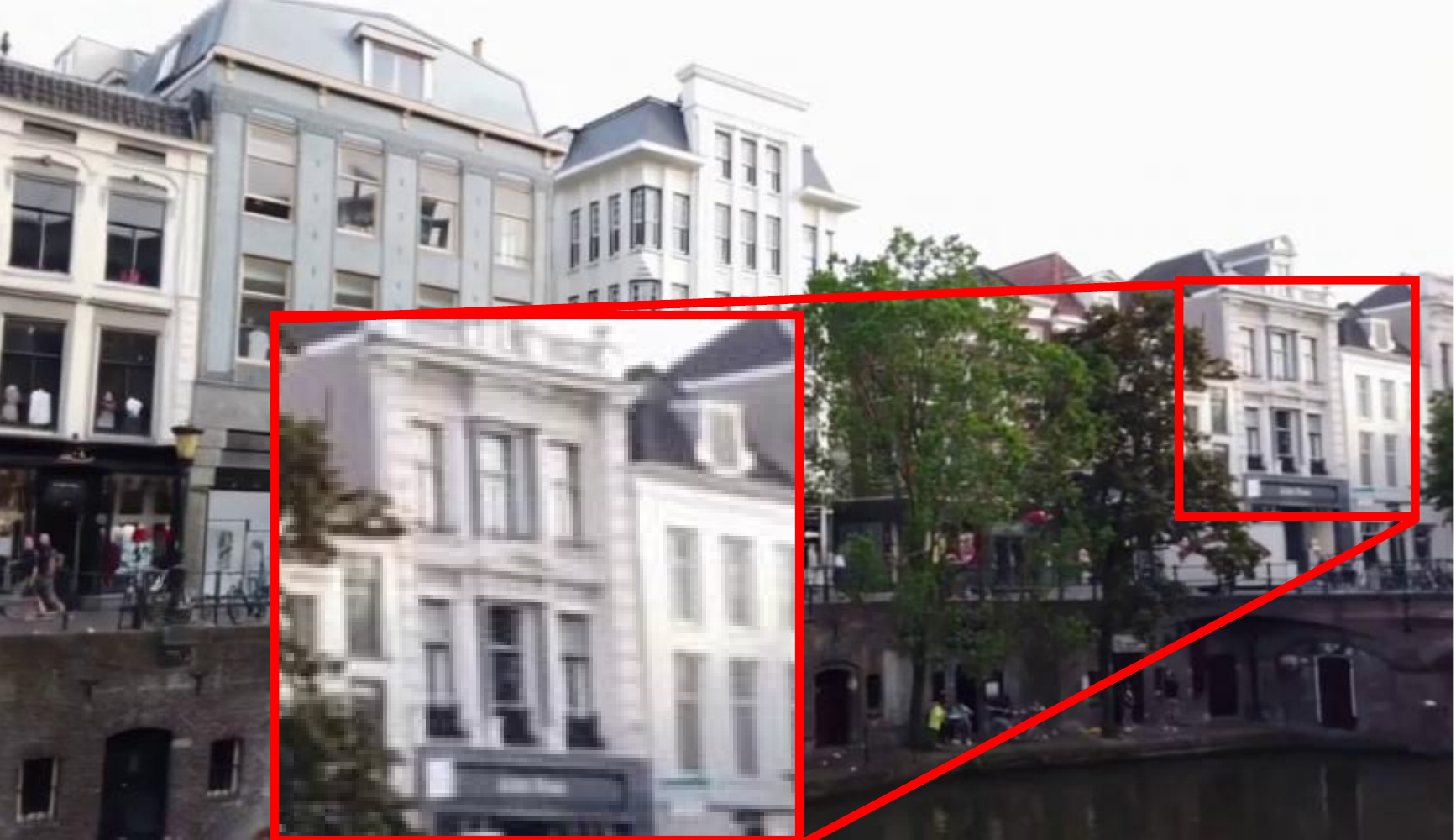} &
    \includegraphics[width=0.235\textwidth]{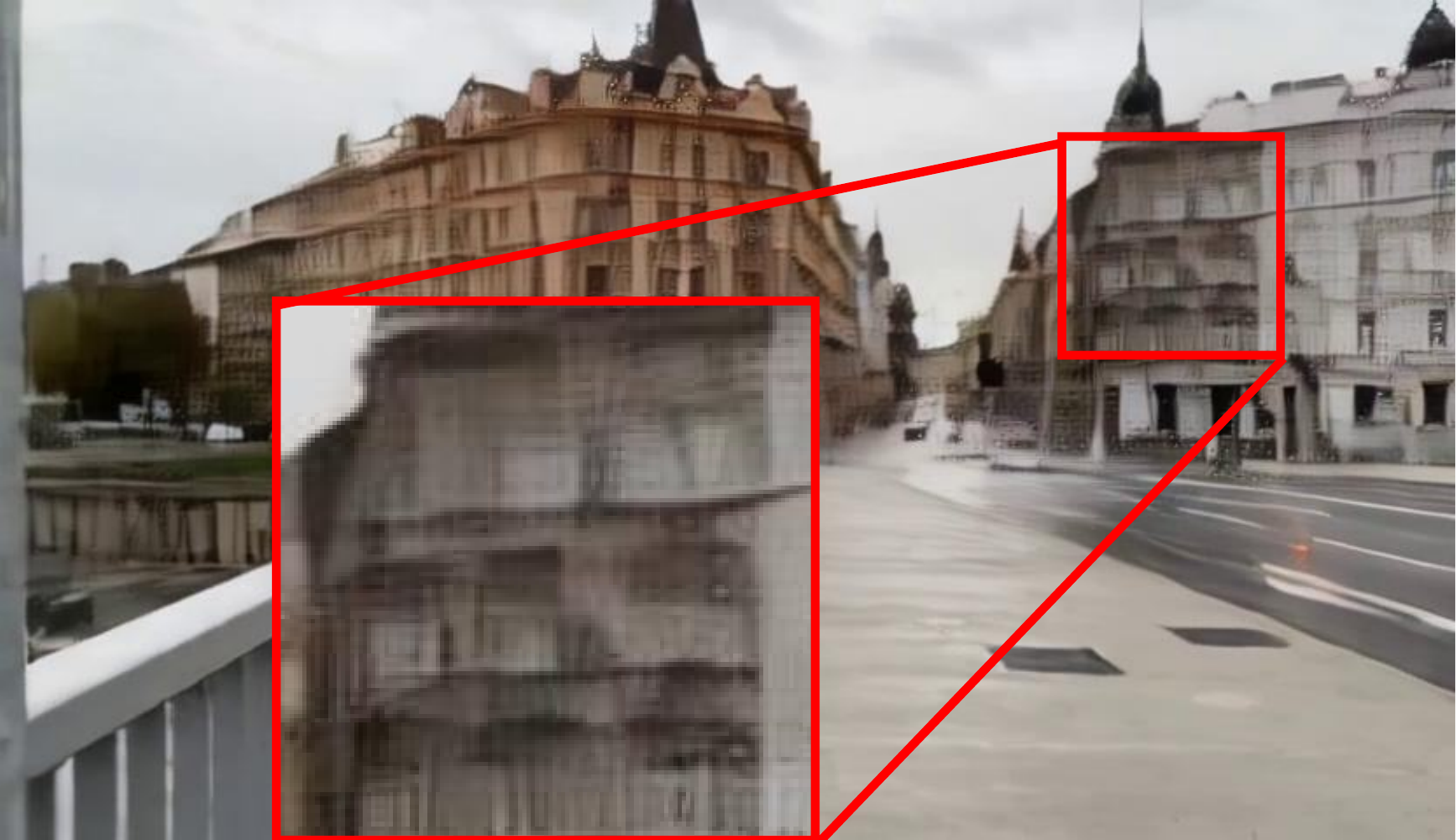} &
    \includegraphics[width=0.235\textwidth]{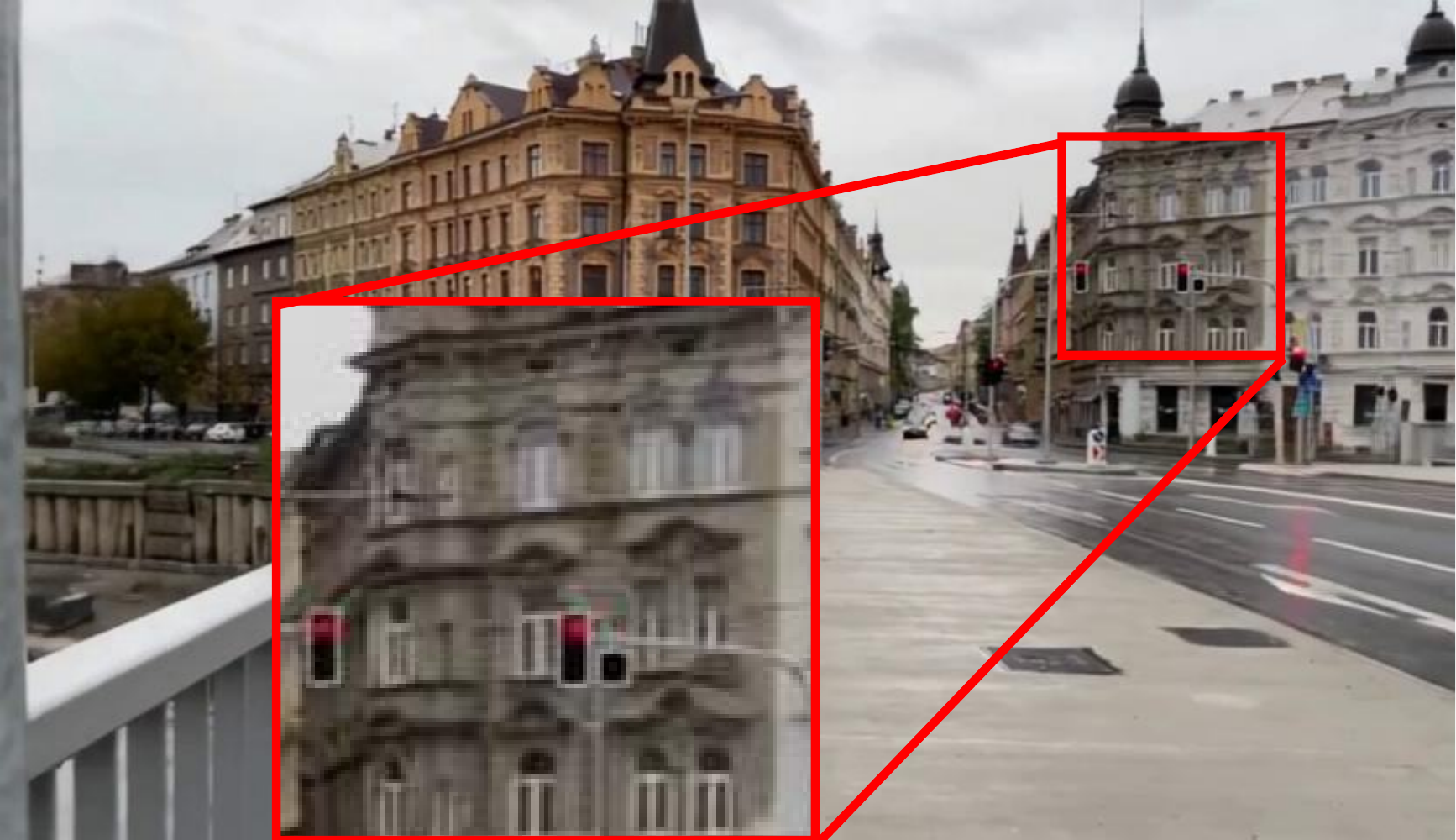} \\
    \includegraphics[width=0.235\textwidth]{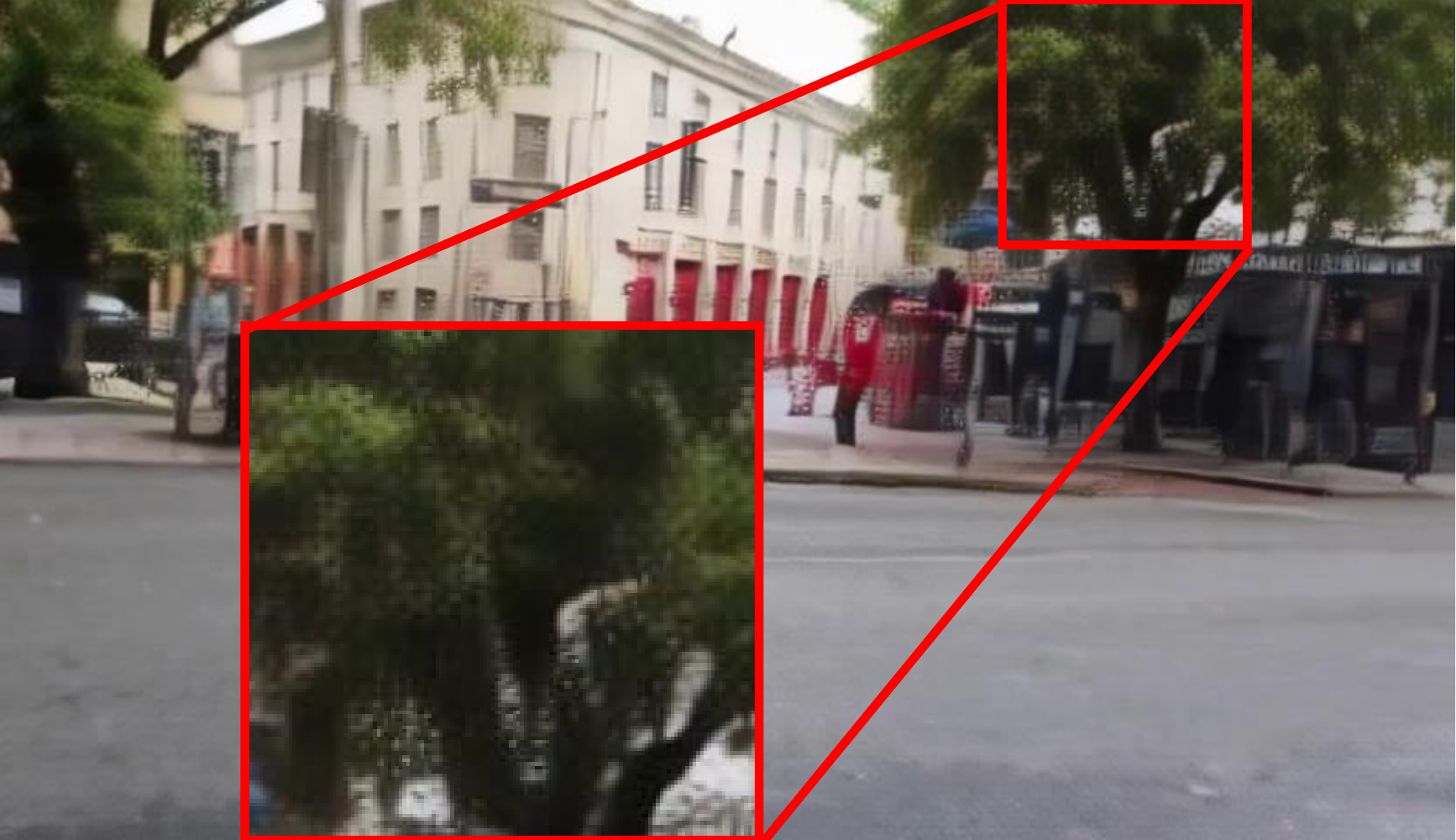} &
    \includegraphics[width=0.235\textwidth]{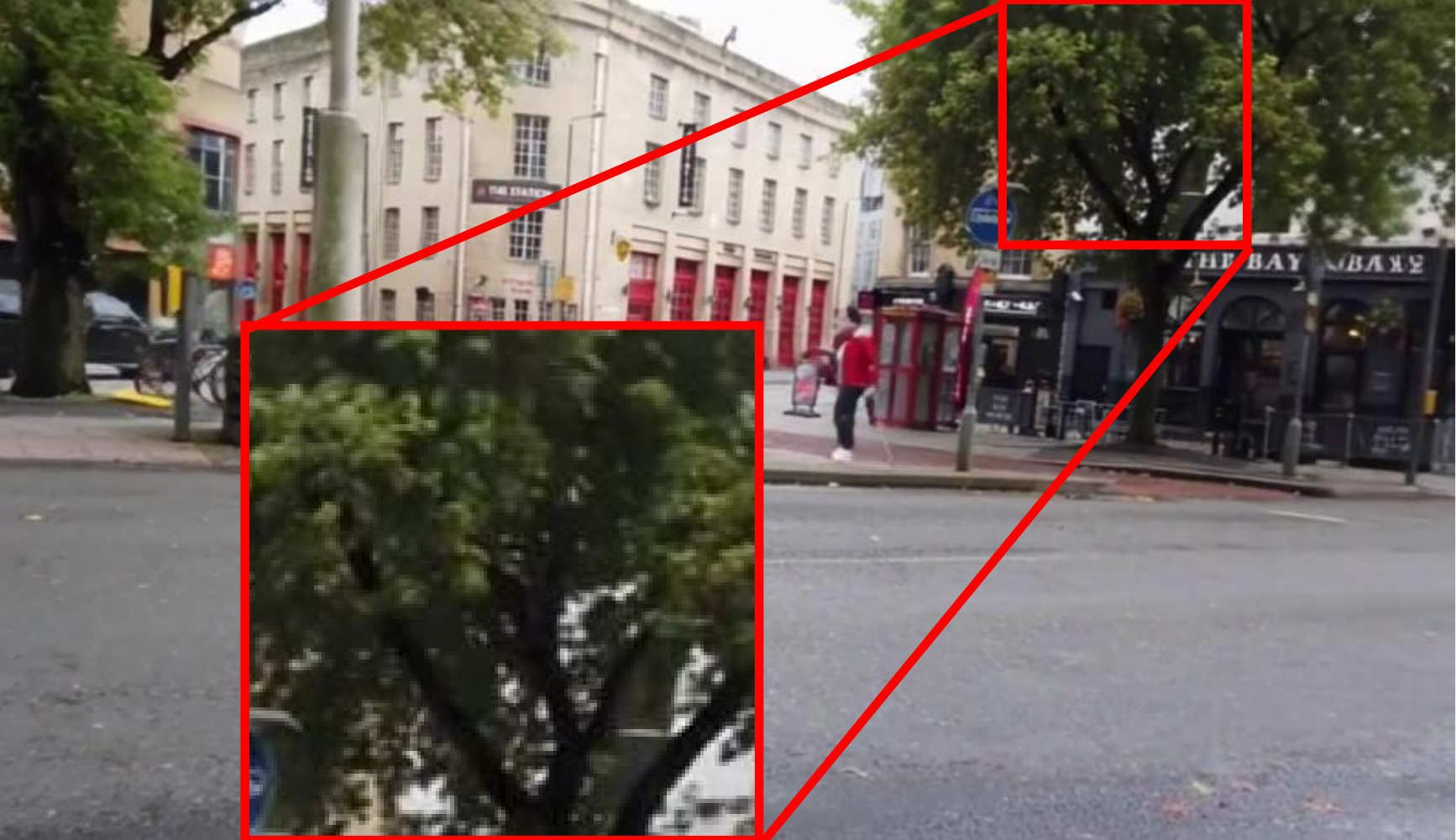} &
    \includegraphics[width=0.235\textwidth]{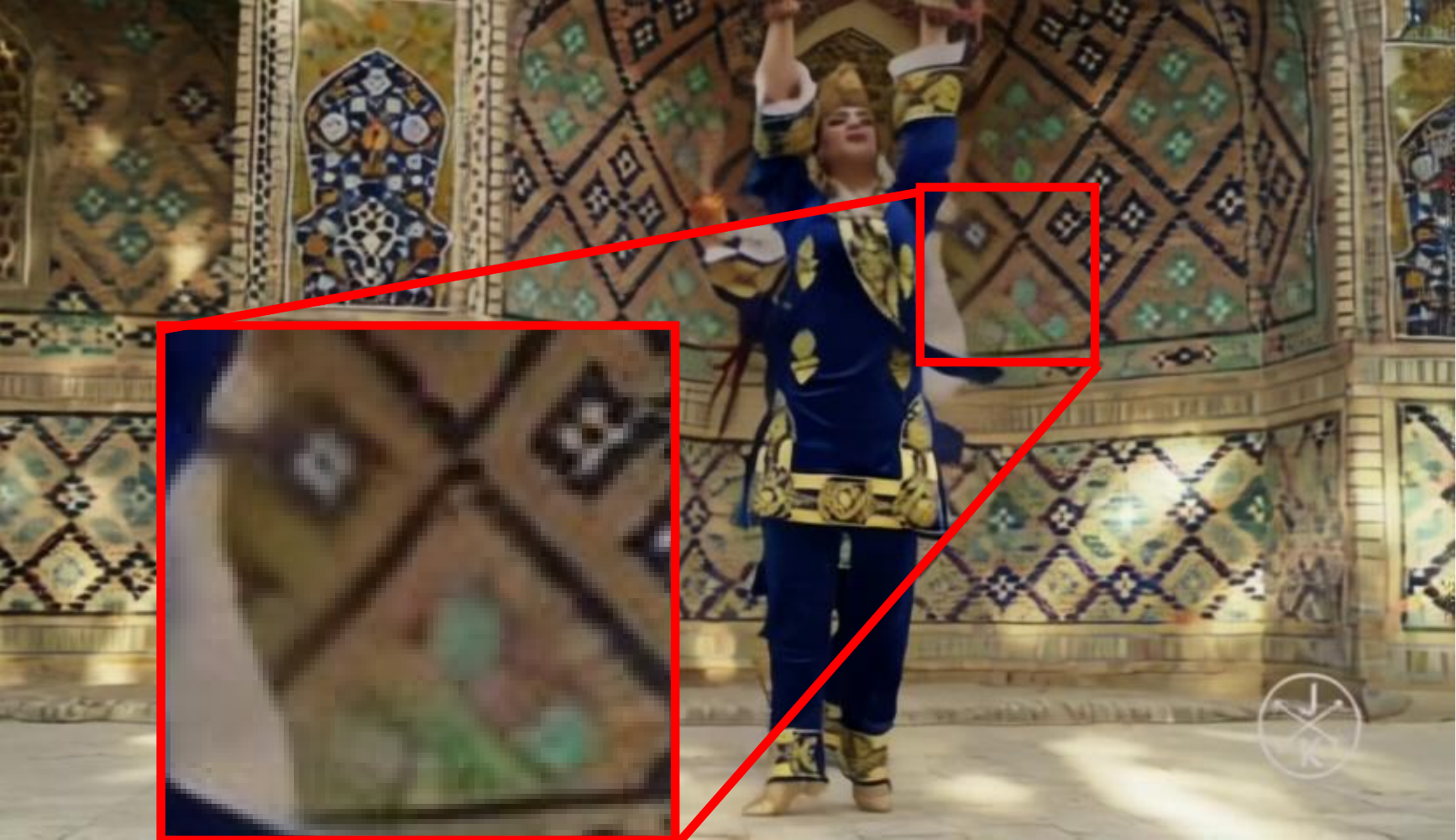} &
    \includegraphics[width=0.235\textwidth]{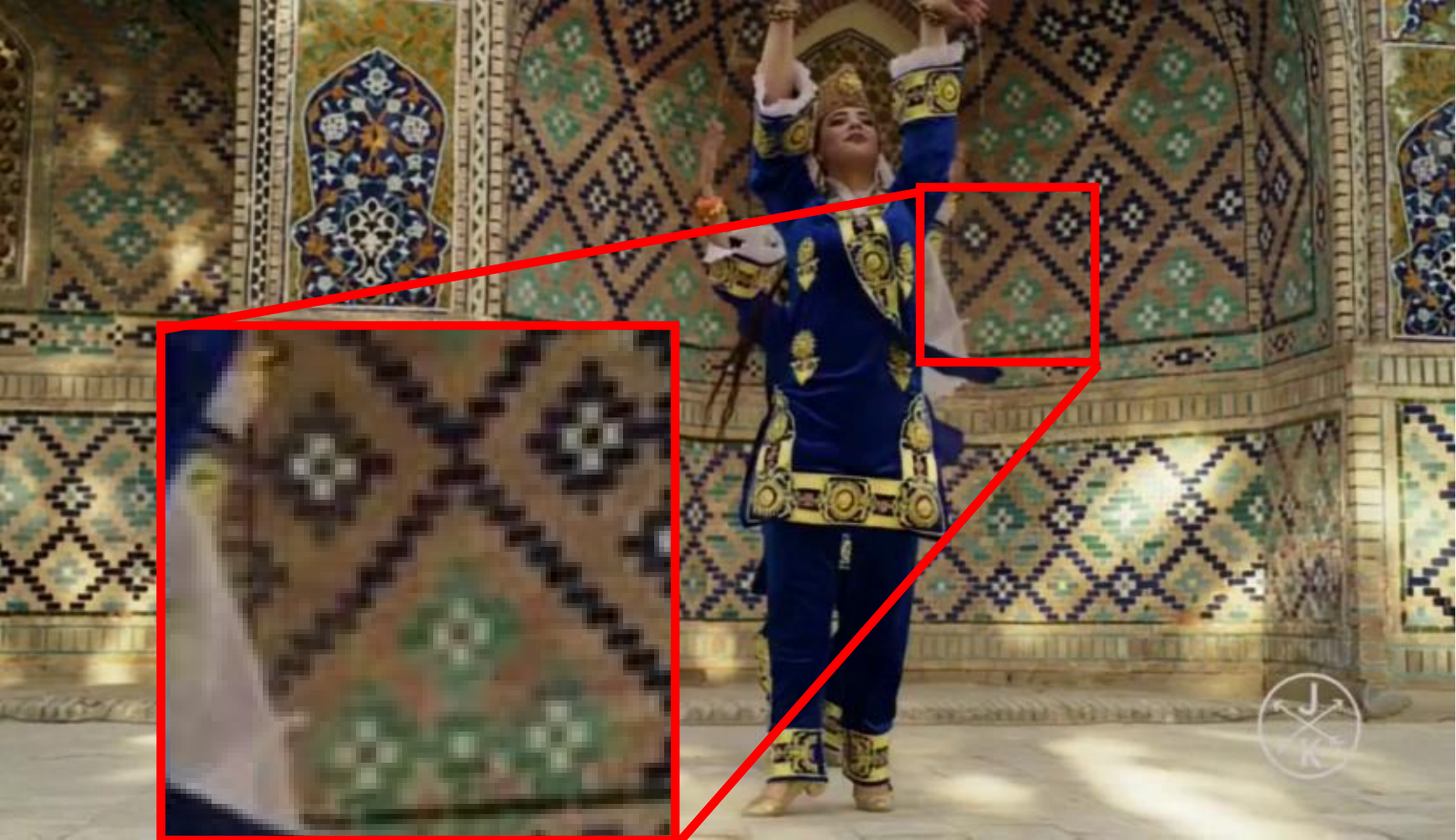} \\
    \includegraphics[width=0.235\textwidth]{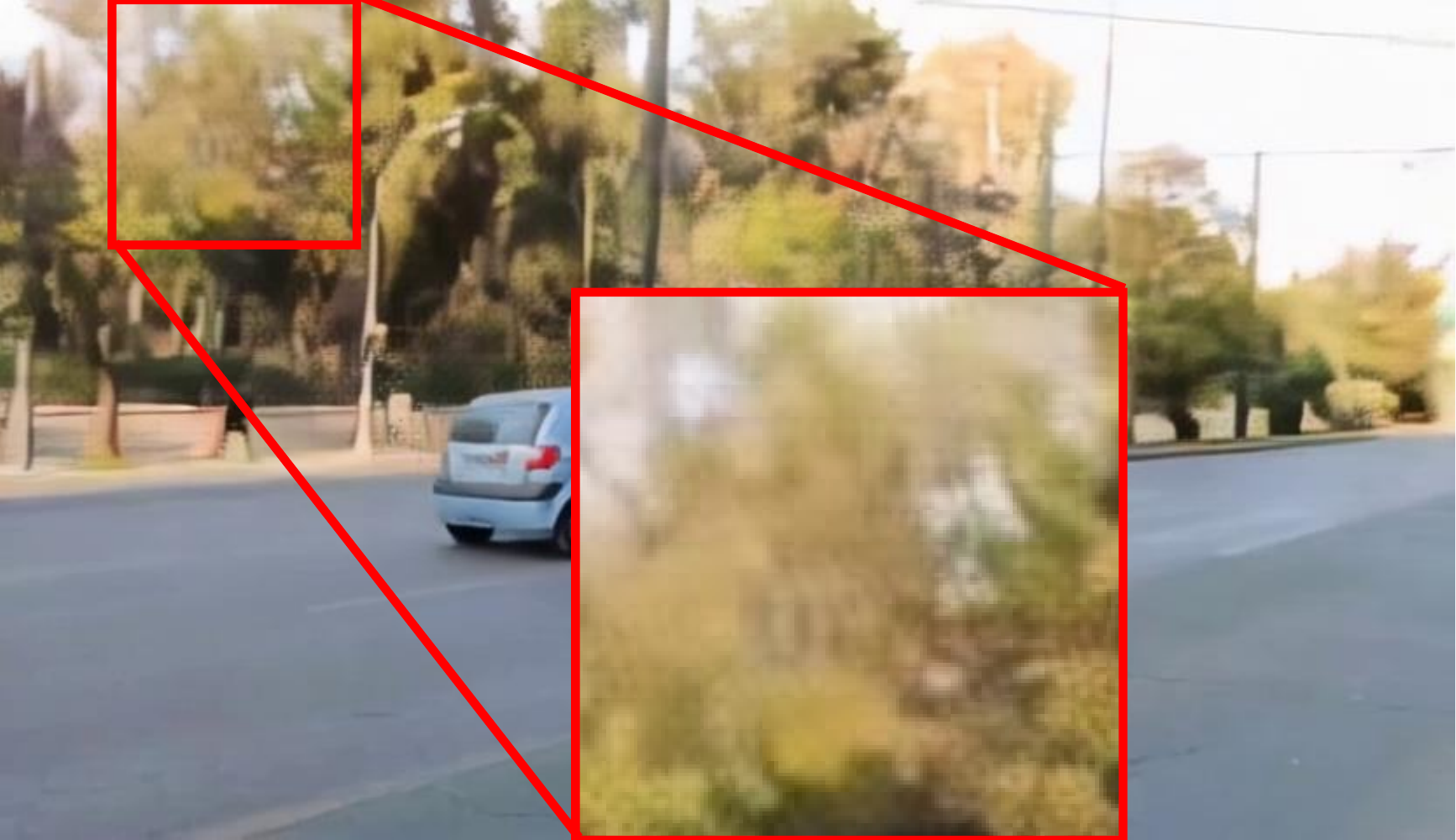} &
    \includegraphics[width=0.235\textwidth]{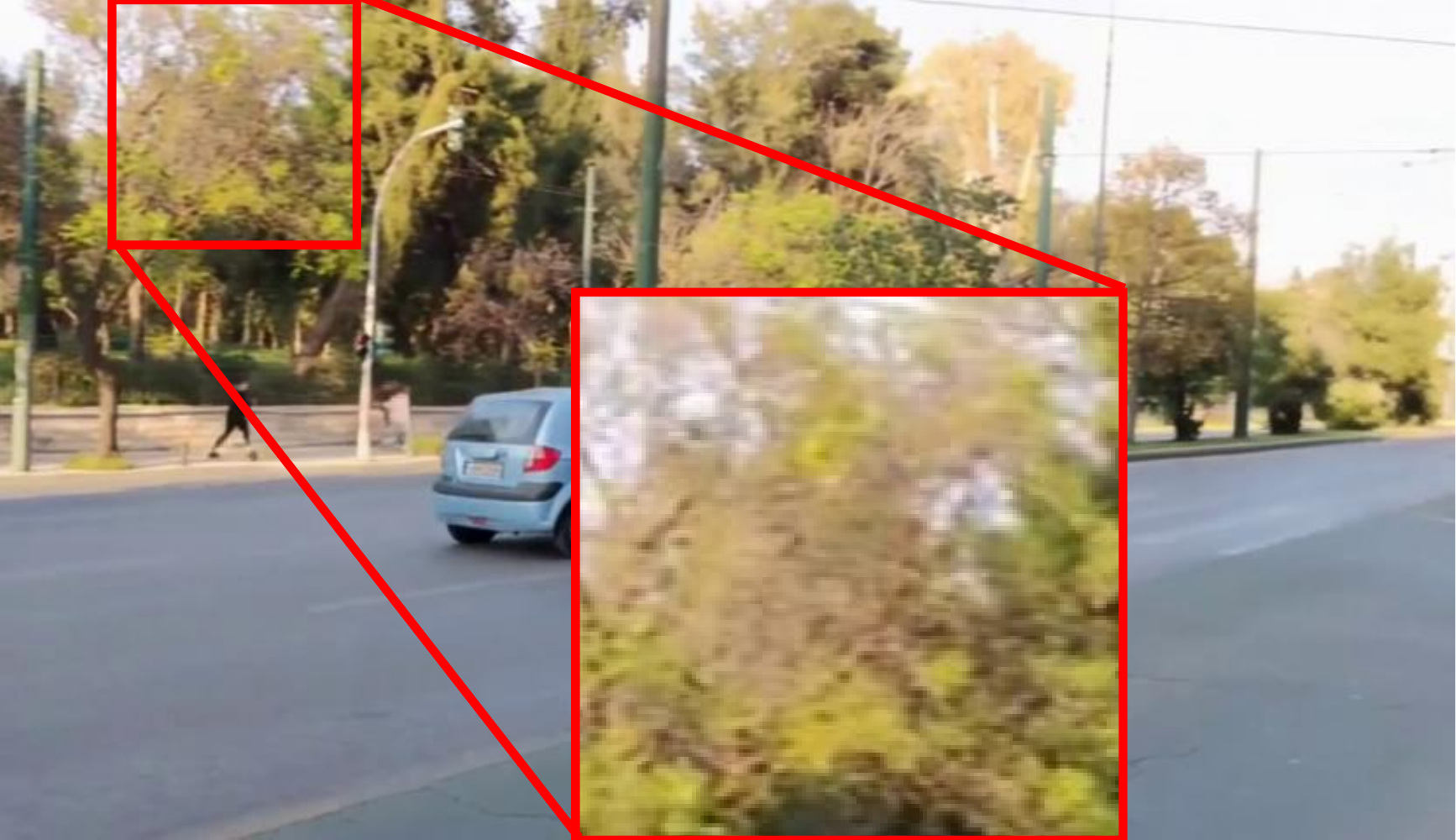} &
    \includegraphics[width=0.235\textwidth]{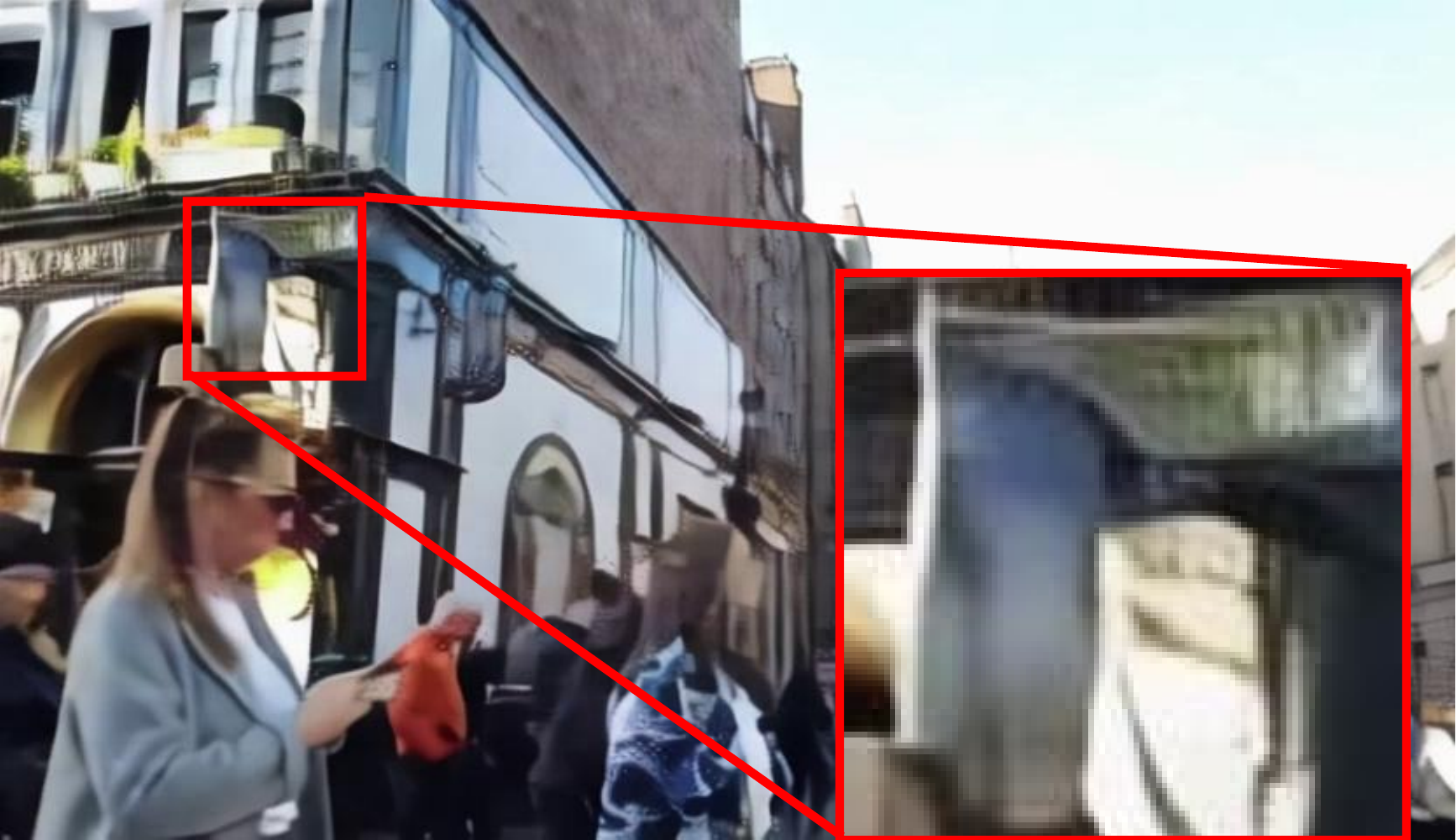} &
    \includegraphics[width=0.235\textwidth]{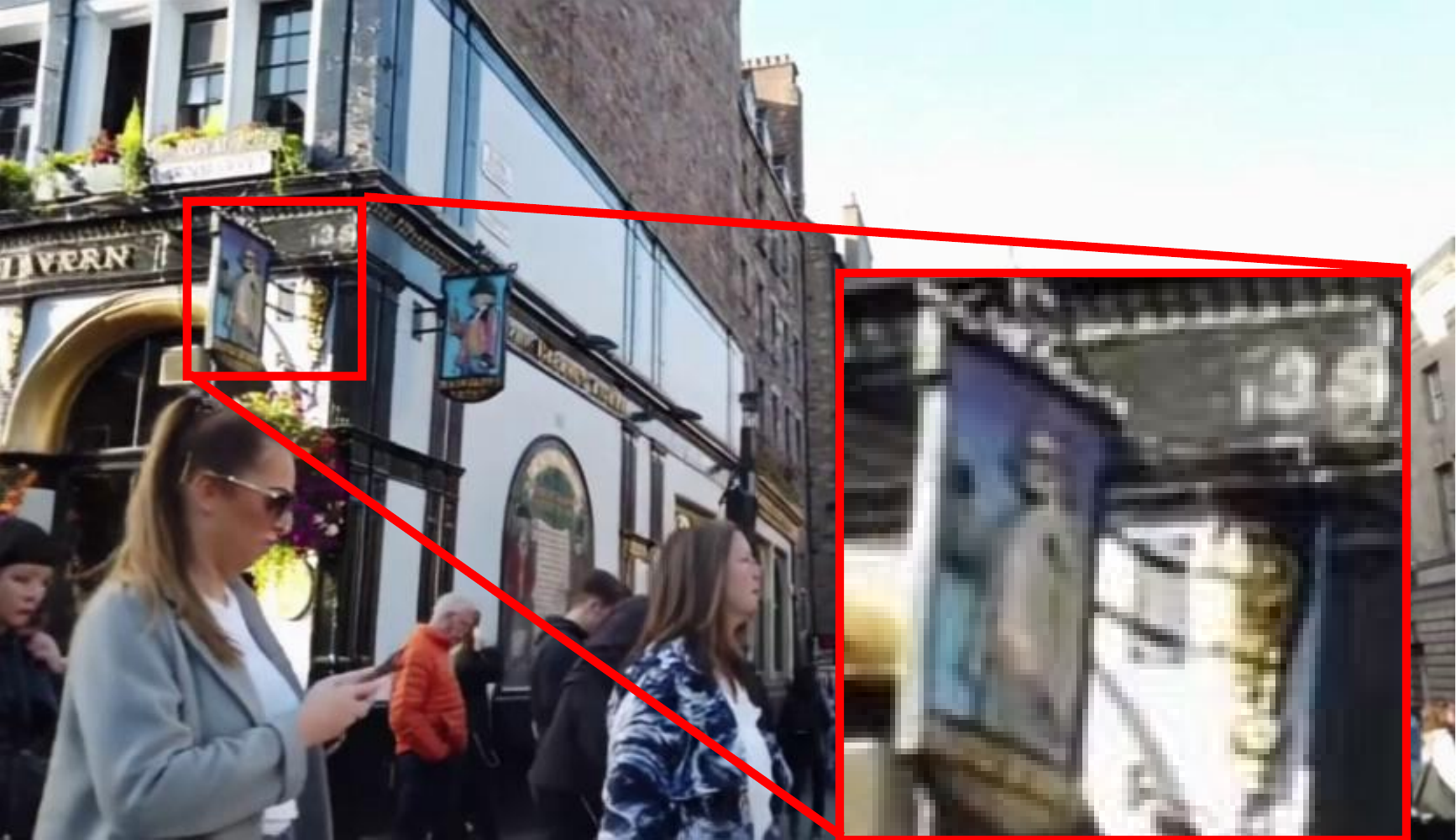} \\
    \includegraphics[width=0.235\textwidth]{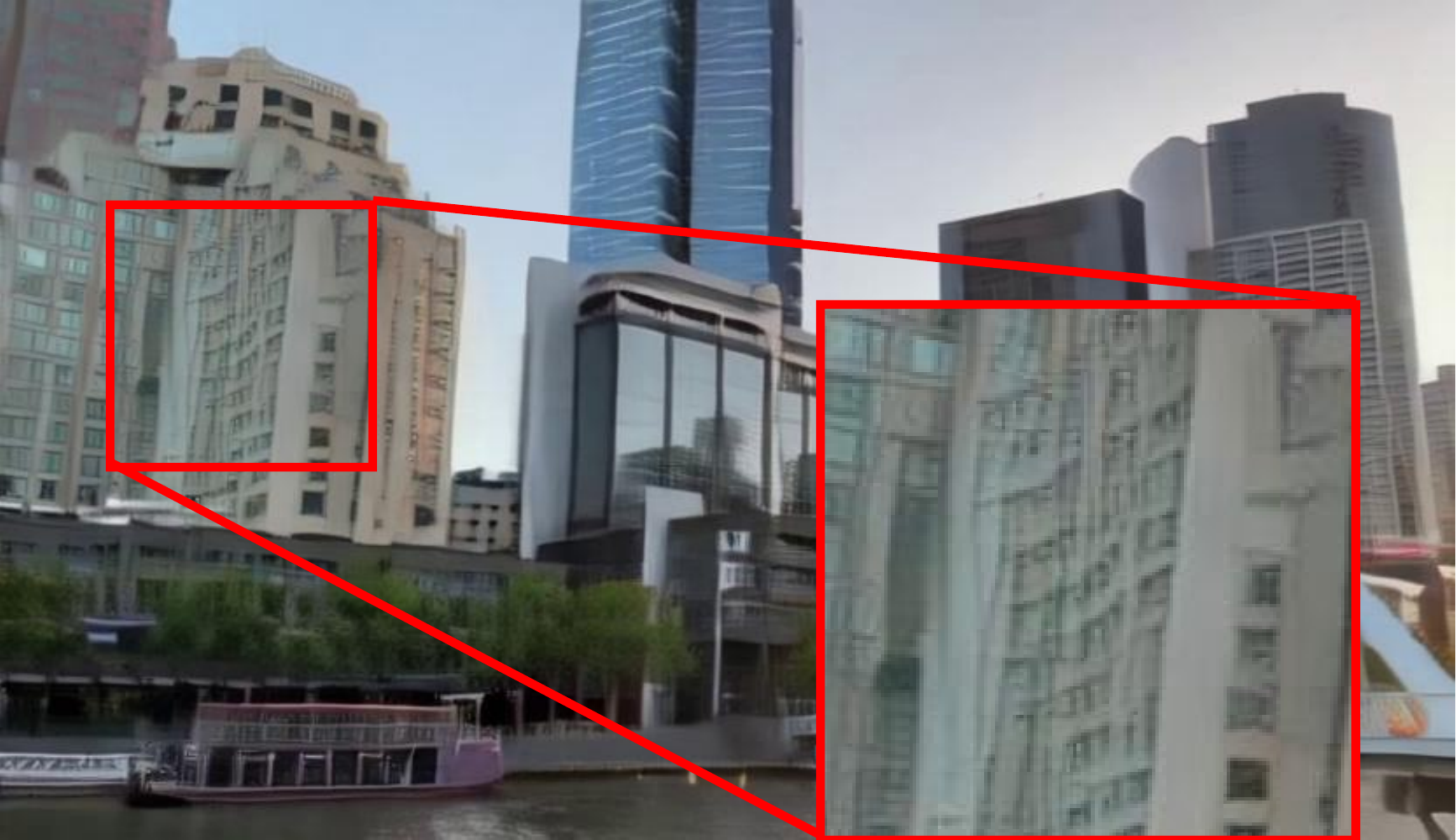} &
    \includegraphics[width=0.235\textwidth]{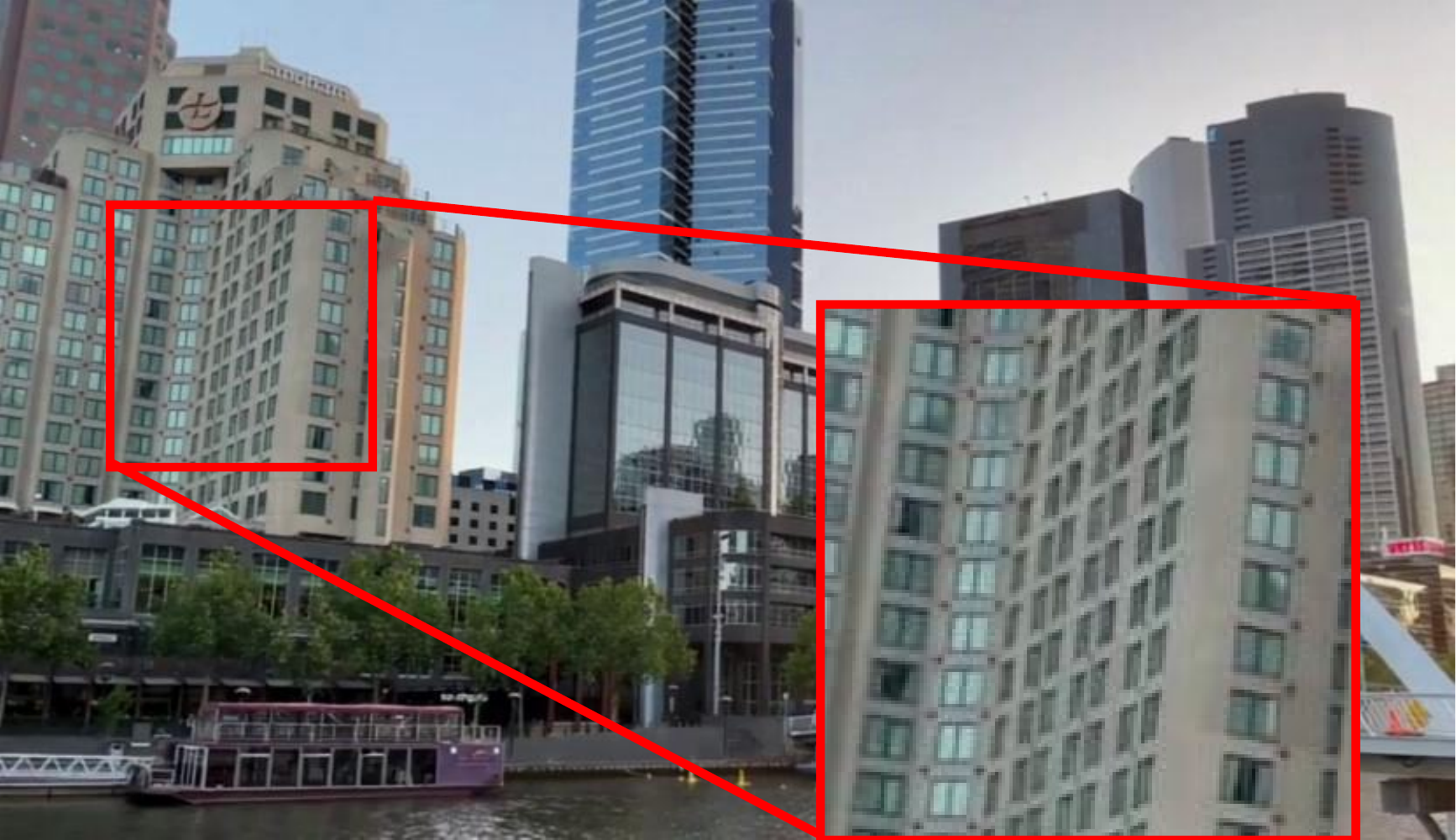} &
    \includegraphics[width=0.235\textwidth]{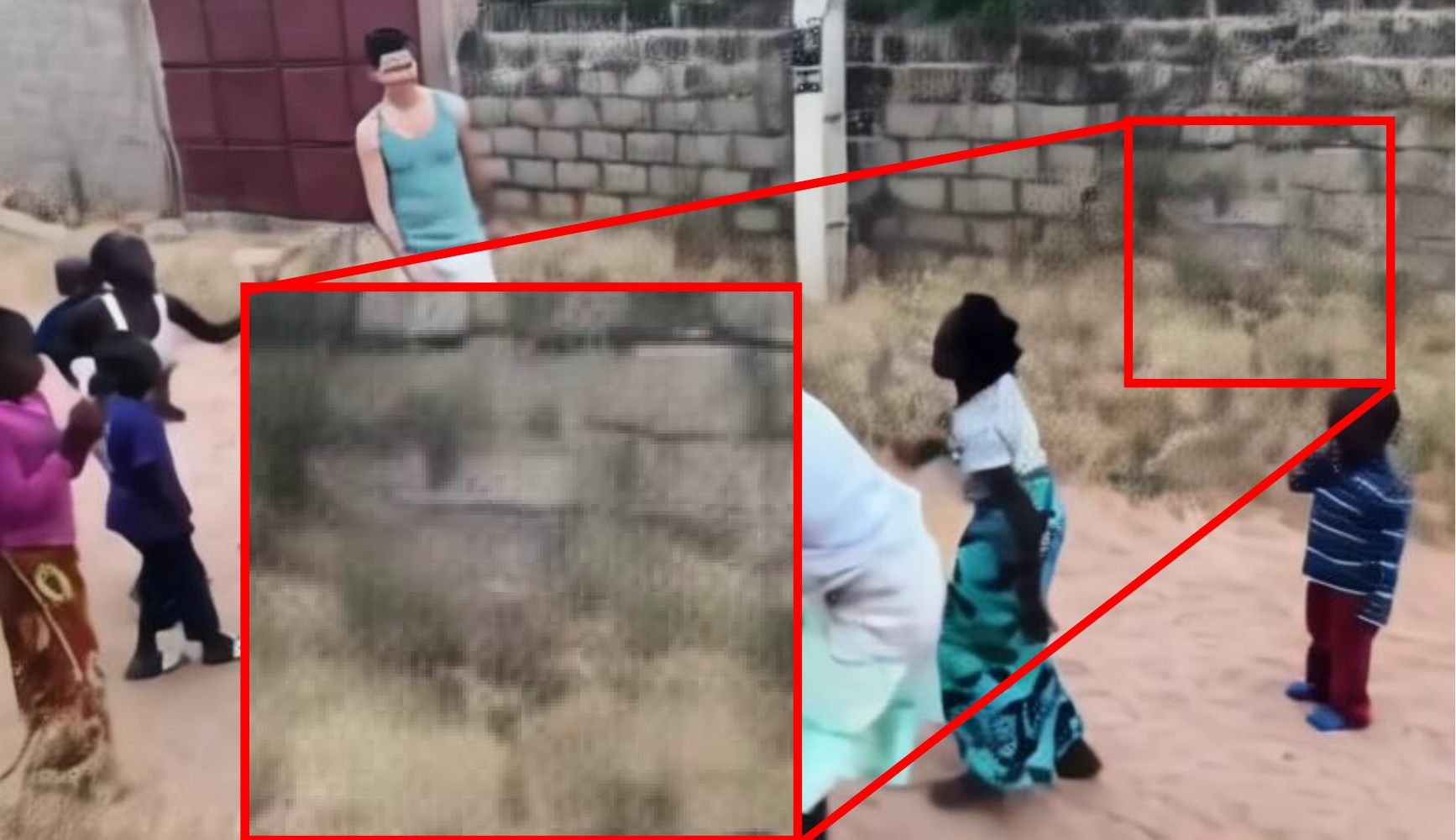} &
    \includegraphics[width=0.235\textwidth]{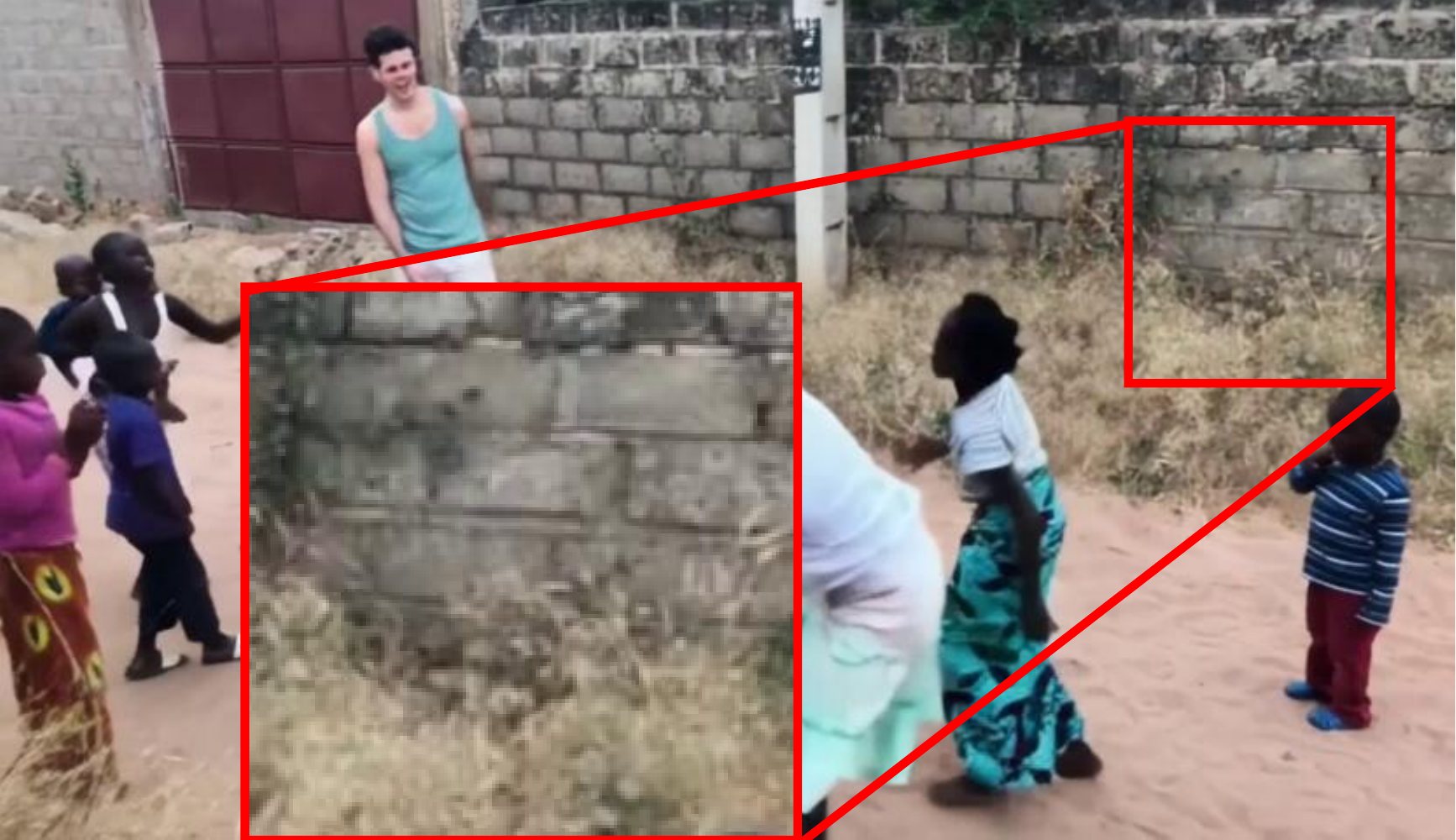} \\
    \includegraphics[width=0.235\textwidth]{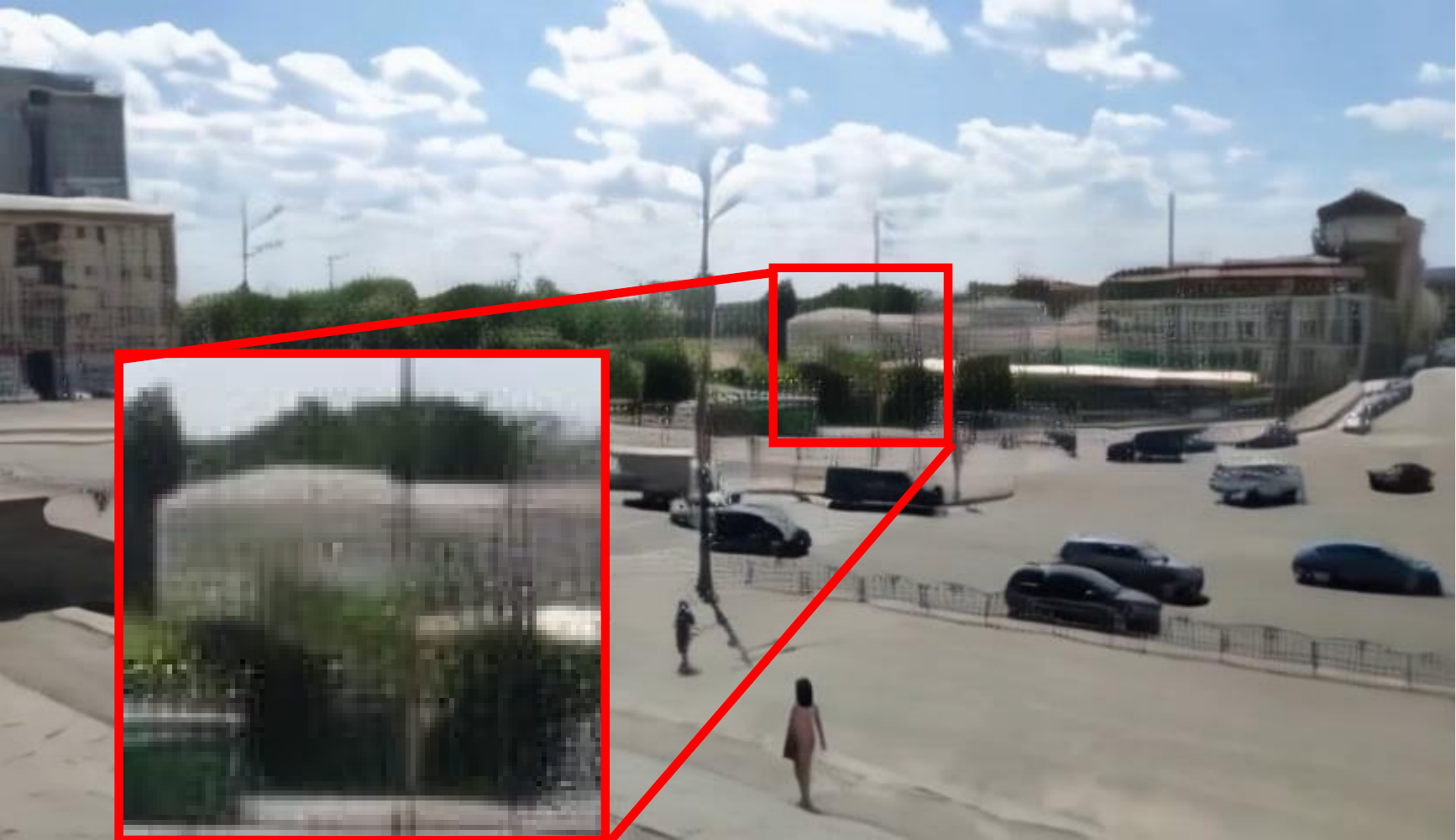} &
    \includegraphics[width=0.235\textwidth]{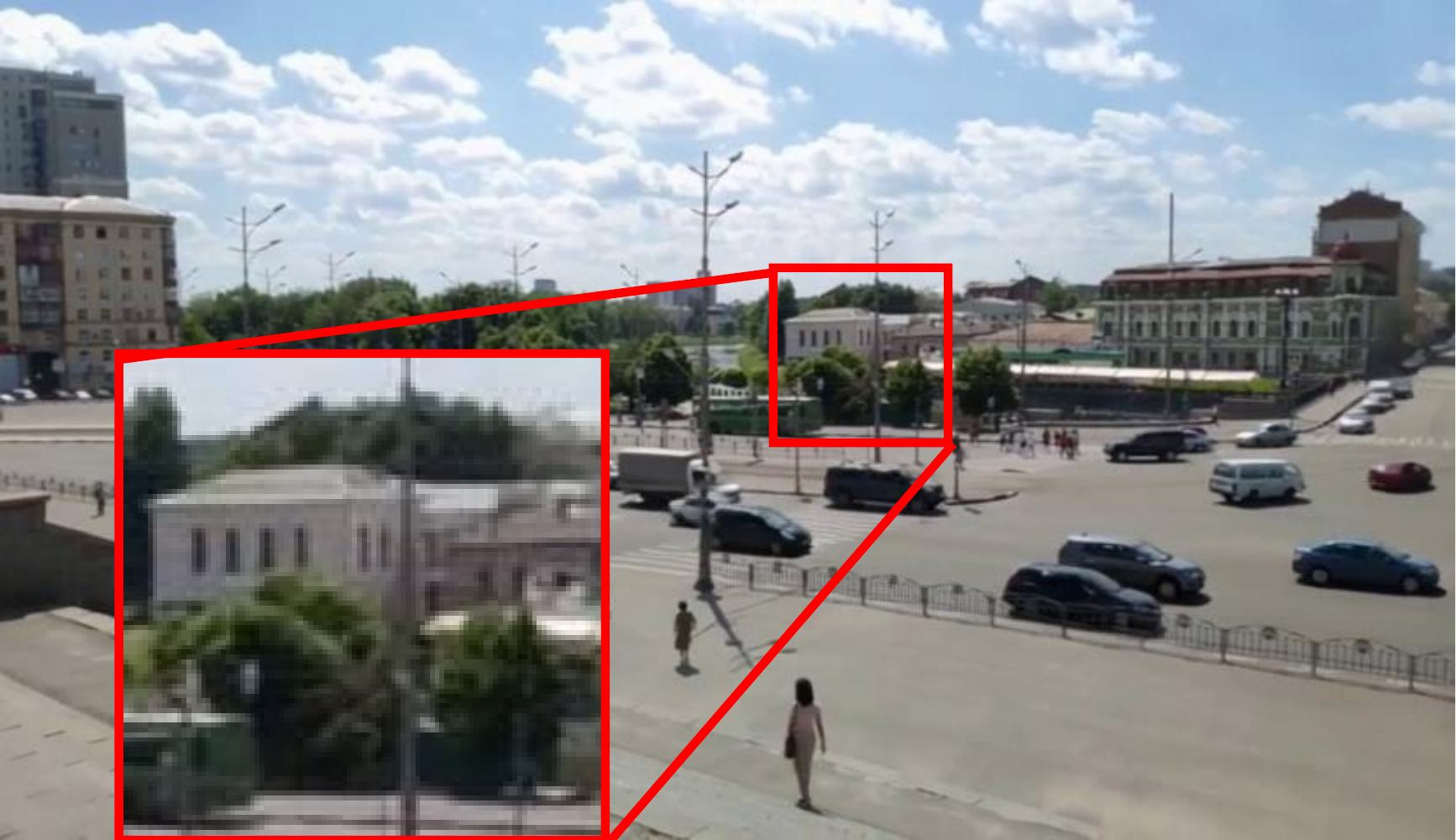} &
    \includegraphics[width=0.235\textwidth]{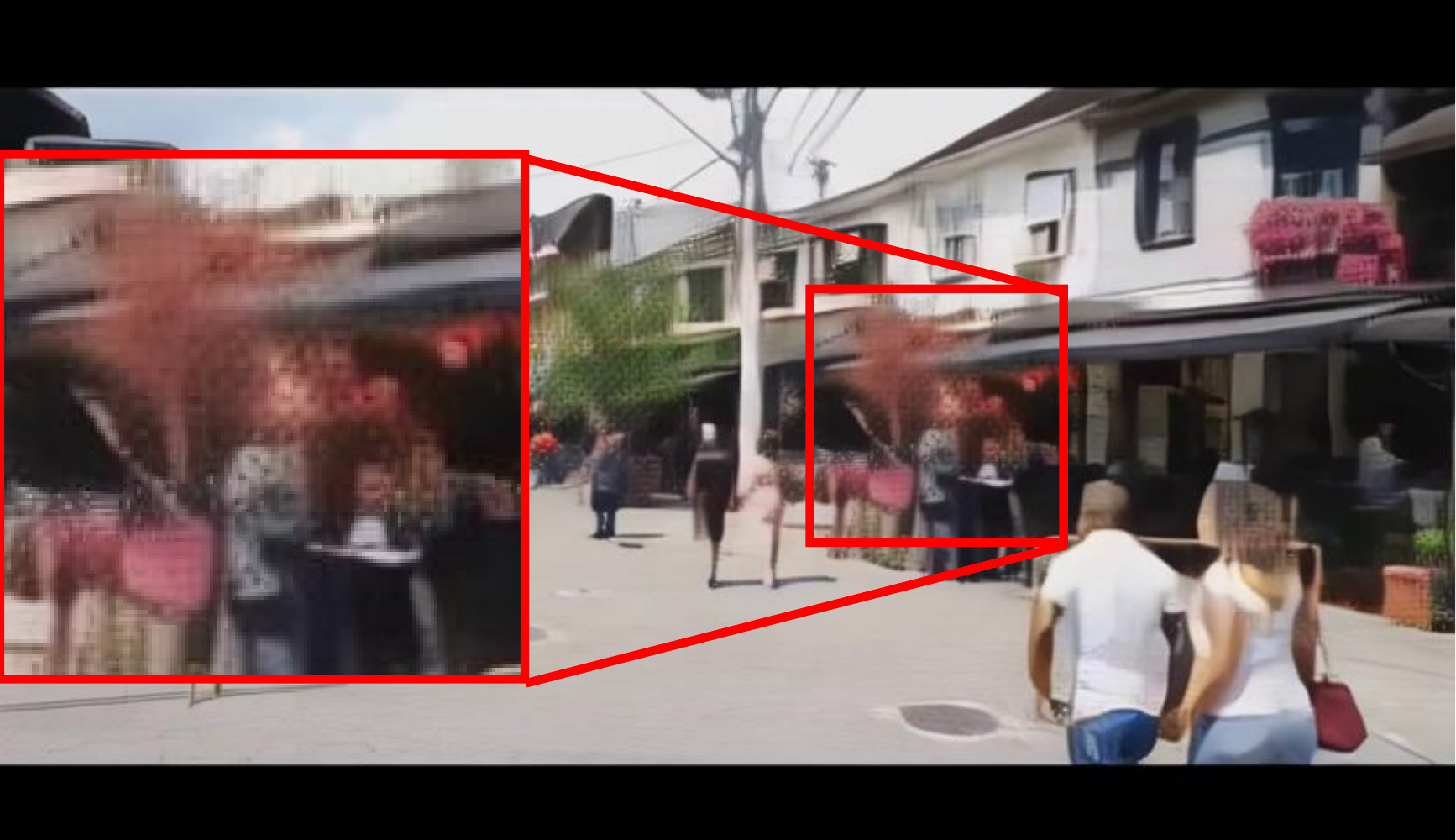} &
    \includegraphics[width=0.235\textwidth]{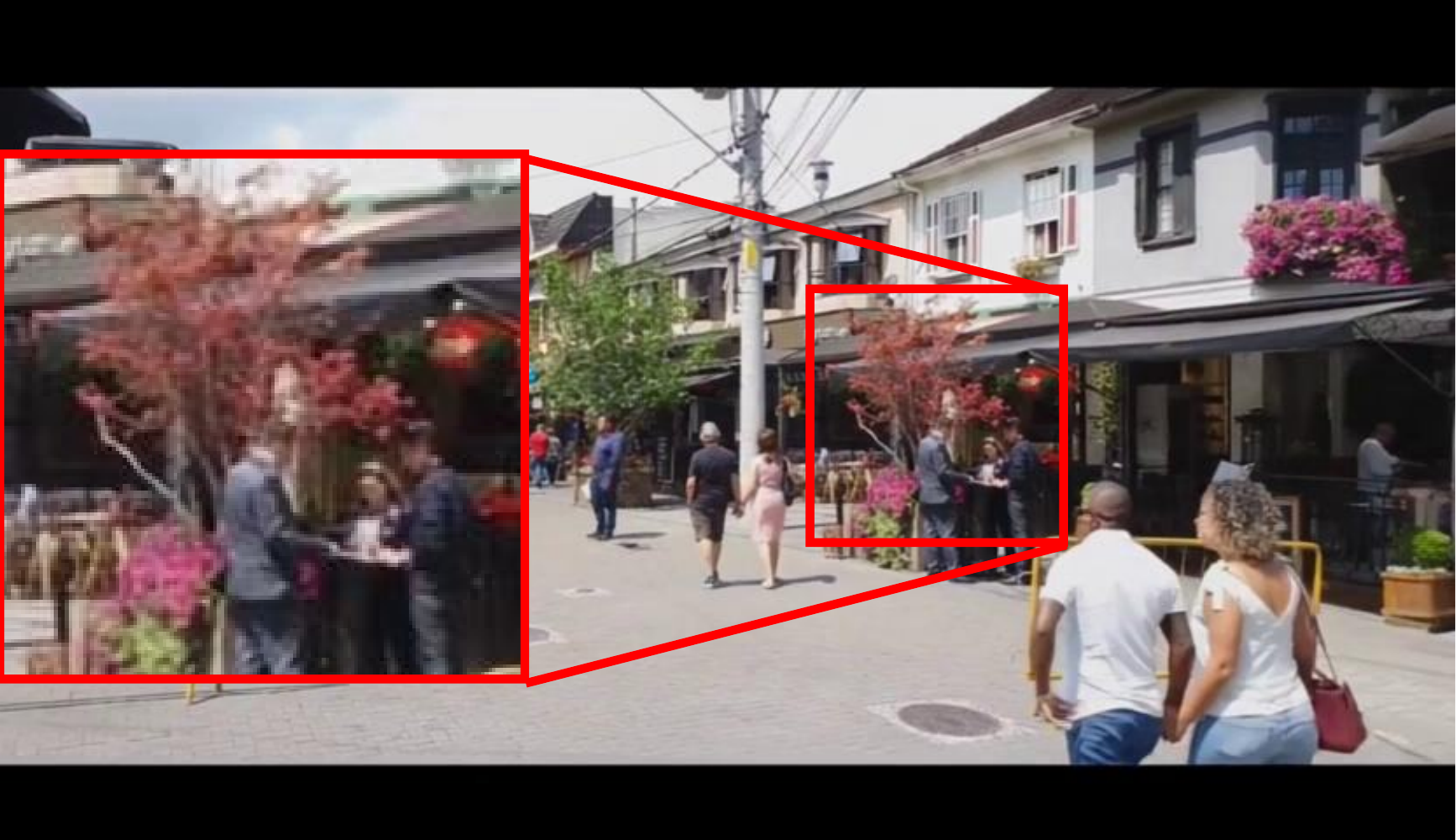} \\
    \includegraphics[width=0.235\textwidth]{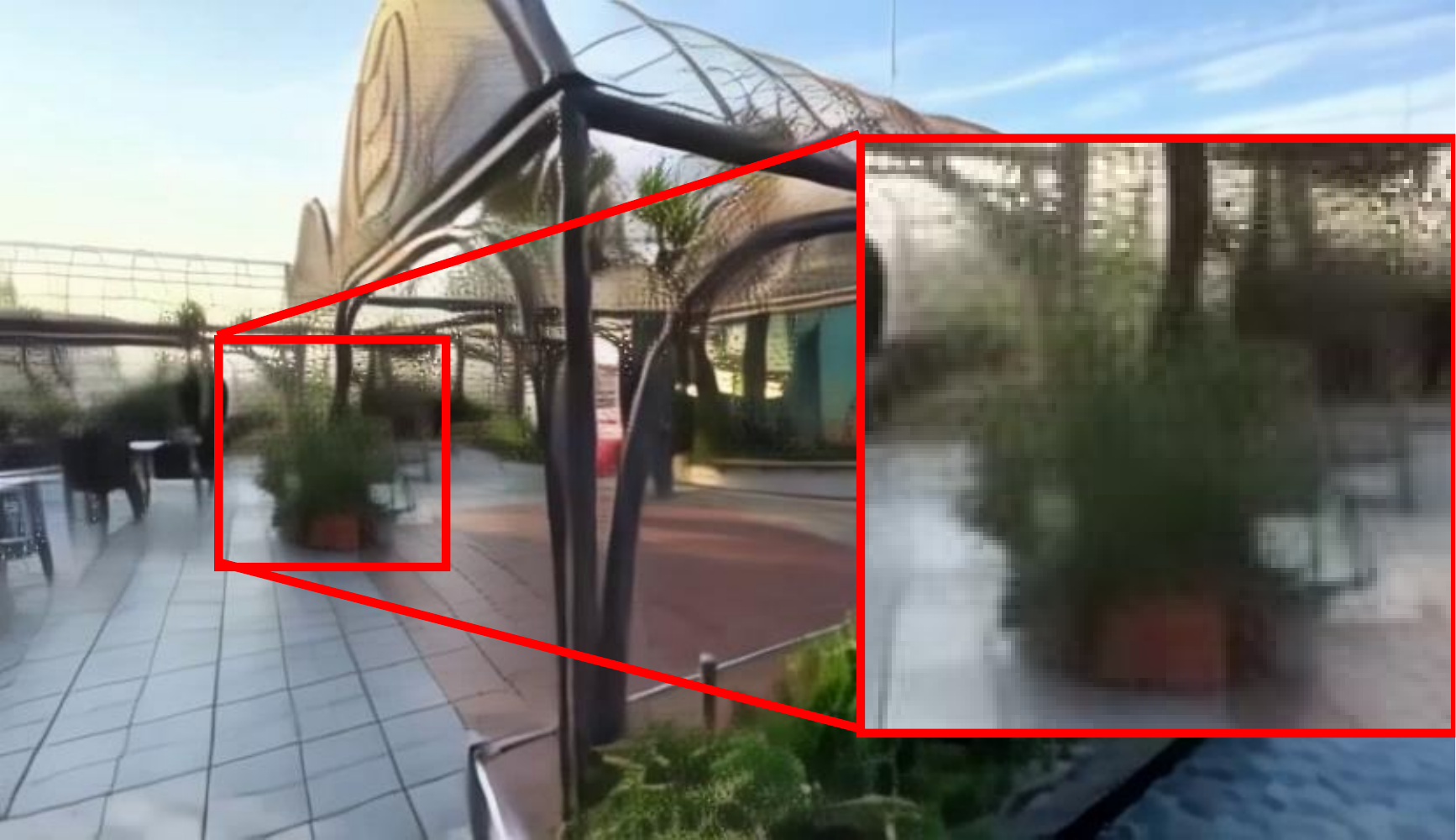} &
    \includegraphics[width=0.235\textwidth]{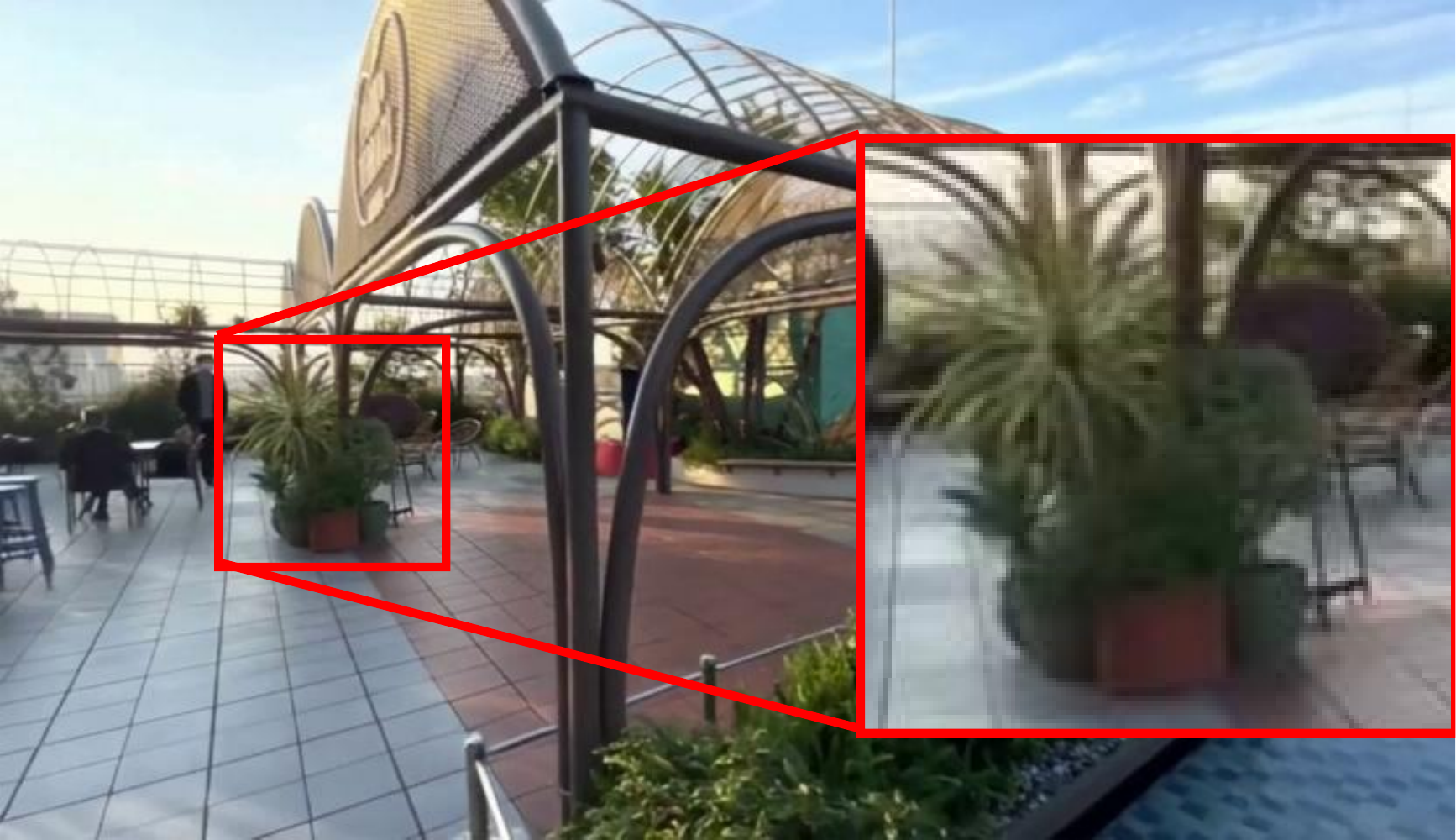} &
    \includegraphics[width=0.235\textwidth]{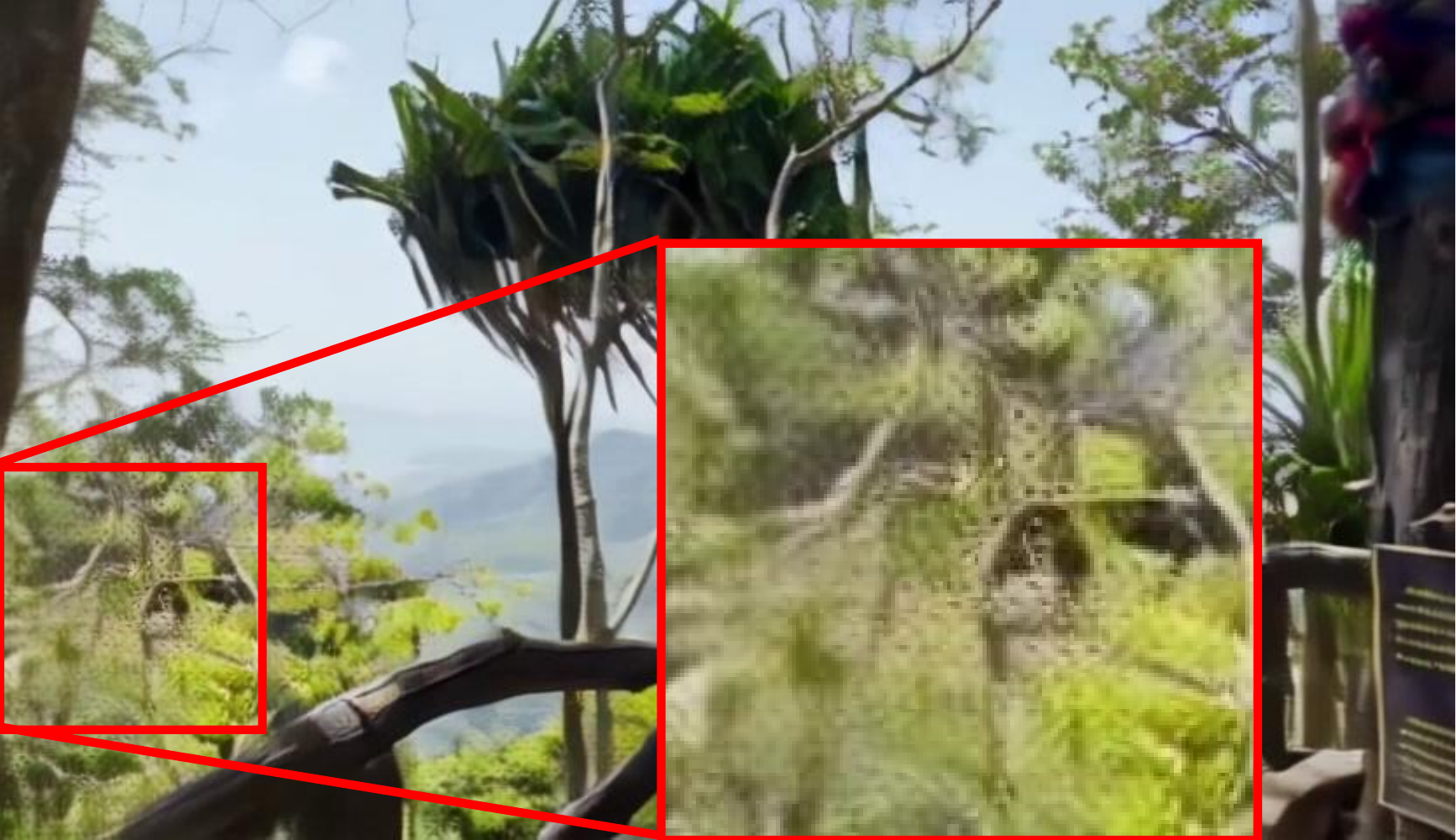} &
    \includegraphics[width=0.235\textwidth]{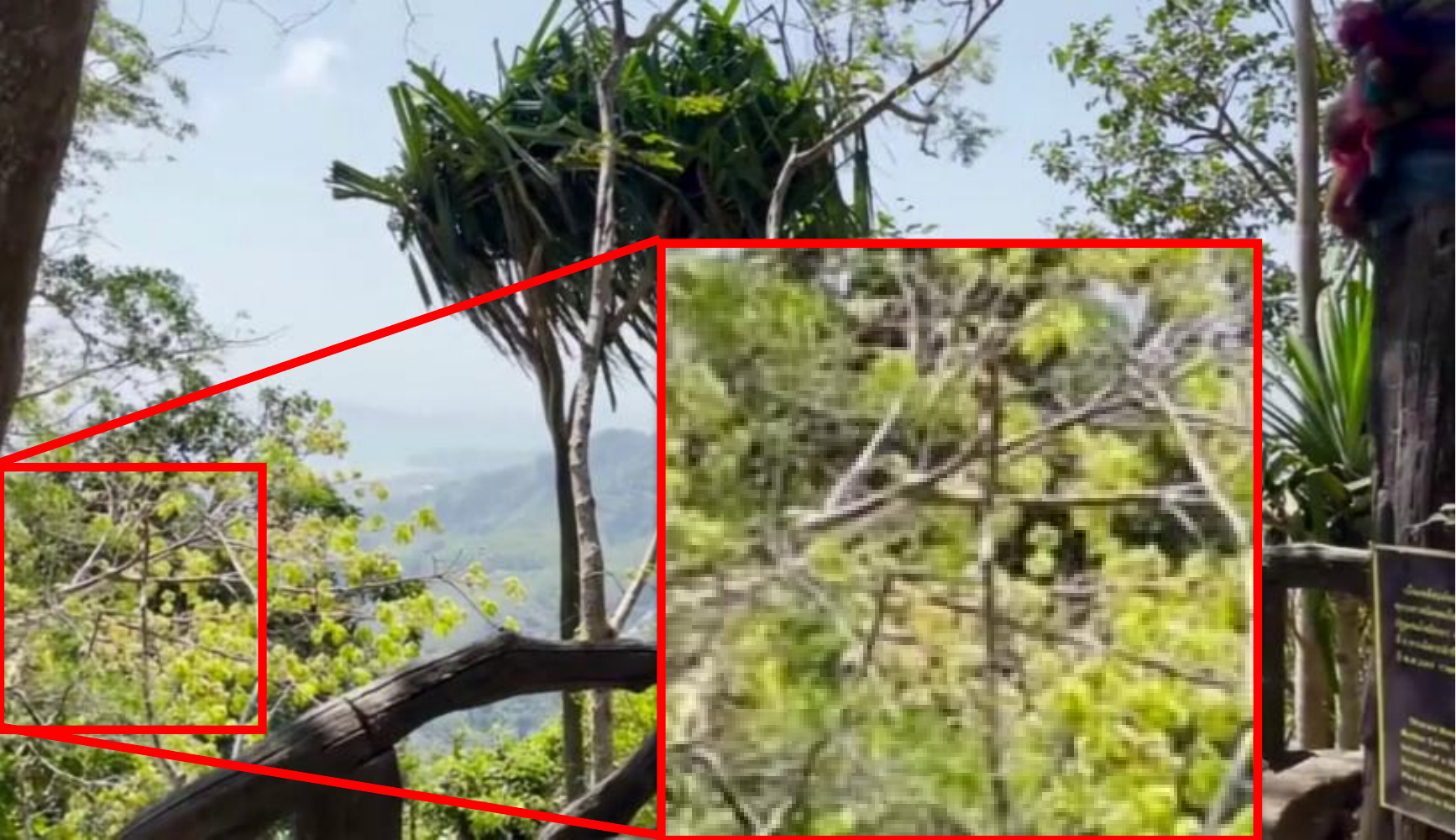} \\
    \includegraphics[width=0.235\textwidth]{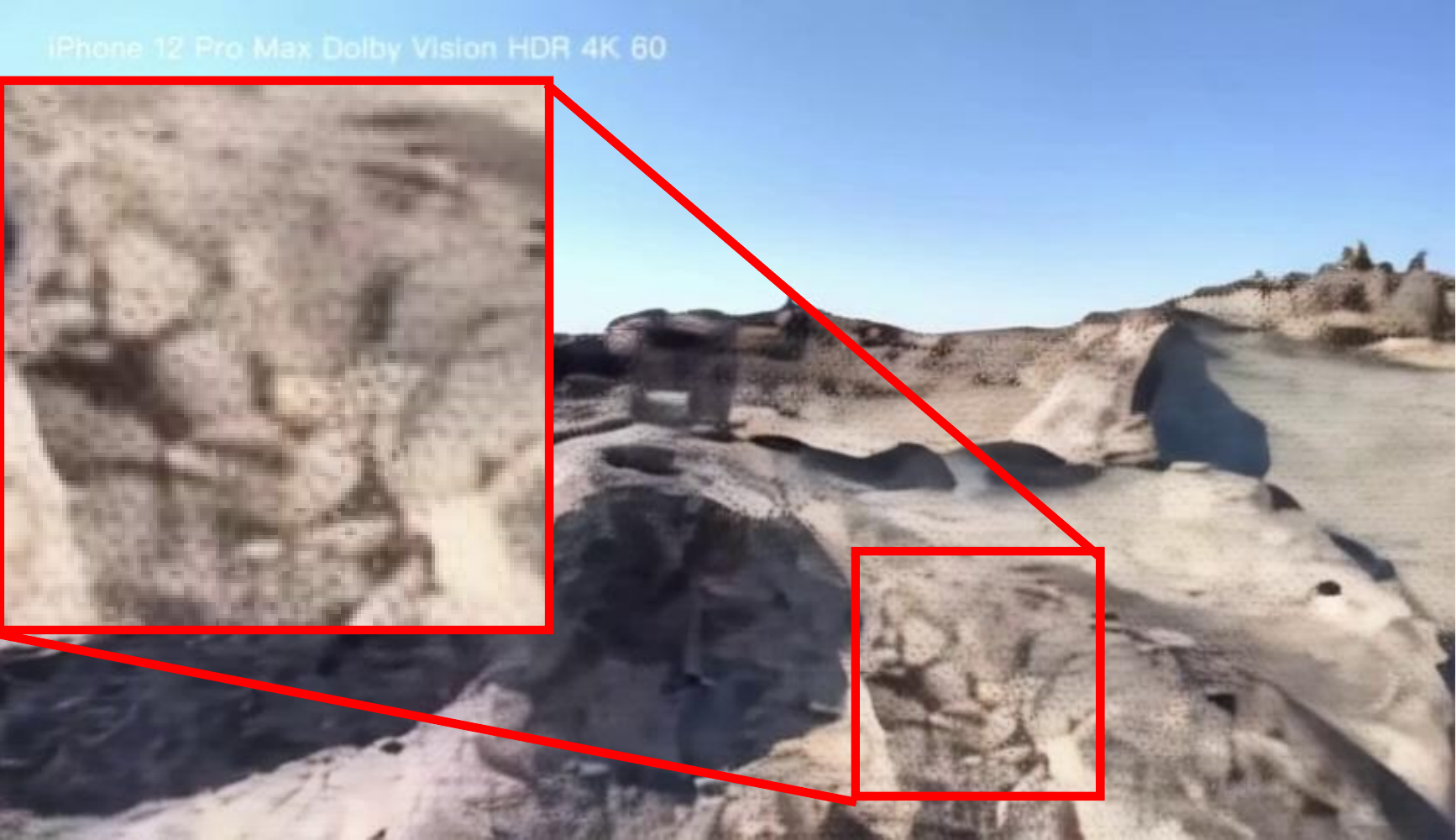} &
    \includegraphics[width=0.235\textwidth]{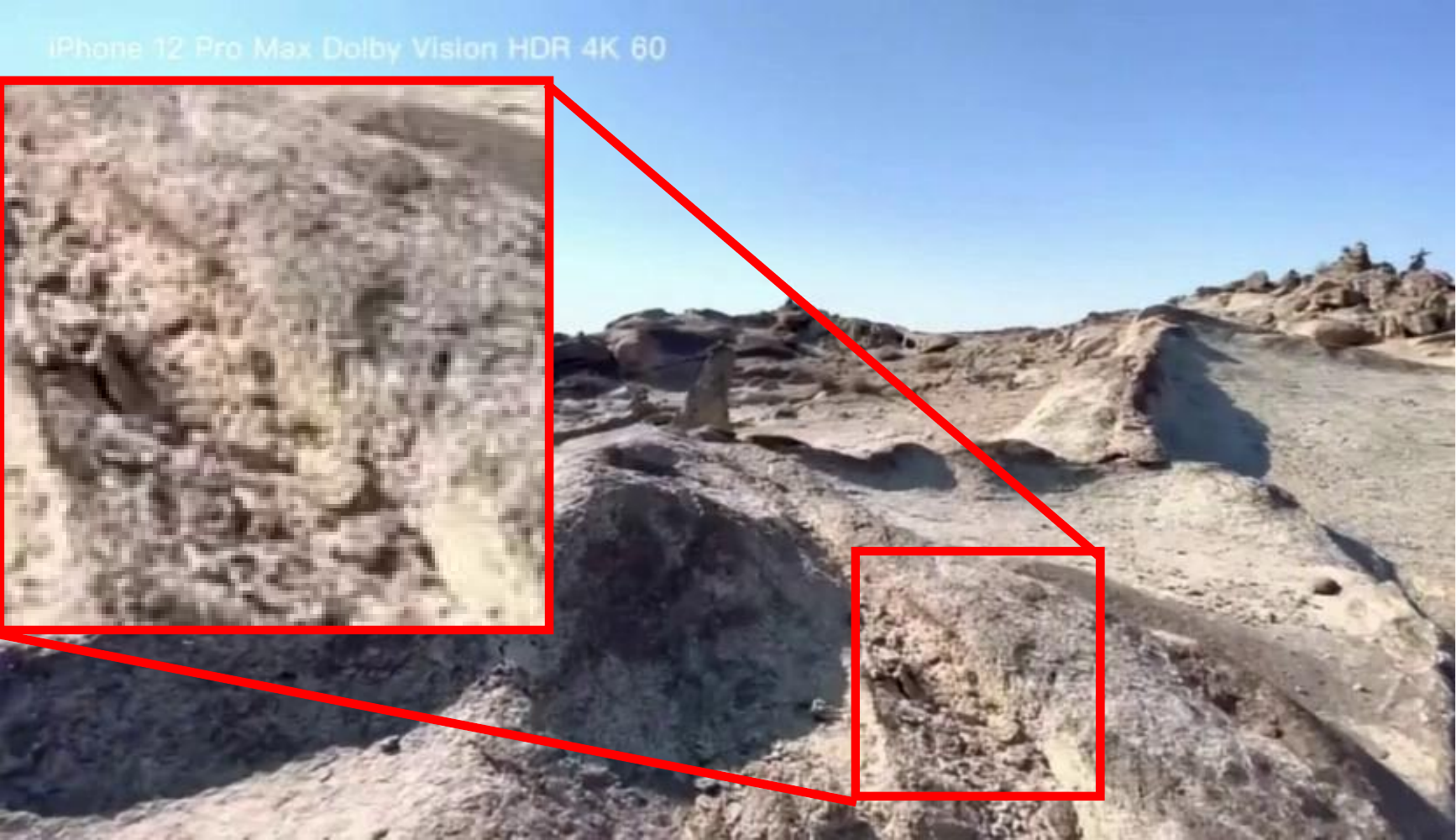} &
    \includegraphics[width=0.235\textwidth]{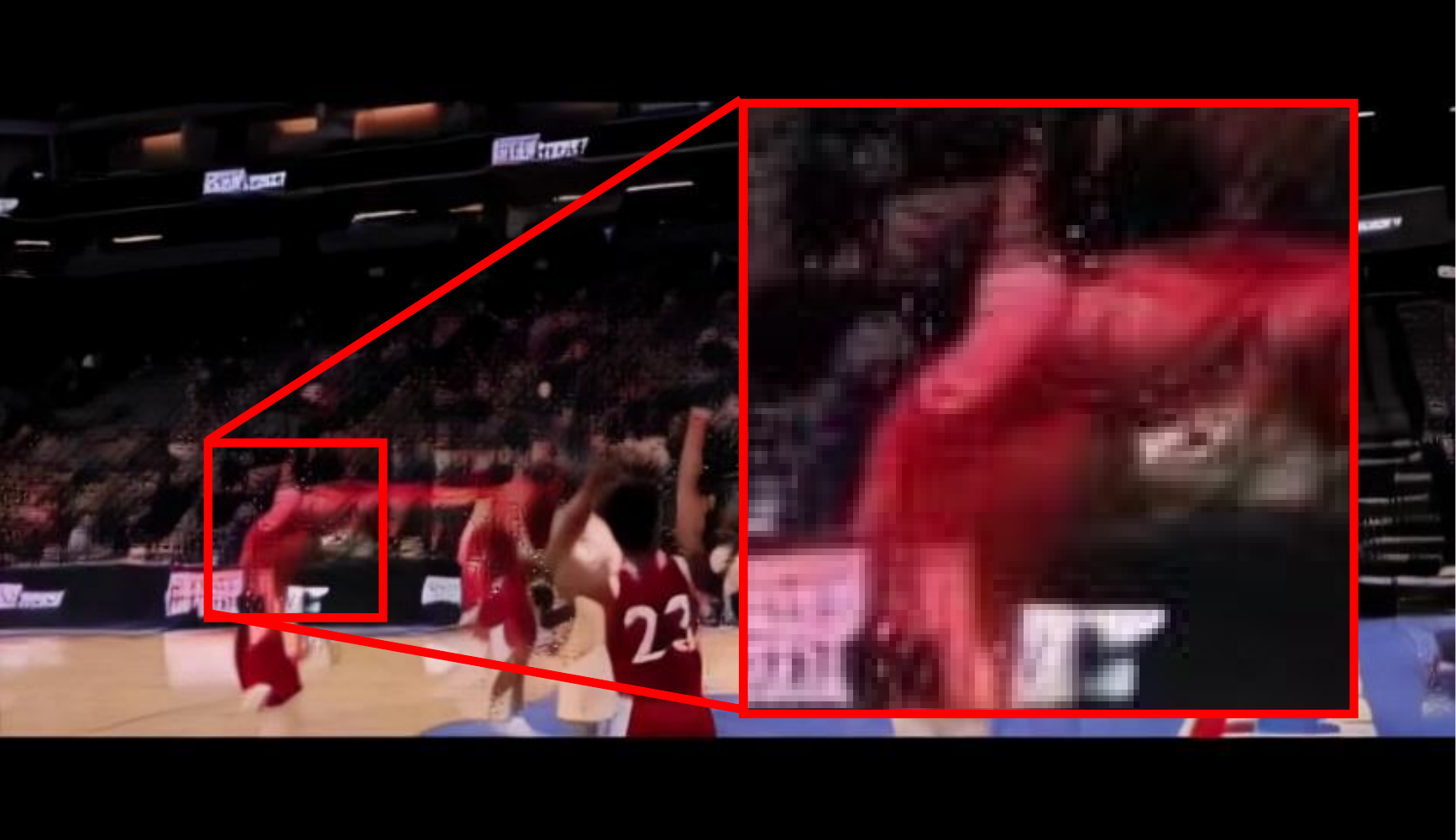} &
    \includegraphics[width=0.235\textwidth]{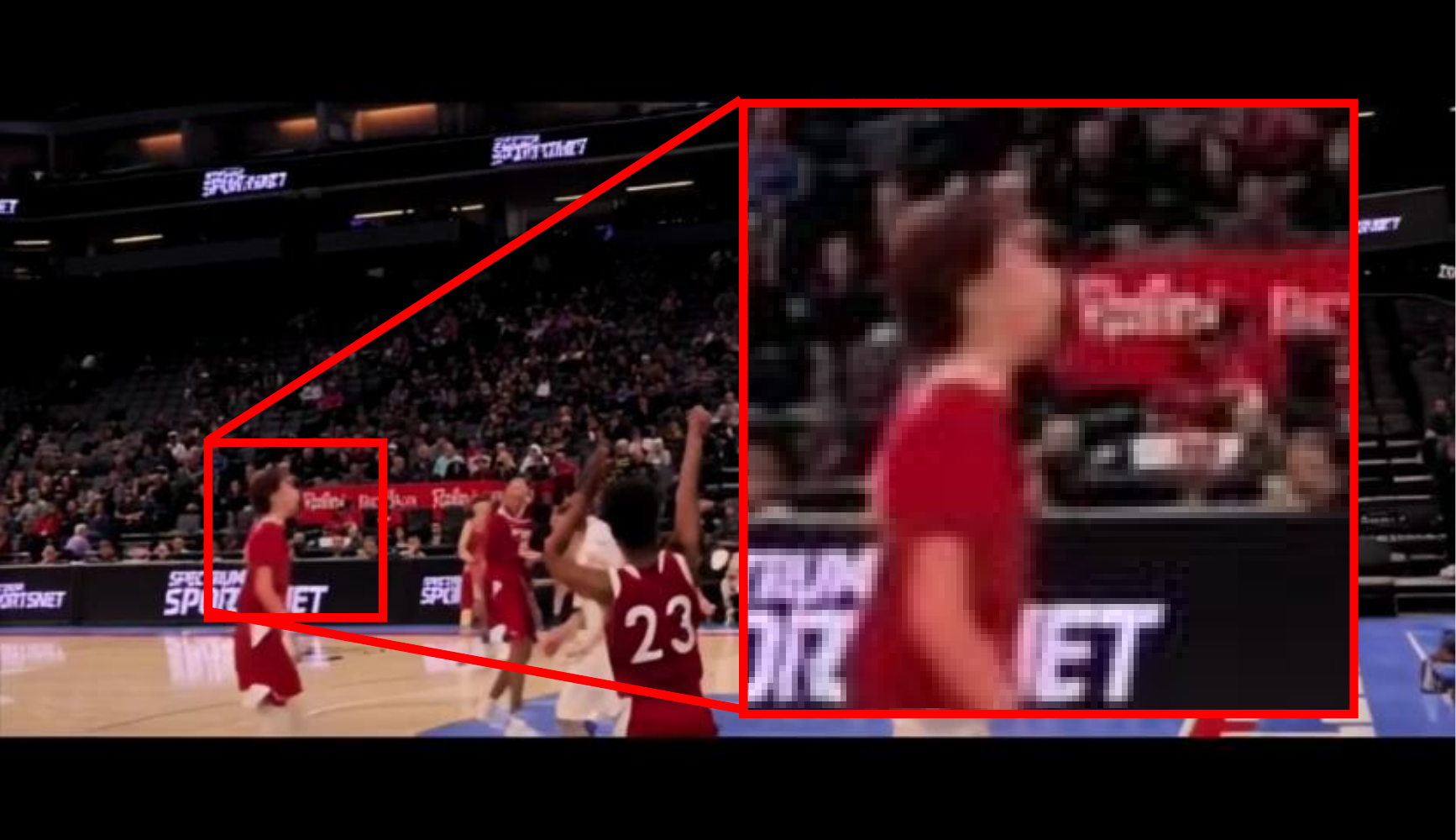} \\[2pt]
    \small{Baseline} & \small{Ours} & \small{Baseline} & \small{Ours} \\
\end{tabular}
\caption{\textbf{Additional Wan~2.1 reconstruction comparisons.} Highlighted regions are shown for baseline vs.\ \model. Our method consistently recovers sharper textures and finer details across diverse scenes.}
\label{fig:supp_wan_recon}
\end{figure*}

\begin{figure*}[t]
\centering
\setlength{\tabcolsep}{1pt}
\renewcommand{\arraystretch}{0.5}
\begin{tabular}{cccc}
    \includegraphics[width=0.235\textwidth]{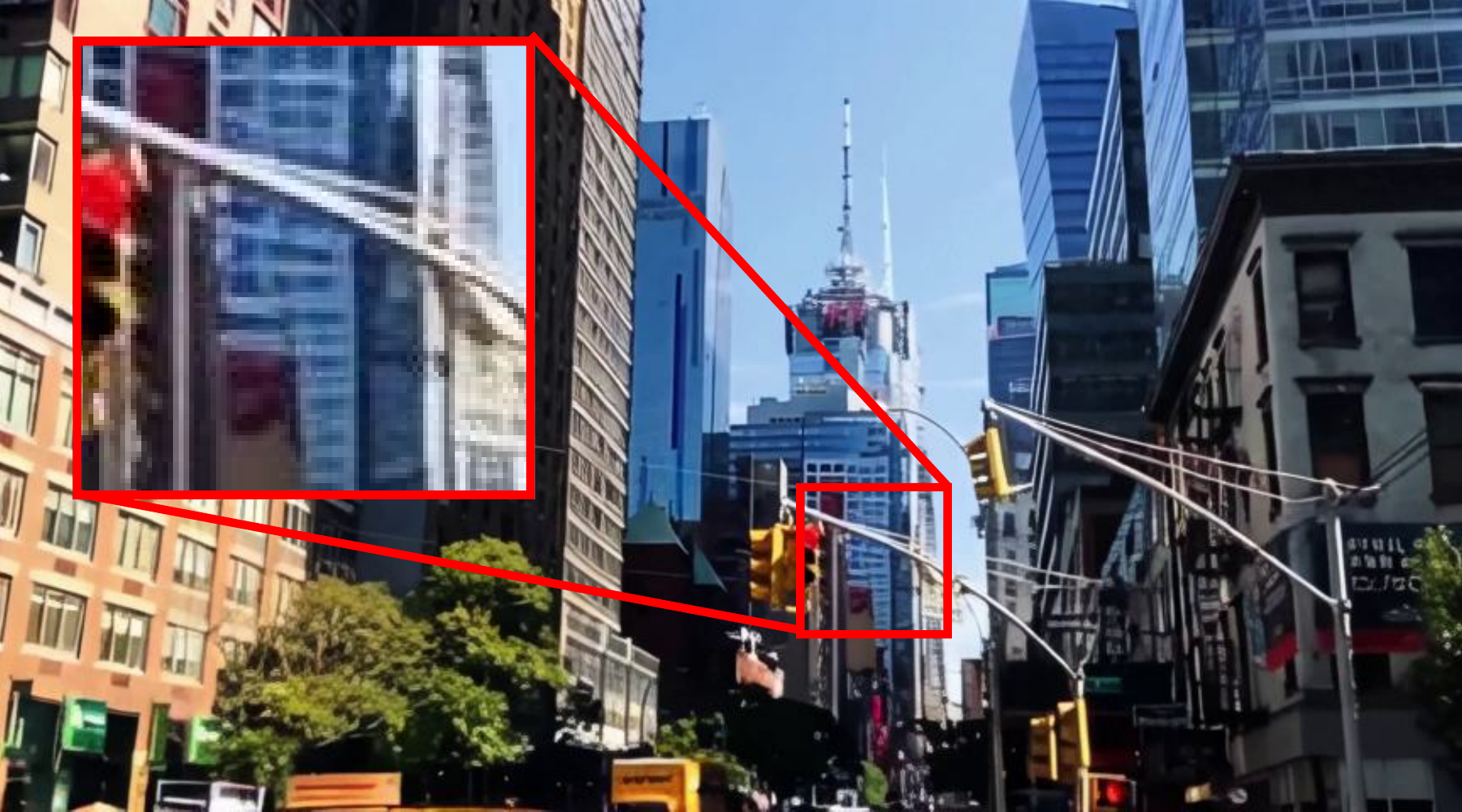} &
    \includegraphics[width=0.235\textwidth]{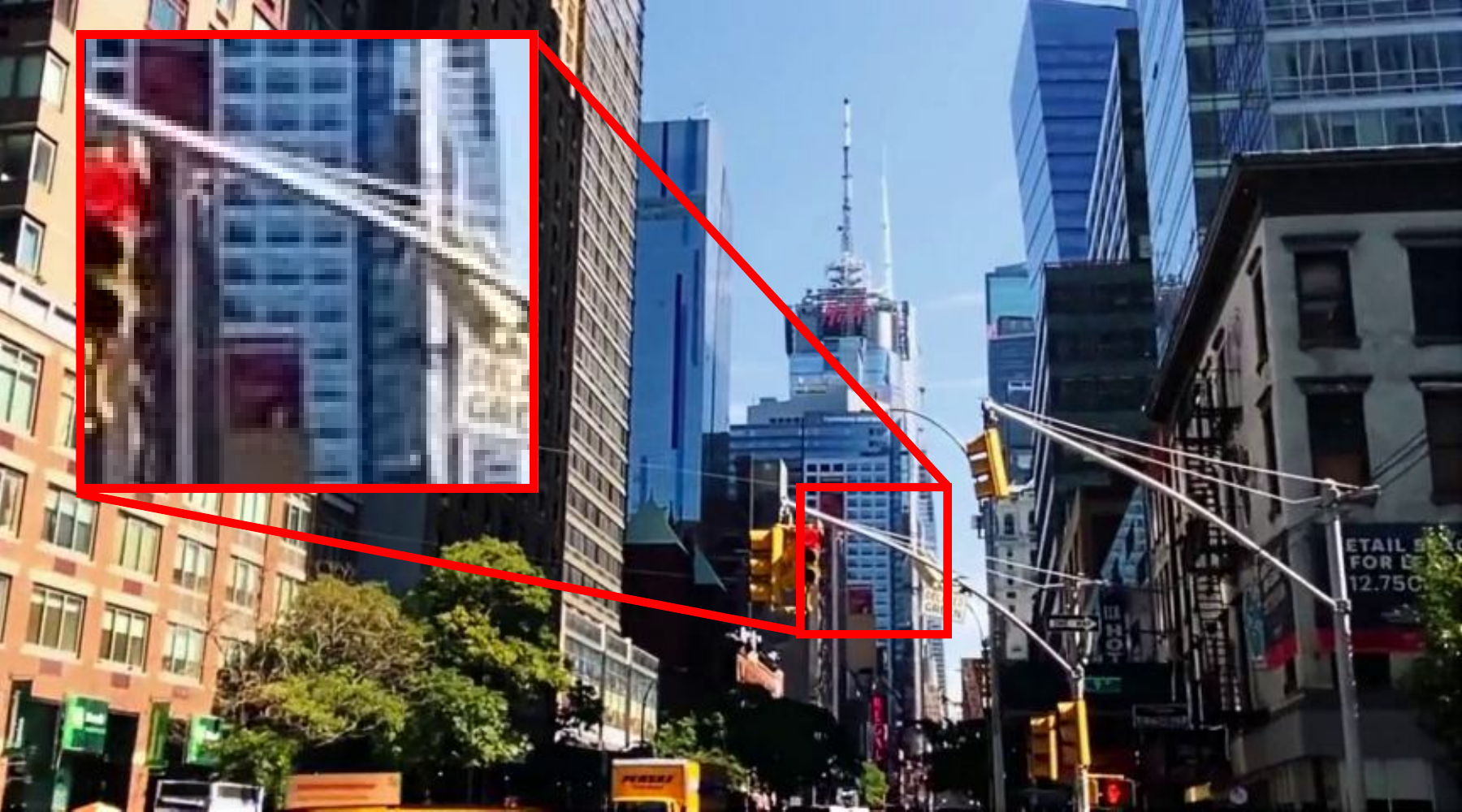} &
    \includegraphics[width=0.235\textwidth]{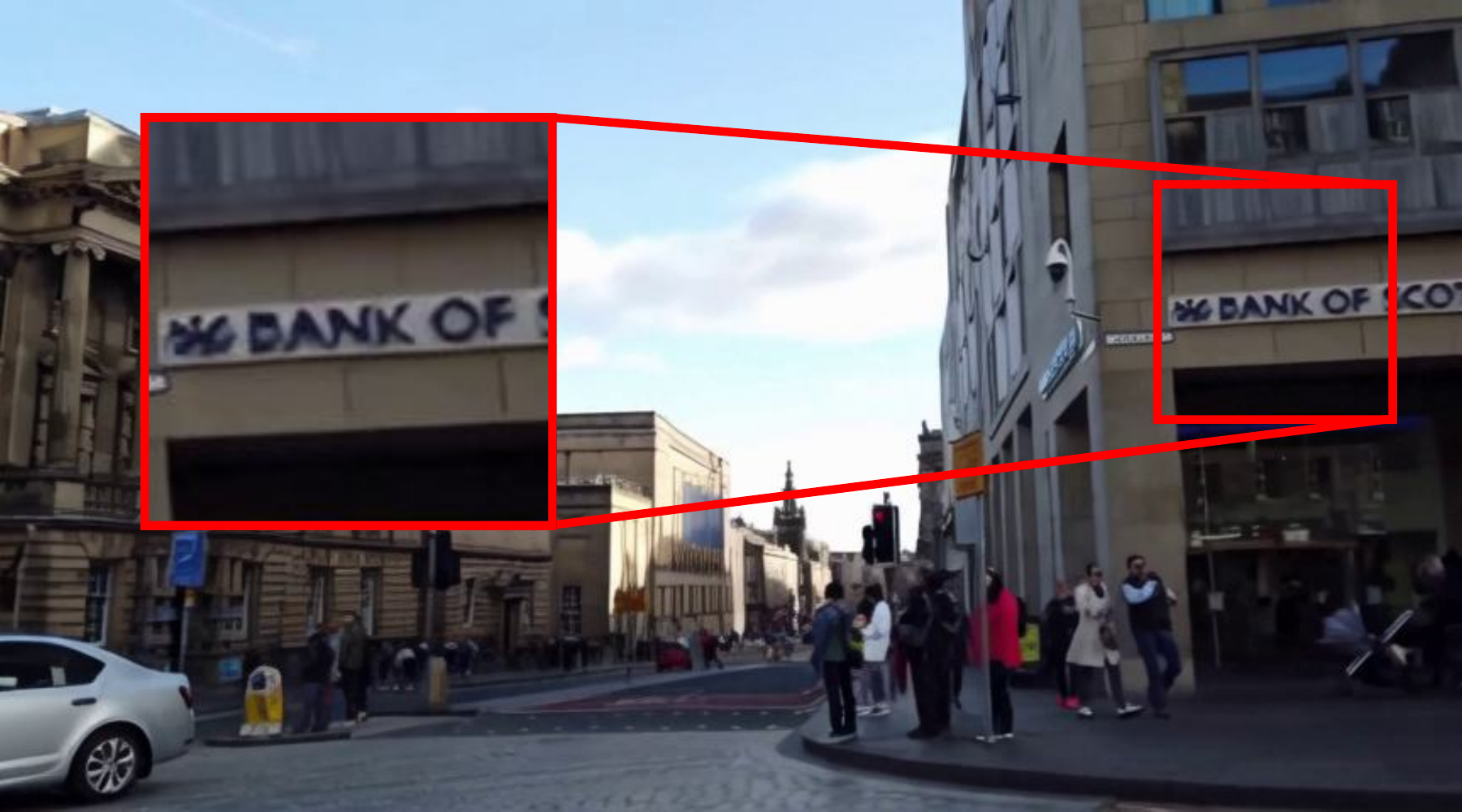} &
    \includegraphics[width=0.235\textwidth]{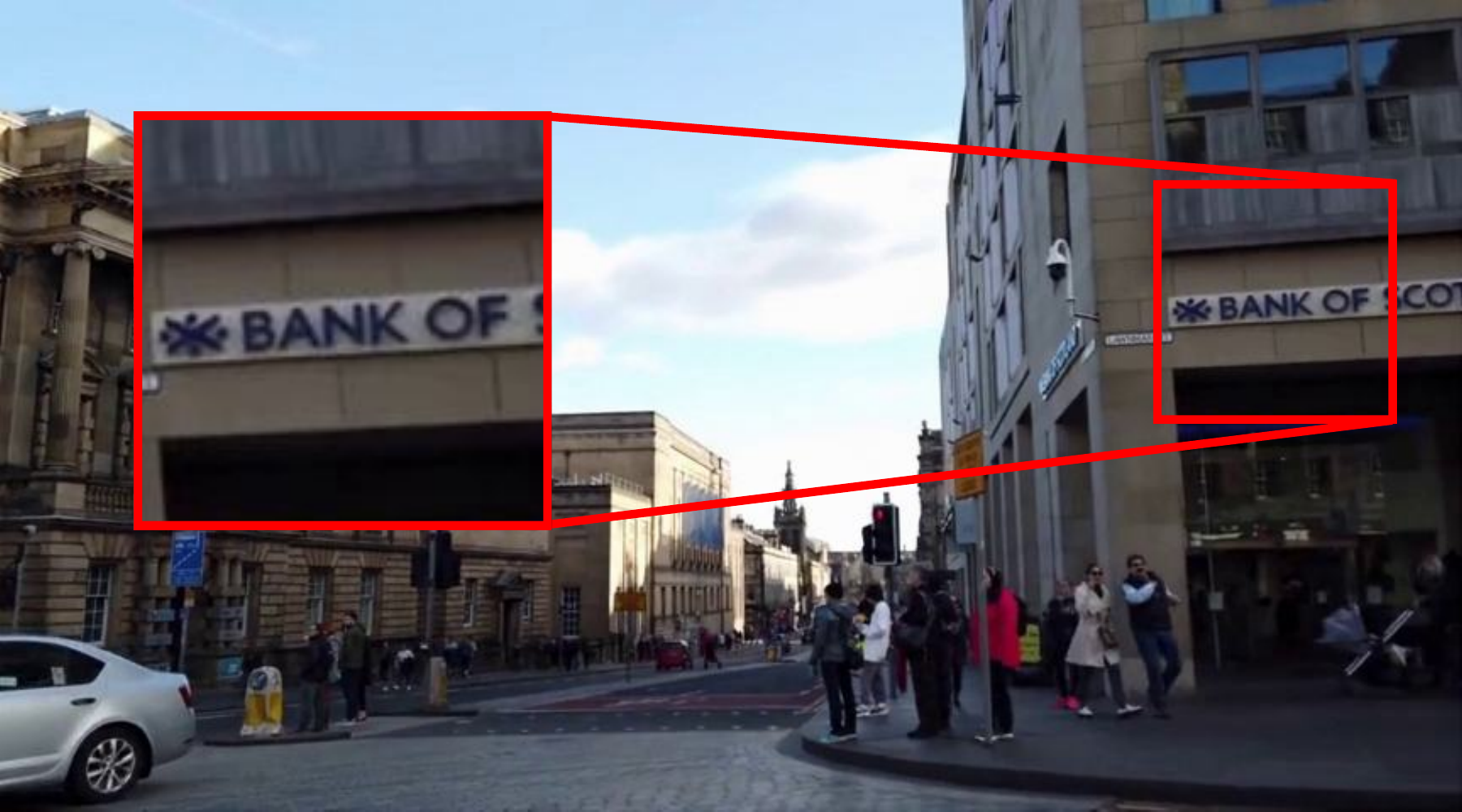} \\
    \includegraphics[width=0.235\textwidth]{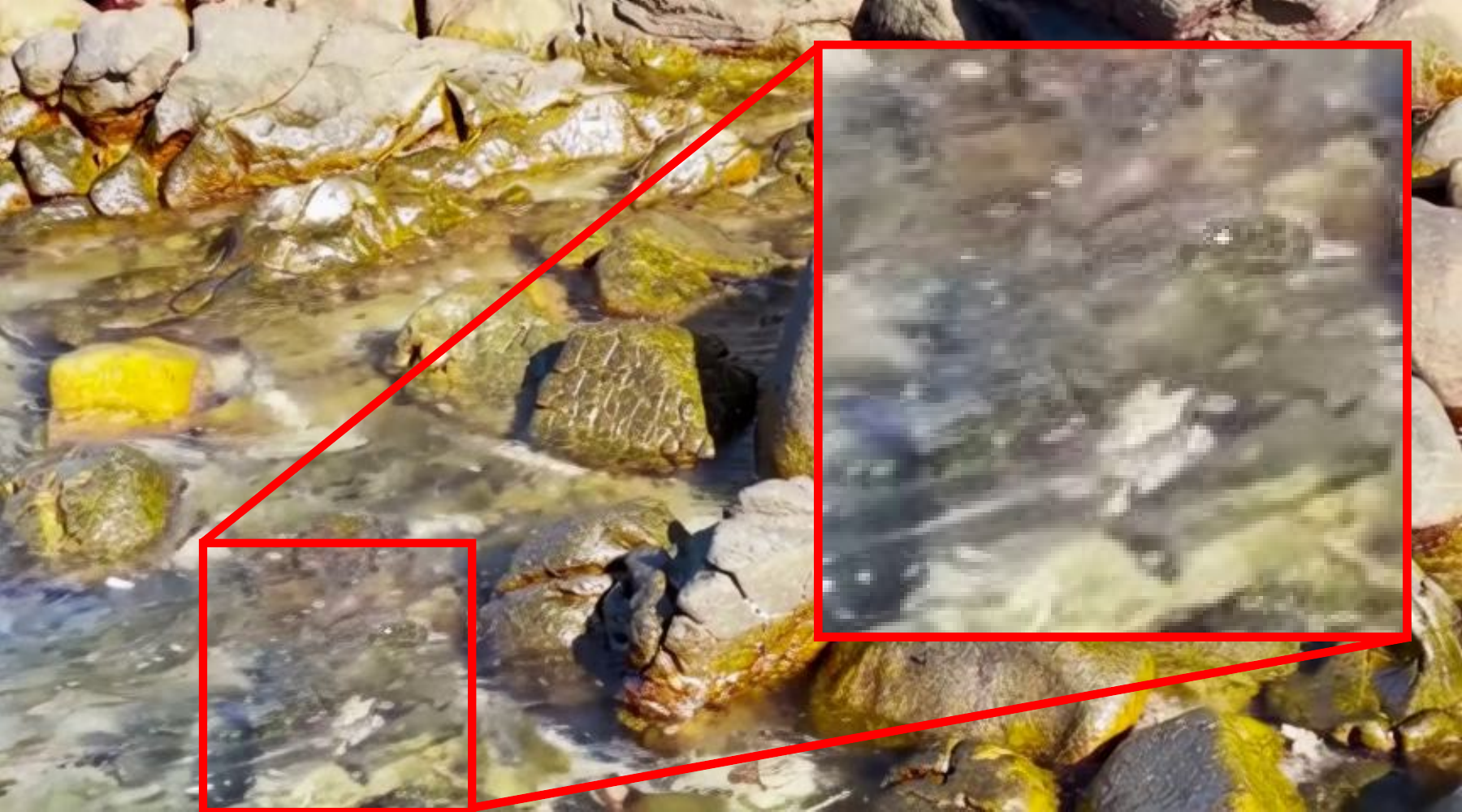} &
    \includegraphics[width=0.235\textwidth]{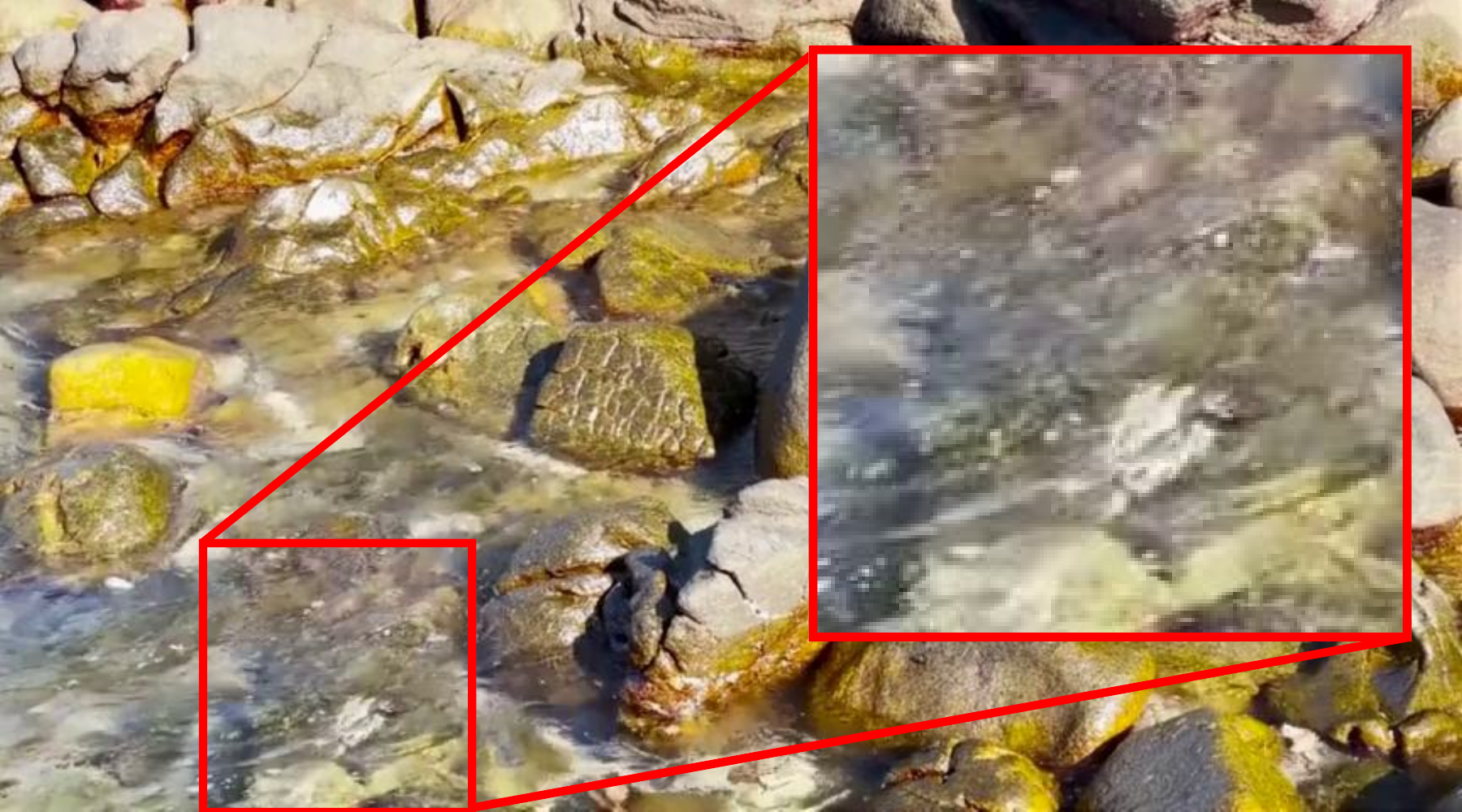} &
    \includegraphics[width=0.235\textwidth]{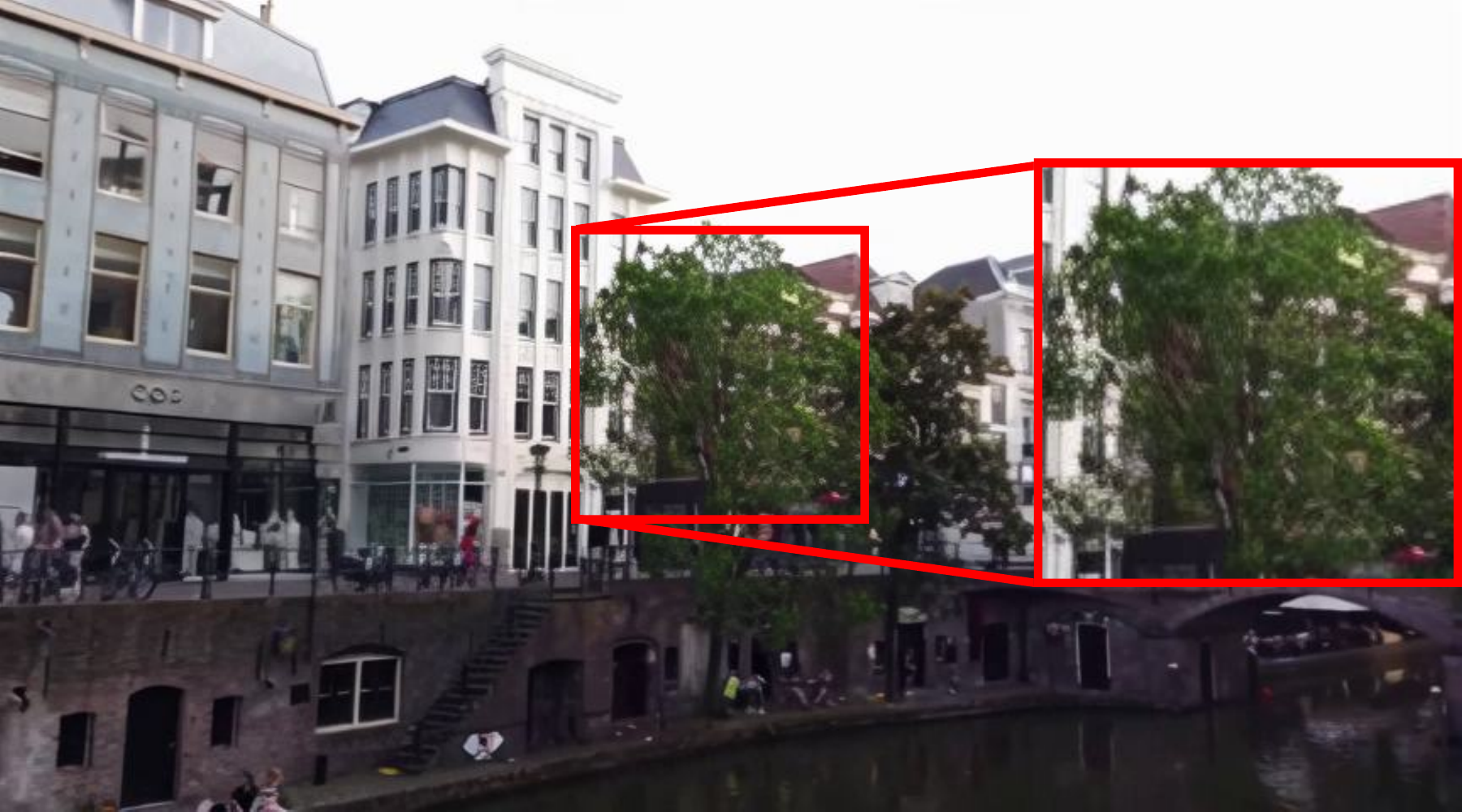} &
    \includegraphics[width=0.235\textwidth]{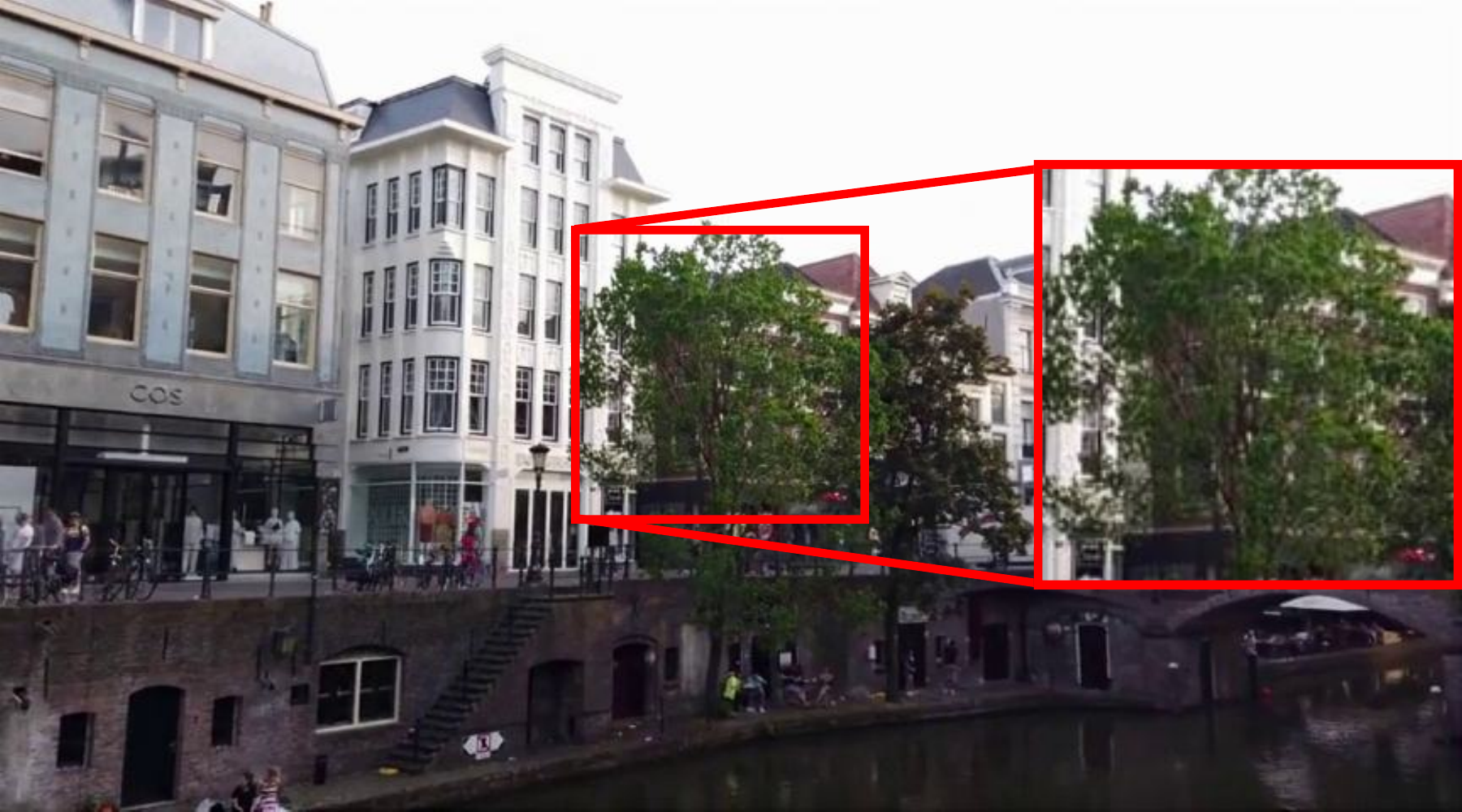} \\
    \includegraphics[width=0.235\textwidth]{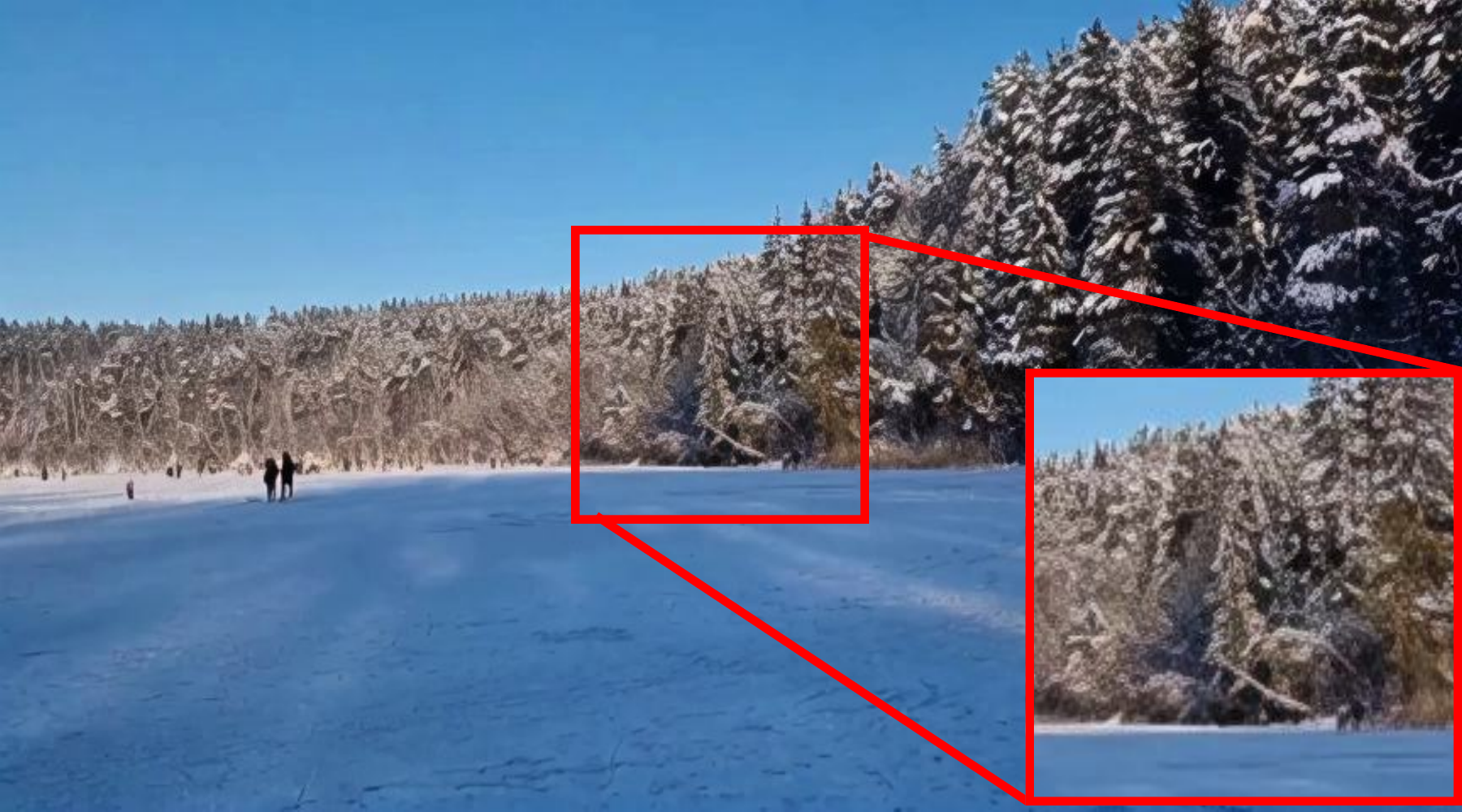} &
    \includegraphics[width=0.235\textwidth]{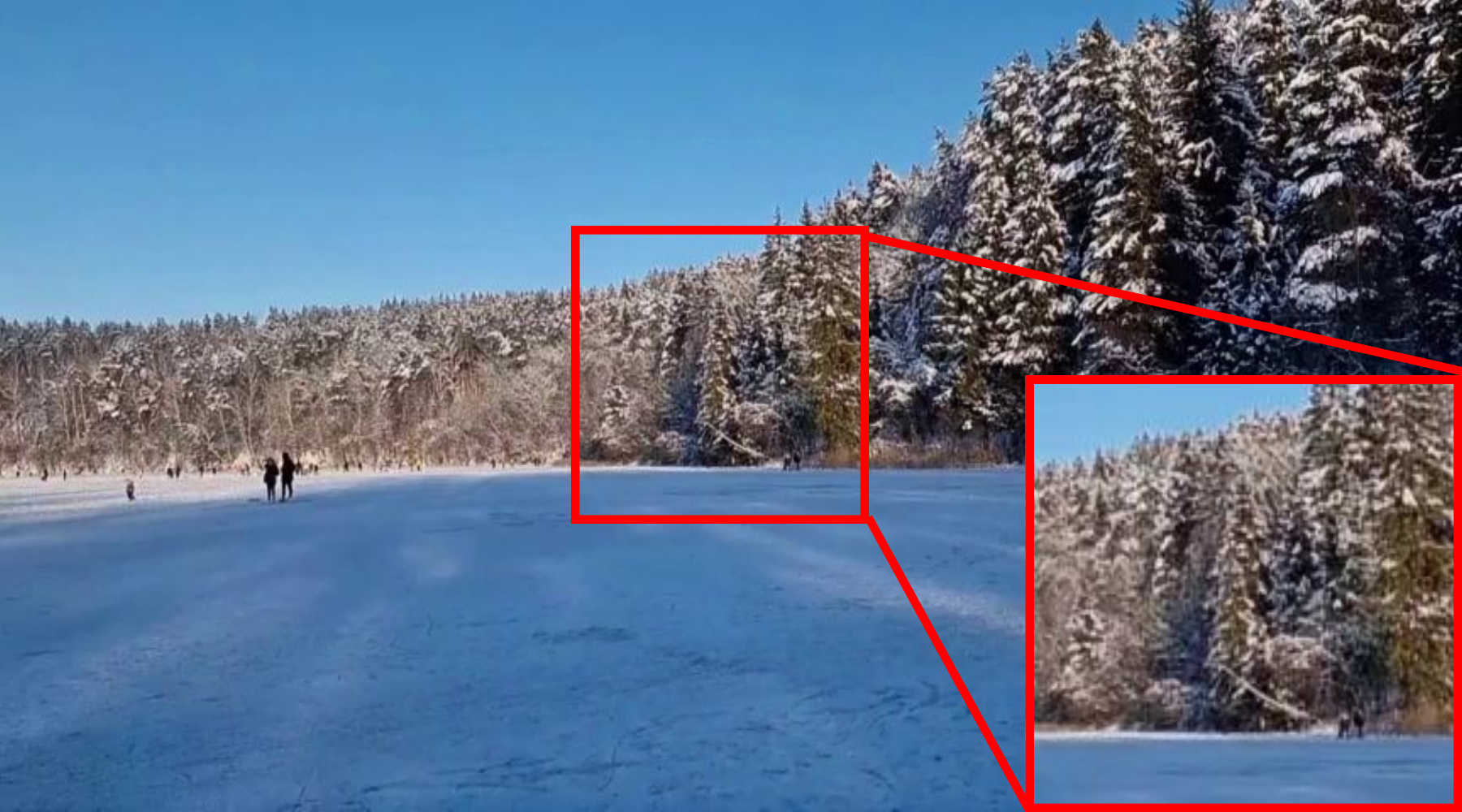} &
    \includegraphics[width=0.235\textwidth]{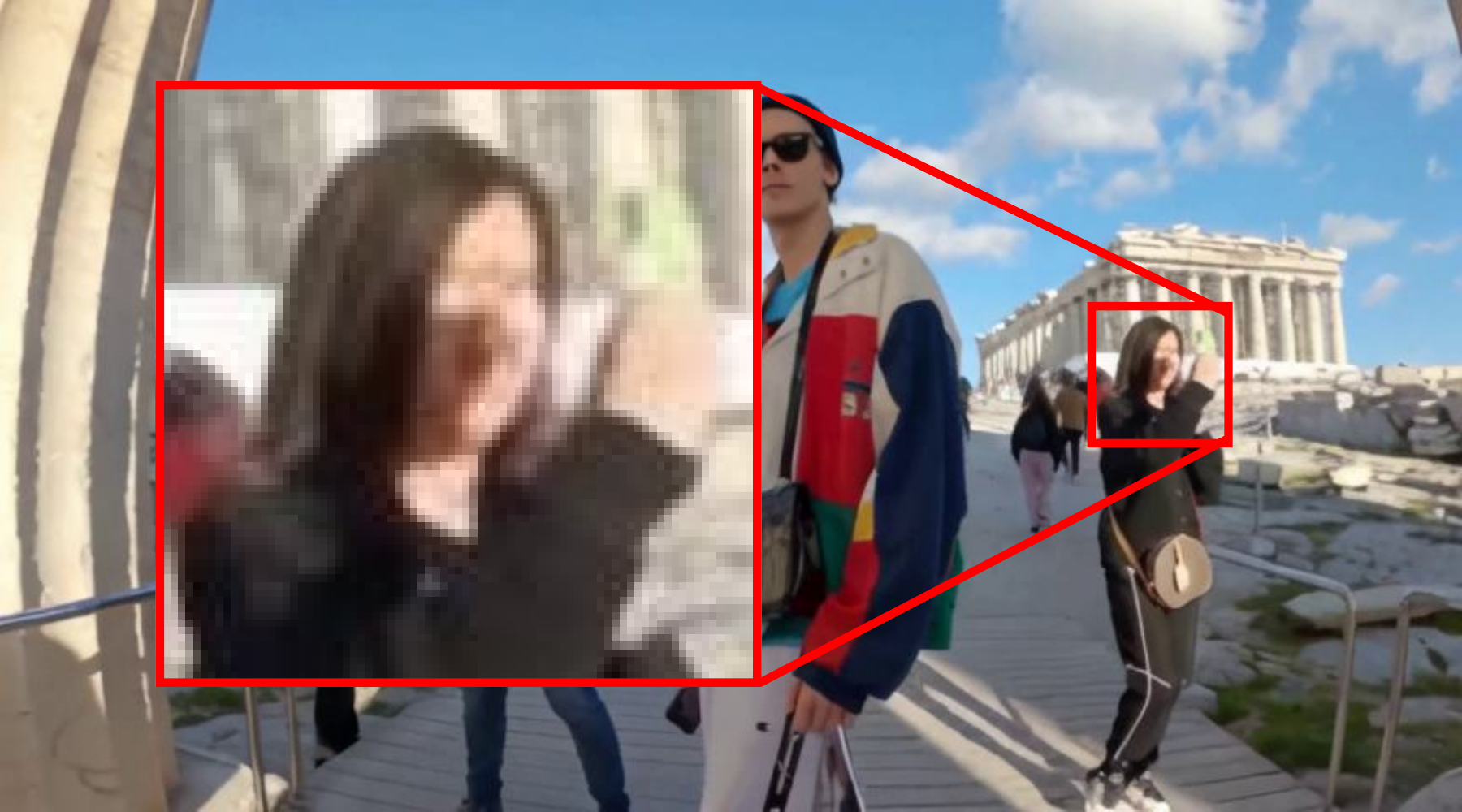} &
    \includegraphics[width=0.235\textwidth]{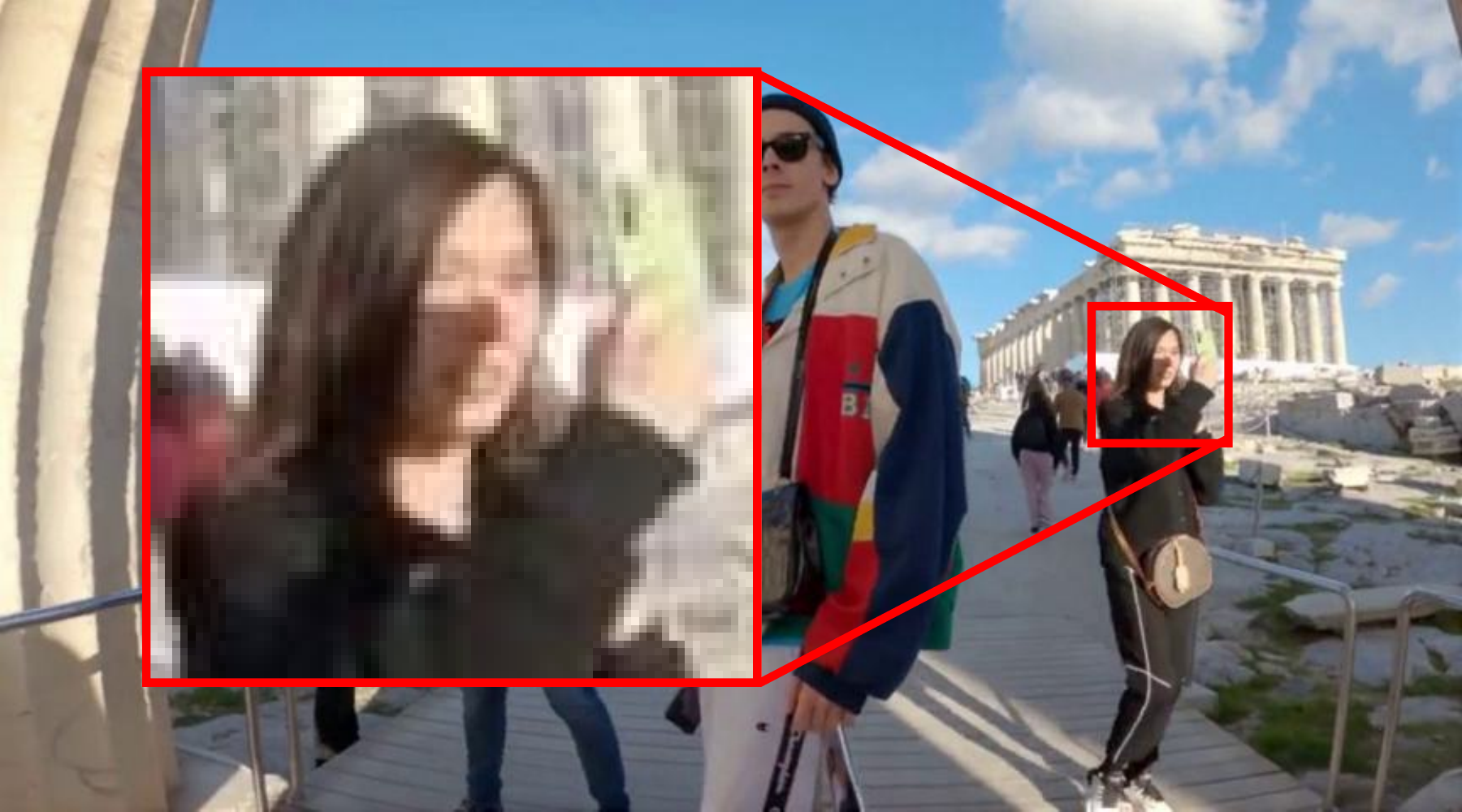} \\
    \includegraphics[width=0.235\textwidth]{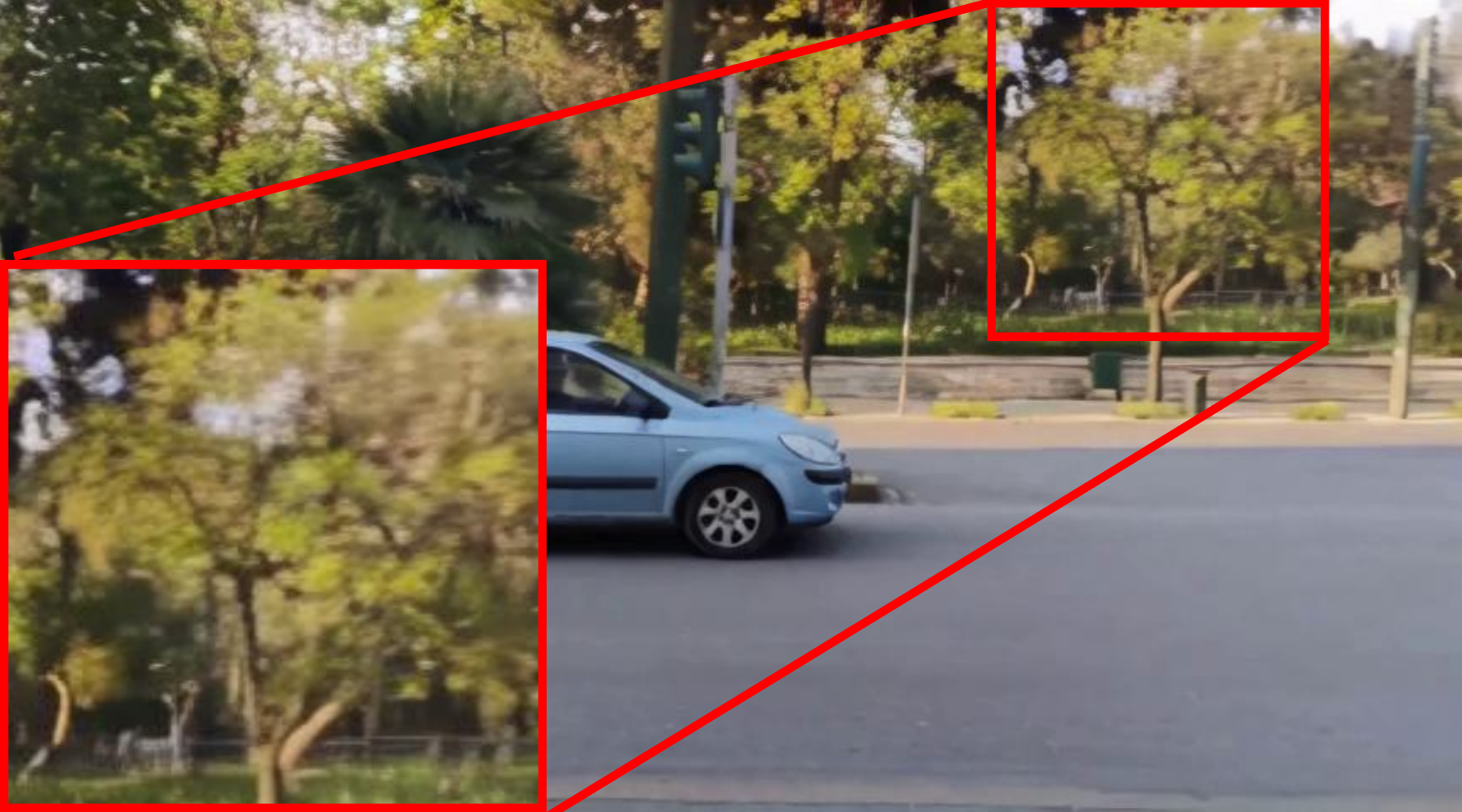} &
    \includegraphics[width=0.235\textwidth]{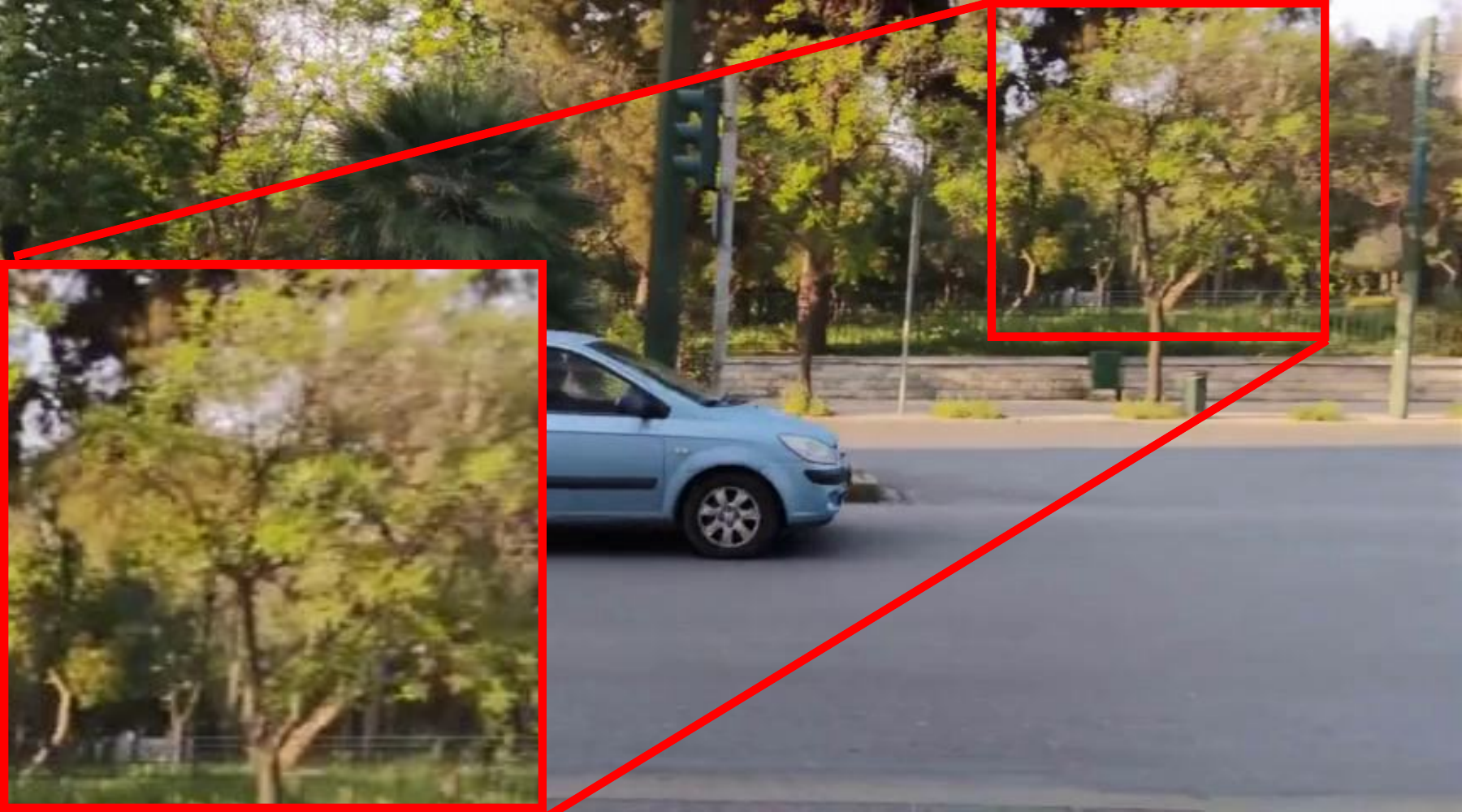} &
    \includegraphics[width=0.235\textwidth]{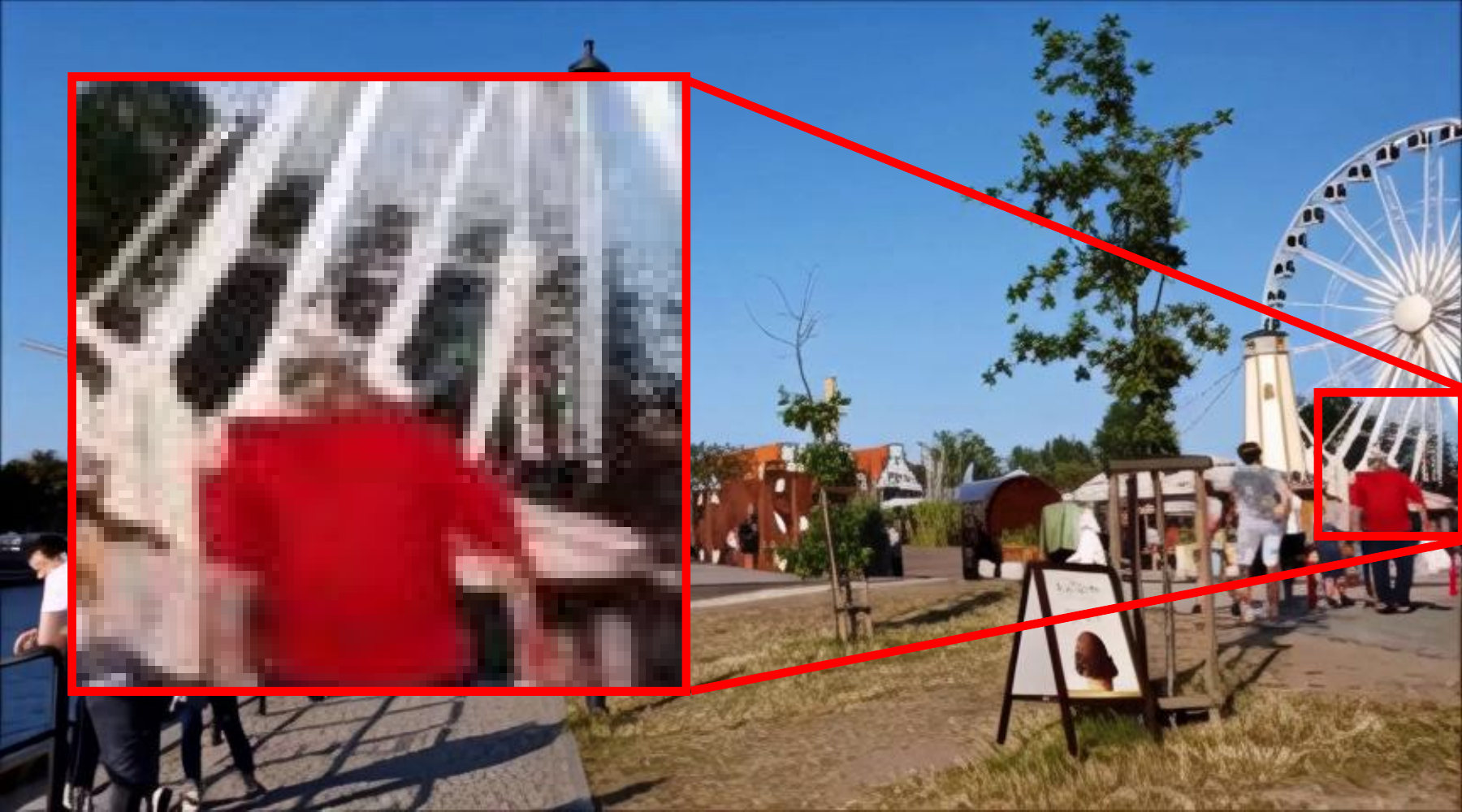} &
    \includegraphics[width=0.235\textwidth]{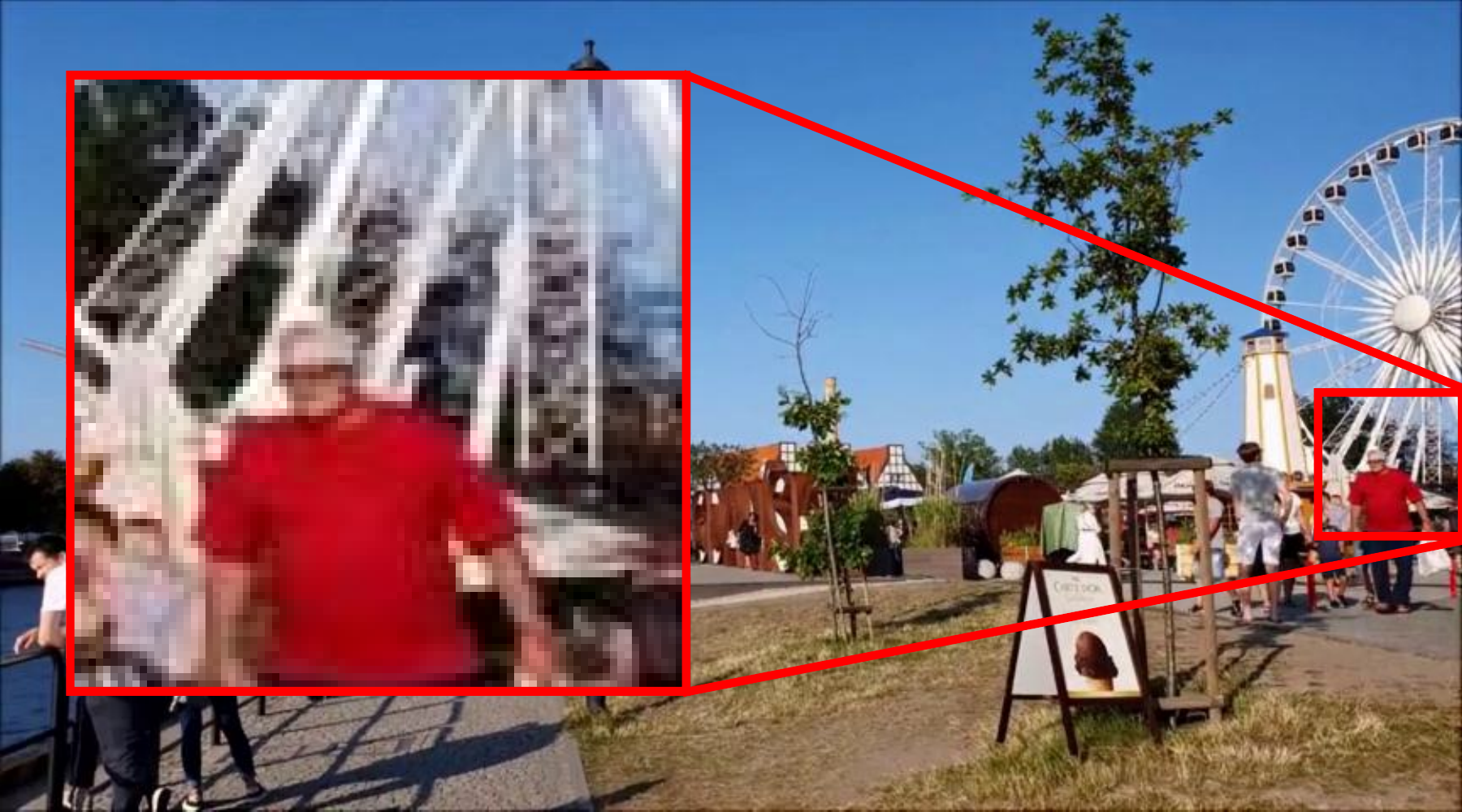} \\
    \includegraphics[width=0.235\textwidth]{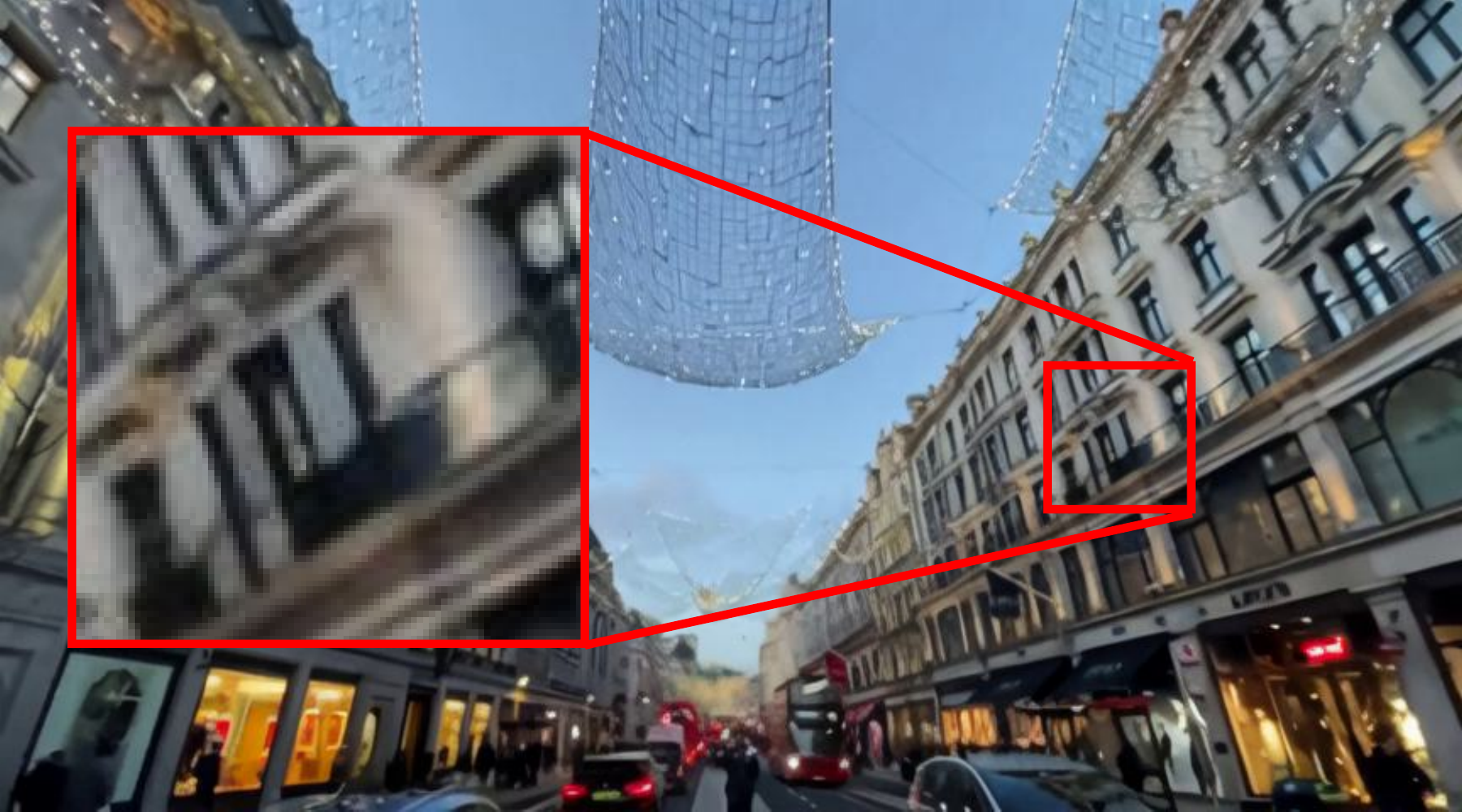} &
    \includegraphics[width=0.235\textwidth]{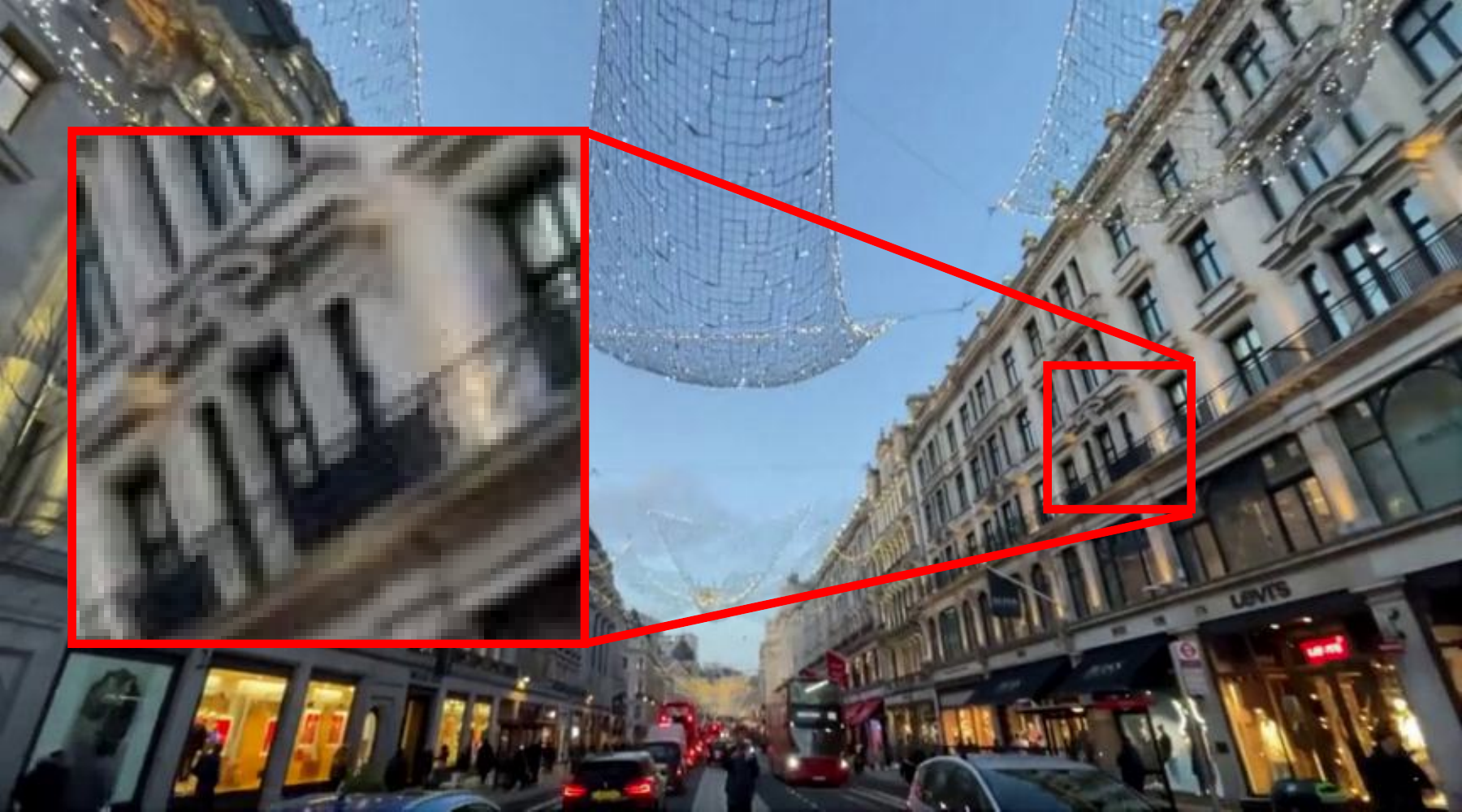} &
    \includegraphics[width=0.235\textwidth]{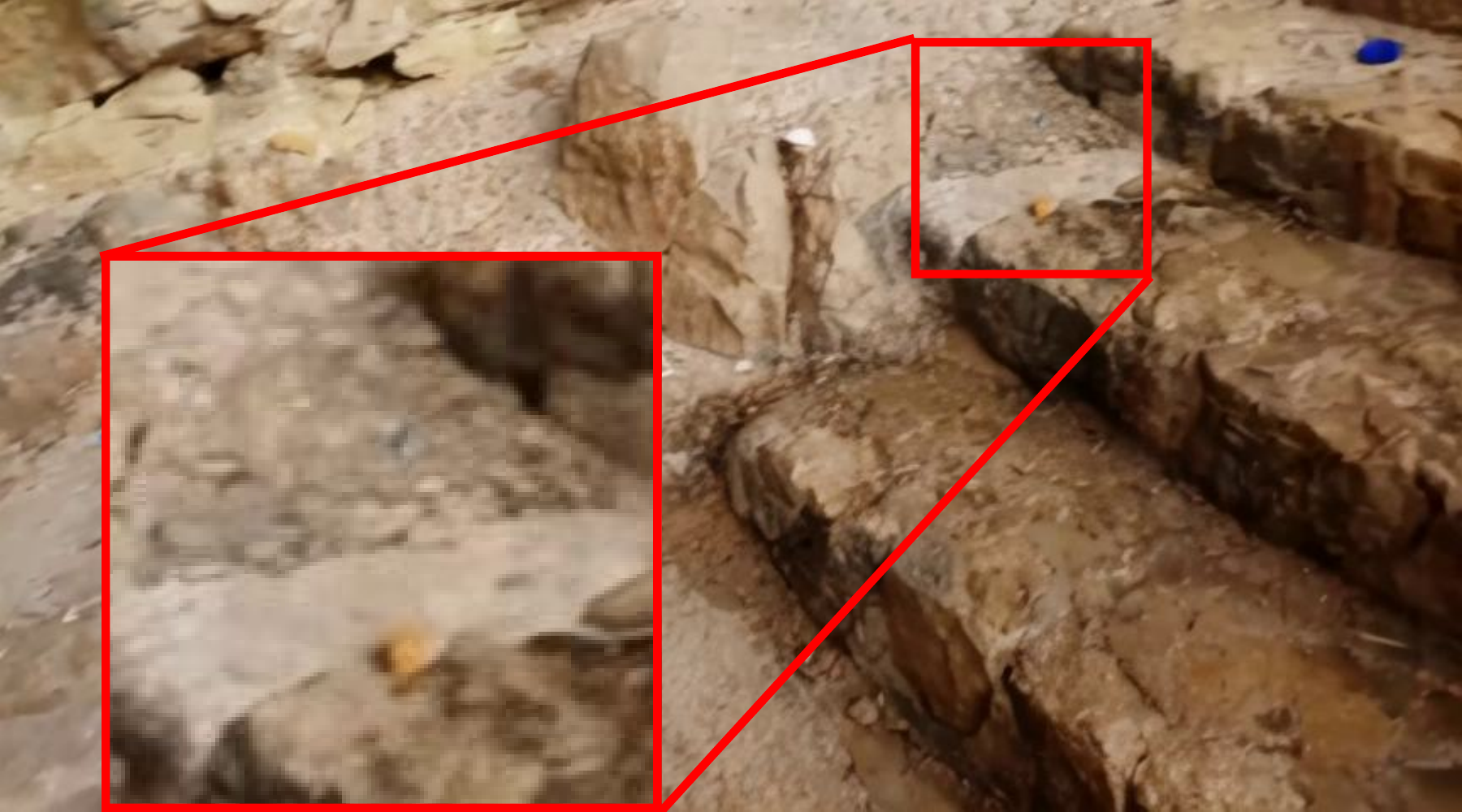} &
    \includegraphics[width=0.235\textwidth]{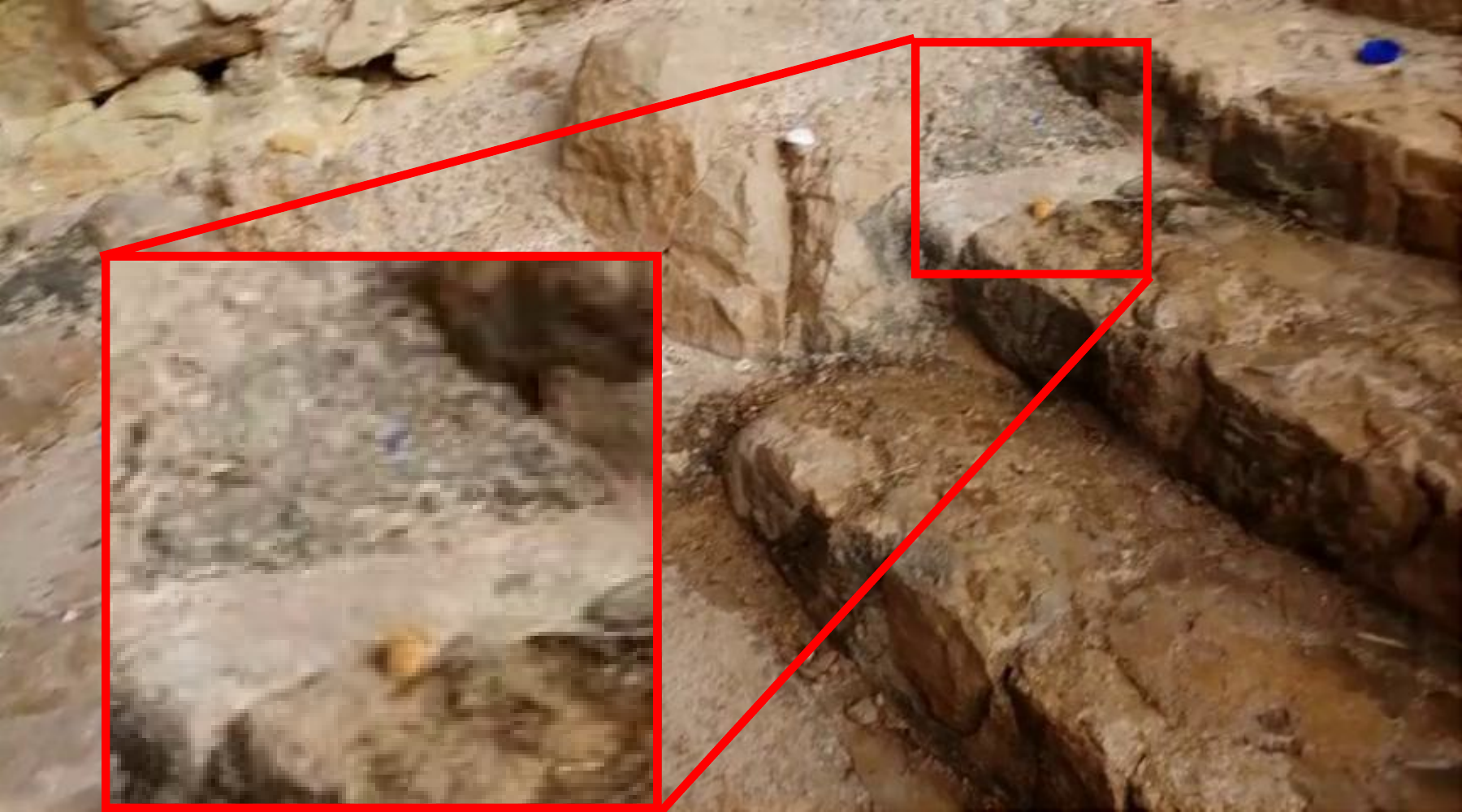} \\
    \includegraphics[width=0.235\textwidth]{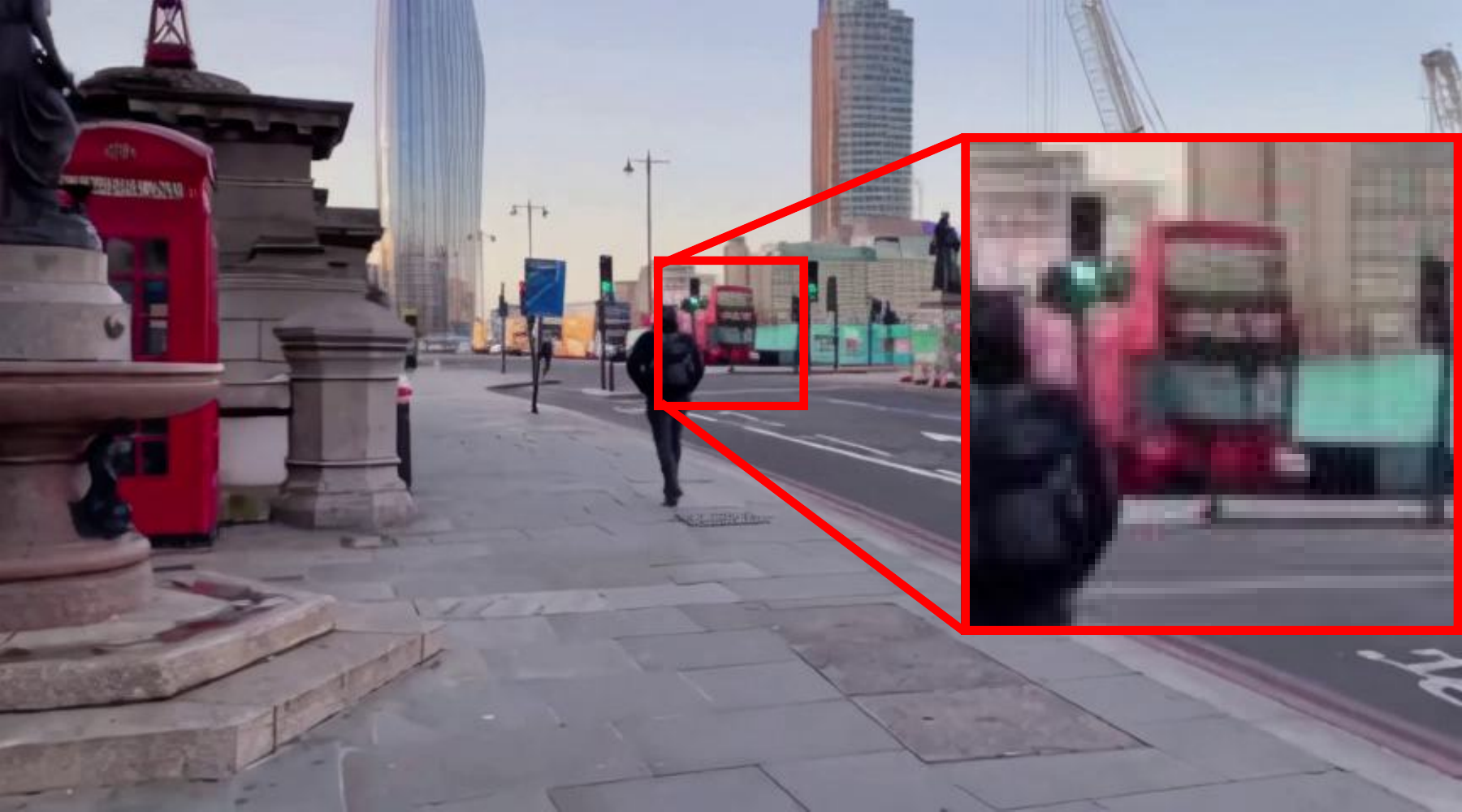} &
    \includegraphics[width=0.235\textwidth]{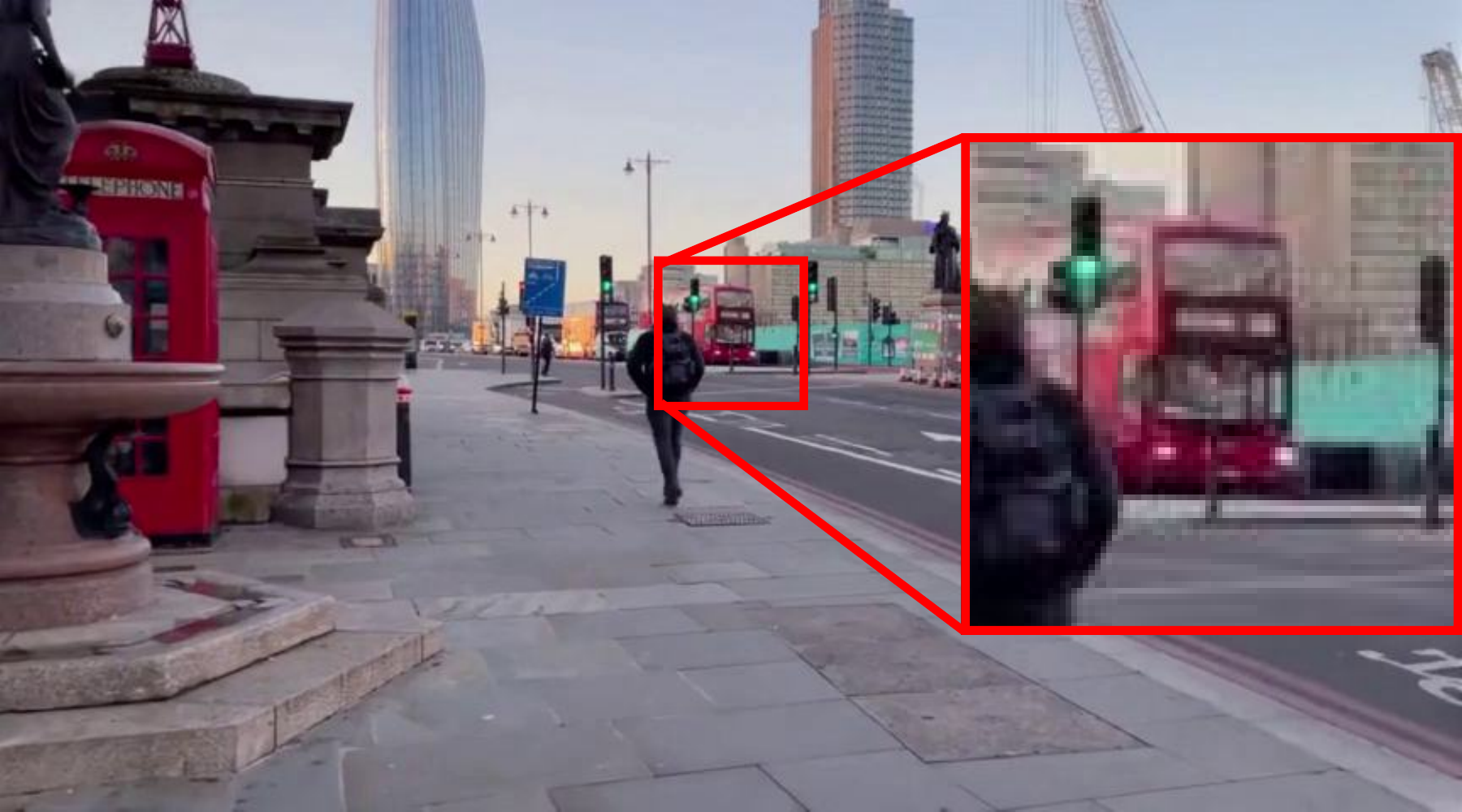} &
    \includegraphics[width=0.235\textwidth]{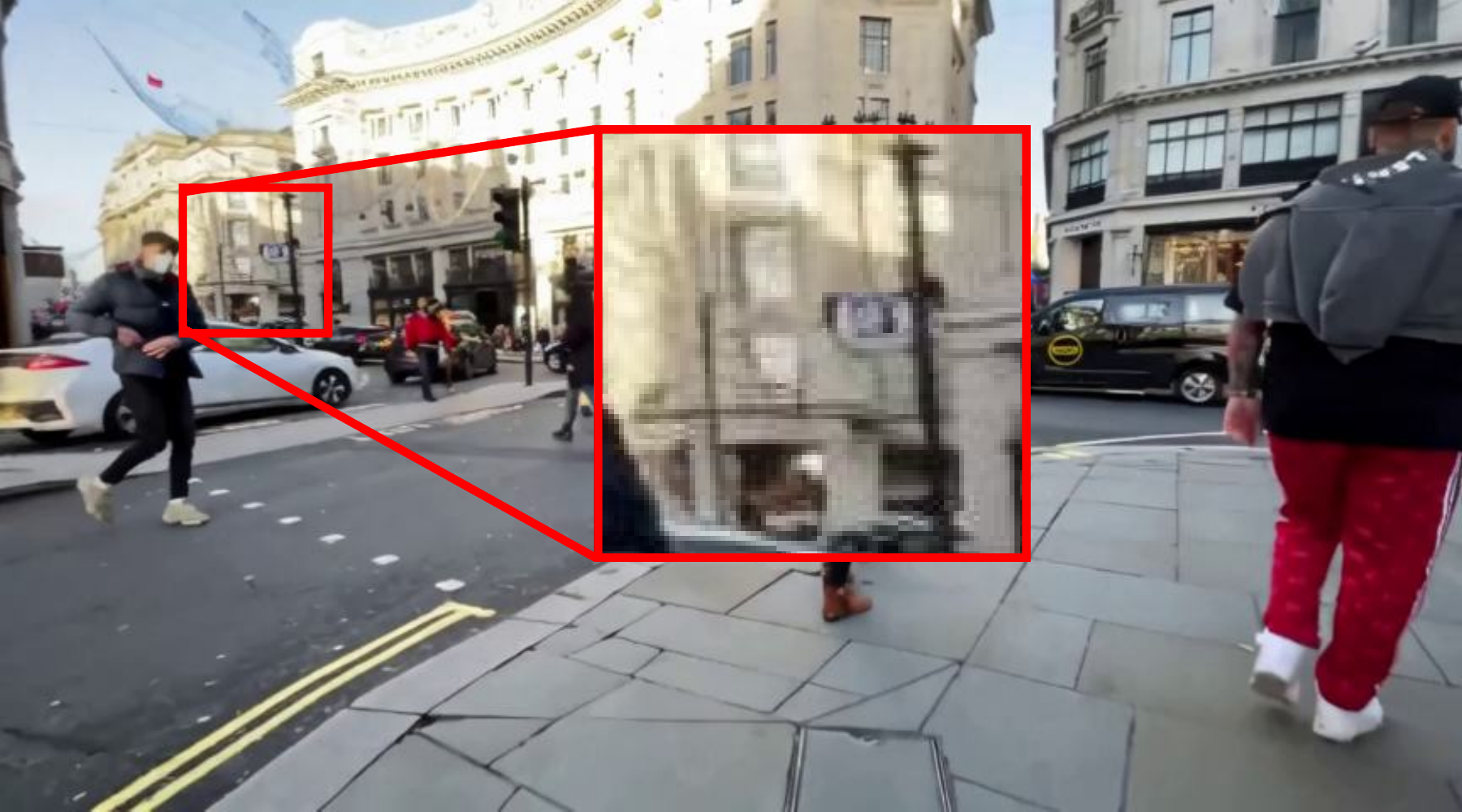} &
    \includegraphics[width=0.235\textwidth]{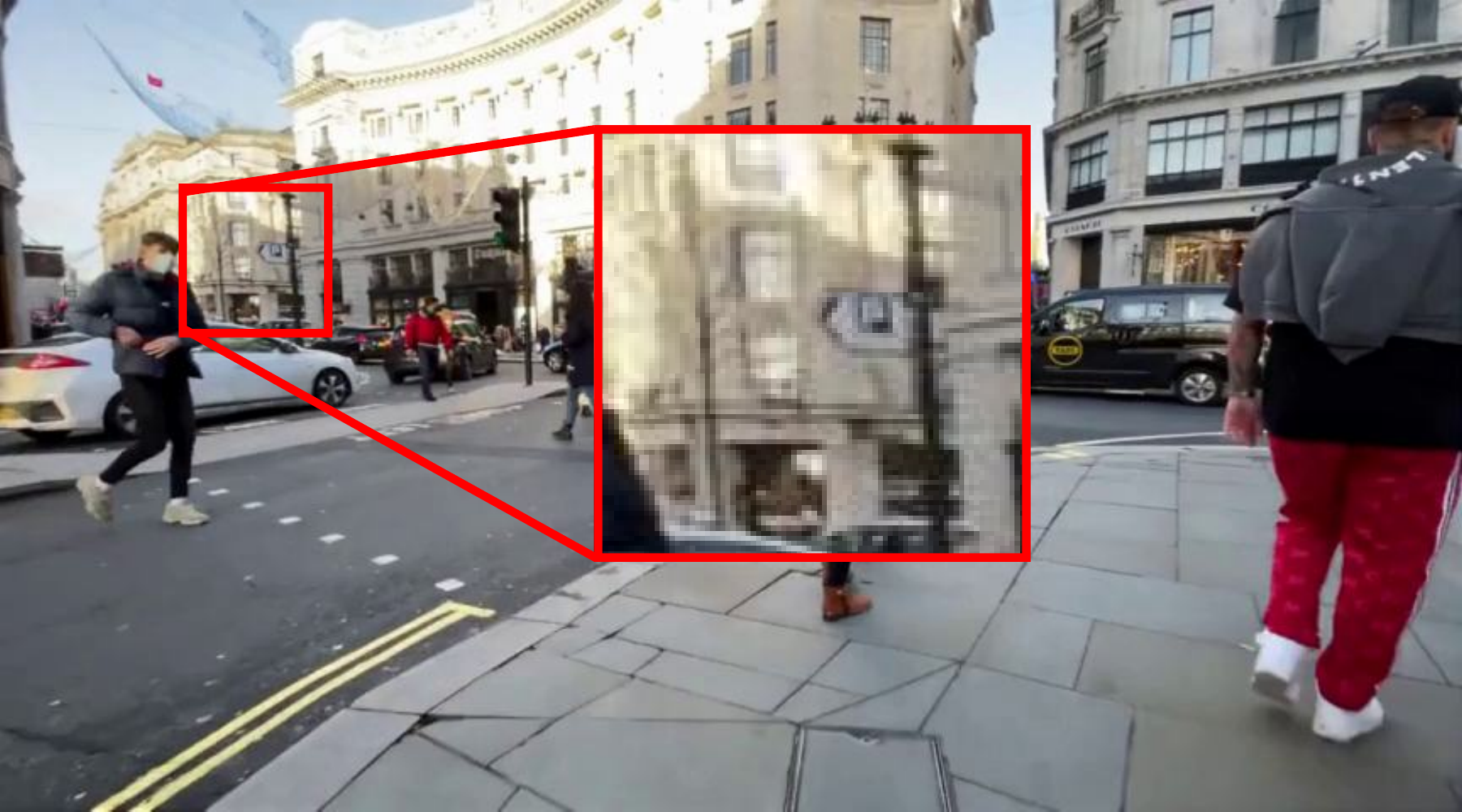} \\
    \includegraphics[width=0.235\textwidth]{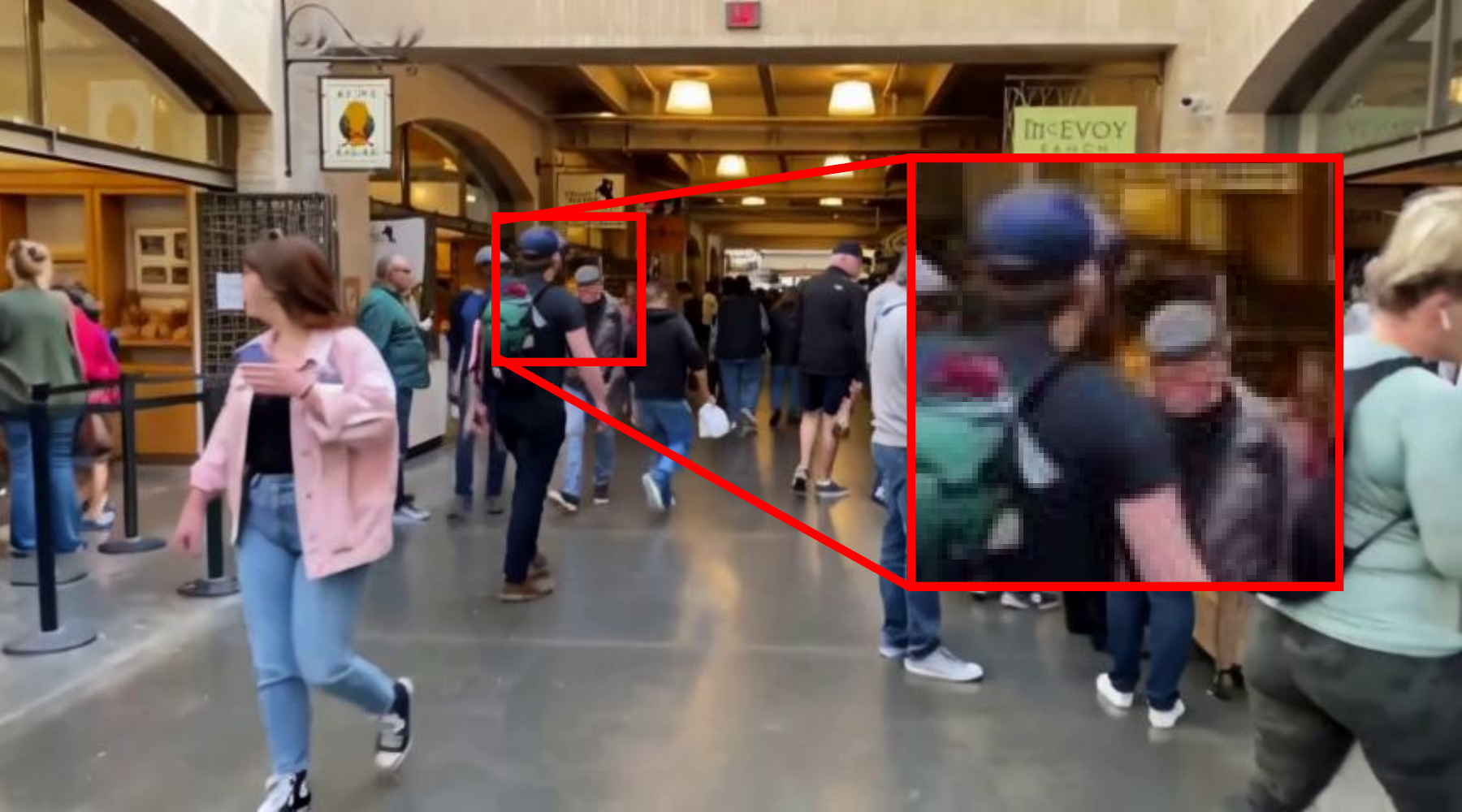} &
    \includegraphics[width=0.235\textwidth]{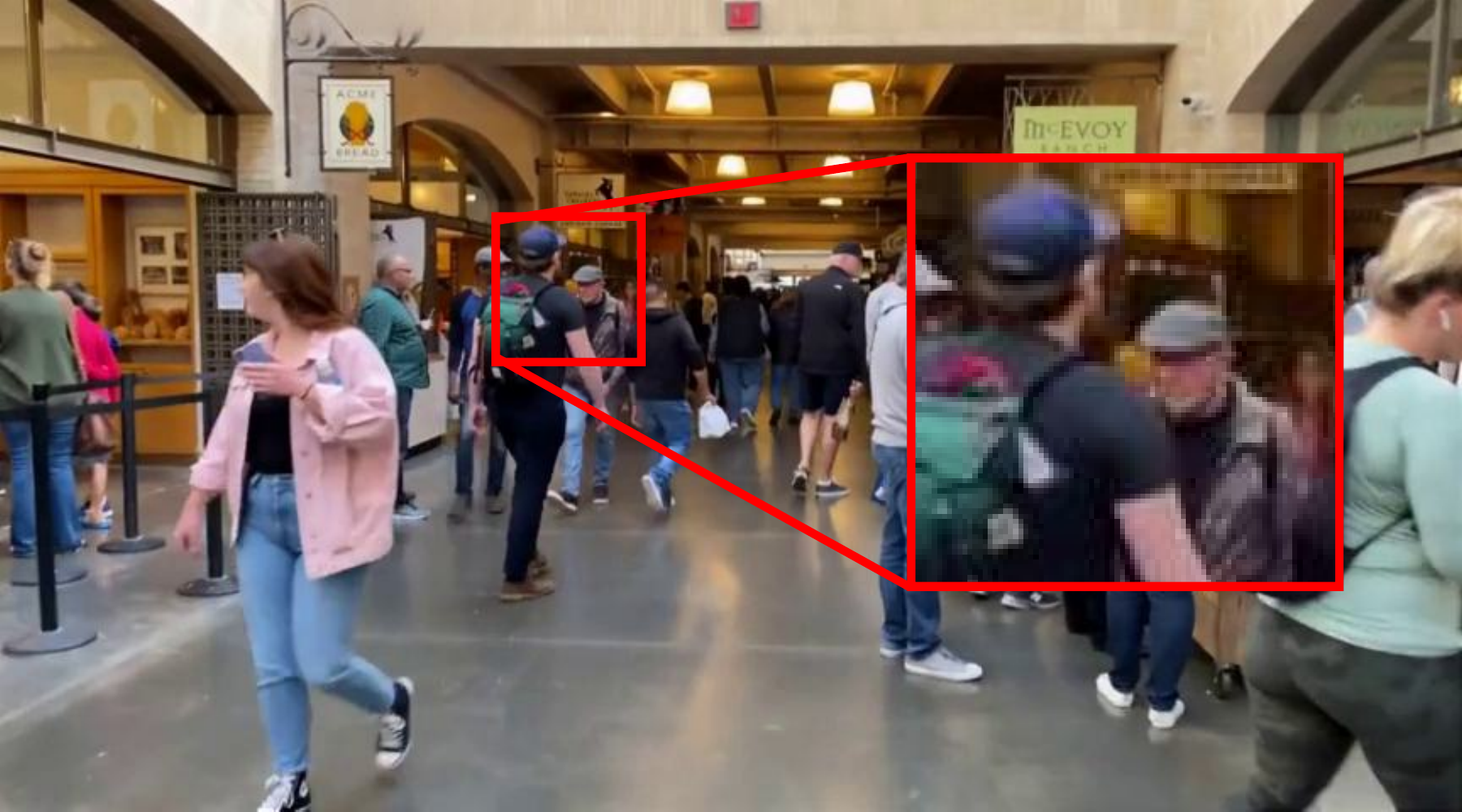} &
    \includegraphics[width=0.235\textwidth]{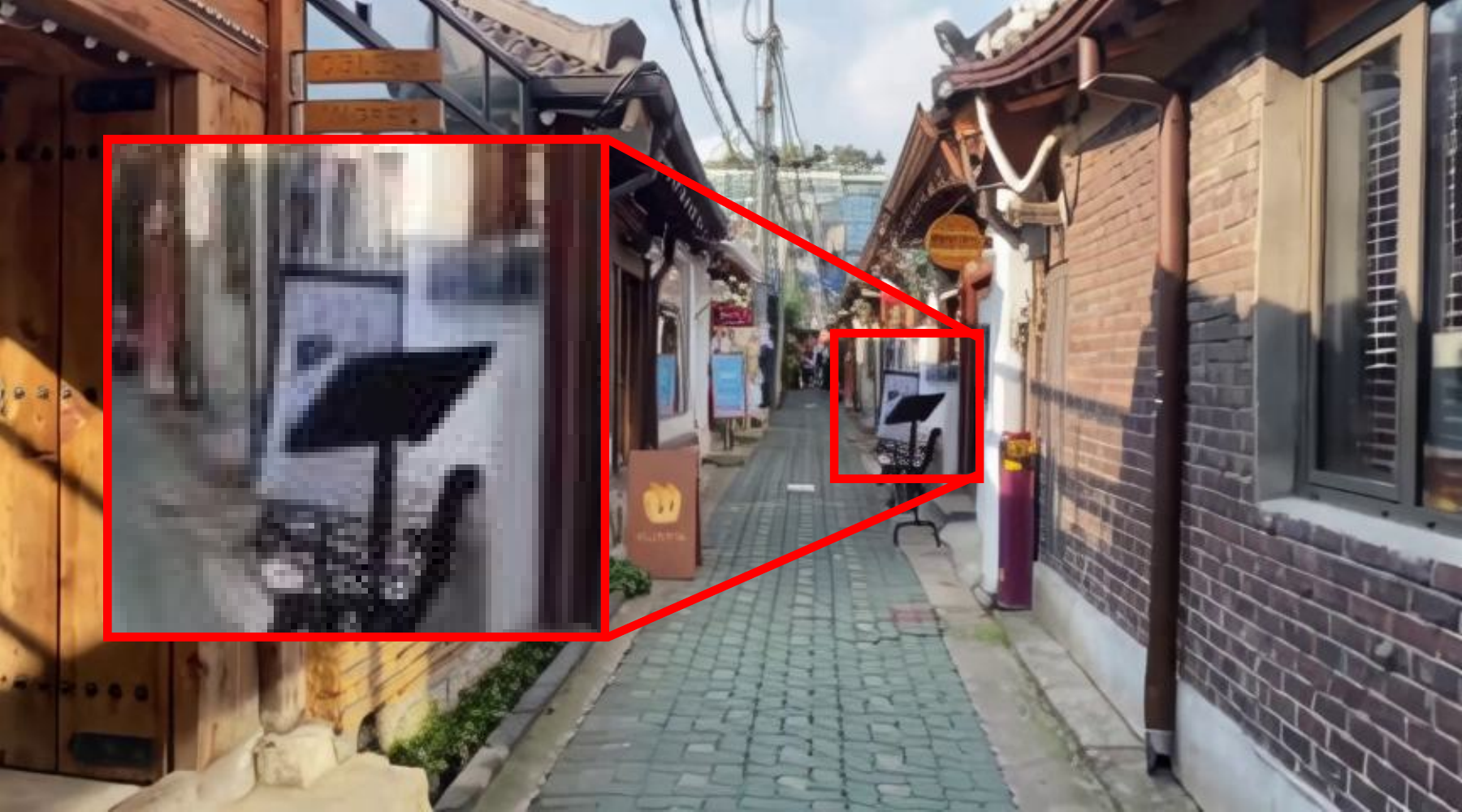} &
    \includegraphics[width=0.235\textwidth]{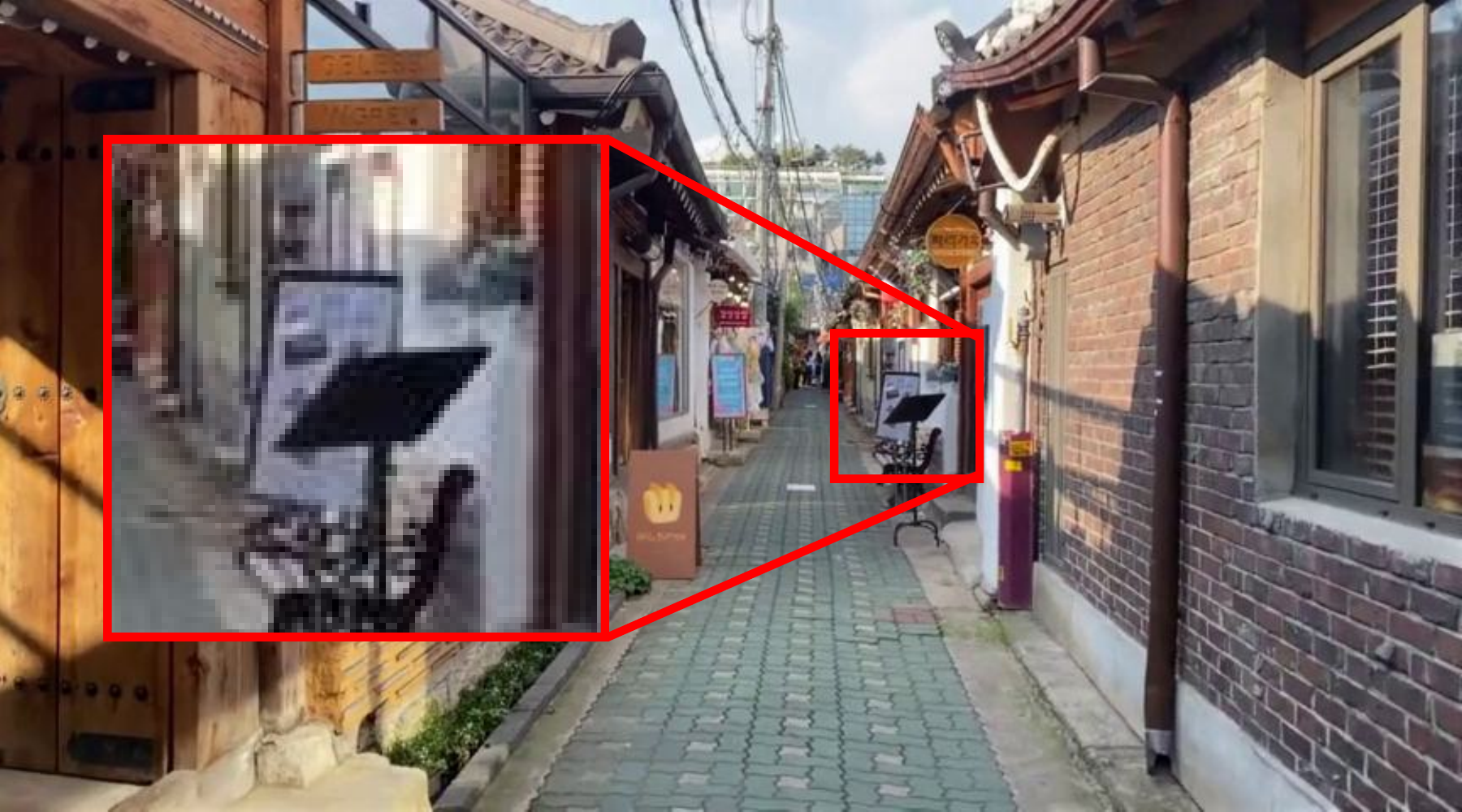} \\
    \includegraphics[width=0.235\textwidth]{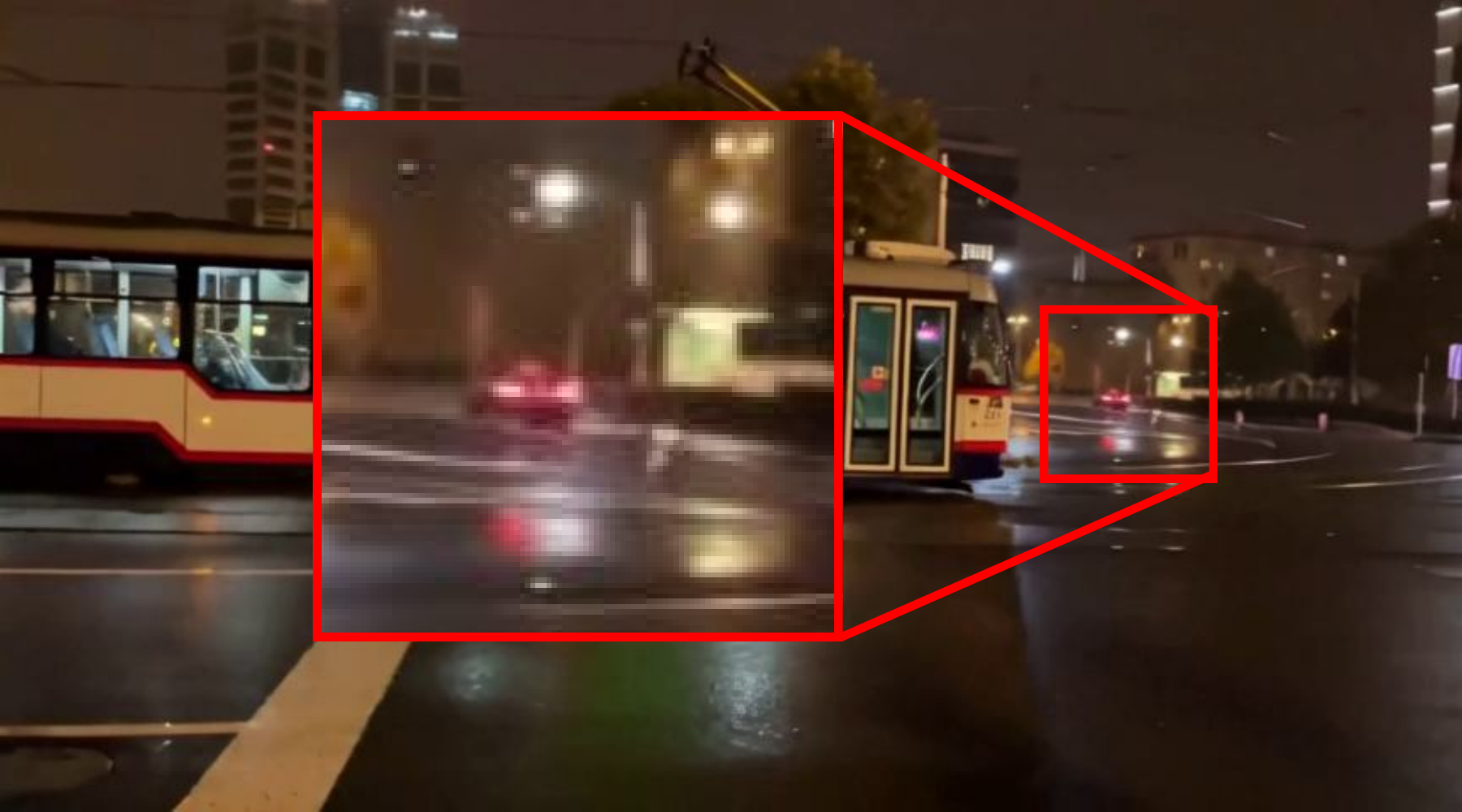} &
    \includegraphics[width=0.235\textwidth]{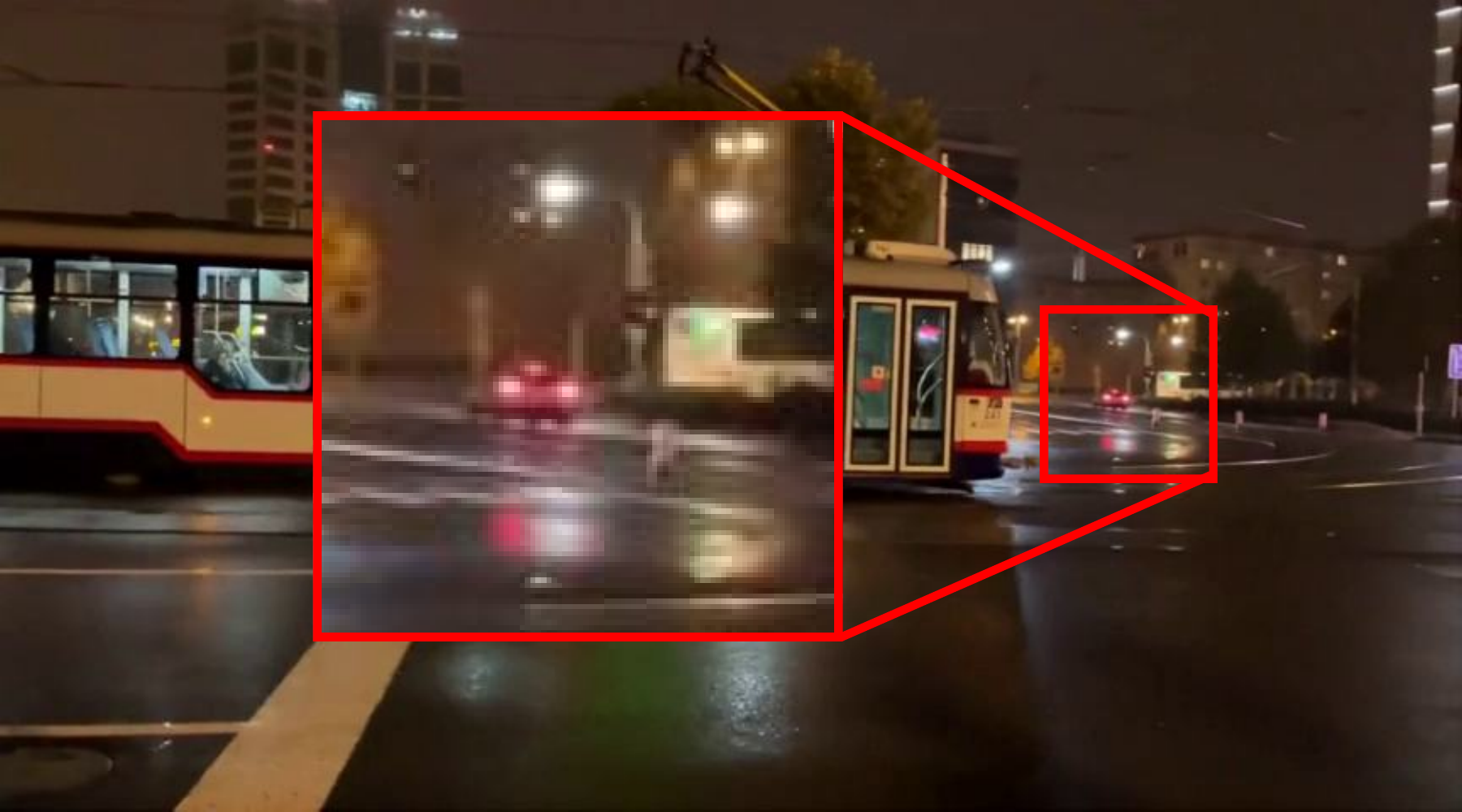} &
    \includegraphics[width=0.235\textwidth]{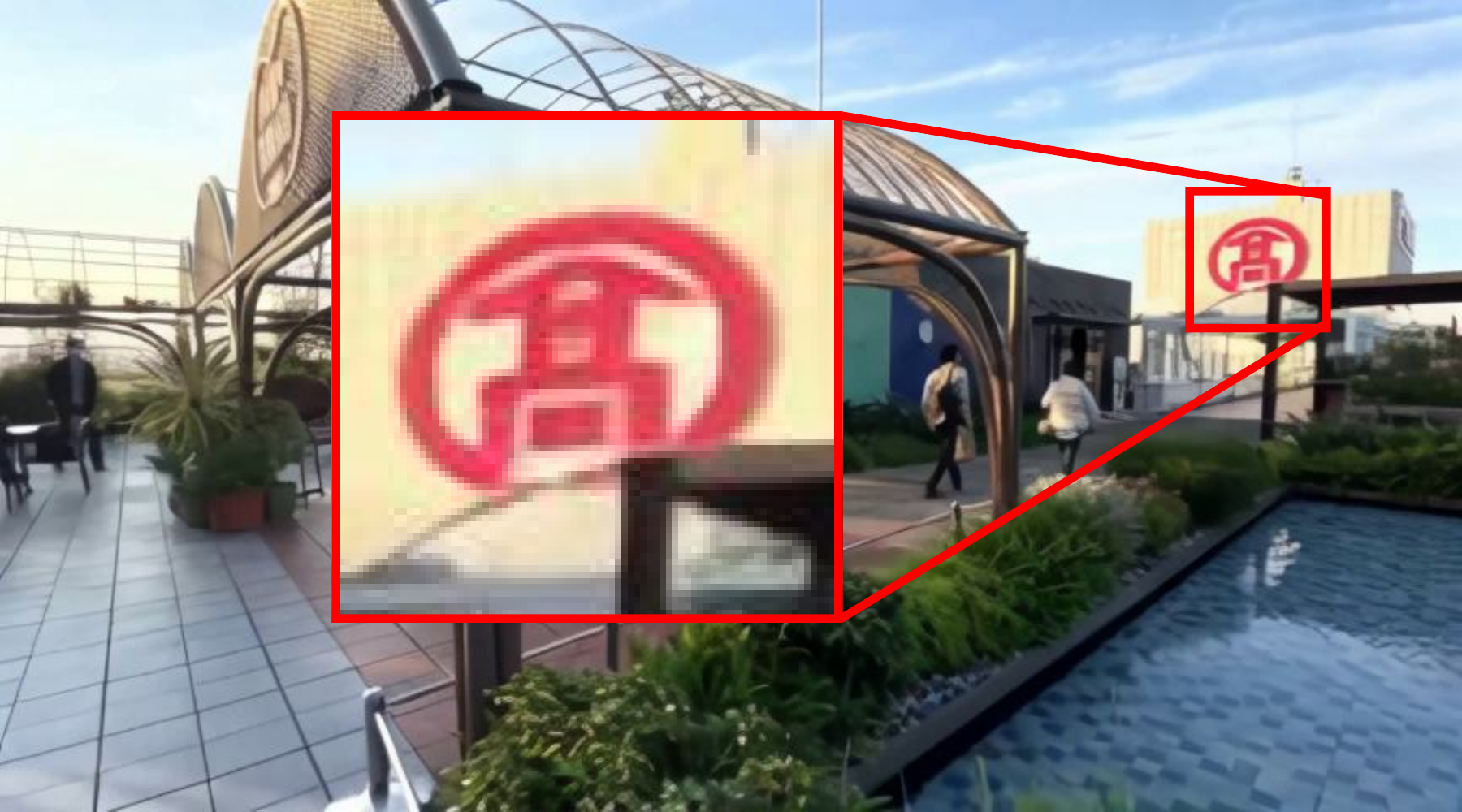} &
    \includegraphics[width=0.235\textwidth]{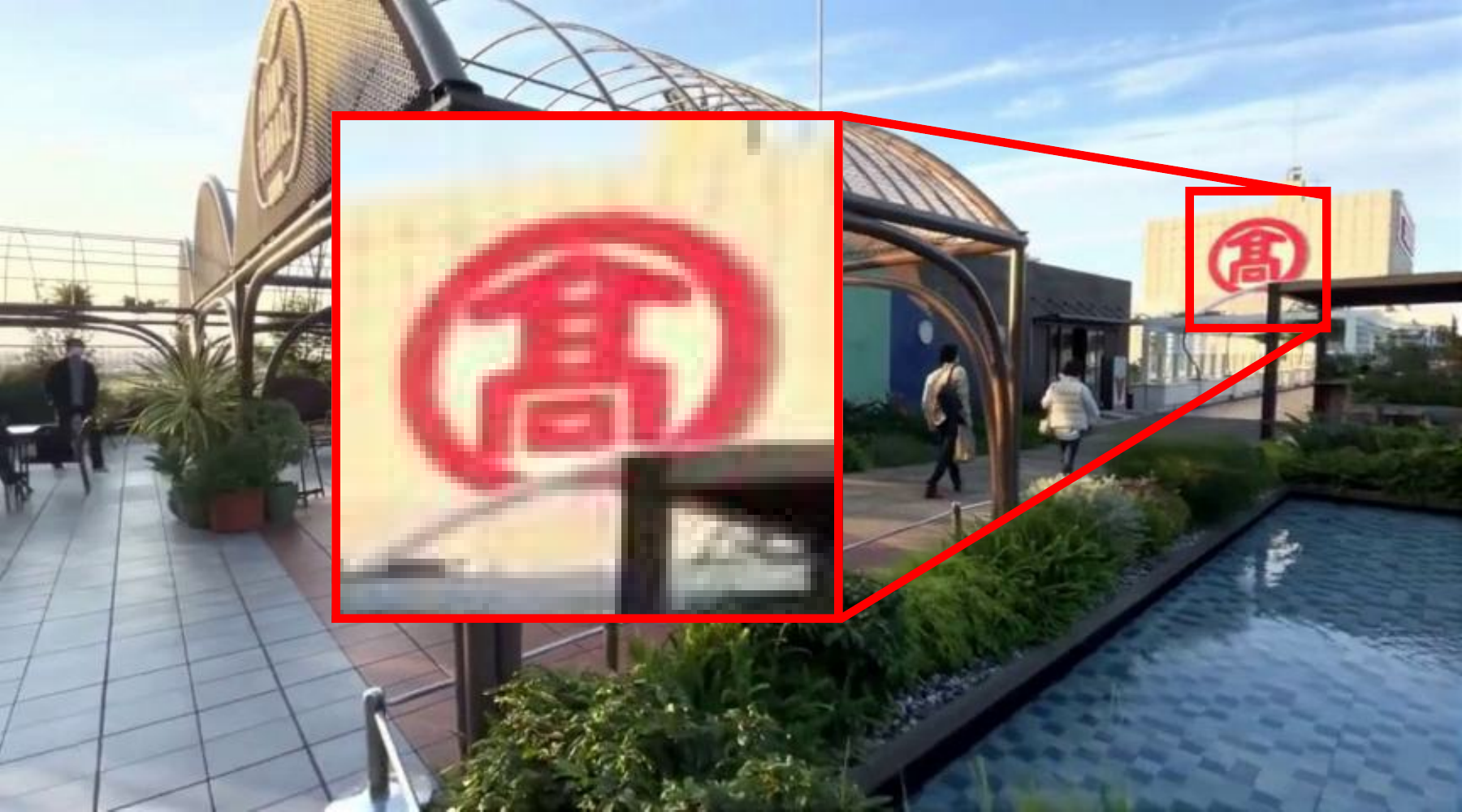} \\
    \includegraphics[width=0.235\textwidth]{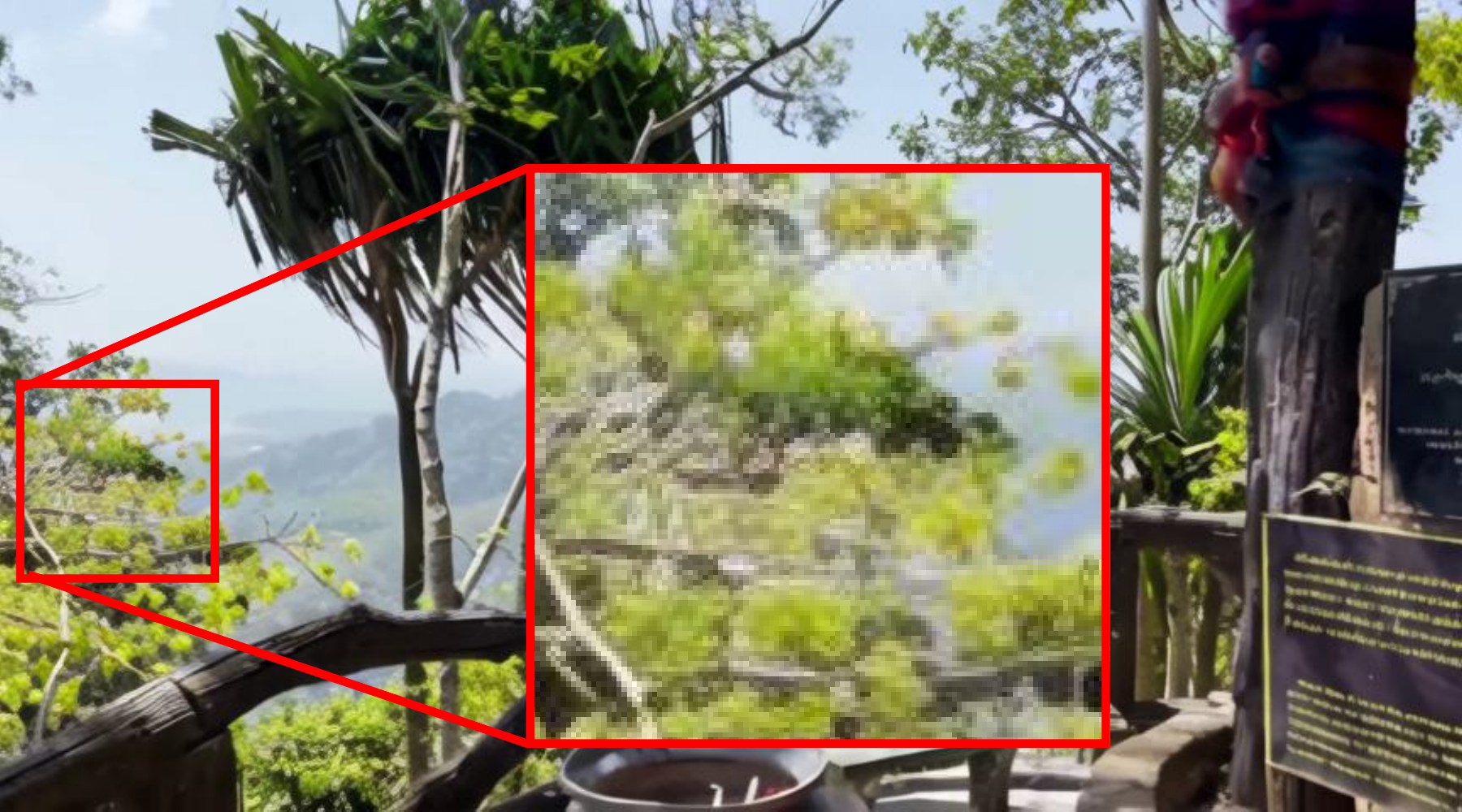} &
    \includegraphics[width=0.235\textwidth]{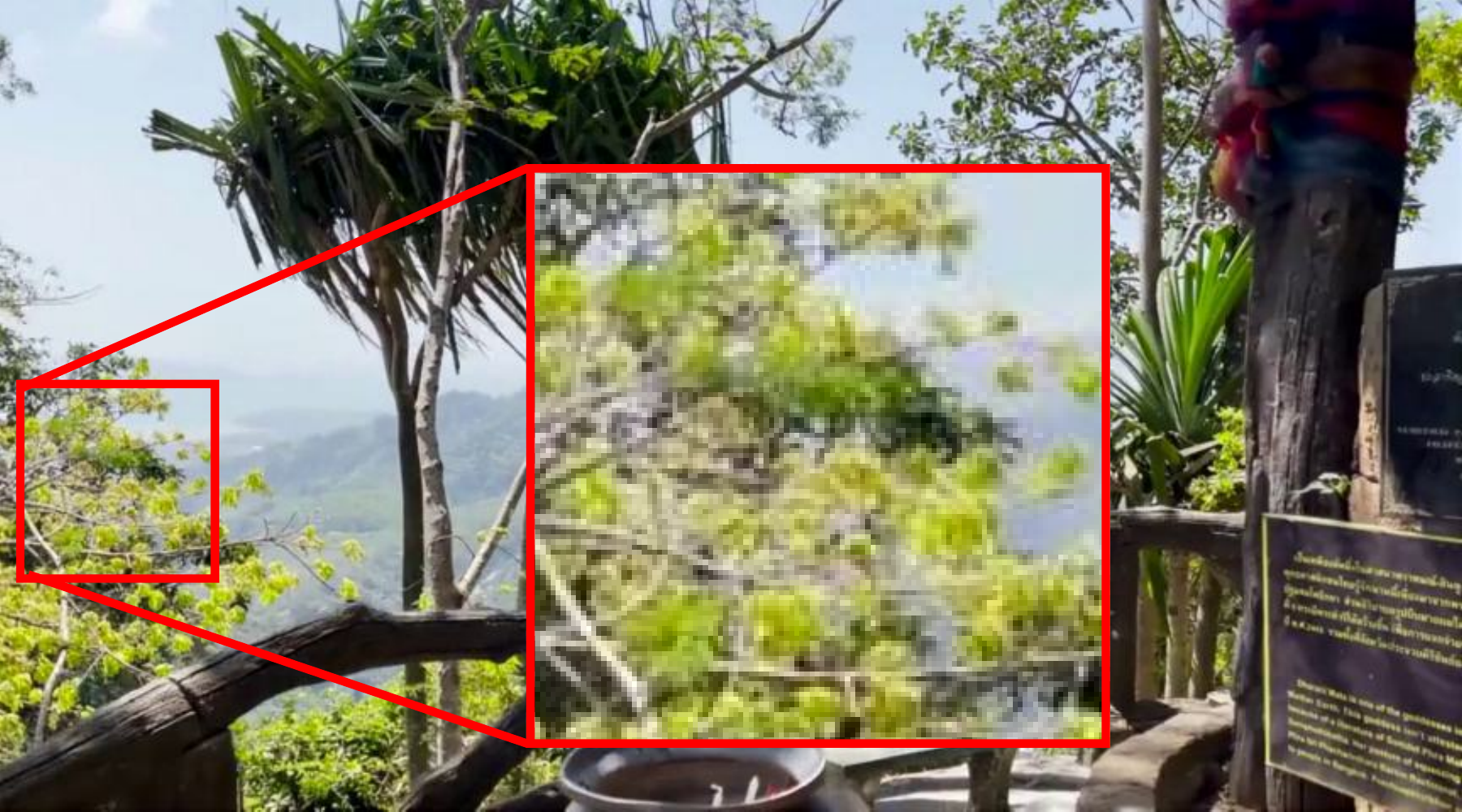} &
    \includegraphics[width=0.235\textwidth]{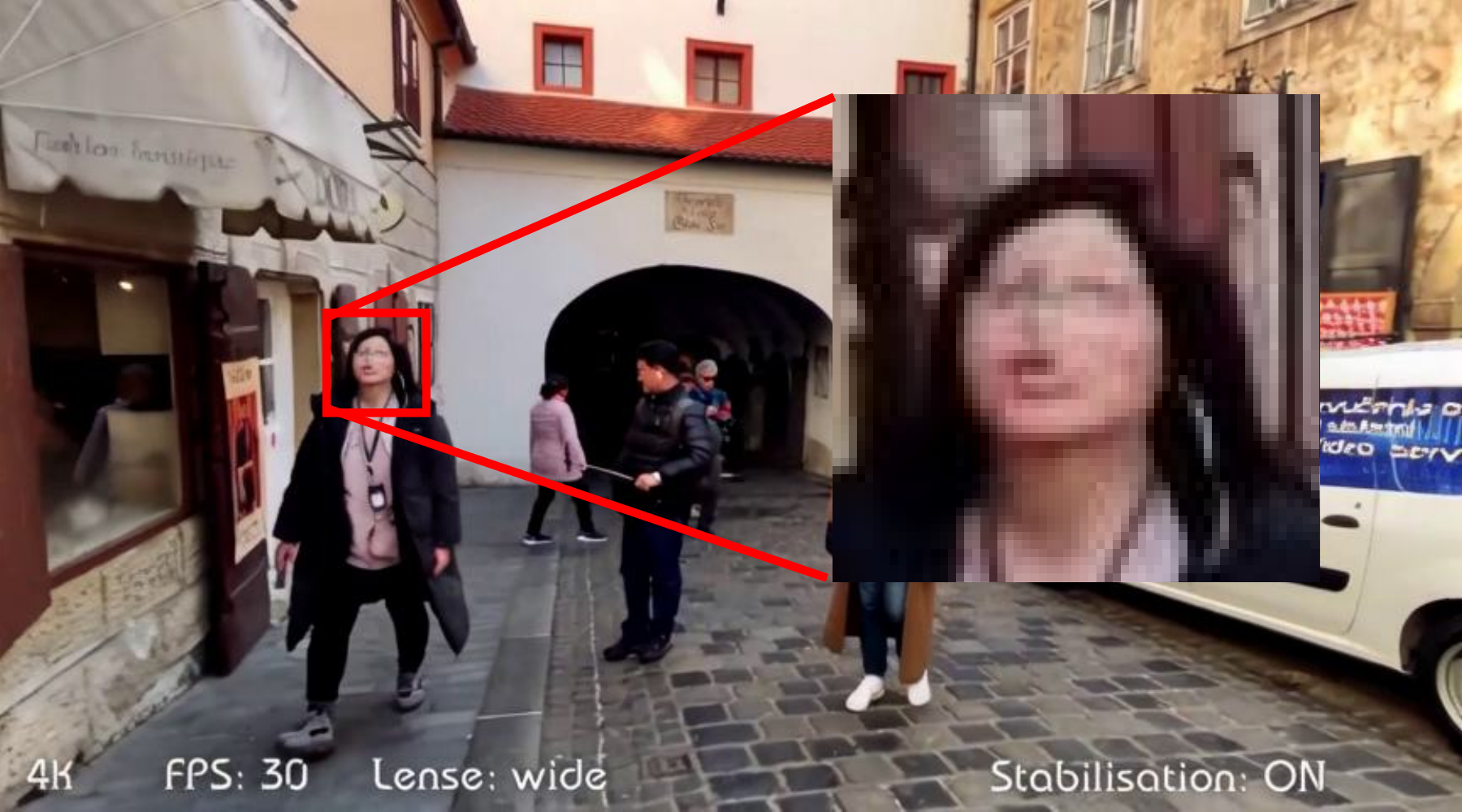} &
    \includegraphics[width=0.235\textwidth]{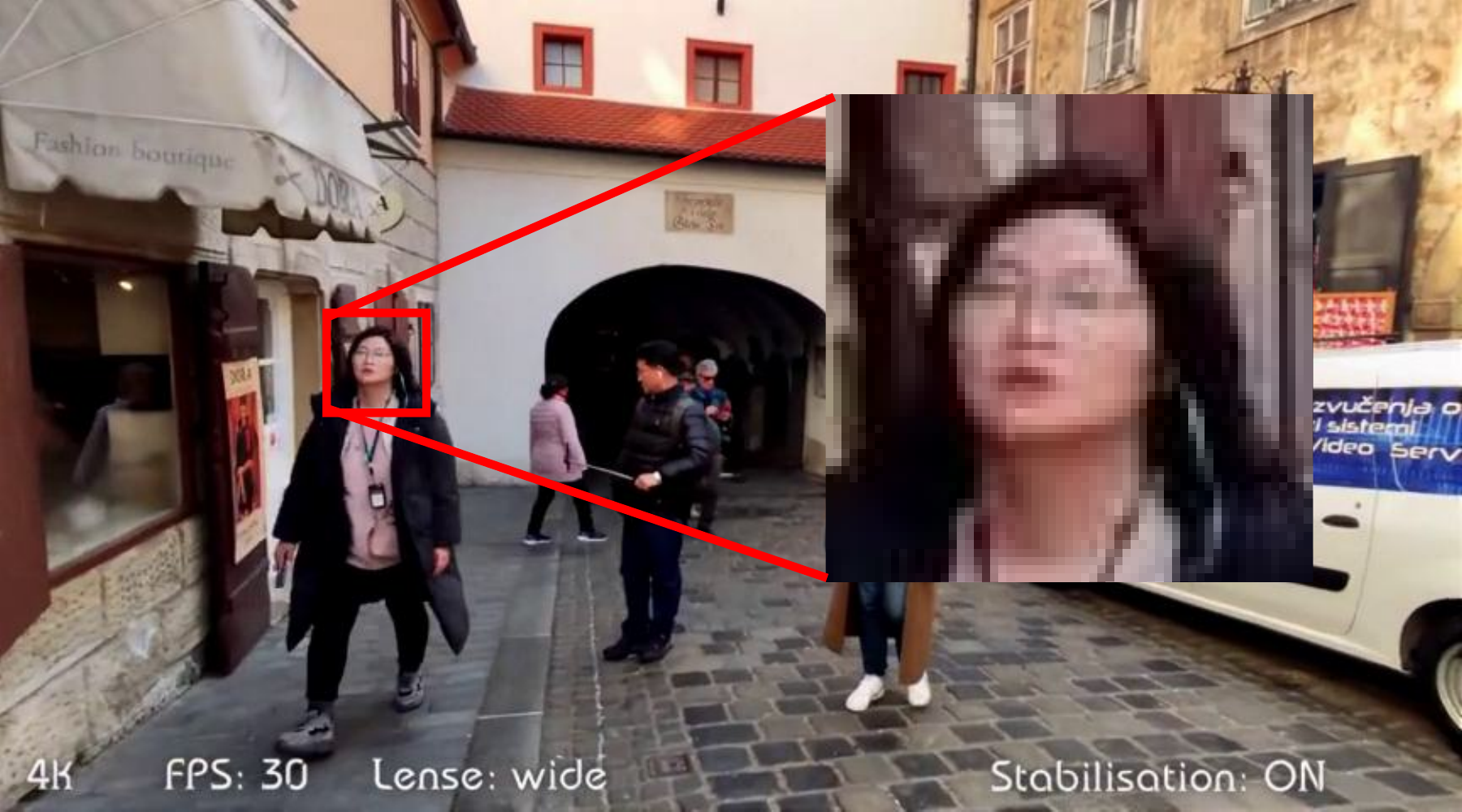} \\
    \includegraphics[width=0.235\textwidth]{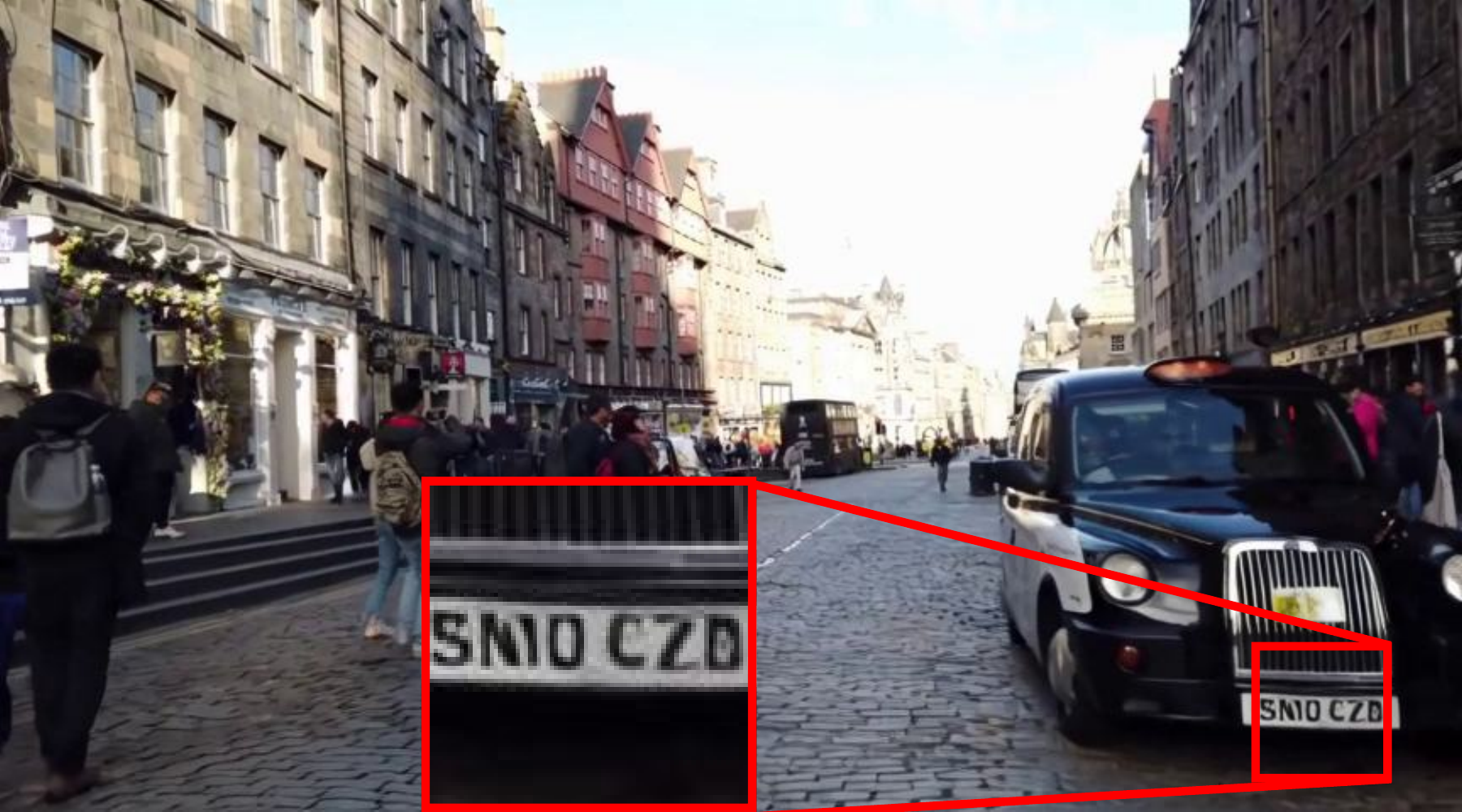} &
    \includegraphics[width=0.235\textwidth]{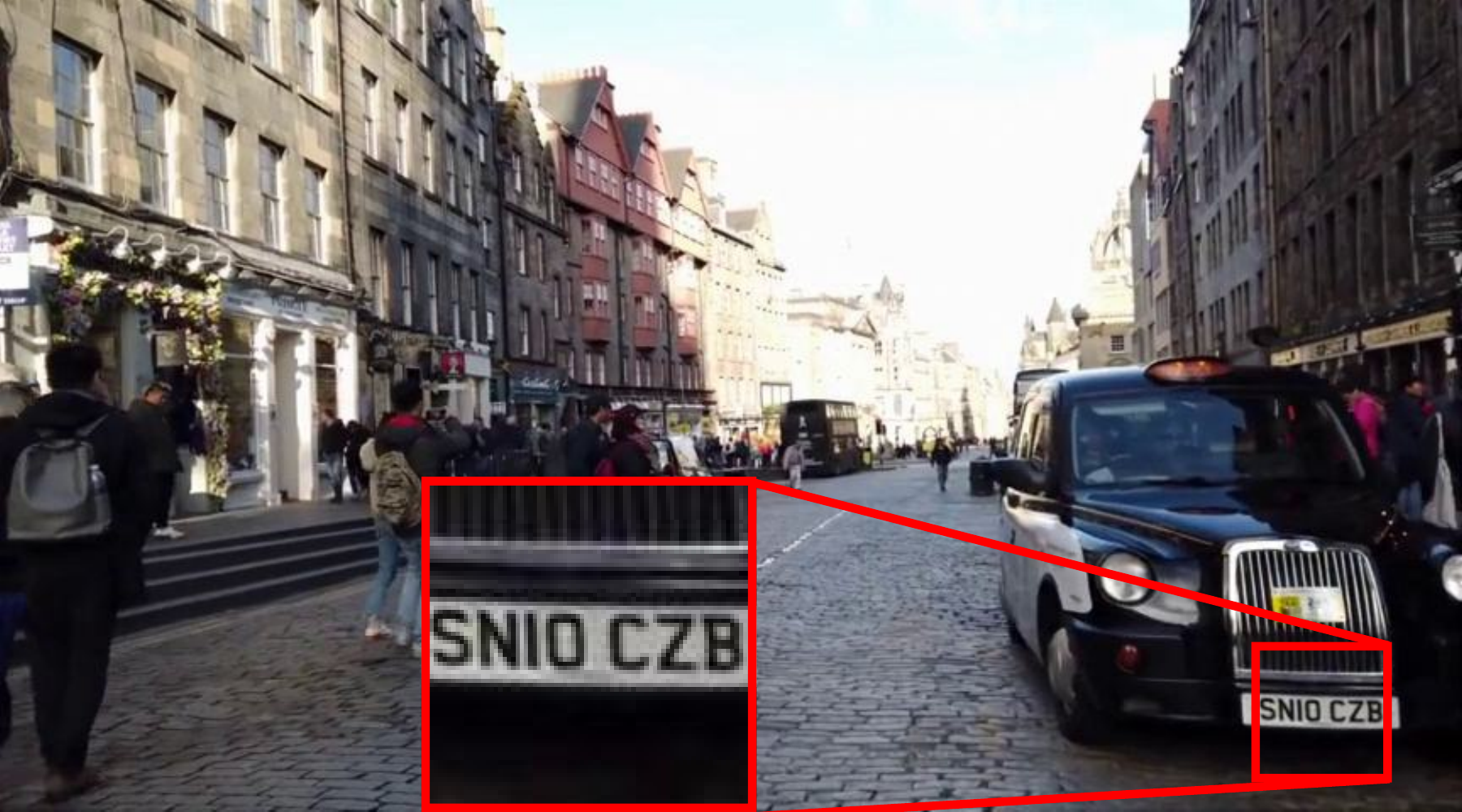} &
    \includegraphics[width=0.235\textwidth]{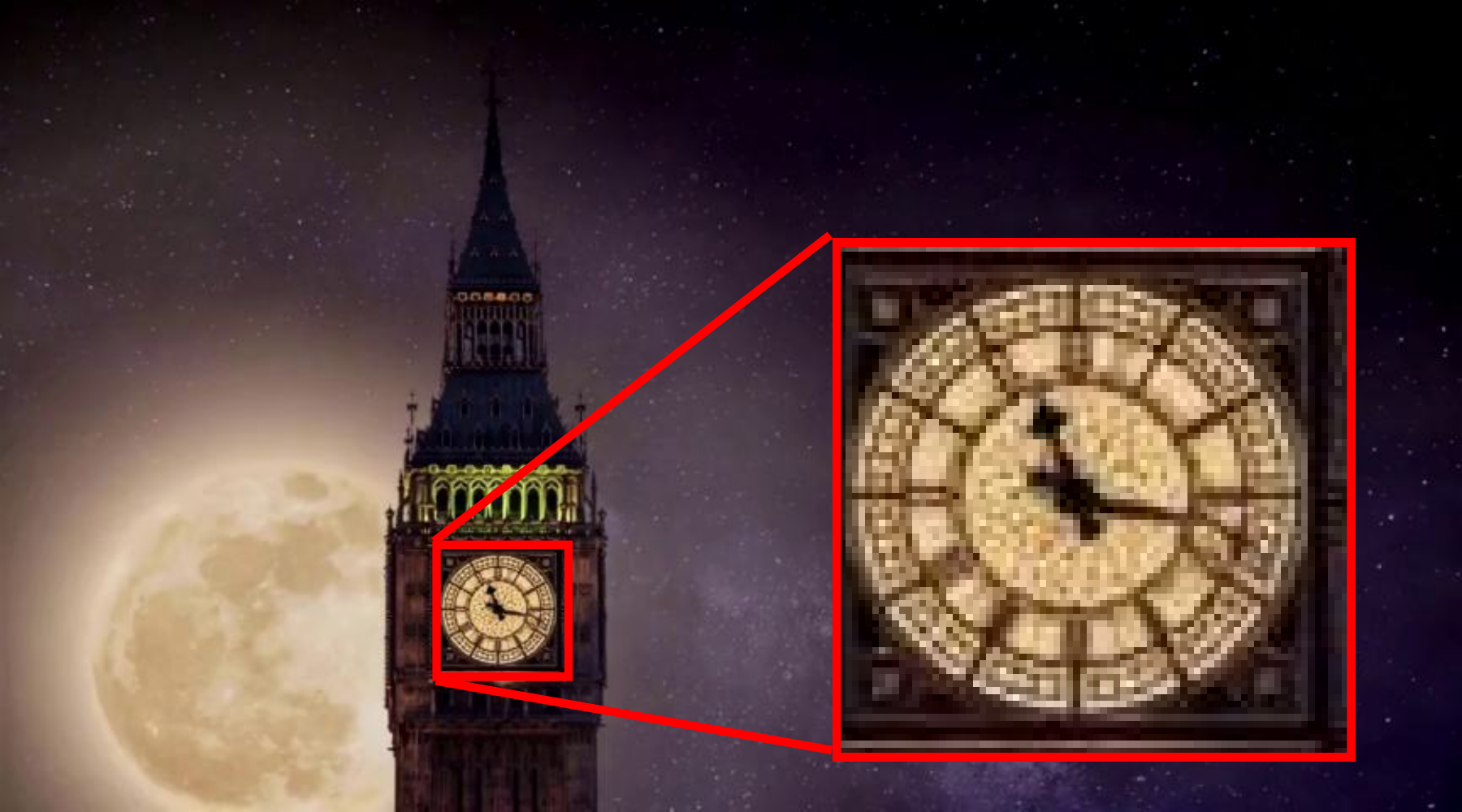} &
    \includegraphics[width=0.235\textwidth]{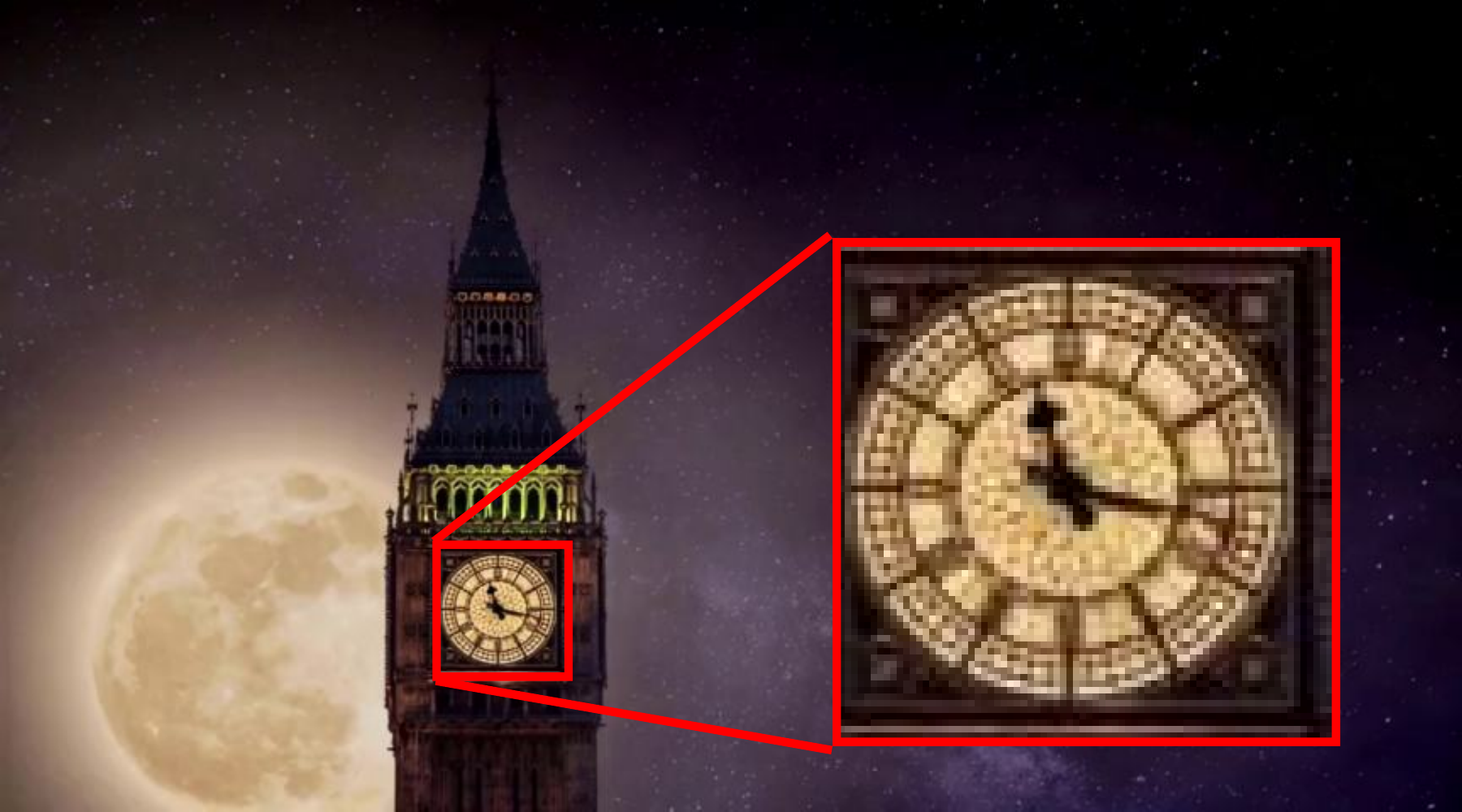} \\[2pt]
    \small{Baseline} & \small{Ours} & \small{Baseline} & \small{Ours} \\
\end{tabular}
\caption{\textbf{Additional VideoVAE\textbf{+} reconstruction comparisons.} Highlighted regions are shown for baseline vs.\ \model. Our method preserves fine-grained details such as text, facial features, and structural patterns more faithfully.}
\label{fig:supp_videovaeplus_recon}
\end{figure*}

Figure~\ref{fig:supp_wan_recon} shows reconstruction comparisons on the Wan~2.1 backbone. Each pair displays a cropped region from the same video frame, with the baseline decoder on the left and \model\ on the right. Figure~\ref{fig:supp_videovaeplus_recon} provides an analogous comparison on the VideoVAE\textbf{+} backbone with cropped region pairs. In both cases, \model\ recovers finer textures, sharper edges, and more faithful structural details such as text and facial features.

\begin{figure*}[t]
\centering
\setlength{\tabcolsep}{1pt}
\renewcommand{\arraystretch}{0.5}
\begin{tabular}{cc@{\hspace{10pt}}cccc}
    \multirow{2}{*}[-2pt]{\rotatebox{90}{\normalsize\textbf{GT}}} &
    \multirow{2}{*}[15pt]{\includegraphics[width=0.245\textwidth]{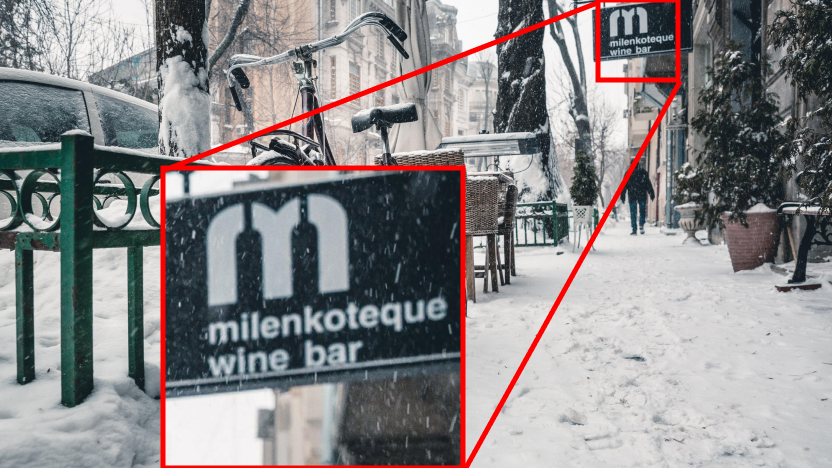}} &
    \rotatebox{90}{\small\hspace{15pt}Wan} &
    \includegraphics[width=0.245\textwidth]{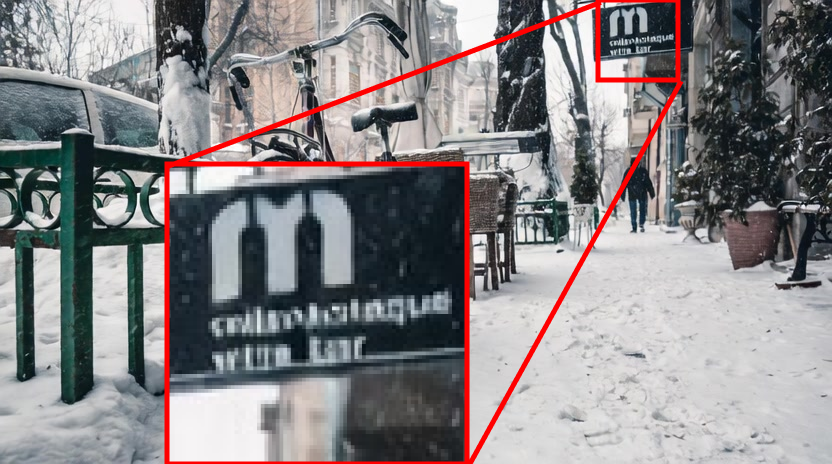} &
    \includegraphics[width=0.245\textwidth]{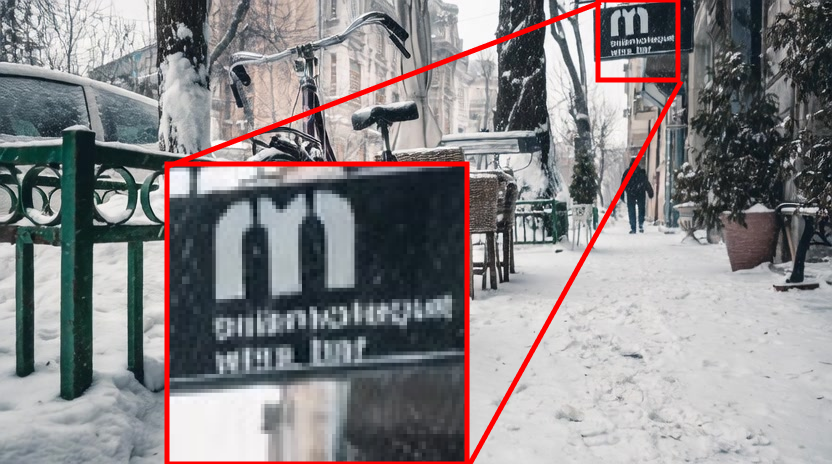} &
    \includegraphics[width=0.245\textwidth]{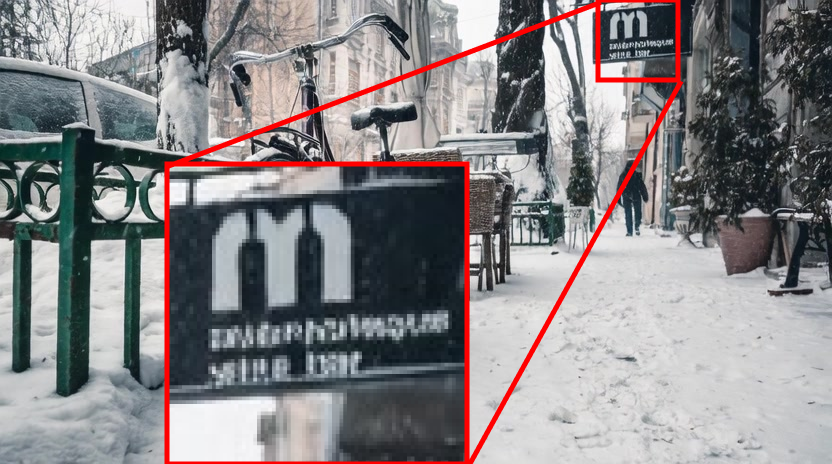} \\

    &
    &
    \rotatebox{90}{\small\hspace{8pt}Wan Ours} &
    \includegraphics[width=0.245\textwidth]{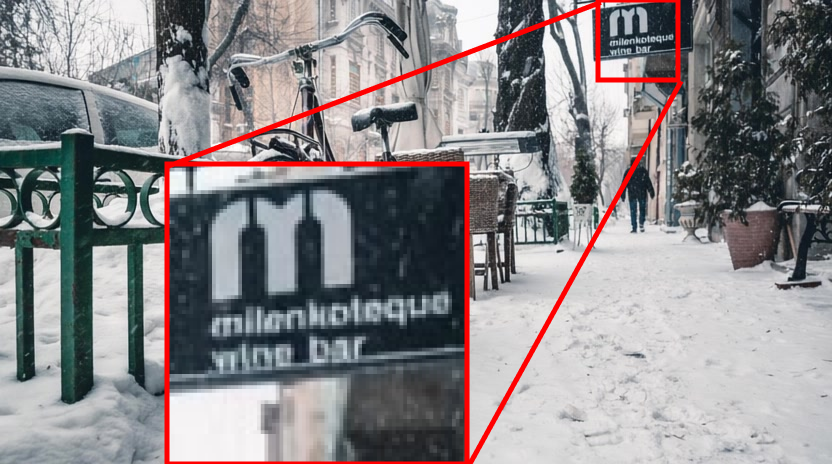} &
    \includegraphics[width=0.245\textwidth]{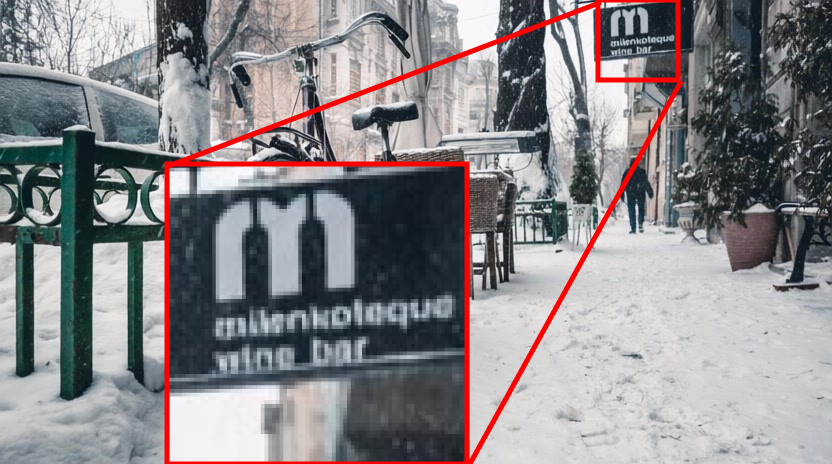} &
    \includegraphics[width=0.245\textwidth]{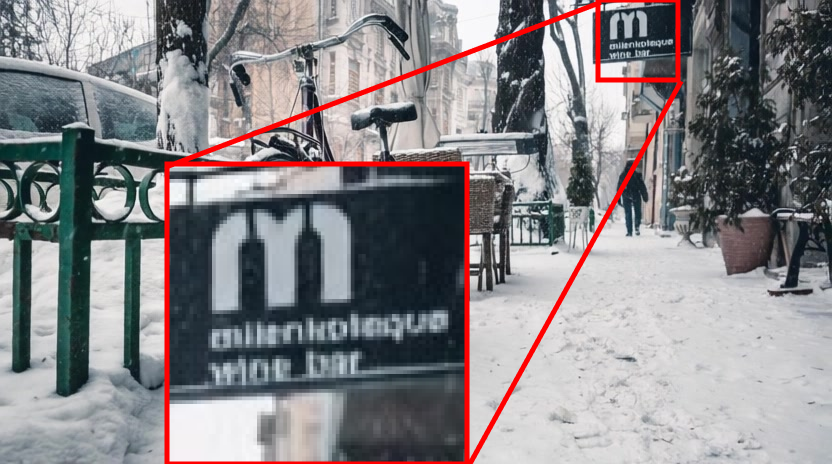} \\

    \multirow{2}{*}[-2pt]{\rotatebox{90}{\normalsize\textbf{GT}}} &
    \multirow{2}{*}[15pt]{\includegraphics[width=0.245\textwidth]{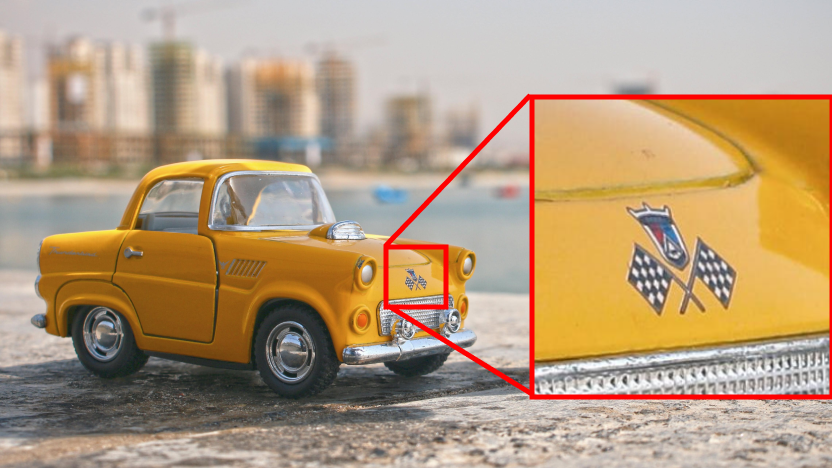}} &
    \rotatebox{90}{\small\hspace{15pt}Wan} &
    \includegraphics[width=0.245\textwidth]{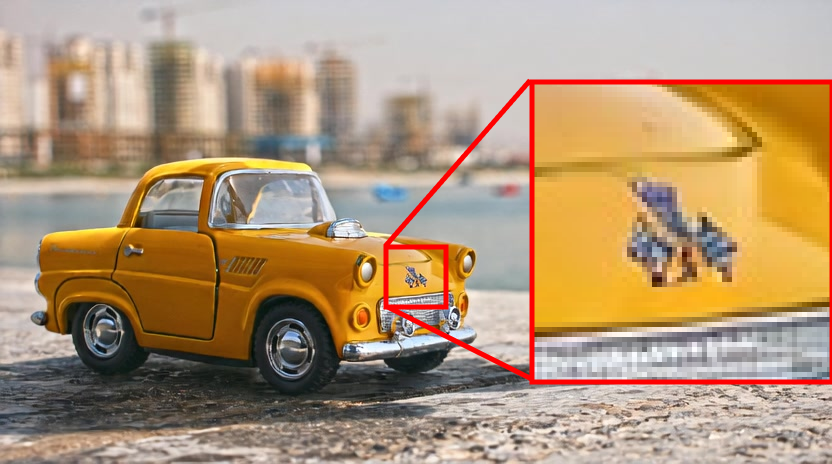} &
    \includegraphics[width=0.245\textwidth]{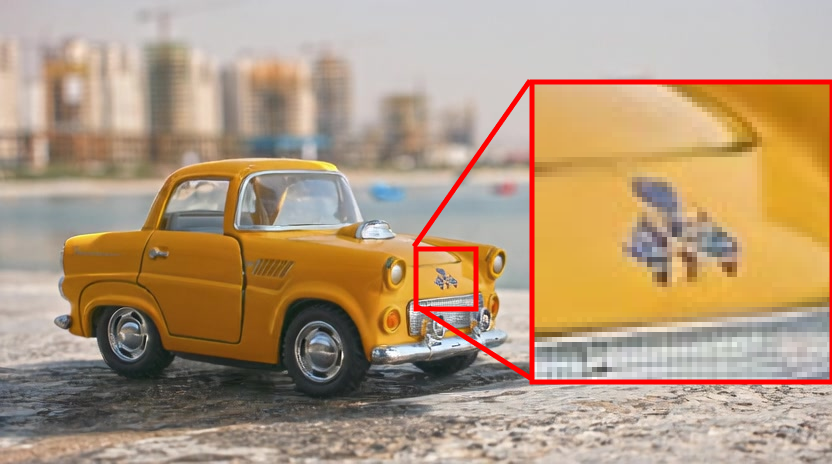} &
    \includegraphics[width=0.245\textwidth]{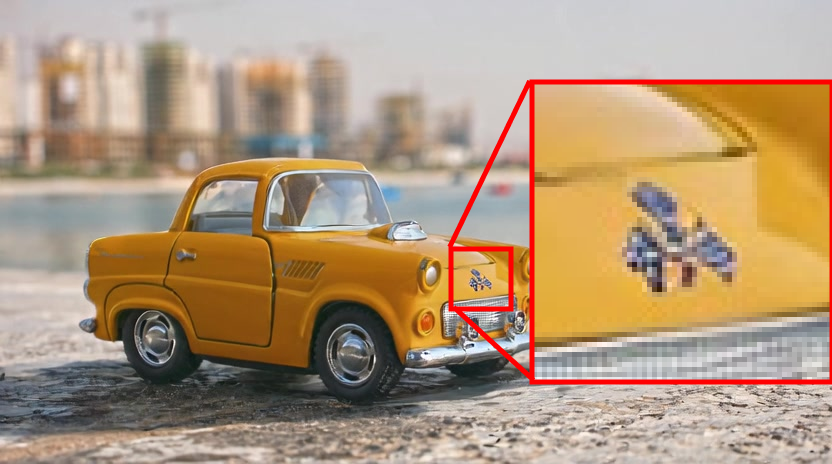} \\

    &
    &
    \rotatebox{90}{\small\hspace{8pt}Wan Ours} &
    \includegraphics[width=0.245\textwidth]{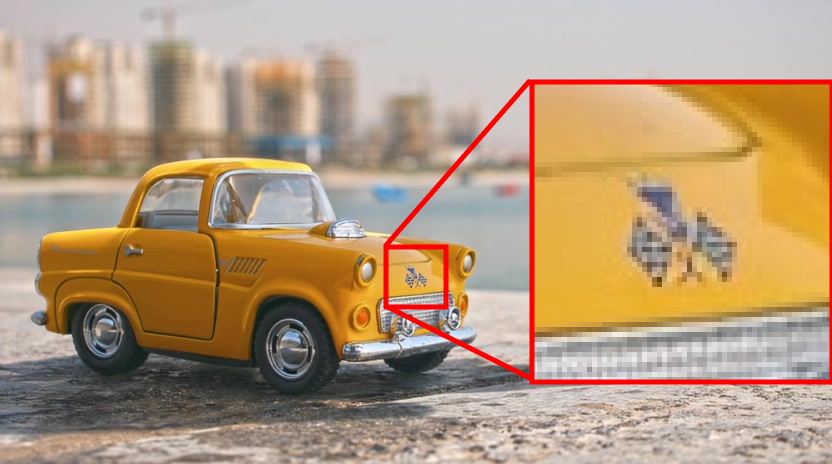} &
    \includegraphics[width=0.245\textwidth]{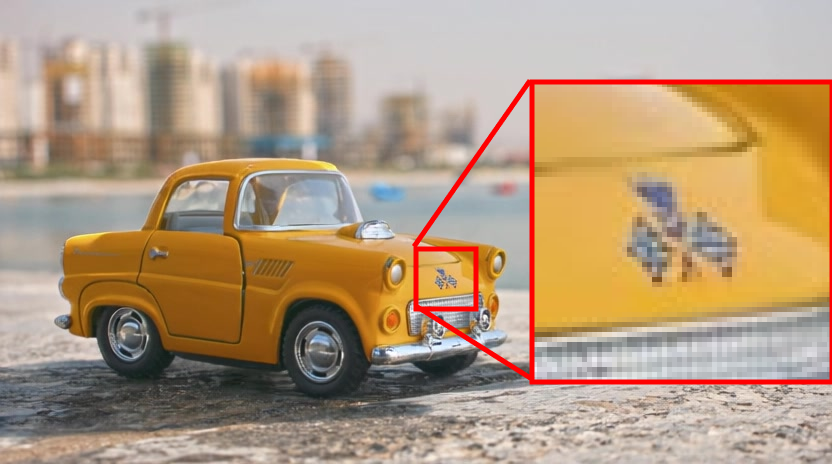} &
    \includegraphics[width=0.245\textwidth]{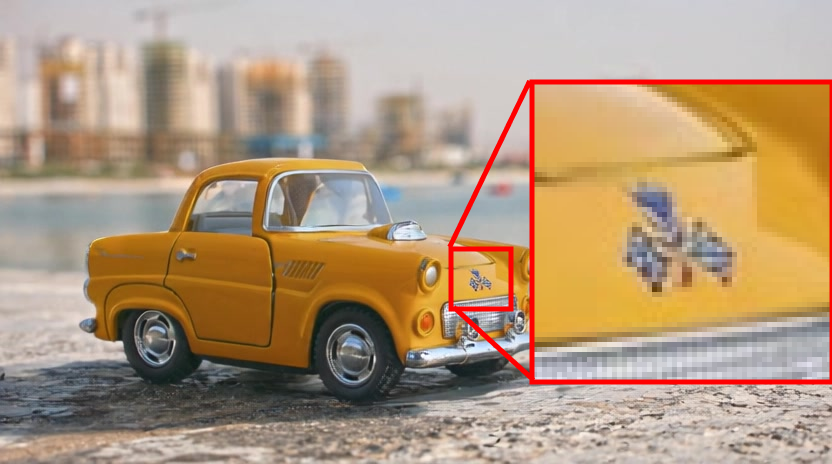} \\

    \multirow{2}{*}[-2pt]{\rotatebox{90}{\normalsize\textbf{GT}}} &
    \multirow{2}{*}[15pt]{\includegraphics[width=0.245\textwidth]{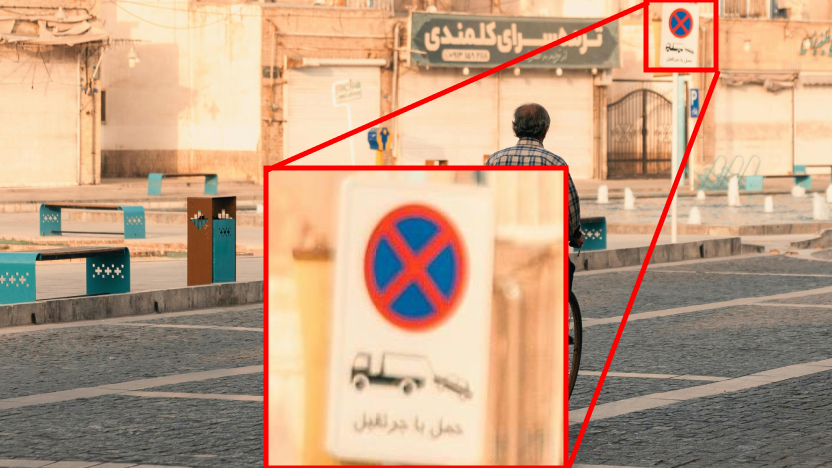}} &
    \rotatebox{90}{\small\hspace{15pt}Wan} &
    \includegraphics[width=0.245\textwidth]{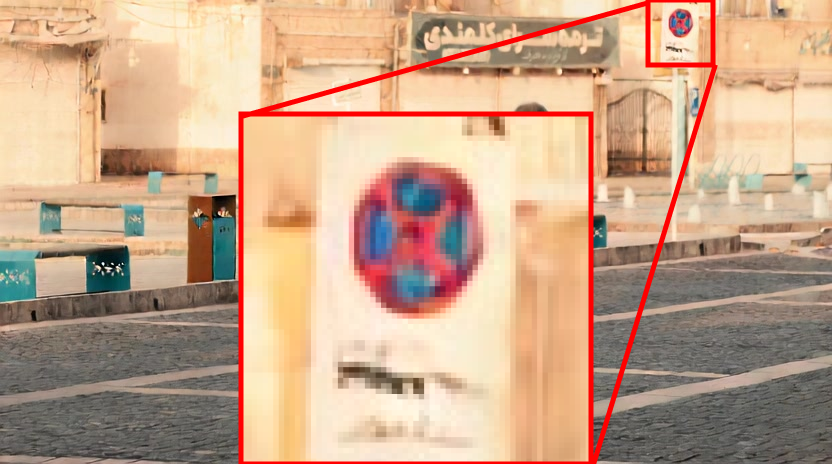} &
    \includegraphics[width=0.245\textwidth]{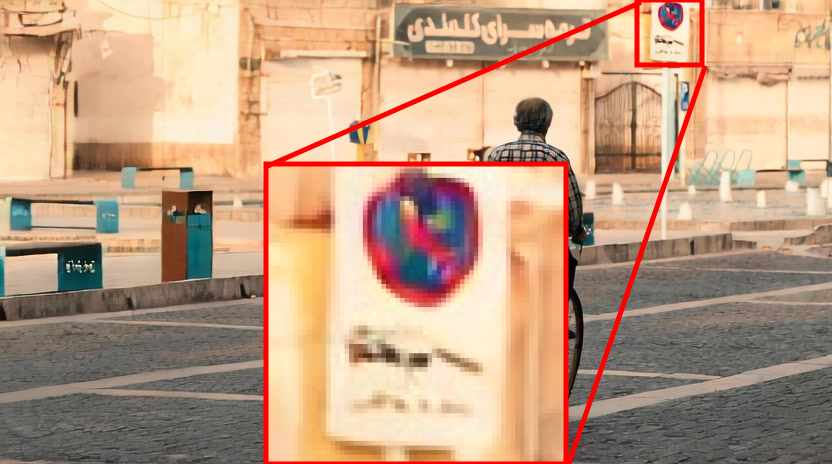} &
    \includegraphics[width=0.245\textwidth]{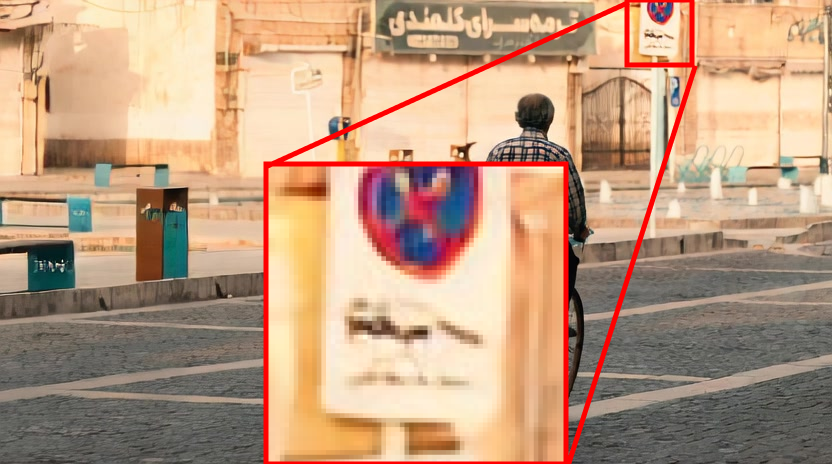} \\

    &
    &
    \rotatebox{90}{\small\hspace{8pt}Wan Ours} &
    \includegraphics[width=0.245\textwidth]{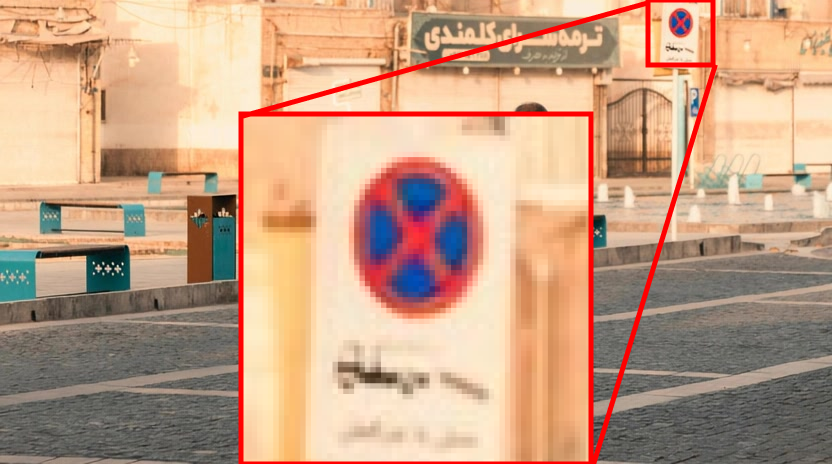} &
    \includegraphics[width=0.245\textwidth]{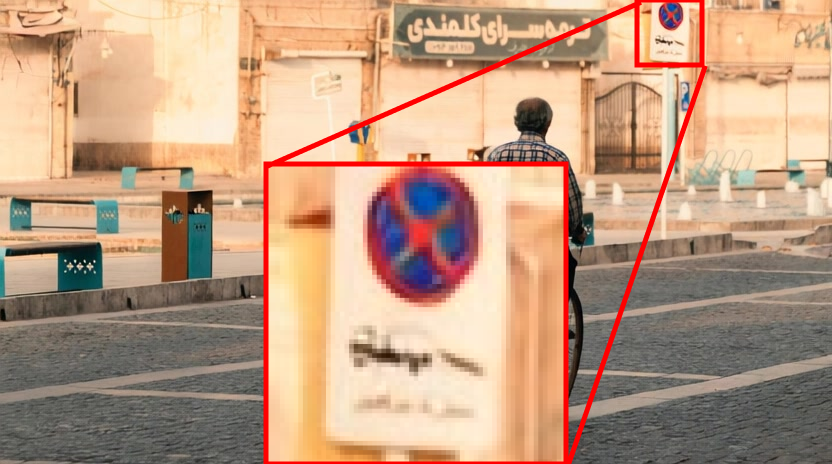} &
    \includegraphics[width=0.245\textwidth]{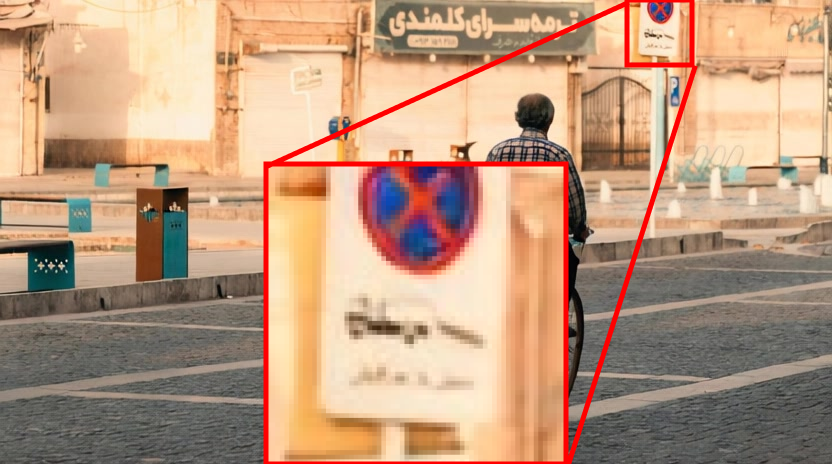} \\

    \multirow{2}{*}[-2pt]{\rotatebox{90}{\normalsize\textbf{GT}}} &
    \multirow{2}{*}[15pt]{\includegraphics[width=0.245\textwidth]{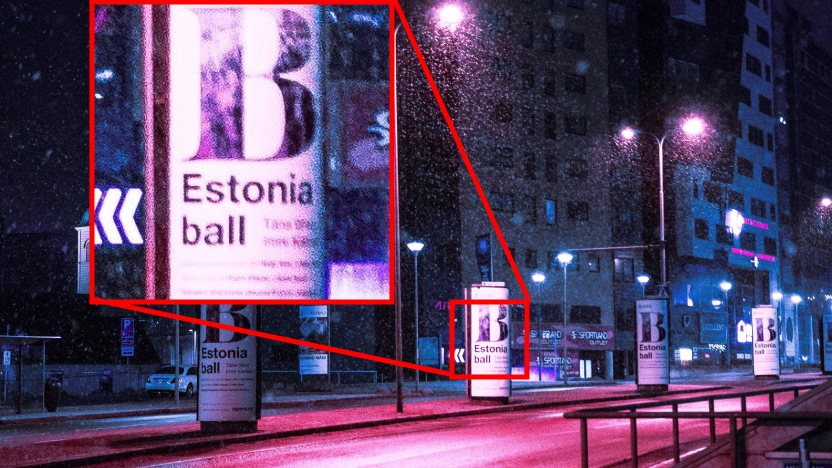}} &
    \rotatebox{90}{\small\hspace{15pt}Wan} &
    \includegraphics[width=0.245\textwidth]{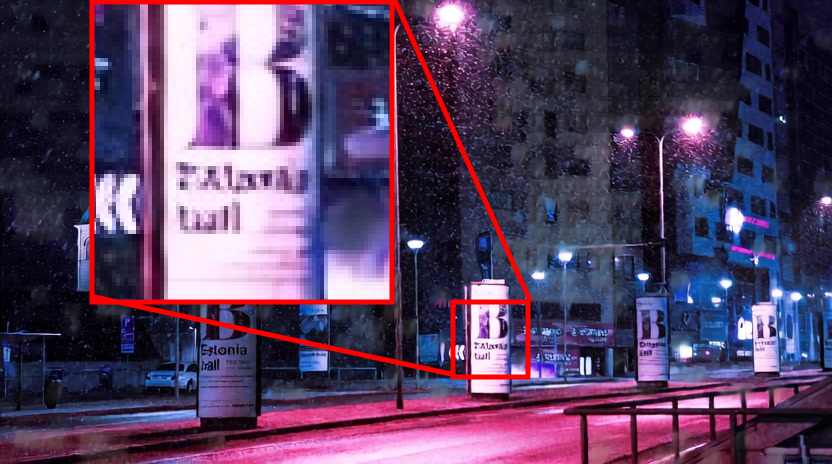} &
    \includegraphics[width=0.245\textwidth]{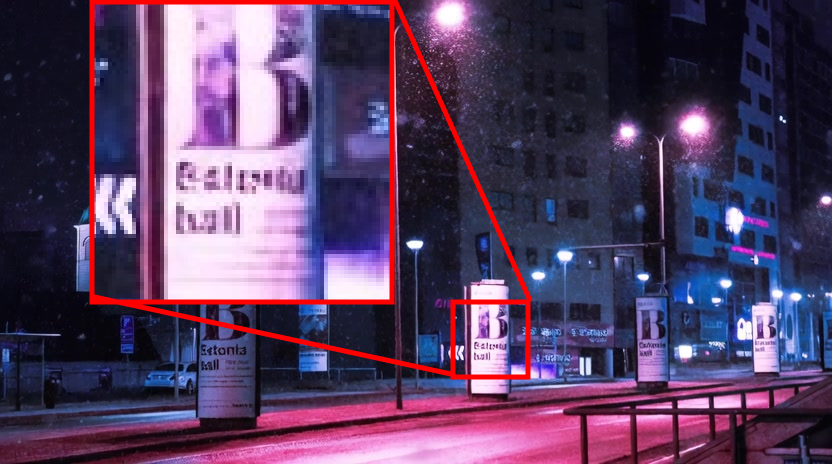} &
    \includegraphics[width=0.245\textwidth]{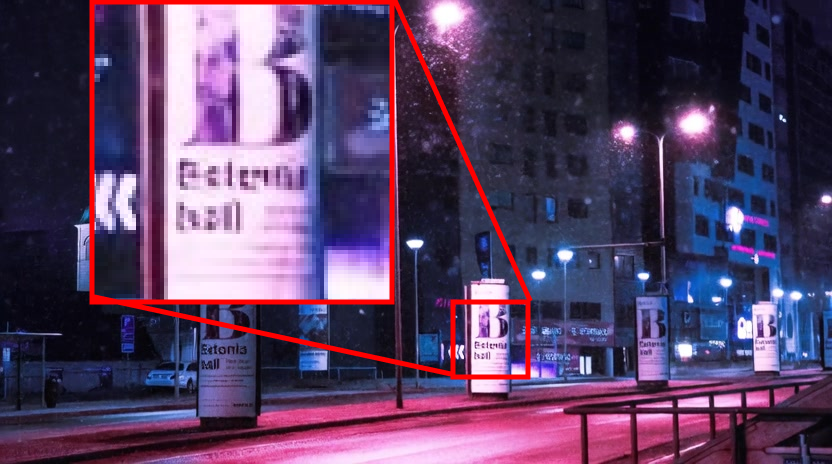} \\

    &
    &
    \rotatebox{90}{\small\hspace{8pt}Wan Ours} &
    \includegraphics[width=0.245\textwidth]{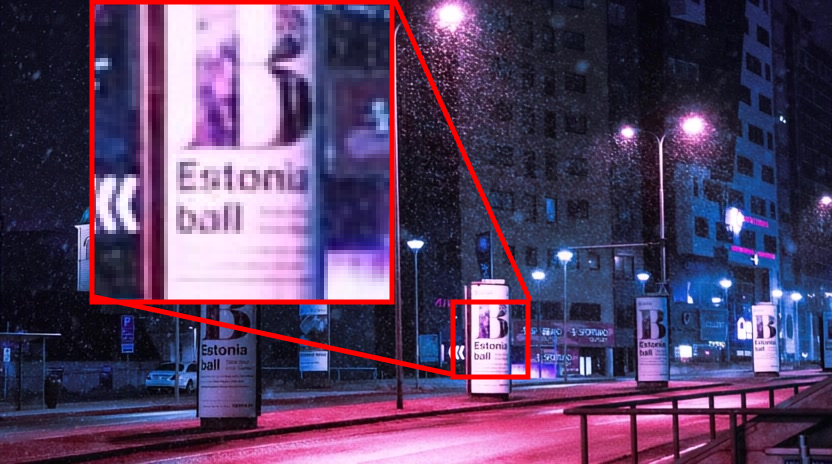} &
    \includegraphics[width=0.245\textwidth]{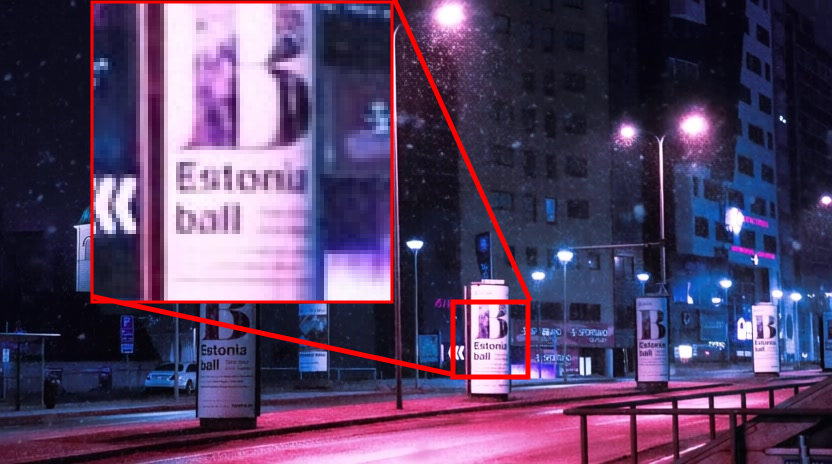} &
    \includegraphics[width=0.245\textwidth]{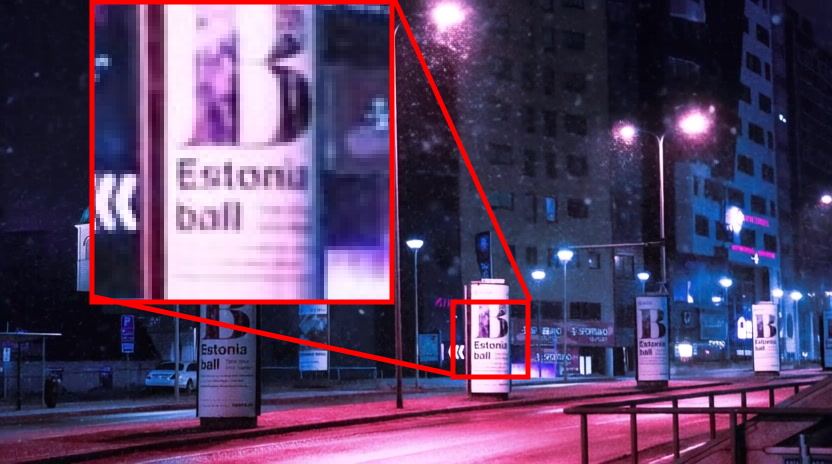} \\
\end{tabular}
\caption{\textbf{Additional video generation comparisons.} For each example, the GT reference image and three generated frames at different timesteps are shown. \model\ preserves scene structure and produces sharper, more consistent details across frames compared to the baseline.}
\label{fig:supp_wan_i2v_gen}
\end{figure*}

Figure~\ref{fig:supp_wan_i2v_gen} shows video generation comparisons on the Wan~2.1 backbone. For each example, the ground-truth reference image is shown alongside three generated frames (first, middle, and last) from both the baseline model and \model. Our method better preserves scene structure and fine-grained appearance from the reference image while maintaining temporal consistency across frames.

\clearpage

\end{document}